\documentclass[oneside, openright, titlepage, numbers=noenddot, headinclude, letterpaper, footinclude=true, cleardoublepage=empty, abstractoff, fontsize=12pt, american, dottedtoc]{scrreprt}

\pdfoutput=1
\PassOptionsToPackage{eulerchapternumbers,listings,%drafting,
				      pdfspacing,%floatperchapter,%linedheaders,%
				      subfig,beramono,eulermath,parts}{classicthesis}								

\usepackage{ifthen}
\newboolean{enable-backrefs} % enable backrefs in the bibliography
\setboolean{enable-backrefs}{false} % true false
\newcommand{\myTitle}{Curved Surface Patches for Rough Terrain Perception\xspace}
\newcommand{\myName}{Dimitrios Kanoulas\xspace}
\newcommand{\myTime}{July 24, 2014\xspace}
\newcommand{\myUni}{Northeastern University\xspace}

\newcounter{dummy} % necessary for correct hyperlinks (to index, bib, etc.)
\providecommand{\mLyX}{L\kern-.1667em\lower.25em\hbox{Y}\kern-.125emX\@}

\PassOptionsToPackage{latin9}{inputenc}	% latin9 (ISO-8859-9) = latin1+"Euro sign"
 \usepackage{inputenc}				

 \usepackage{babel}					

\PassOptionsToPackage{square,numbers}{natbib}
 \usepackage{natbib}				

 \usepackage{amsmath}

\PassOptionsToPackage{T1}{fontenc} % T2A for cyrillics
	\usepackage{fontenc}     
\usepackage{textcomp} % fix warning with missing font shapes
\usepackage{scrhack} % fix warnings when using KOMA with listings package          
\usepackage{xspace} % to get the spacing after macros right  
\usepackage{mparhack} % get marginpar right
\usepackage{fixltx2e} % fixes some LaTeX stuff 
\PassOptionsToPackage{printonlyused,smaller}{acronym}
\usepackage{acronym} % nice macros for handling all acronyms in the thesis
\usepackage{tabularx} % better tables
\usepackage{caption}
\usepackage{subfig}

\usepackage{listings} 
\usepackage{enumerate}

\PassOptionsToPackage{pdftex,hyperfootnotes=false,pdfpagelabels}{hyperref}
	\usepackage{hyperref}  % backref linktocpage pagebackref
\PassOptionsToPackage{pdftex}{graphicx}
	\usepackage{graphicx} 

\newcommand{\backrefnotcitedstring}{\relax}%(Not cited.)
\newcommand{\backrefcitedsinglestring}[1]{(Cited on page~#1.)}
\newcommand{\backrefcitedmultistring}[1]{(Cited on pages~#1.)}
\ifthenelse{\boolean{enable-backrefs}}%
{%
		\PassOptionsToPackage{hyperpageref}{backref}
		\usepackage{backref} % to be loaded after hyperref package 
		   \renewcommand*{\backref}[1]{}  % disable standard
		   \renewcommand*{\backrefalt}[4]{% detailed backref
		      \ifcase #1 %
		         \backrefnotcitedstring%
		      \or%
		         \backrefcitedsinglestring{#2}%
		      \else%
		         \backrefcitedmultistring{#2}%
		      \fi}%
}{\relax}    

\hypersetup{%
    colorlinks=true, linktocpage=true, pdfstartpage=3, pdfstartview=FitV,%
    breaklinks=true, pdfpagemode=UseNone, pageanchor=true, pdfpagemode=UseOutlines,%
    plainpages=false, bookmarksnumbered, bookmarksopen=true, bookmarksopenlevel=1,%
    hypertexnames=true, pdfhighlight=/O,%nesting=true,%frenchlinks,%
    urlcolor=webbrown, linkcolor=RoyalBlue, citecolor=webgreen, %pagecolor=RoyalBlue,%
    pdftitle={\myTitle},%
    pdfauthor={\textcopyright\ \myName, \myUni},%
    pdfsubject={},%
    pdfkeywords={},%
    pdfcreator={pdfLaTeX},%
    pdfproducer={LaTeX with hyperref and classicthesis}%
}   

\makeatletter
\@ifpackageloaded{babel}%
    {%
       \addto\extrasamerican{%
				}%
       \addto\extrasngerman{% 
				}%	
    }{\relax}
\makeatother

\usepackage{classicthesis} 
\areaset[current]{422pt}{761pt} % 686 (factor 2.2) + 33 head + 42 head \the\footskip
\graphicspath{{./figures/}}
\usepackage{amssymb}
\usepackage{multirow}
\usepackage{algorithm}
\usepackage{algpseudocode}
\usepackage{setspace}
\newcommand{\specialcell}[2][c]{\begin{tabular}[#1]{@{}c@{}}#2\end{tabular}}
\usepackage{pdfpages}
\begin{document}
\frenchspacing
\raggedbottom
\selectlanguage{american}
%\renewcommand*{\bibname}{new name}
%\setbibpreamble{}
\pagenumbering{roman}
\pagestyle{plain}

%******************************************************************************
% Frontmatter
%******************************************************************************
% Macros
\newcommand\postpubfixes{true}

\def\beqn#1\eeqn{\begin{eqnarray}#1\end{eqnarray}}
\def\beq#1\eeq{\begin{equation}#1\end{equation}}
\def\bea#1\eea{\begin{align}#1\end{align}}
\def\beg#1\eeg{\begin{gather}#1\end{gather}}
\def\beqs#1\eeqs{\begin{equation*}#1\end{equation*}}
\def\beas#1\eeas{\begin{align*}#1\end{align*}}
\def\begs#1\eegs{\begin{gather*}#1\end{gather*}}
\def\bpm#1\epm{\begin{pmatrix}#1\end{pmatrix}}

\newcommand{\step}[1]{\vspace{0.5em}\noindent{#1}}
\newcommand{\poly}{\mathrm{poly}}
\newcommand{\eps}{\epsilon}
\newcommand{\e}{\epsilon}
\newcommand{\polylog}{\mathrm{polylog}}
\newcommand{\rob}[1]{\left( #1 \right)} %Round Brackets
\newcommand{\sqb}[1]{\left[ #1 \right]} %square Brackets
\newcommand{\cub}[1]{\left\{ #1 \right\} } %curly brackets
\newcommand{\rb}[1]{\left( #1 \right)} %Round
\newcommand{\abs}[1]{\left| #1 \right|} %| |
\newcommand{\zo}{\{0, 1\}}
\newcommand{\zonzo}{\zo^n \to \zo}
\newcommand{\zokzo}{\zo^k \to \zo}
\newcommand{\zot}{\{0,1,2\}}

\newcommand{\en}[1]{\marginpar{\textbf{#1}}}
\newcommand{\efn}[1]{\footnote{\textbf{#1}}}

\renewcommand{\vec}[1]{\mathbf{#1}} %looks better
\newcommand{\vecbm}[1]{\boldmath{#1}} %more general (handles greek letters)
\newcommand{\uvec}[1]{\hat{\vec{#1}}}
\newcommand{\vecmb}[1]{\mathbold{#1}} %handles better zero vectors

\newcommand{\thv}{\vecbm{\theta}}

\newcommand{\var}{\mathop{\mathrm{var}}}
\newcommand{\rank}{\mathop{\mathrm{rank}}}
\newcommand{\diag}{\mathop{\mathrm{diag}}}
\newcommand{\tr}{\mathop{\mathrm{tr}}}
\newcommand{\acos}{\mathop{\mathrm{acos}}}
\newcommand{\atantwo}{\mathop{\mathrm{atan2}}}
\newcommand{\SVD}{\mathop{\mathrm{SVD}}}
\newcommand{\quadf}{\mathop{\mathrm{q}}}
\newcommand{\linterp}{\mathop{\mathrm{l}}}
\newcommand{\sgn}{\mathop{\mathrm{sign}}}
\newcommand{\sym}{\mathop{\mathrm{sym}}}
\newcommand{\avg}{\mathop{\mathrm{avg}}}
\newcommand{\mean}{\mathop{\mathrm{mean}}}
\newcommand{\erf}{\mathop{\mathrm{erf}}}
\newcommand{\grad}{\nabla}
\newcommand{\R}{\mathbb{R}}
\newcommand{\defeq}{\triangleq}
\newcommand{\dims}[2]{[#1\!\times\!#2]}
\newcommand{\sdims}[2]{\mathsmaller{#1\!\times\!#2}}
\newcommand{\udims}[3]{#1}
\newcommand{\udimst}[4]{#1}

\newcommand{\com}[1]{\rhd\text{\emph{#1}}}
\newcommand{\ind}{\hspace{1em}}

\newcommand{\argmin}[1]{\underset{#1}{\operatorname{argmin}}}
\newcommand{\floor}[1]{\left\lfloor{#1}\right\rfloor}

% useful comment to temporarily remove unwanted text
\newcommand{\junk}[1]{}

%% Better referencing 
\newcommand{\eref}[1] {(\ref{#1})} % simple equation references
\newcommand{\Eref}[1] {Equation (\ref{#1})} % equation references
\newcommand{\fref}[1] {Figure \ref{#1}} % figure references
\newcommand{\fpath}[1]{figures/pdfs/#1} % path to the graphics

\newcommand{\quat}[1]{\ensuremath{\mathring{\mathbf{#1}}}}

\newcommand{\pard}[2]{\frac{\partial{#1}}{\partial{#2}}}
% Titlepage

\begin{titlepage}
  %\begin{addmargin}[-1cm]{-3cm}
    \begin{center}
    \large
        
    \hfill \vfill

    {\color{Maroon} {\huge Curved Surface Patches}}\\ \medskip \medskip
    {\huge {\it for}}\\ \medskip
    {\color{Maroon} {\huge Rough Terrain Perception}}
		
    \vfill
        
    \spacedlowsmallcaps{\myName} \\ \medskip \medskip
    \myTime \\ \medskip
		
    \vfill
    \end{center}
  %\end{addmargin}
  
  %\begin{addmargin}[-1cm]{-3cm}
    \begin{center}
    \large 
        
    Submitted in partial fulfillment of the requirements\\
	for the degree of Doctor of Philosophy\\ \medskip \medskip
    
    {\it to the}\\ \medskip \medskip
    
    Faculty of the College\\
    of Computer and Information Science\\
    Northeastern University\\
    Boston, Massachusetts

    \vfill                      
    \end{center}
  %\end{addmargin}
\end{titlepage}
% Titleback & Colophon

\thispagestyle{empty}

\hfill \vfill

\pdfbookmark[0]{Colophon}{colophon}
\section*{Colophon}
This document was typeset using the \texttt{classicthesis} template ({\url{http://code.google.com/p/classicthesis/}}) in \LaTeX\ developed by Andr\'e Miede and inspired by Robert Bringhurst's seminal book on typography ``\emph{The Elements of Typographic Style}''.

\hfill

This material is based upon work supported by the National Science Foundation under Grant No. 1149235.  The portion of this research related to human subjects was approved by the Northeastern University Institutional Review Board (IRB \# 13-06-21).

\hfill

\noindent {\it \myTitle}\\
\textcopyright\ \myTime, \myName
\includepdf[pages={1}]{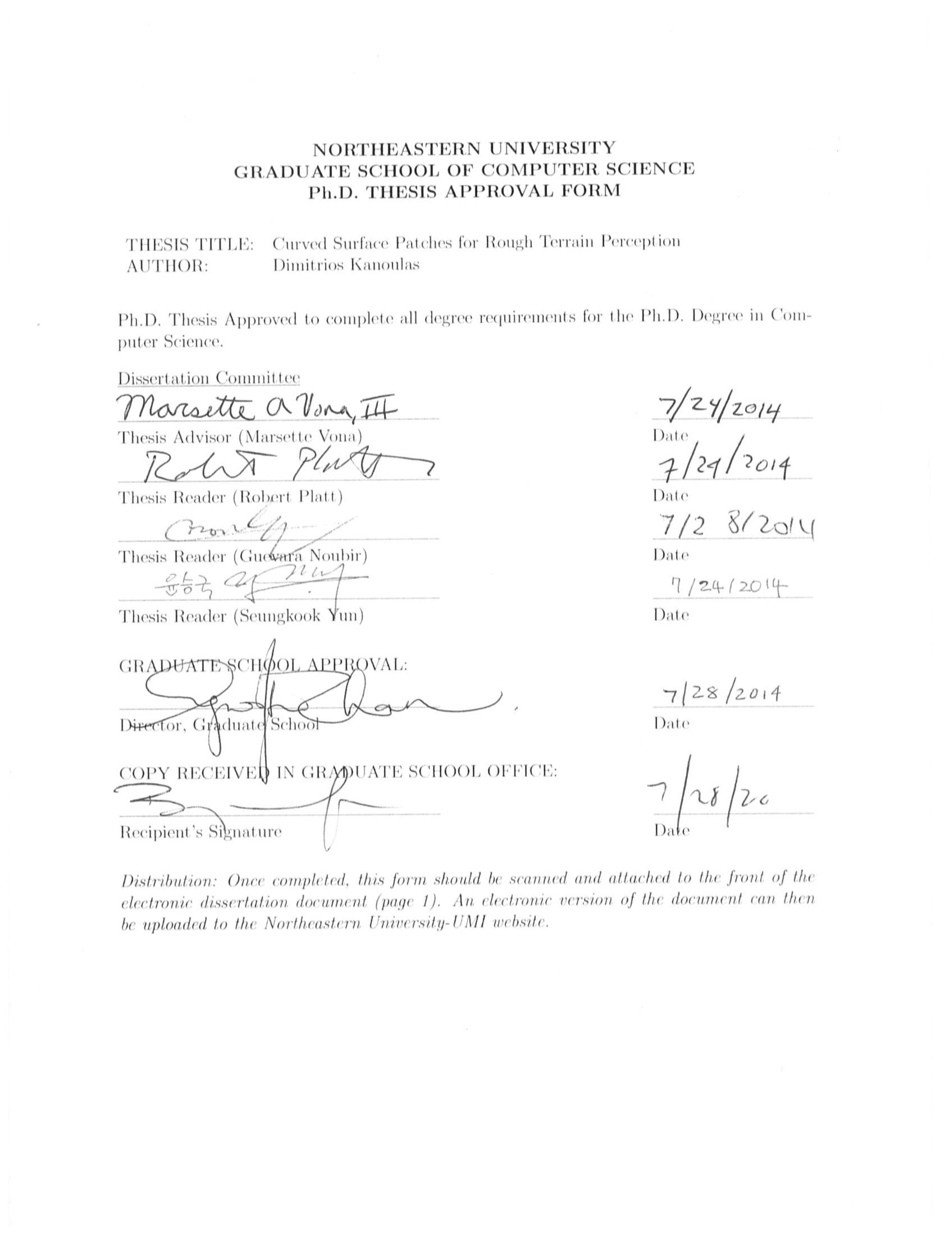}
%*******************************************************
% Dedication
%*******************************************************
\thispagestyle{empty}
\phantomsection 
\refstepcounter{dummy}
\pdfbookmark[1]{Dedication}{Dedication}

\vspace*{3cm}

\begin{center}
{\small {\it You have no responsibility to live up to what other people think you ought to accomplish.  I have no responsibility to be like they expect me to be.  It's their mistake, not my failing.}\\ \medskip
--- Richard Feynman, {\it Surely You're Joking, Mr. Feynman!}}
\end{center}

\cleardoublepage%*******************************************************
% Abstract
%*******************************************************
%\renewcommand{\abstractname}{Abstract}
\pdfbookmark[1]{Abstract}{Abstract}
\begingroup
\let\clearpage\relax
\let\cleardoublepage\relax
\let\cleardoublepage\relax

\chapter*{Abstract}
Attaining animal-like legged locomotion on rough outdoor terrain with sparse foothold affordances --- a primary use-case for legs vs other forms of locomotion --- is a largely open problem.  New advancements in control and perception have enabled bipeds to walk on flat and uneven indoor environments.  But tasks that require reliable contact with unstructured world surfaces, for example walking on natural rocky terrain, need new perception and control algorithms.

This thesis introduces 3D perception algorithms for contact tasks such as foot placement in rough terrain environments.  We introduce a new method to identify and model potential contact areas between the robot's foot and a surface using a set of bounded curved patches.  We present a patch parameterization model and an algorithm to fit and perceptually validate patches to 3D point samples.  Having defined the environment representation using the patch model, we introduce a way to assemble patches into a spatial map.  This map represents a sparse set of local areas potentially appropriate for contact between the robot and the surface.  The process of creating such a map includes sparse seed point sampling, neighborhood searching, as well as patch fitting and validation.  Various ways of sampling are introduced including a real time bio-inspired system for finding patches statistically similar to those that humans select while traversing rocky trails.  These sparse patch algorithms are integrated with a dense volumetric fusion of range data from a moving depth camera, maintaining a dynamic patch map of relevant contact surfaces around a robot in real time.  We integrate and test the algorithms as part of a real-time foothold perception system on a mini-biped robot, performing foot placements on rocks.

\endgroup \vfill
\cleardoublepage% Publications

\pdfbookmark[1]{Publications}{publications}
\chapter*{Publications}
This dissertation is based on the work that has been presented in the following conference/workshop papers and posters:

\bigskip

\begin{itemize}
  \item Dimitrios Kanoulas, Marsette Vona. {\it Bio-Inspired Rough Terrain Contact Patch Perception}. In the 2014 IEEE International Conference on Robotics and Automation, ICRA 2014.
  \item Dimitrios Kanoulas, Marsette Vona. {\it The Surface Patch Library (SPL)}. In the 2014 IEEE International Conference on Robotics and Automation Workshop: MATLAB/Simulink for Robotics Education and Research, ICRA 2014.
  \item Dimitrios Kanoulas. {\it Surface Patches for Rough Terrain Perception}.  In the Northeast Robotics Colloquium, Second Edition (poster), NERC 2013.
  \item Dimitrios Kanoulas, Marsette Vona. {\it Sparse Surface Modeling with Curved Patches}. In the 2014 IEEE International Conference on Robotics and Automation, ICRA 2013.
  \item Marsette Vona, Dimitrios Kanoulas. {\it Curved Surface Contact Patches with Quantified Uncertainty}. In the IEEE/RSJ International Conference on Intelligent Robots and Systems, IROS 2011.
\end{itemize}
\pagestyle{scrheadings}
\cleardoublepage%*******************************************************
% Table of Contents
%*******************************************************
%\phantomsection
\refstepcounter{dummy}
\pdfbookmark[1]{\contentsname}{tableofcontents}
\setcounter{tocdepth}{2} % <-- 2 includes up to subsections in the ToC
\setcounter{secnumdepth}{3} % <-- 3 numbers up to subsubsections
\manualmark
\markboth{\spacedlowsmallcaps{\contentsname}}{\spacedlowsmallcaps{\contentsname}}
\tableofcontents 
\automark[section]{chapter}
\renewcommand{\chaptermark}[1]{\markboth{\spacedlowsmallcaps{#1}}{\spacedlowsmallcaps{#1}}}
\renewcommand{\sectionmark}[1]{\markright{\thesection\enspace\spacedlowsmallcaps{#1}}}
%*******************************************************
% List of Figures and of the Tables
%*******************************************************
\clearpage

\begingroup 
    \let\clearpage\relax
    \let\cleardoublepage\relax
    \let\cleardoublepage\relax
    %*******************************************************
    % List of Figures
    %*******************************************************    
    %\phantomsection 
    \refstepcounter{dummy}
    %\addcontentsline{toc}{chapter}{\listfigurename}
    \pdfbookmark[1]{\listfigurename}{lof}
    \listoffigures

    \vspace*{8ex}

    %*******************************************************
    % List of Tables
    %*******************************************************
    %\phantomsection 
    \refstepcounter{dummy}
    %\addcontentsline{toc}{chapter}{\listtablename}
    \pdfbookmark[1]{\listtablename}{lot}
    %\listoftables
        
    \vspace*{8ex}
%   \newpage
    
    %*******************************************************
    % List of Listings
    %*******************************************************      
	  %\phantomsection 
    %\refstepcounter{dummy}
    %\addcontentsline{toc}{chapter}{\lstlistlistingname}
    %\pdfbookmark[1]{\lstlistlistingname}{lol}
    %\lstlistoflistings

    %\vspace*{8ex}
    
    %*******************************************************
    % List of Algorithms
    %*******************************************************      
       
    %*******************************************************
    % Acronyms
    %*******************************************************
    %\phantomsection 
    %\refstepcounter{dummy}
    %\pdfbookmark[1]{Acronyms}{acronyms}
    %\markboth{\spacedlowsmallcaps{Acronyms}}{\spacedlowsmallcaps{Acronyms}}
    %\input{FrontBackmatter/Acronyms}                    
\endgroup

\cleardoublepage

%******************************************************************************
% Mainmatter
%******************************************************************************
\pagenumbering{arabic}
%\setcounter{page}{90}
% use \cleardoublepage here to avoid problems with pdfbookmark
\cleardoublepage

%\begin{doublespacing}
\begin{onehalfspacing}
%******************************************************************************
\chapter{Introduction} \label{Ch:intro}
%******************************************************************************

In 1986, Daniel Whitney in his article ``Real Robots Don't Need Jigs'' \cite{Whitney86} highlighted the need for redesigning robots to complete tasks in very unstructured environments, under significant uncertainty.  Almost three decades later, robots have achieved high efficiency in well-structured environments like factories and labs, but still are not flexible enough to reliably deal with real-world tasks.  Interest in uncertainty goes back to the beginning of robotics \cite{Mason12}, but only over the last few years have mobile manipulators (e.g. \cite{RHDBB09, mobilemanipulation13}) and rough terrain robots (e.g. \cite{MLB07,RBNP08}) started dealing with it efficiently, both in the environment and in their own state.

The Fukushima Daiichi nuclear disaster in 2011 had a profound impact on robotics.  Despite rapid advancements in actuation and control, robots were unable to directly replace humans in hazardous tasks, like climbing in a damaged nuclear plant, searching rubble piles after a disaster \cite{KS09}, or operating in human-traversable rough terrain.  Legged locomotion in uneven 3D terrain is a key aspect for completing these and similar tasks, because of the primary advantage of legs to efficiently negotiate highly faceted 3D trails with more flexibility and mobility than other forms of locomotion such as wheels or tracks.

\begin{figure*}[b]
\begin{center}
\includegraphics[width=\textwidth]{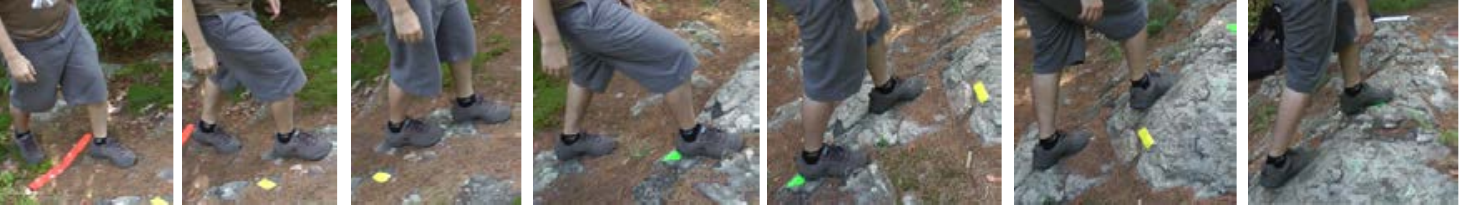}
\end{center}
\caption[Human locomotion considering a sparse set of footholds]{Humans and animals locomote reliably even under significant uncertainty about the environment and their own state, considering only a sparse set of footholds.}
\label{Fig:hiking}
\end{figure*}

Recent advancements in control and perception have enabled bipeds to walk on flat \cite{CLCKHK05} and uneven indoor terrains \cite{NCK12}.  Major advances have also been made for outdoor quadrupeds and bipeds in rough terrain where the probability of blindly landing footholds is high \cite{RBNP08} and uncertainty can be tolerated by low-level feedback control.  Online footfall selection has been considered for quadrupeds and hexapods~\cite{KS96, BW99, KKN09, PMPKRB09, KBPMS10}, but still, to the best of our knowledge, no physical humanoid has previously been shown to walk autonomously on unmodeled sparse 3D terrain.  New advances in both perception and control~\cite{KC14} are required; here we attempt to disentangle these two aspects to a degree and focus primarily on perception.  The foothold selection problem is particularly interesting for bipeds with non-point feet that make contact with patches of terrain.  Perception is one of the key enablers for finding such patches in the environment \cite{KKB08,Behnke08,Brooks09}.  This brings us to our main hypothesis (Figure~\ref{Fig:hiking}):
\begin{quote}
	\textit{Main Hypothesis: Robots operating in many unstructured environments need to perceive sparse areas for potential contact.  These can be detected and modeled using curved surface patches, and spatially mapped in real-time.}
\end{quote}

\section{Thesis Outline and Contributions}
Sparsity of footholds for bipedal robots requires i) a model formulation for the local contact surface areas, ii) an online perception algorithm for finding them, iii) techniques for handling uncertainty and reliability, and iv) a method for creating a map of the detected local contact areas around the robot and localizing within it during motion.  This thesis presents algorithms to address each of these four requirements.  We have also developed and released the Surface Patch Library (SPL)~\cite{SPL} which contains the software implementations we used to evaluate the algorithms in our experiments.

In Chapter~\ref{Ch:input} we describe the sensing system we are using for acquiring data from the environment.  This includes both a range sensor that produces a set of 3D point clouds over time and an Inertial Measurement Unit (IMU) that gives the corresponding gravity vector.  We also discuss uncertainty models for the input 3D sample points associated with the sensor, along with some types of point cloud filtering, including outlier removal and smoothing.  We also introduce a way for calibrating the IMU sensor with respect to the range sensor to which it is attached.

In Chapter~\ref{Ch:EnvRep} we describe the system for representing the environment.  We introduce a set of 10 bounded curved-surface patch types (Figure~\ref{Fig:patches} left, \cite{VK11}) suitable for modeling local contact regions both in the environment and on a robot.  We present minimal geometric parameterizations using the exponential map for spatial pose both in the usual 6DoF case and also for patches with revolute symmetry that have only 5DoF.  We then give an algorithm to fit any patch type to 3D point samples of a surface, with quantified uncertainty both in the input points (including nonuniform variance, common in data from range sensors) and in the output patch.  We also introduce an algorithm for validating the fitted patch for fit quality and fidelity to the actual data --- extrapolations (like hole-filling) which are not directly supported by data are avoided (\cite{KV13}).

In Chapter~\ref{Ch:patch_mapping} we define the notion of a volumetric working space around the robot and we describe the patch mapping system.  A dynamic map of bounded curved patches fit randomly over an environment surface that has been sampled by a range sensor is developed.  The mapping algorithm is divided into four main steps after data acquisition.  The first is a data pre-processing step, where both a bilateral filter is applied to the cloud to reduce noise and a sample decimation filter for performance purposes.  A bio-inspired saliency filter is also introduced for detecting points in a hiking-task scenario, so only relevant parts of the environment are considered for patch fitting.  Recordings of human subjects traversing rough rocky trails were analyzed to give a baseline for target surface properties for foot placement.  After filtering, the second step is the selection of seed points, where a random grid-based approach is introduced and applied to the filtered samples.  Next is a neighborhood search around these points.  Three different approaches for finding local neighborhoods were analyzed, which have different properties near surface discontinuities.  The last step is to fit the pose, curvatures, and boundary of patches to the neighborhoods and validate them to quantify fit quality and to ensure that the patch is sufficiently representative of the actual data.  We finally highlight the construction of a spatial map of the fitted patches around a robot.

In Chapter~\ref{Ch:patch_tracking} we present the patch tracking method that completes the whole Patch Mapping and Tracking system.  For tracking the camera pose at each frame an adapted version of the Moving Volume KinectFusion~\cite{NIHMKDKSHF11, RV12} algorithm is applied.  It is the first time that this camera tracking method is used for a bipedal locomotion application on physical hardware (Kinect Fusion without the moving volume algorithm is used in~\cite{RGMSHS13}, though in simulation only).  We improve the original algorithm for our particular application both by using the gravity vector from the IMU to keep the local map in a pose aligned to gravity, and also by using a virtual camera, which lies above the robot looking down in the direction of gravity, for acquiring a point cloud from a synthetic birds-eye viewpoint during walking.  In contrast to the original real camera raycasting method that considers upcoming surfaces only, the advantage of our virtual camera version is that the raycasting considers the environment around and under the robot's feet, even portions that were previously visible but currently occluded by the robot itself.

\begin{figure*}[!h]
  \begin{center}
    \includegraphics[width=\textwidth]{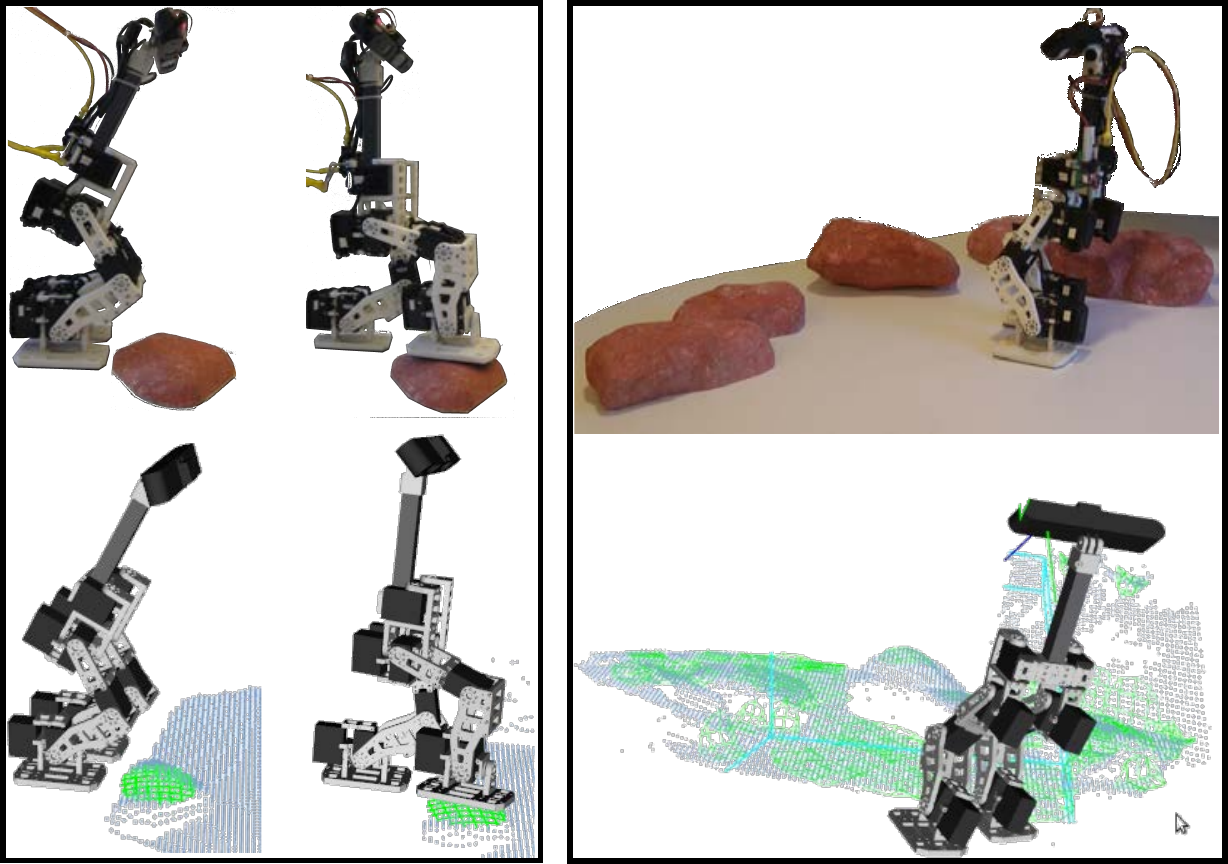}
  \end{center}
\caption[Patch mapping and tracking on the RPBP biped robot]{\textbf{Left}: The RPBP mini-biped robot detecting a patch on a rock and placing its foot on it (\textbf{upper}: the RPBP robot; \textbf{lower}: the detected patch in the point cloud).  \textbf{Right}: The patch map system integrated on the RPBP robot (\textbf{upper}: the RPBP robot walking on a table with four rocks; \textbf{lower}: patches mapped and tracked in the environment using the moving volume Kinect Fusion system).}
\label{Fig:intro_rpbp}
\end{figure*}

In Chapter~\ref{Ch:biped_exp} we test the patch mapping and tracking system on a mini-biped robot platform developed in our lab called RPBP (Rapid Prototyped Biped).  We ran two experiments.  In the first one (Figure~\ref{Fig:intro_rpbp}, right) we test the system integrated on the robot independently of its control, making sure that shaking and vibration while the robot is walking do not hinder the tracking process.  In the second one (Figure~\ref{Fig:intro_rpbp}, left) we first train the robot to place its foot on patches that were manually fitted on four different rocks.  Then we let the robot, starting from a fixed position, detect patches in the environment and if any of them matches one of the trained patches it executes the corresponding foot placement motion.  These experiments conclude the thesis, whose main contributions are as follows.

\subsection*{Contributions}
\begin{enumerate}
  \item A new sparse environment surface representation using a set of bounded curved patches suitable for modeling local contact regions both in the environment and on the robot.
  \item A fast algorithm to fit these patches to 3D point samples of a surface, with quantified uncertainty both in the input points and in the output patch.
  \item Fast residual, coverage, and curvature patch validation tests for evaluating the fidelity of fitted patches.
  \item Bio-inspired rules for finding patches statistically similar to those selected by humans for hiking in rough terrain.
  \item Real-time mapping of hundreds of patches near a walking biped in combination with dense volumetric depth map
fusion and inertial sensing.
\end{enumerate}

%******************************************************************************
\section{Related Work} \label{Sec:Intro_Related_Work}
%******************************************************************************
Visual odometry has been used on several current walking robots including BigDog \cite{Howard08, WMBHRR10} and the DLR Crawler \cite{SHG12}, though mainly for obstacle avoidance and traversability analysis, not detailed 3D foot placement or contact planning.  Some steps have been made in that direction in~\cite{LPB07}, where terrain is modeled using a Gaussian Process, but this was not applied for legged locomotion.

On-line perception for foot placement has been recently implemented for quadrupeds and hexapods.  In \cite{PMPKRB08, PMPKRB09} a continuous surface model is used for LittleDog locomotion, whereas in \cite{BPP10} a local decision surface was used on a hexapod walking robot.  In \cite{KBPS09,KBPMS10} a system learns optimal foothold choices from expert demonstration using terrain templates.  Recently in \cite{BS12} a PTAM approach was used for updating the elevation map during locomotion.

In some cases foot placement has been done without perception by using a known 3D terrain map and on-line motion capture (e.g.~\cite{DBSKRTR07, KBPMS10}).  It is also common here to use perception for obstacle avoidance, terrain categorization, or gait selection without specific 3D foot placement~\cite{WMBHRR10,SHG12}.  Quadrupedal and hexapedal studies are related to the bipedal case but often use a point-like contact model, whereas many bipeds have extended feet to support torques for balance and may need to consider foot-sized terrain patches.

To date only a few works have used on-line perception for bipedal foot placement in uneven or rough terrain.  In~\cite{OII03, GFF04, GFF08} planar segments were fitted to point cloud data for indoor scenes with slopes and steps, and in~\cite{NCK12} a laser scanning system is used in a similar context.  In~\cite{RGMSHS13} KinectFusion~\cite{NIHMKDKSHF11} was used in simulation to avoid harsh impacts.  A number of other works (e.g.~\cite{MHB12, HMB13, MLB13}) introduced perception for obstacle detection and navigation in cluttered surfaces, where the foot placement involves stepping over or climbing up/down flat obstacles.  Recently \cite{BVKEK13} presented preliminary results in multi-contact planning for a full-size humanoid using 3D perception for extracting planar contact surfaces for navigation.

This thesis introduces a novel way to detect curved contact patches in the environment, whereas most prior work has focused on flat surfaces.  We integrate this perception system with foot placement on rocks for a physical free-standing biped robot.  Though other rough-terrain walking robots have been developed, there is little prior work in realtime on-board 3D perception for biped foot placement.  Finally, our approach to map and track the patches as the robot locomotes is based on a novel combination of our sparse patch map with a dense point cloud from newly available real-time depth map fusion algorithms.
\cleardoublepage

\part{Sparse Surface Modeling with Curved Patches}
%******************************************************************************
\chapter{Input Data} \label{Ch:input}
%******************************************************************************

Both perceiving the environment around a robot ({\em exteroceptive perception}) and sensing the robot's own internal state ({\em proprioceptive perception}) are important aspects for driving planning and control actions in a real world scenario.  Various perception sensors can be used for acquiring these important measurements (see \cite{Everett95, SNS11}).  In this thesis we use both exteroceptive range sensing for detecting upcoming 3D terrain contacts from a distance and proprioceptive inertial measurement unit (IMU) sensing for acquiring the robot's orientation relative to gravity.  In the next two sections we summarize the range and IMU sensors that provide the main inputs to our system.

\section{Range Sensing} \label{Sec:range_sensing}
3D perception has gained a lot of interest over the last few years \cite{RC11}, mainly because low cost but high quality range sensors are now commonly available.  Stereo and structured light systems, time-of-flight cameras, and laser scanners produce clouds of 3D sample points of environment surfaces in real time.  Here we focus on {\em organized} point cloud data in the form of an image grid acquired from a single point of view.  Initially we take such images directly from a depth camera.  Then in Chapter~\ref{Ch:patch_mapping} a considerable level of indirection is added: the actual depth sensor images are fused (over space and time) into a volumetric model, from which a simulated sensor extracts virtual depth images for patch mapping.

\begin{figure}[ht]
\begin{center}
\includegraphics{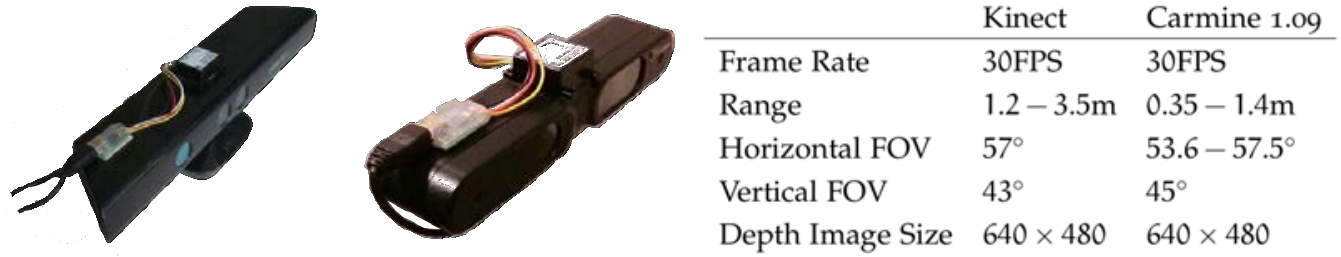}
\end{center}
\caption[Our sensing apparatus: an RGB-D camera with an affixed IMU]{Our sensing apparatus is either a Microsoft Kinect (left) or a PrimeSense Carmine 1.09 (right) RGB-D camera with a CH Robotics UM6 9-DoF IMU attached.}
\label{Fig:range_sensors}
\end{figure}

%\begin{table}[h]
%\begin{center}
%\centering
%\begin{tabular}{lllll}
% 					& Kinect  		& Carmine 1.09 \\ \hline
%Frame Rate			& $30$FPS		& $30$FPS \\ 
%Range 				& $1.2-3.5$m	& $0.35-1.4$m \\
%Horizontal FOV 		& $57^{\circ}$	& $53.6-57.5^{\circ}$\\
%Vertical FOV 		& $43^{\circ}$	& $45^{\circ}$ \\
%Depth Image Size	& $640\times480$& $640\times480$
%\end{tabular}
%\end{center}
%\end{table}

In this thesis either the Microsoft Kinect or the Primesense Carmine 1.09 (see Figure~\ref{Fig:range_sensors}) have been used  for acquiring 3D point clouds, depending on different experimental requirements (mainly involving range limits when the sensor is hand-held or on the mini-biped).  Both the Kinect and the Carmine sensor consists of three parts: 1) an infrared (IR) projector, 2) an infrared (IR) camera, and 3) an RGB camera.  For estimating the 3D point cloud a triangulation method is applied using the IR emitter and detector that are separated by a baseline.  As described in \cite{BK08, Khoshelham11, SJP11, KE12} in detail, given an image pixel with coordinates $(u,v)$ and disparity $d$ from triangulation, the corresponding 3D point $(x,y,z)$ expressed in the camera frame is:
\beg
z = \frac{f_x b}{d}\\
x = \frac{z}{f_x}(u-c_x)\\
y = \frac{z}{f_y}(v-c_y)
\eeg
using:

\begin{tabular}{ll}
$(u,v,d)$		& image pixel coordinates and disparity of the point (in pixels)\\
$(x,y,z)$		& 3D point coordinates in camera frame (in physical units, e.g. m)\\
$b$				& the baseline between IR camera and projector (in physical units)\\
$f_x, f_y$		& IR camera focal length (in pixels) \\
$(c_x, c_y)$  	& the principal point (in pixels)
\end{tabular}\\

The origin of camera frame is the center of projection, the $z$ axis points into the scene through the principal point $(c_x, c_y)$, the $x$ axis points to the right in the camera image, and the $y$ axis points down in the image.  The 3D sample point coordinates $(x,y,z)$ in camera frame can be also expressed as a function of the coordinates of the measurement ray direction vector $\vec{m} = (m_x,m_y,m_z)$ through pixel $(u,v)$ and the range $r$ of the data point along that vector as:
\begin{eqnarray}
	[x\ y\ z] = [m_x\ m_y\ m_z]\ r \label{Eq:mr2xyz}
\end{eqnarray}

From the above equations, the backprojected 2D $(u,v)$ pixel corresponding to an $(x,y,z)$ 3D point can be calculated as:
\begin{eqnarray}
u = \frac{x f_x}{z} + c_x \\
v = \frac{y f_y}{z} +c_y
\end{eqnarray}

\begin{figure}[h,t]
\begin{center}
\includegraphics[width=\textwidth]{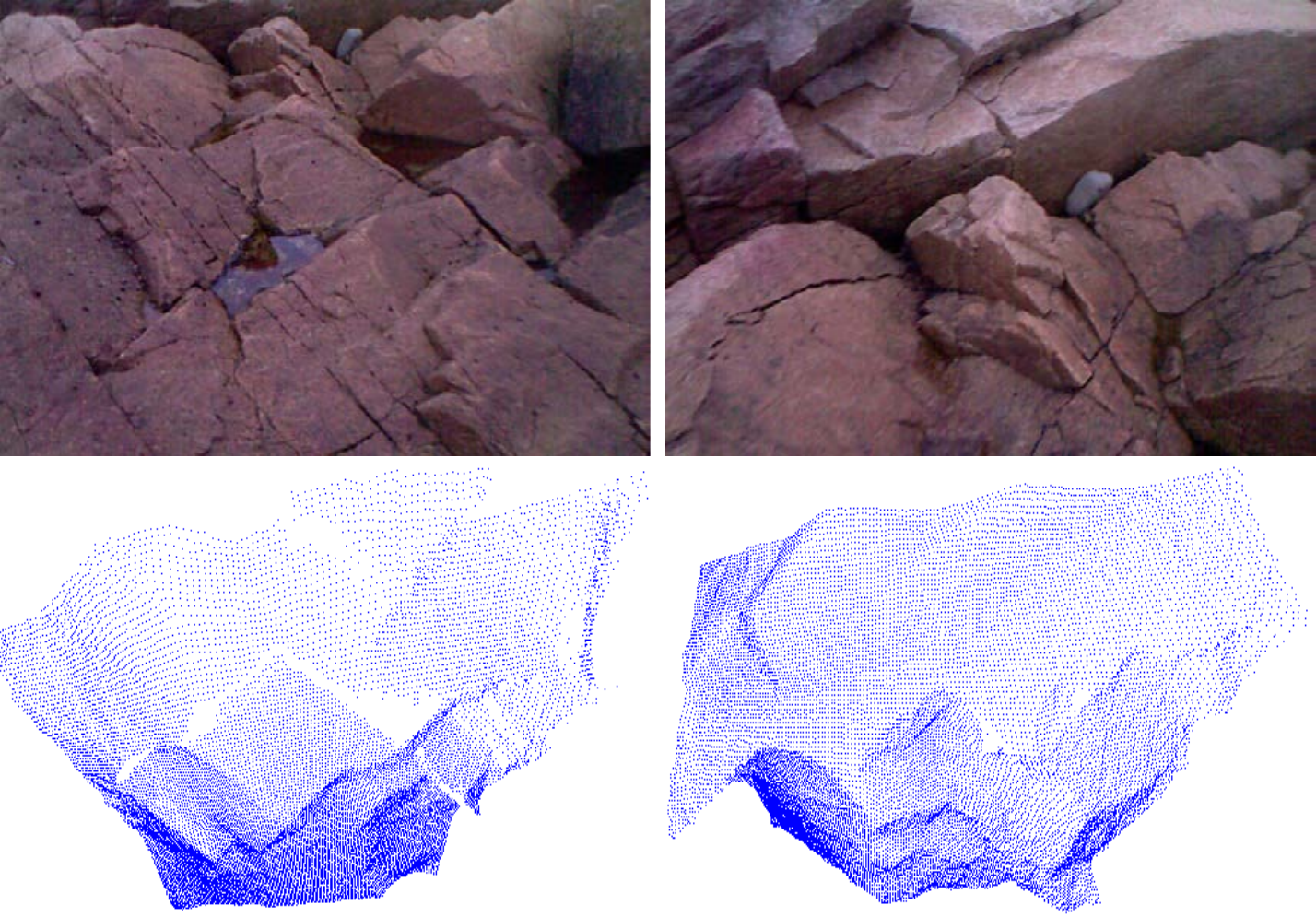}
\end{center}
\caption[A dense point cloud input from an RGB-D camera]{A $640\times480$ dense point cloud input from a Microsoft Kinect RGB-D camera.}
\label{Fig:rgbd_rocks_input}
\end{figure}

Using either of these two range sensors we receive 30Hz $640\times480$ RGB-D (red, green, blue, depth) data \footnote{Spatial registration of the depth and RGB data used built-in calibration in the range sensor (RGB data is only used for visualization purposes in this thesis).}. The structured light method used by the Kinect and Carmine does not work well in full sunlight, so when outdoor data were needed they were taken at twilight.  Sunlight operation could be possible with other types of depth camera or stereo vision.  Two point cloud examples of rocks along with their RGB images appear in Figure~\ref{Fig:rgbd_rocks_input}.

\subsection{3D Point Cloud Uncertainty}
A big challenge with range sensors is to quantify the uncertainty of the acquired data.  The uncertainty could either be due to inaccuracies in the sensor system or due to triangulation errors (i.e. the correspondence problem \cite{SS02}) and it can be twofold; the data may include outlier points and noisy inliers.

\subsubsection*{Outlier Points}
An outlier point is distant from the others and does not represent the underlying surface from where it was sampled.  Ideally such points would be removed from the data.  In many cases outliers appear along depth discontinuities due to occlusions, jumps in surfaces, or reflections.  Veil points \cite{MDHWF08, SBKK11}, which are interpolated across a depth discontinuity, usually appear in data acquired by lidars (which we do not use in the thesis).

There are various methods in the literature for detecting outliers.  One simple approach is to consider as inliers only the points that have a minimum number of neighbors in a fixed distance.  A heuristic has been introduced  in \cite{SBKK11} for finding points that belong to borders both in foreground and background and removing those in between as veil points.  Other methods, for instance the one introduced in \cite{RBMBD08-1}, use statistical analysis for removing neighborhood points that are more than a fixed number of standard deviations away from the median.  Similarly in \cite{YTRS11} another statistical method is proposed to identify and remove outliers by checking for big residuals during plane fitting.  When dealing with static environments either data fusion over time~\cite{NIHMKDKSHF11, RV12}, or outlier removal using octree raycasting as proposed in \cite{BRBB09} can also be used.

In this thesis we address outliers both in a preprocessing step where a real-time discontinuity-preserving bilateral filter removes some outliers from the data (Section~\ref{Sec:map_preprocess}), and also when Kinect Fusion is used (Chapter~\ref{Ch:patch_tracking}) for tracking and inherently ignores some outliers when data fusion over time is applied.

\subsubsection*{Noisy Inlier Points} \label{Sec:noisy_inliers}
\begin{figure*}[h]
  \begin{center}
    \includegraphics[width=0.8\textwidth]{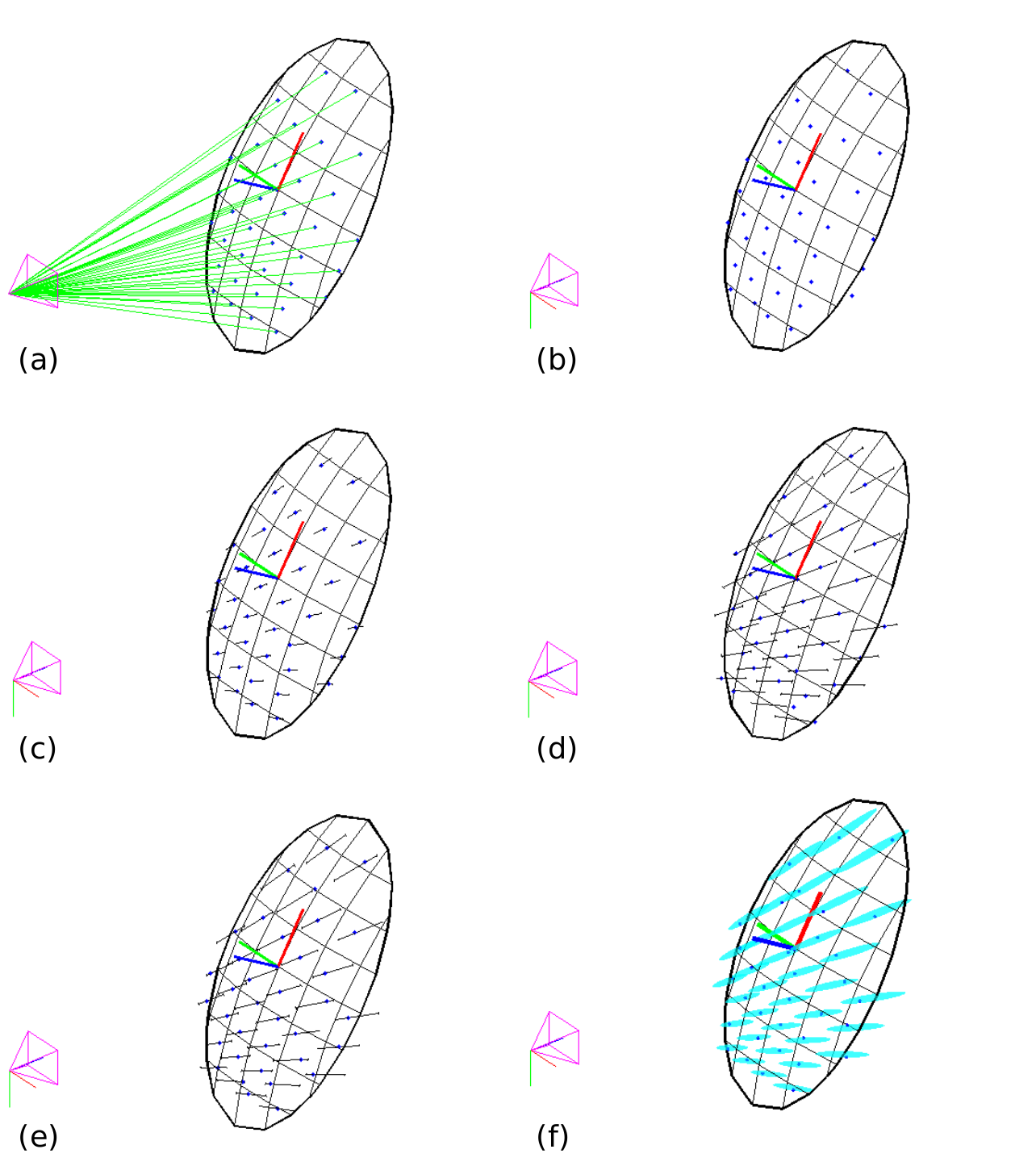}
  \end{center}
\caption[Different types of error modeling for a 3D point cloud]{Different types of error modeling for a 3D point cloud. (a) Stereo range sensing sampling from a simulated surface (black paraboloid) along measurement rays (green) from a depth camera whose field of view is defined from the viewing frustum (pink); (b) The sampled 3D point data (blue dots) deviated from their original position by adding white Gaussian noise (using the stereo error model in (f)); (c) constant error modeling; (d) linear error modeling; (e) quadric error modeling; (f) stereo error modeling, visualizing the 95\% probability error ellipsoids (pointing error exaggerated for illustration)}
\label{Fig:error_model}
\end{figure*}

A noisy inlier point deviates from the ground truth that represents the underlying surface.  To express the data noise we use Gaussian modeling with $3\times3$ covariance matrices.  Though this is not the only way to represent uncertainty, it does cover common situations\footnote{The covariance matrices may also enable data fusion based on the Kalman filter, but in this thesis we do not explore that further.}.  There are various ways to estimate these covariance matrices, depending on the error model assumptions.  Some assumptions can be the following (Figure~\ref{Fig:error_model}):

\begin{itemize}
	\item \textbf{Constant Error (Figure~\ref{Fig:error_model}-c):} with constant nonnegative uncertainty $k$ in range, independent of the sample range, and no uncertainty in pointing direction, the covariance matrix for a sample with measurement vector $\vec{m}$ is:
	\begin{eqnarray}
		\Sigma = k \vec{m} \vec{m}^T
	\end{eqnarray}

	\item \textbf{Linear Error (Figure~\ref{Fig:error_model}-d):} with a nonnegative factor $k$ that scales uncertainty linearly with the sample range, and no uncertainty in pointing direction, the covariance matrix for a sample with range $r$ and measurement vector $\vec{m}$ is:
	\begin{eqnarray}
		\Sigma = k r \vec{m} \vec{m}^T
	\end{eqnarray}

	\item \textbf{Quadratic Error (Figure~\ref{Fig:error_model}-e):} with a nonnegative factor $k$ that scales uncertainty quadratically with the sample range, and no uncertainty in pointing direction, the covariance matrix for a sample with range $r$ and measurement vector $\vec{m}$ is:
	\begin{eqnarray}
		\Sigma = k r^2 \vec{m} \vec{m}^T
	\end{eqnarray}

	\item \textbf{Stereo Error (Figure~\ref{Fig:error_model}-f):} in Murray and Little's \cite{ML05} two-parameter error model for stereo disparity uncertainty is represented by two nonnegative parameters $\sigma_p$ and $\sigma_{\vec{m}}$:
		\begin{itemize}
			\item $\sigma_p$ is the variance of the pointing error of the measurement vectors, represented as the variance in pixels of their intersections with the image plane at $z=f_x$.
			\item $\sigma_{\vec{m}}$ is the variance in the disparity matching error, also measured in pixels.
		\end{itemize}
	The covariance matrix for a 3D point in physical units is:
	\begin{eqnarray}
		\Sigma = J E J^T
	\end{eqnarray}
	where:
	\begin{eqnarray}
		E =  \begin{bmatrix}
	  		 \sigma_p & 0 & 0 \\
  			 0 & \sigma_p & 0 \\
  			 0 & 0 & \sigma_{\vec{m}}
 		     \end{bmatrix} and \
 		J =  \begin{bmatrix}
	  		 \frac{b}{d} & 0 & -\frac{b u}{d^2} \\
  			 0 & \frac{b}{d} & -\frac{b v}{d^2} \\
  			 0 & 0 & -\frac{f_x b}{d^2}
 		     \end{bmatrix}
	\end{eqnarray}
	$b$ is the baseline (in physical units), $d$ the disparity (in pixels), $(u,v)$ the image pixel coordinates, and $f_x$ the IR camera focal length (in pixels).
\end{itemize}

Whether based on stereo or time-of-flight, range data exhibits heteroskedasticity (non-uniform variance) --- typically there is much more uncertainty in range than aim~\cite{PVB09,ML05}, the variance changes with range, and because the measurement rays usually have a single center of projection, the error ellipsoids for the sampled points are not co-oriented: each is elongated in the direction of its own measurement ray (Fig~\ref{Fig:error_model}).

Thus, to estimate range data covariances we apply the two-parameter pointing/disparity stereo error model proposed by Murray and Little in~\cite{ML05} (based on earlier work by others such as~\cite{MS87}) to estimate input sample covariances $\Sigma_i$.  The error model parameters we used for the Kinect are $\sigma_p = 0.35$px, $\sigma_{\vec{m}} = 0.17$px; the former is from~\cite{KM10}, the latter was determined experimentally following~\cite{ML05}\footnote{For the Bumblebee2 camera $\sigma_p= 0.05px$ and $\sigma_{\vec{m}} = 0.1px$ (from the specifications document)}.

We use this error model when fitting patches to points sampled from a single sensor viewpoint (i.e. a single captured range image).  In Chapter~\ref{Ch:patch_tracking} we apply KinectFusion to the range data, which provides an alternate approach to handling inlier noise by averaging over many re-samplings of the same environment surfaces.  In some cases we also use either discontinuity-preserving bilateral \cite{TM98} or median filters \cite{HYT79} to reduce noise effects:
\begin{itemize}
	\item \textit{Median Filter} The median filter replaces the central pixel of a fixed size window in the image with the median inside the window.  The method can be very efficient \cite{PH07} and effective for reducing noise and removing outliers from the data, while preserving discontinuities.

	\item \textit{Bilateral Filter} The bilateral filter is similar to the median filter with the difference that central pixel's neighbors are weighted making the filter non-linear~\cite{PF06}.
\end{itemize}

\subsection{3D Point Cloud Filtering}
There are various other filtering methods for the acquired point clouds serving different purposes~\cite{RC11}.  Some used in this thesis are the following:
\begin{itemize}
	\item \textit{Passthrough:} The passthrough filter removes points whose specified properties (e.g. x,y,z-coordinates, intensity, etc) are outside of some limits. 
	
	\item \textit{Radius Outlier Removal:} Removes outliers by checking the number of points in a predefined radius neighborhood.  
	
	\item \textit{Decimation:} Decimates the points by a given factor, discarding rows and columns of pixels in the image, e.g. a factor of 2 will discard all the even rows and columns.
	
	\item \textit{Lower Resolution:} Lowers the resolution by a given factor by block averaging, e.g. a factor of 2 will replace each 2-by-2 submatrix with its average value.  It is similar to the median filter, but the latter can be more robust to outliers.

	\item \textit{Voxel Grid:} The approximate voxel grid filter downsamples the cloud by creating a 3D voxel grid and replacing all the points in each voxel with their centroid.  This method leaves the point cloud unorganized.  Some fast approximations have been introduced \cite{RC11} to improve the efficiency of this filter.
\end{itemize}

\section{Inertial Measurement Unit (IMU)}
The use of proprioceptive Inertial Measurement Units (IMUs) for sensing the direction of gravity is very useful for locomotion.  Using a CH Robotics UM6 9-DoF IMU mounted on the top of our range sensors (Figure~\ref{Fig:range_sensors}), we receive 100Hz IMU data spatiotemporally coregistered with the 30Hz RGB-D data received from the depth sensor.  Though an IMU can also sense velocities and accelerations, in this work we use only the gravity direction as input to our algorithms.  In this thesis temporal registration of the RGB, depth, and IMU datastreams is based on timestamps, and is approximate because the underlying operating systems used were not hard real-time.  Spatial registration of the RGB and depth data is based on manufacturer hardware calibration and image warping implemented in the hardware driver.  Spatial registration of the depth and IMU data uses a custom calibration algorithm described next.

\subsection{IMU Calibration} \label{Sec:imu_cal}
\begin{figure*}[!htb]
  \begin{center}
    \includegraphics[width=0.7\textwidth]{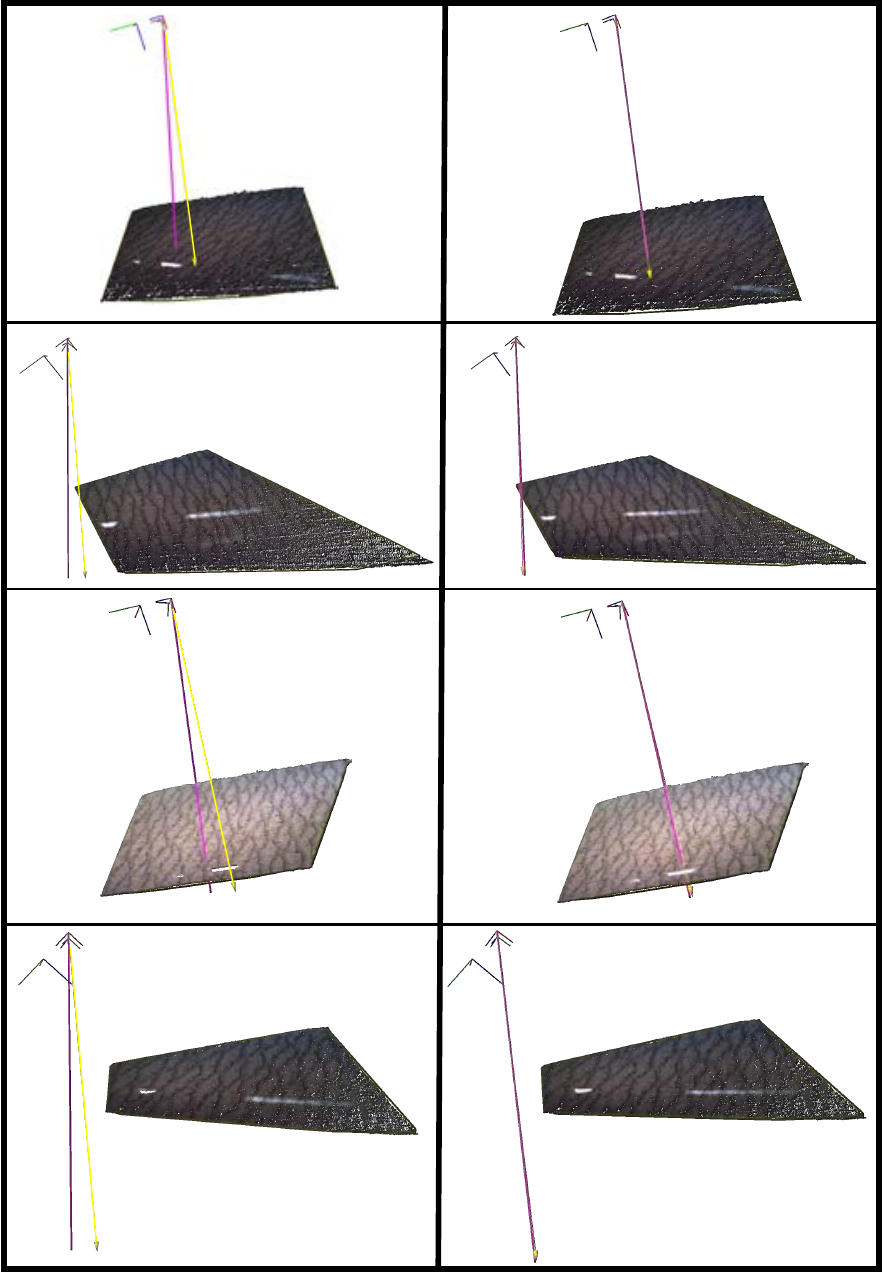}
  \end{center}
\caption[IMU calibration instances]{IMU calibration instances for 4 different frames, where the gravity vector (yellow) and the ground plane normal (magenta) appear before (left) and after (right) calibration.  Before calibration the two vectors have some angle difference between them, but after calibration they are nearly on top of each other.}
\label{Fig:imu_calibration}
\end{figure*}

Calibration is required for calculating the rotation transform that gives the orientation of the UM6 relative to the range sensor.  Given a dataset of depth images of a flat horizontal surface that includes a dominant plane (e.g. a flat floor) and the corresponding UM6 orientation data, the gravity vector is calculated for each depth image in the UM6 coordinate frame from the UM6 orientation data.  We pair each gravity vector with the corresponding one in the depth camera coordinate frame, which is estimated as the downward facing normal of the dominant plane.  For all these pairs of gravity vectors we solve the orthogonal Procrustes problem \cite{ELF97} that gives the UM6 to Camera transform (Figure~\ref{Fig:imu_calibration}).

\section{Related Work}
\subsection*{Perception and Sensing}
Much research on locomotion focuses on control or path planning and assumes known terrain models or uses motion capture systems to extract information about the robot and its position with respect to the terrain (e.g. \cite{DBSKRTR07}).  Other systems, such as \cite{CPP99}, use only proprioceptive sensors for driving locomotion actions.  For an autonomous real world task, where there is no prior information about the environment, driving actions from high-level---but quantitative---perception using exteroceptive sensors is essential.  Tactile sensors are helpful only when actual contact is taking place.  For \emph{a priori} information about the surface, range sensing is required.  There are systems that only use color cameras \cite{FB12} and others that use laser scanners \cite{NCK12}, stereo \cite{SHG12}, or time-of-flight \cite{RKD10} cameras to extract depth data.  Several other walking robots have used depth or RGB-D sensors, as we do, including stereo vision on QRIO~\cite{GFF08}, Xtion on NAO~\cite{MHB12}, and depth camera on HRP-2~\cite{NCK12, RGMSHS13, BVKEK13}.  Since sensors are noisy and the uncertainty of the measurements is high, perception using range sensing and IMU is a very challenging task, but it is rapidly advancing \cite{RC11}.

\subsection*{Uncertainty Representation}
The importance of representing 3D range data uncertainty has been considered at least since the 80's~\cite{MS87}, where 2D \cite{Gennery80} and 3D \cite{Hallam83, BC86} normal distributions were used, as well as additive zero mean Gaussian noise modeling \cite{FAF86} for 3D stereo measurements.  In \cite{MS87} non-Gaussian noise was considered for errors in the non-linear triangulation operation, which are approximated with 3D Gaussian distributions, while later in \cite{JK97} a cylindrical Gaussian distribution centered at the 3D point and oriented along the measurement ray was used for modeling the stereo error uncertainty.  In \cite{DWJ97} uncertainty was modeled in the depth measurement using ellipses \cite{HLP93}.  Tasdizen and Whitaker \cite{TW03} assumed a Gaussian distribution for representing the depth noise with zero angular error.  Gaussian modeling is not the only way to represent 3D point cloud uncertainty.  Pauly, Mitra, and Guibas \cite{PMG04} considered the point cloud as a result of a stochastic process corrupted by zero-mean additive noise to come up with a likelihood and a confidence map for the data.  Closed form variance formulations \cite{BBL09} and non-Gaussian distributions \cite{PCM11} are also alternative ways to represent the uncertainty of the range data.  Recently an uncertainty model for the Kinect sensor has been introduced \cite{Khoshelham11, SJP11, KE12}, while a mixture of Gaussians has been used in \cite{DVX13}.

\section{Summary and Future Work}
In this chapter we introduced the input data acquisition process for both exteroceptive (range sensor) and proprioceptive (IMU sensor) data.  We discussed error modeling for the range data and different types of filtering for 3D point clouds.  We also described a method for calibrating the IMU sensor with respect to the RGB-D camera on which it is mounted.

Data fusion is a key aspect in robotics and has been studied exhaustively.  An interesting direction for bipedal locomotion perception is to fuse exteroceptive (range sensing) and proprioceptive (kinematics and tactile sensing) data for detecting contact areas in the environment.  Exteroception can detect upcoming terrain contact areas from a distance, but with relatively high uncertainty.  Kinematic proprioception senses the pose of contact areas on the robot itself---e.g. heel, toe, foot sole---potentially with relatively low uncertainty.  Once a contact is established, the environment contact area can be re-measured \emph{exproprioceptively} through kinematics and touch, possibly with reduced uncertainty compared to prior exteroception.  Finally, the uncertainty representation plays an important role in 3D perception.  Various ways of modeling uncertainty have been introduced, but there is not yet a generally accepted model and further investigation is needed.

\junk{
Range sensors are now commonly available. Stereo and structured light sensors like the Kinect, time-of-flight cameras, and laser scanners produce clouds of 3D sample points of environment surfaces.  Here we focus on point cloud data in the form of an image acquired from a single point of view, with $N$ 3D sample points organized in an $N_r \times N_c$ grid, $N_r N_c = N$.  Due to occlusions, limited sensor range, and other factors, there may be grid cells map where depth information is invalid.  (Grid-organized data can also be synthesized from other representations, including from the volumetric fusion mentioned above, by raycasting.)
}
%******************************************************************************
\chapter{Environment Representation} \label{Ch:EnvRep}
%******************************************************************************

\begin{figure*}[htb]
  \begin{center}
    \includegraphics[width=\textwidth]{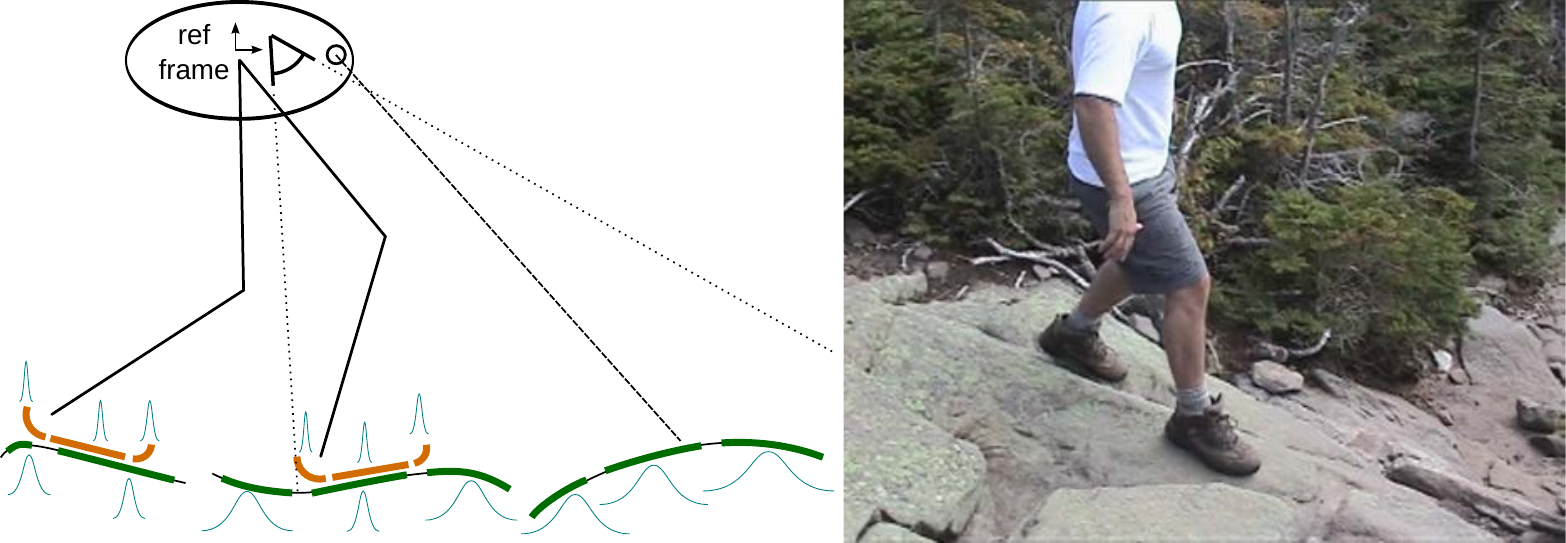}
  \end{center}
  \caption[Biped approximates contact areas with bounded curved patches]{\textbf{Left:} a biped considering a set of bounded curved patches that locally approximate both the environment surfaces (green) and the key contact surfaces on the robot (brown), all with quantified uncertainty (blue Gaussians). \textbf{Right:} robots will be required to perform tasks similar to those of humans hiking down rocky trails.}
  \label{Fig:envrep}
\end{figure*}

Some of the most challenging open problems in robotics are those which require reliable contact with unstructured world surfaces when locomoting (Figure~\ref{Fig:envrep} right).  To enable rough-terrain walking and climbing, a perception system that can spatially model and finely quantify potential 3D contact patches may be needed.  Contact is well-studied (e.g.~\cite{MS85}) but, arguably, there is not yet any accepted general system for modeling the shape and pose of potential contact surface patches, including both patches on the robot (e.g. finger tips, foot soles, etc) and also in the surrounding environment.  This is especially true when (a) curved, bounded patches with (b) geometrically meaningful minimal parameterizations and (c) quantified uncertainty are desired (Figure~\ref{Fig:envrep} left).

Why \textbf{curved patches}?  Our interest is legged locomotion on large rocks.  Flat areas can be rare in such natural environments.  More broadly, contact surfaces in man-made environments are also often curved---railings, doorknobs, steering wheels, knobs, etc.  Though curved surfaces can be approximated by sets of smaller planar patches~\cite{VBPS10}, the job can often be done with fewer and larger curved patches.  Curved surface geometry is more complex, but it may still be an advantageous trade-off to reason about fewer and larger patches.  For example, a spherical robot foot stepping into a divot on a rock might be modeled as the interaction between just one spherical and one elliptic paraboloid patch (on foot and rock, respectively).  If the surfaces were approximated using collections of smaller planar patches the interaction could require combinatorial reasoning about many possible contacting pairs.

By \textbf{``geometrically meaningful minimal parameterizations''} we mean that each patch is defined by the fewest possible parameters, and that these have direct geometric interpretations---rotations, translations, curvatures, lengths, and angles.  Geometric (vs. algebraic) parameterizations also support reasoning~\cite{DNC07} about possible actions with patches, and allow some representation of spatial uncertainty with geometric error ellipsoids.  Minimality is desirable because redundant (non-minimal) parameterizations can slow the numerical optimizations used in surface fitting~\cite{Gra98} and must be handled specially in uncertainty modeling~\cite{PVB09}.  

It is often important to get both a best estimate of patch parameters and a \textbf{quantification of the uncertainty} therein.  We develop full uncertainty quantifications based on Gaussian modeling with covariance matrices as were described in Chapter~\ref{Ch:input}, by propagating the input 3D point cloud uncertainty~\cite{Meyer92, TBF05} to the output patch.  Though in this thesis we use dense volumetric depth map fusion for mapping (Chapter~\ref{Ch:patch_mapping}), we also intend our models to be usable in sparse Kalman-type SLAM (Simultaneous Localization and Mapping \cite{SC86,SSC90,DB06}) algorithms that maintain a dynamic local patch map, Figure~\ref{Fig:envrep}, of contact patch features around a robot.  Such a map could potentially include both environment surfaces and contact pads on the robot itself, which may themselves be potentially uncertain due to kinematic error.

We first give the details of the patch models for representing the environment in local contact areas, followed by an algorithm to fit and validate a patch to noisy point cloud data from common types of range sensor.  This fitting is the main step in using patches to represent surface shape and pose.  We also demonstrate the algorithms in experiments with simulated and real range data.  More experiments are presented in Chapters~\ref{Ch:patch_mapping} and~\ref{Ch:biped_exp} in practical contexts including humans walking on rocky natural terrain and a biped robot walking near and stepping on rocks.

\section{Patch Modeling} \label{Sec:patch_modeling}

\begin{figure*}[htb]
  \begin{center}
    \includegraphics[width=\textwidth]{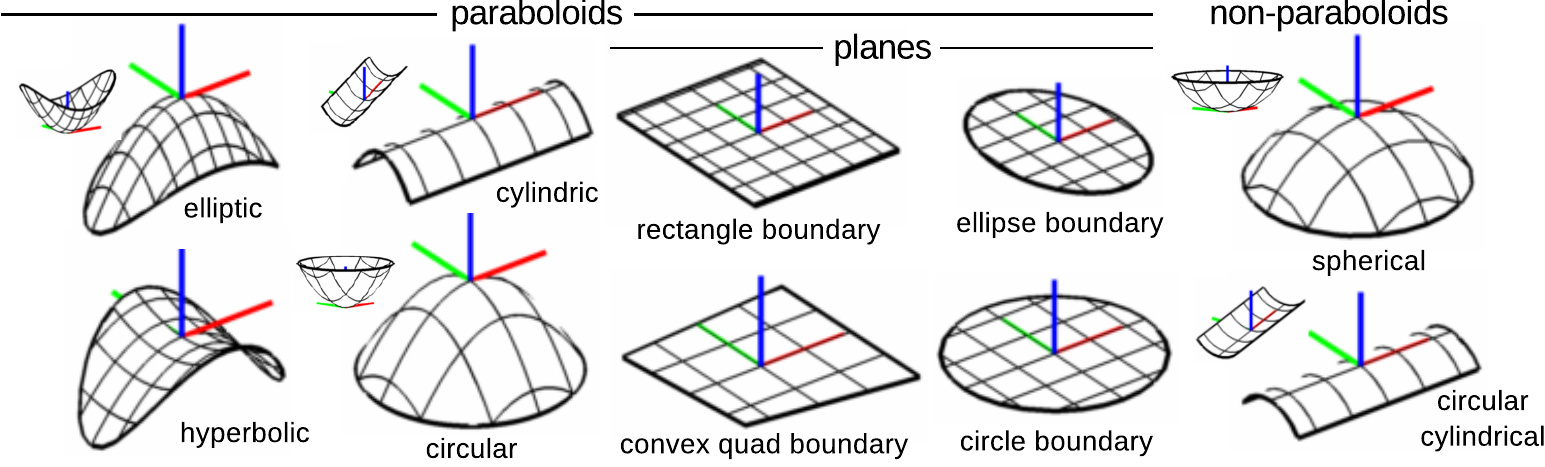}
  \end{center}
  \caption[Examples of all patch types]{Examples of all patch types, each with axes of the local coordinate frame.  Concave variants shown inset.}
  \label{Fig:patches}
\end{figure*}

In \cite{VK11}, we introduced a general-purpose set of ten curved and flat patch types (Figure~\ref{Fig:patches}, Table~\ref{Tb:patches}) suitable for both natural and man-made surfaces and balancing expressiveness with compactness of representation.  Eight come from the general second-order polynomial approximation to a smooth surface at a given point---the principal quadric---which is always a paraboloid, possibly degenerated to a plane~\cite{Petitjean02}.  We add two non-paraboloid types to better model common man-made spherical and cylindrical surfaces, and we pair each surface type with a specific boundary curve to capture useful symmetries and asymmetries.  Each patch is parametrized using \emph{extrinsic} and \emph{intrinsic} parameters (see parametric surfaces in~\cite{Mortenson06}) for its shape and spatial pose.

\begin{table}
\begin{center}
{\setlength{\tabcolsep}{2pt}
\begin{tabular}{|l|l|r|r|c|l|}\hline
{\bf surface}         &{\bf bound}&\multicolumn{2}{|l|}{\bf parameters}   & {\bf DoF} & {\bf world frame equations}\\\cline{3-4}
                      &       &{\bf intrin.}       & {\bf extrin.}        &    & \\\hline\hline
elliptic paraboloid             &ellipse&$\vec{d}_e,\vec{k}$ &$\vec{r},\vec{t}$     & 10 & (\ref{Eq:pwi},\ref{Eq:pwe},\ref{Eq:ellipse}); \ \ \ \ \,$\sgn(\kappa_x)\!\!=\!\!\sgn(\kappa_y)$\\\hline
hyperbolic paraboloid             &ellipse&$\vec{d}_e,\vec{k}$ &$\vec{r},\vec{t}$     & 10 & (\ref{Eq:pwi},\ref{Eq:pwe},\ref{Eq:ellipse}); \ \ \ \ \,$\sgn(\kappa_x)\!\!\neq\!\!\sgn(\kappa_y)$\\\hline
cylindric paraboloid             &aa rect&$\vec{d}_r,\kappa$  &$\vec{r},\vec{t}$     & 9 & (\ref{Eq:pwi},\ref{Eq:pwe},\ref{eq-rect},\ref{eq-quad}); \,$\vec{k}\!\!=\!\![0\ \kappa]^T$\\\hline
circular paraboloid            &circle &$d_c,\kappa$    &$\vec{r}_{xy},\vec{t}$    & 7 & (\ref{Eq:pwi},\ref{Eq:pwe},\ref{Eq:ellipse}); \ \ \ \ \ $\vec{k}\!\!=\!\![\kappa\, \kappa]^T$,\ $\vec{d}_e\!\!=\!\![d_c\, d_c]^T$\\\hline
\multirow{4}{*}{plane}&ellipse&$\vec{d}_e$         &$\vec{r},\vec{t}$     & 8 & (\ref{Eq:pwi},\ref{Eq:pwe},\ref{Eq:ellipse}); \ \ \ \ \,$\vec{k}\!\!=\!\!\vecmb{0}$\\\cline{2-6}
                      &circle &$d_c$               &$\vec{r}_{xy},\vec{t}$& 6 & (\ref{Eq:pwi},\ref{Eq:pwe},\ref{Eq:ellipse}); \ \ \ \ \,$\vec{k}\!\!=\!\!\vecmb{0}$, $\vec{d}_e\!\!=\!\![d_c\, d_c]^T$\\\cline{2-6}
                      &aa rect&$\vec{d}_r$         &$\vec{r},\vec{t}$     & 8 & (\ref{Eq:pwi},\ref{Eq:pwe},\ref{eq-rect},\ref{eq-quad}); \,$\vec{k}\!\!=\!\!\vecmb{0}$\\\cline{2-6}
                      &c quad &$\vec{d}_q$         &$\vec{r},\vec{t}$     & 11 & (\ref{Eq:pwi},\ref{Eq:pwe},\ref{eq-cquad},\ref{eq-quad}); \,$\vec{k}\!\!=\!\!\vecmb{0}$\\\hline
sphere                &circle &$d_c,\kappa$        &$\vec{r}_{xy},\vec{t}$& 7 & (\ref{Eq:swi},\ref{Eq:swe},\ref{Eq:ellipse}); \ \ \ \ \,$\vec{d}_e\!\!=\!\![d_c\, d_c]^T$\!\!, $|\kappa|d_c\!\!\leq\!\!1$\\\hline
circular cylinder           &aa rect&$\vec{d}_r,\kappa$  &$\vec{r},\vec{t}$     & 9 & (\ref{Eq:cwi},\ref{Eq:cwe},\ref{eq-rect},\ref{eq-quad}); \,$|\kappa|d_y\!\!\leq\!\!1$\\\hline
\end{tabular}
}
\end{center}
\caption{The 10 patch types shown in Figure~\ref{Fig:patches}.}
\label{Tb:patches}
\end{table}

\subsection{Extrinsic and Intrinsic Surface Parameters} \label{Sec:params}
\begin{figure*}[htb]
  \begin{center}
  \includegraphics{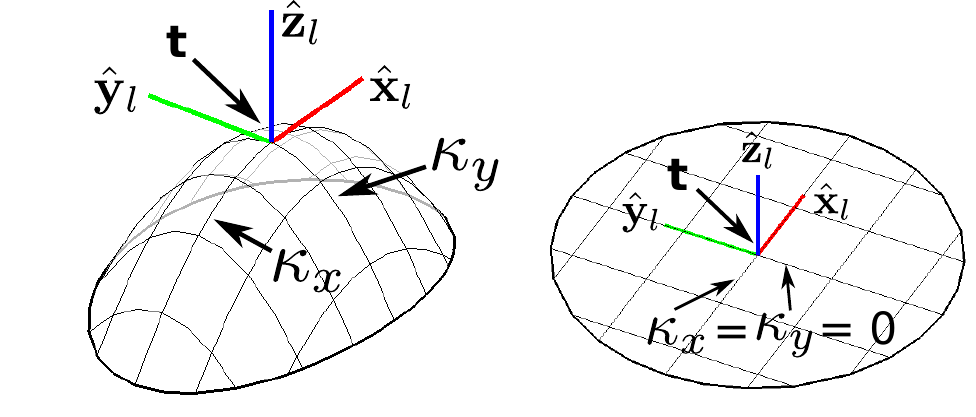}
  \end{center}
  \caption[A paraboloid and a planar patch example]{A paraboloid patch with two negative curvatures $(\kappa_x,\kappa_y)$ (left), a planar patch with zero curvatures (right), the symmetry point $\vec{t}$, and the local frame basis $[\uvec{x}_l\ \uvec{y}_l\ \uvec{z}_l]$.}
  \label{Fig:patch-details}
\end{figure*}

An instance of a patch will be a vector of real parameters which define both its shape (curvature and boundary) and its 3D rigid-body pose.  We call the former \emph{intrinsic} and the latter \emph{extrinsic} parameters~\cite{Srinivasan03}.  We must consider different issues to achieve minimal parametrization for each, and the distinction also enables the option to model shape (intrinsic) and pose (extrinsic) uncertainty separately.  Minimal intrinsic parametrization for the proposed patches will be given by (a) one parameter for each variable curvature, and (b) a minimal parametrization of the boundary curve.  However, minimal extrinsic parametrization depends on the \emph{continuous symmetry class} of the patch.  For example, a patch with two different curvatures (Figure~\ref{Fig:patch-details} left) has no continuous symmetry: its rigid body pose---here any element in the special Euclidean group $SE(3)$---has six degrees of freedom (DoF).

But a planar patch with a circular boundary (Figure~\ref{Fig:patch-details} right) has a continuous rotation symmetry and only five extrinsic DoF.  Remarkably, it has been shown that there are exactly seven continuous symmetry classes in 3D~\cite{Srinivasan03}: revolute, prismatic, planar, spherical, cylindrical, helical, and general (the first six correspond to the lower kinematic pairs; the last represents \emph{no} continuous symmetry).  Since we only consider patches with boundaries, we need only the general (no continuous symmetry, 6 DoF pose) and revolute (one continuous rotation symmetry, 5 DoF pose) classes---continuous translation symmetries are not possible for bounded patches.

\subsection{Pose Representation with the Exponential Map} \label{Sec:pose}
We require two extrinsic parameterizations: one with six parameters for asymmetric patches and one with five parameters for patches with revolute symmetry.  It is well known that, because the Lie-manifold of the special orthogonal group $SO(3)$ (the rotation subgroup of $SE(3)$) is non-Euclidean, there is no singularity-free minimal parametrization of $SE(3)$.  For the general 6-DoF case we thus select a minimal parametrization with singularities that are easiest to handle for our application.  One of the core computations will be patch fitting by iterative optimization, and for this Grassia showed in~\cite{Gra98} that a useful pose representation is\footnote{We explicitly notate transposes; orientation is crucial esp. for Jacobians.}
\beg
  [\vec{r}^T\ \vec{t}^T]^T\in\R^6\text{ with }(\vec{r},\vec{t})\in\R^3\times\R^3 \label{Eq:rt}
\eeg
where $\vec{t}$ is a translation and $\vec{r}$ is an \emph{orientation vector} giving an element of $SO(3)$ via an exponential map.  Grassia observed that in this parametrization singularities are avoidable by a fast dynamic reparameterization, reviewed below.

We use Rodrigues' rotation formula for the exponential map $R(\vec{r})\!:\!\R^3\!\rightarrow\!SO(3)\!\subset\!\R^{3\times3}$ (Grassia used quaternions):

\beg
  R(\vec{r})=I+[\vec{r}]_\times\alpha+[\vec{r}]_\times^2\beta \label{Eq:rexp}\\ \label{Eq:R}
  \theta\defeq\|\vec{r}\|,\ %
  \alpha\defeq\frac{\sin\theta}{\theta},\ %
  \beta\defeq\frac{1-\cos\theta}{\theta^2}\nonumber\\
  \vec{r}=\left[\begin{smallmatrix}r_x\\r_y\\r_z\end{smallmatrix}\right],\ %
  [\vec{r}]_\times\defeq
  \left[\begin{smallmatrix}0&-r_z&r_y\\r_z&0&-r_x\\-r_y&r_x&0\end{smallmatrix}\right].\nonumber
\eeg

Despite division by $\theta=\|\vec{r}\|$, (\ref{Eq:rexp}) converges to $I$ as $\theta\rightarrow0$.  For numerical stability we use the series expansion approximations $\alpha\approx1-\theta^2/6$ and $\beta\approx1/2-\theta^2/24$ for small $\theta$ (e.g. for $\theta\leq\sqrt[4]{\text{machine precision}}$).  As promised, the $(\vec{r},\vec{t})$ representation has a direct geometric interpretation: $\vec{t}$ is just a translation, and (wlog for $\theta\neq0$) $\theta$ gives the right-hand-rule rotation angle about the spatial axis defined by the unit vector $\vec{r}/\theta$.  While exponential map approaches are not new~\cite{Bro83,Par94}, matrices in $se(3)\subset\R^{4\times4}$, the Lie algebra of $SE(3)$, are typically used instead of $(\vec{r},\vec{t})$.  Though elegant, the former do not satisfy our goals of minimal parametrization and direct geometric interpretation.\footnote{It is true that there is a 1:1 correspondence between matrices in $se(3)$ and elements of the $(\vec{r},\vec{t})$ parametrization; conceptually, we have invoked this correspondence and simplified the results directly in terms of $(\vec{r},\vec{t})$.}

Using the fact that counterclockwise rotation by $\theta$ is equivalent to clockwise rotation by $2\pi-\theta$, Grassia's reparametrization converts any $\vec{r}$ into a canonical \footnote{For $\theta=\pi$ there is still ambiguity between $\vec{r}$ and $-\vec{r}$; this can be resolved by a consistent sign policy.} one with $\|\vec{r}\|\leq\pi$:

\beg \label{Eq:rreparam}
  \theta'\defeq\theta\bmod2\pi,\ %
  \vec{r}'\defeq
  \begin{cases}
    \vecmb{0}&\text{if}\ \theta=0\\
    \vec{r}\theta'/\theta&\text{if}\ 0<\theta'\leq\pi\\
    \vec{r}(\theta'-2\pi)/\theta&\text{otherwise.}
  \end{cases}
\eeg

$\vec{r}'$ represents the same rotation as $\vec{r}$, but stays away from the singularity surfaces where $\theta$ is a multiple of $2\pi$.

Algebraically, $(\vec{r},\vec{t})$ corresponds to an element
\beq \label{Eq:SE3}
  \begin{bmatrix}R(\vec{r})&\vec{t}\\\vecmb{0}^T&1\end{bmatrix}\nonumber
\eeq
of $SE(3)$, a $4\times4$ homogeneous rigid body transform, and can thus define the pose of a local coordinate frame $L$ (and a patch therein) relative to a world frame $W$: $R(\vec{r})$ is a basis for $L$ and $\vec{t}$ is its origin.  The transformation of a point $\vec{q}_l$ in $L$ to $\vec{q}_w$ in $W$, and the reverse, are familiar functions $X_{f,r}:\R^3\times\R^3\times\R^3\rightarrow\R^3$
\bea
  \vec{q}_w &= X_f(\vec{q}_l,\vec{r},\vec{t}) \defeq R(\vec{r})\vec{q}_l+\vec{t} \label{Eq:xf}\\
  \vec{q}_l &= X_r(\vec{q}_w,\vec{r},\vec{t}) \defeq R(-\vec{r})(\vec{q}_w\!\!-\!\vec{t})=R(\vec{r})^T(\vec{q}_w\!\!-\!\vec{t}) \label{Eq:xr}
\eea
where (\ref{Eq:xr}) makes use of the inverse transform
\beq\label{eq-xinv}
  (\vec{r},\vec{t})^{-1}\defeq(-\vec{r},-R(-\vec{r})\vec{t})=(-\vec{r},-R(\vec{r})^T\vec{t}).
\eeq

Equations (\ref{Eq:rt}--\ref{Eq:xr}) constitute our 6 DoF pose parametrization.  For the 5 DoF case, observe that only one of the three basis vectors of $L$ need be specified; rotation symmetry allows the others to make any mutually orthogonal triple.  Only two DoF are required, equivalent to specifying a point on the unit sphere.  We do this by re-using (\ref{Eq:rt}--\ref{Eq:xr}) with $r_z$ fixed at 0:
\beq \label{Eq:rt2}
  (\vec{r}_{xy},\vec{t})\in\R^2\times\R^3\ \text{corresp.}\ ([\vec{r}_{xy}^T\ 0]^T,\vec{t})\in\R^3\times\R^3.
\eeq

The geometric interpretation of $(\vec{r}_{xy},\vec{t})$ is the same as for $(\vec{r},\vec{t})$, except that $\vec{r}_{xy}$ is constrained to the $xy$ plane.  In some contexts we may want to calculate an $\vec{r}_{xy}\in\R^2$ that induces a local coordinate frame with the same $\uvec{z}_l$ as a given $\vec{r}\in\R^3$.  For any given canonical $\vec{r}$, a canonical $\vec{r}_{xy}$ always exists that satisfies
\beq \label{Eq:r2-condition}
  R([\vec{r}_{xy}^T\ 0]^T)\uvec{z} = R(\vec{r})\uvec{z}\text{ with }[\uvec{x}\ \uvec{y}\ \uvec{z}]\defeq I_{3\times3}.
\eeq
$\vec{r}_{xy}$ can be calculated as
\beg \label{Eq:r2}
  \vec{r}_{xy}(\vec{r}) = 
  %\begin{bmatrix}1&0&0\\0&1&0\end{bmatrix}
  \begin{bmatrix}\uvec{x}^T\\\uvec{y}^T\end{bmatrix}
  \begin{cases}
    \vec{r}&\text{if }\theta_{xy}\approx\pi\\
    (\uvec{z}\times\uvec{z}_l)/\alpha_{xy}&\text{otherwise}
  \end{cases}\\
  \uvec{z}_l\defeq R(\vec{r})\uvec{z},
  \theta_{xy}\defeq\atantwo(\|\uvec{z}\times\uvec{z}_l\|,\uvec{z}^T\uvec{z}_l),
  \alpha_{xy}\defeq\frac{\sin\theta_{xy}}{\theta_{xy}}\nonumber
\eeg

As in Brockett's product of exponentials~\cite{Bro83}, 6 DoF poses can be composed to make any kinematic chain.  Let
\beq\label{eq-chain}
(\vec{r}_n,\vec{t}_n)^{\phi_n},\ldots,(\vec{r}_1,\vec{t}_1)^{\phi_1}\text{ with }\phi_i\in\{+1,-1\}
\eeq
be the poses (equiv. transforms) in the chain from end to base in order from right to left.  Then the pose $(\vec{r}_c,\vec{t}_c)$ of a patch attached to the end of the chain relative to the base is
\beg \label{Eq:xc}
  (\vec{r}_c,\vec{t}_c)=(\vec{r}(R_n\cdots R_1),(X_n\!\circ\cdots\circ\!X_1)(\vecmb{0}))\\
  R_j\defeq R(\phi_j\vec{r}_j),\ %
  X_j(\vec{q})\defeq
  \begin{cases}
    X_f(\vec{q},\vec{r}_j,\vec{t}_j)&\text{if }\phi_j=+1\\
    X_r(\vec{q},\vec{r}_j,\vec{t}_j)&\text{if }\phi_j=-1
  \end{cases}\nonumber
\eeg
substituting $\vec{r}_{xy}(\vec{r_c})$ for 5 DoF patches, and using the \emph{log map} $\vec{r}(R):SO(3)\rightarrow\R^3$ corresponding to the inverse of (\ref{Eq:rexp}).  We give an algorithm to compute $\vec{r}(R)$ in Appendix~\ref{Sec:rlog}.

We will need the partial derivatives of (\ref{Eq:xr}) 
\beq \label{Eq:dxr}
  \udims{\frac{\partial\vec{q}_l}{\partial\vec{q}_w}}{3}{3}\!=\!R^T\!\!,%
  \udims{\frac{\partial\vec{q}_l}{\partial\vec{r}}}{3}{3}\!=\!\frac{\partial R^T}{\partial\vec{r}}(\vec{q}_w\!\!-\!\vec{t}),%
  \udims{\frac{\partial\vec{q}_l}{\partial\vec{t}}}{3}{3}\!=\!-R^T\!\!,%
  R\!\defeq\!R(\vec{r})
\eeq
the Jacobian of (\ref{Eq:rexp})---including its use as part of ${\partial\vec{q}_l}/{\partial\vec{r}}$ in (\ref{Eq:dxr})---and the Jacobians of (\ref{Eq:r2}) and (\ref{Eq:xc}):
\beq
  \udimst{\frac{\partial R}{\partial\vec{r}}}{3}{3}{3},\ %
  \udims{\frac{\partial\vec{r}_{xy}}{\partial\vec{r}}}{2}{3},\ %
  \udims{\frac{\partial(\vec{r}_c,\vec{t}_c)}{\partial(\vec{r}_1,\vec{t}_1),\ldots,(\vec{r}_n,\vec{t}_n)}}{6}{6n}.\nonumber
\eeq
The latter three are given in Appendix~\ref{Sec:jacobians}.

\subsection{Patch Models} \label{Sec:patch_models}
We now present the details of ten surface patch models (Figure~\ref{Fig:patches}, Table~\ref{Tb:patches}) based on seven curved surface types.  Five of these partition the paraboloids, including the important degenerate case of a plane; the other two add true spherical and circular cylinder patches, non-paraboloids that are common in man-made environments and on robots.  For non-planar surfaces we select one specific parametrized boundary shape which trims the surface into a local patch.  For planes we allow a choice of four boundary shapes.

The next two sections give the details of the paraboloid and non-paraboloid patch models.  This particular system is not the only possible taxonomy; it reflects our design choices in an attempt to balance expressiveness vs minimality.

\subsubsection{Paraboloids} \label{Sec:paraboloids}
The best-fit degree-two local polynomial approximation to any smooth surface $S\subset\R^3$ at a given point $\vec{t}\in S$, called the \emph{principal quadric}, is always a \emph{paraboloid}---a quadric of one sheet with a central point of symmetry about which the surface has two independent curvatures $\kappa_x, \kappa_y$ in orthogonal directions (Figure~\ref{Fig:patch-details} left).  These are the \emph{principal curvatures} of $S$ at $\vec{t}$, and $\vec{t}$ is the symmetry point.  Defining $\uvec{x}_l$ and $\uvec{y}_l$ as unit vectors in the directions of the principal curvatures in the tangent plane to $S$ at $\vec{t}$, the surface normal to $S$ at $\vec{t}$ is $\uvec{z}_l\defeq\uvec{x}_l\times\uvec{y}_l$.  If $S$ is considered to be embedded in a world coordinate frame $W$, then $\vec{t}\in\R^3$ is the origin and
\beq \label{Eq:parab_local_basis}
  R\defeq[\uvec{x}_l\ \uvec{y}_l\ \uvec{z}_l]\nonumber
\eeq
is a basis for the \emph{principal coordinate frame} (all standard terms) of $S$ at $\vec{t}$, which we also call \emph{local frame} $L$ \footnote{L is also the \emph{Darboux} frame \cite{HS02} of the paraboloid}.

Using the log map, the transform
\beq
  (\vec{r},\vec{t})\defeq(\vec{r}(R),\vec{t})\nonumber
\eeq
takes points from $L$ to $W$\!\!, enabling a short derivation of equations for a general paraboloid parametrized by $\vec{k}\defeq[\kappa_x\ \kappa_y]^T$\!\!, $\vec{r}$, and $\vec{t}$.  Starting in $L$ where the paraboloid is in standard position, with $p_{li}\!:\!\R^3\!\times\!\R^2\!\!\rightarrow\R$ and $p_{le}\!:\!\R^2\!\times\!\R^2\!\!\rightarrow\R^3$\!\!\!,
\bea
  0=p_{li}(\vec{q}_l,\vec{k})&\defeq\vec{q}_l^T\diag([\vec{k}^T\ 0]^T)\vec{q}_l-2\vec{q}_l^T\uvec{z} \label{Eq:pli}\\
  \vec{q}_l=p_{le}(\vec{u},\vec{k})&\defeq[\uvec{x}\ \uvec{y}]\vec{u}+\frac{1}{2}\vec{u}^T \diag(\vec{k})\vec{u}\uvec{z} \label{Eq:ple}
\eea
are the implicit and explicit forms for the surface equation, respectively, with $\vec{q}_l\in\R^3$ a point on the patch in $L$ and $\vec{u}\in\R^2$ parameters of the explicit form.  Moving to $\vec{q}_w\in\R^3$ in world frame $W$ is accomplished by composing (\ref{Eq:pli},\ref{Eq:ple}) with (\ref{Eq:xf},\ref{Eq:xr}), yielding 
\bea
  0=p_{wi}(\vec{q}_w,\vec{k},\vec{r},\vec{t})&\defeq p_{li}(X_r(\vec{q}_w,\vec{r},\vec{t}),\vec{k}) \label{Eq:pwi}\\
  \vec{q}_w=p_{we}(\vec{u},\vec{k},\vec{r},\vec{t})&\defeq X_f(p_{le}(\vec{u},\vec{k}),\vec{r},\vec{t}) \label{Eq:pwe}
\eea

\vspace{-2em}

\beg
  p_{wi}:\R^3\!\!\times\!\!\R^2\!\!\times\!\!\R^3\!\!\times\!\!\R^3\rightarrow\R,\ p_{we}:\R^2\!\!\times\!\!\R^2\!\!\times\!\!\R^3\!\!\times\!\!\R^3\rightarrow\R^3.\nonumber
\eeg
Note that in this formulation $\vec{u}$ is always the projection of $\vec{q}_l$ onto the local frame $xy$ plane:
\beq \label{Eq:u}
  \vec{u}\defeq\Pi_{xy}\vec{q}_l=\Pi_{xy}X_r(\vec{q}_w,\vec{r},\vec{t}),\ \Pi_{xy}\defeq
  %\begin{bmatrix}1&0&0\\0&1&0\end{bmatrix}.
  [\uvec{x}\ \uvec{y}]^T.
\eeq

In the general case $\kappa_x\!\neq\!\kappa_y$, giving 7 or 8 DoF paraboloids---6 pose DoF plus up to two curvatures (boundary parameterizations will add DoF).  Six DoF pose is required because $\kappa_x\!\neq\!\kappa_y$ implies no continuous rotation symmetries, only discrete symmetries about $\vec{t}$.  It is standard to separate three surface types where $\kappa_x\!\neq\!\kappa_y$ (Figure~\ref{Fig:patches}): \emph{elliptic paraboloids} have two nonzero curvatures with equal signs, \emph{hyperbolic paraboloids} have two nonzero curvatures with opposite signs, and \emph{cylindric paraboloids} have one nonzero curvature.  In all cases $\uvec{z}_l$ is the outward pointing surface normal and positive/negative curvatures correspond to concave/convex directions on the patch, respectively\footnote{To reduce ambiguity wlog choose $|\kappa_x|<|\kappa_y|$, though some ambiguity is unavoidable due to bilateral symmetries.}.

\subsubsection*{Boundaries}
\begin{figure*}[h]
  \begin{center}
  \includegraphics{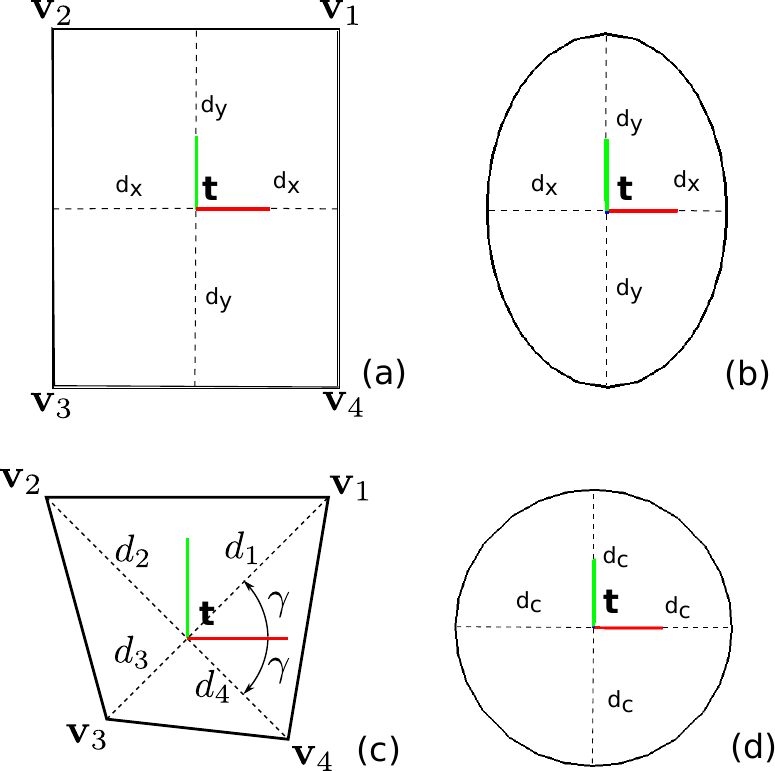}
  \end{center}
  \caption[Examples of all four different types of boundaries]{(a) Rectangle boundary parametrized by $\vec{d}_e\defeq[d_x\ d_y]^T$; (b) Ellipse boundary parametrized by $\vec{d}_r=[d_x\ d_y]^T$; (c) Convex quadrilateral boundary parametrized by $\vec{d}_q\!\!\defeq\!\![d_1\ d_2\ d_3\ d_4\ \gamma]^T$; (d) Circle boundary parametrized by $d_c$}
  \label{Fig:paraboloids}
\end{figure*}

We bound \emph{elliptic and hyperbolic paraboloid} patches with \textbf{ellipses} in the $xy$ plane of the local frame $L$, axis aligned and centered at $\vec{t}$ (Figure~\ref{Fig:paraboloids} (b)).  If $\vec{d}_e\defeq[d_x\ d_y]^T$ are the ellipse radii then the bounded patch is the subset of the full surface (\ref{Eq:pli}--\ref{Eq:pwe}) where, with $e\!:\!\R^2\!\times\!\R^2\!\!\rightarrow\R$, $\vec{u}$ satisfies
\beq \label{Eq:ellipse}
  0\geq e(\vec{u},\vec{d}_e)\defeq\vec{u}^T\diag([1/d_x^2\ 1/d_y^2])\vec{u}-1.
\eeq

For \emph{cylindric paraboloid} patches, replace the ellipse boundary with an \textbf{axis aligned rectangle} with half-widths $\vec{d}_r=[d_x\ d_y]^T$ (Figure~\ref{Fig:paraboloids} (a)).  In the $xy$ plane of $L$ the vertices are 
\beq \label{eq-rect}
  \vec{v}_1\defeq\vec{d}_r,
  \vec{v}_2\defeq[-d_x\ d_y]^T,
  \vec{v}_3\defeq-\vec{v}_1,
  \vec{v}_4\defeq-\vec{v}_2
\eeq
in counterclockwise order, and the bounding condition can be stated as, with $q:\R^2\times\R^2\times\R^2\times\R^2\times\R^2\rightarrow\R$,
\beg
  0\geq q(\vec{u},\vec{v}_1,\vec{v}_2,\vec{v}_3,\vec{v}_4)\defeq\label{eq-quad}\\
  \max(
    l(\vec{u},\vec{v}_1,\vec{v}_2),
    l(\vec{u},\vec{v}_2,\vec{v}_3),
    l(\vec{u},\vec{v}_3,\vec{v}_4),
    l(\vec{u},\vec{v}_4,\vec{v}_1))\nonumber
\eeg
where $l:\R^2\times\R^2\times\R^2\rightarrow\R$ is the implicit form for a 2D line given two points on the line; $\vec{u}$ is on or to the left of the directed line through $\vec{v}_i$ towards $\vec{v}_j$ iff 
\beq
  0\geq l(\vec{u},\vec{v}_i,\vec{v}_j)\defeq (\vec{u}-\vec{v}_i)^T[\vec{v}_j-\vec{v}_i]_\perp,\
  \begin{bmatrix}x\\y\end{bmatrix}_\perp\!\!\!\!\!\defeq\!\begin{bmatrix}y\\-x\end{bmatrix}.\label{eq-line}
\eeq
 
For the special case $\kappa_x=\kappa_y$ we identify two more surface types (Figure~\ref{Fig:patches}): \emph{circular paraboloids} have both curvatures equal and non-zero, and \emph{planes} have both curvatures zero.  Both of these have continuous rotation symmetry about $\uvec{z}_l$, so we use the 5-DoF pose parametrization $(\vec{r}_{xy},\vec{t})$, provided that the patch boundary also has the same continuous rotation symmetry.  The latter holds for \textbf{circular boundaries}, which we use for \emph{circular paraboloids} (Figure~\ref{Fig:paraboloids} (d)).  Let $\kappa$ be the surface curvature and $d_c$ the bounding circle radius; circular paraboloids are then defined by (\ref{Eq:pli}--\ref{Eq:ellipse}) with $\vec{k}=[\kappa\ \kappa]^T$, $\vec{r}=[\vec{r}_{xy}^T\ 0]^T$, $\vec{d}_e=[d_c\ d_c]^T$, and with the dimensions of the function domains correspondingly reduced.

For the important case of paraboloids degenerated to \emph{planes} we give a choice of four boundary types: \textbf{ellipses, circles, rectangles, or general convex quadrilaterals} (developed next).  For all except circles, the planar patch loses its continuous rotation symmetry and requires full 6-DoF pose parametrization; the patch is defined by (\ref{Eq:pli}--\ref{Eq:pwe}) with $\vec{k}=\vecmb{0}$ (and correspondingly reduced function domains) and either (\ref{Eq:ellipse}) or (\ref{eq-quad}).  Planar patches with circular boundary are the same as circular paraboloids but with $\vec{k}=\vecmb{0}$.

For \textbf{convex quadrilateral} boundaries, keep $\vec{t}$ at the intersection of the diagonals $\overline{\vec{v}_1\vec{v}_3}$ and $\overline{\vec{v}_2\vec{v}_4}$ (Figure~\ref{Fig:paraboloids} (c)), where $\vec{v}_{1\ldots4}$ are the vertices in counterclockwise order in the $xy$ plane of local frame $L$.  Define $\gamma$ as half the angle between the diagonals and $d_{1\ldots4}\geq0$ the half-diagonal lengths such that
\beg
  \vec{v}_i\defeq d_i[\cos\phi_i\ \sin\phi_i]^T\label{eq-cquad}\\
  \phi_1\defeq\gamma,\ \phi_2\defeq\pi-\gamma,\ \phi_3\defeq\pi+\gamma,\ \phi_4\defeq-\gamma\nonumber\\
  0<\gamma<\pi/2\nonumber.
\eeg
Then the quadrilateral is defined by (\ref{eq-quad}) using vertices (\ref{eq-cquad}) parametrized by $\vec{d}_q\!\!\defeq\!\![d_1\ d_2\ d_3\ d_4\ \gamma]^T$.  Convexity is ensured by construction, and only five parameters are needed even though a general planar quadrilateral has 8 DoF---the remaining three (a rotation about the plane normal and two in-plane translations) are contributed by the extrinsic pose.

\subsubsection{Spheres and Circular Cylinders} \label{Sec:spheres-and-circular-cylinders}
Spheres and circular cylinders are common on robots and in man-made environments.  Though still quadrics, neither is a paraboloid, suggesting two additional patch types (Figure~\ref{Fig:patches}).  (We do not model complete spheres or cylinders, only bounded patches of hemispheres and half-cylinders.)

Again starting in local frame $L$, the implicit and explicit equations of an upright hemisphere with apex at the origin and curvature $\kappa$ (hence possibly infinite radius $|1/\kappa|$) are\footnote{In the limit as $\kappa\rightarrow0$ (\ref{Eq:sli}--\ref{Eq:cwe}) all reduce to planes.}, with $s_{li}:\R^3\times\R\rightarrow\R$ and $s_{le}:\R^2\times\R\rightarrow\R^3$,
\bea
  0=s_{li}(\vec{q}_l,\kappa)&\defeq\kappa\vec{q}_l^T\vec{q}_l-2\vec{q}_l^T\uvec{z},\ %
  0\leq\kappa\vec{q}_l^T\uvec{z}\leq1 \label{Eq:sli}\\
  \vec{q}_l=s_{le}(\vec{u},\kappa)&\defeq[\uvec{x}\ \uvec{y}]\vec{u}\!+\!(\uvec{z}/\kappa)\left(1\!-\!\sqrt{1\!-\!\kappa^2\vec{u}^T\vec{u}}\right). \label{Eq:sle}
\eea
Composing these with (\ref{Eq:xf},\ref{Eq:xr}) gives the world frame forms $s_{wi}:\R^3\times\R\times\R^2\times\R^3\rightarrow\R$, $s_{we}:\R^2\times\R\times\R^2\times\R^3\rightarrow\R^3$
\bea
  0=s_{wi}(\vec{q}_w,\kappa,\vec{r}_{xy},\vec{t})&\defeq s_{li}(X_r(\vec{q}_w,[\vec{r}_{xy}^T\ 0]^T,\vec{t}),\kappa) \label{Eq:swi}\\
  0=s_{we}(\vec{u},\kappa,\vec{r}_{xy},\vec{t})&\defeq X_f(s_{le}(\vec{u},\kappa),[\vec{r}_{xy}^T\ 0]^T,\vec{t}). \label{Eq:swe}
\eea

Circular half-cylinder surfaces are similar but (a) have no dependence on $x_l$ and (b) require 6 DoF pose:
\beg
  0=c_{li}(\vec{q}_l,\kappa)\defeq\vec{q}_l^TK\vec{q}_l-2\vec{q}_l^T\uvec{z},\ %
  0\leq\kappa\vec{q}_l^T\uvec{z}\leq1\label{eq-cli}\\
  \vec{q}_l\!=\!c_{le}(\vec{u},\kappa)\!\defeq\![\uvec{x}\ \uvec{y}]\vec{u}\!+\!(\uvec{z}/\!\kappa)\left(\!1\!-\!\sqrt{1\!-\!\kappa^2\vec{u}^TY\vec{u}}\right)\label{eq-cle}\\
  K\defeq\diag([0\ \kappa\ \kappa]^T),\ %
  Y\defeq[0\ 1]^T[0\ 1]\nonumber\\%\begin{bmatrix}0&0\\0&1\end{bmatrix}\nonumber\\
  0=c_{wi}(\vec{q}_w,\kappa,\vec{r},\vec{t})\defeq c_{li}(X_r(\vec{q}_w,\vec{r},\vec{t}),\kappa) \label{Eq:cwi}\\
  0=c_{we}(\vec{u},\kappa,\vec{r},\vec{t})\defeq X_f(c_{le}(\vec{u},\kappa),\vec{r},\vec{t}). \label{Eq:cwe}
\eeg

\subsubsection*{Boundaries}
To maintain revolute symmetry we use circular boundary for spherical patches: $\vec{u}$ must satisfy (\ref{Eq:ellipse}) with $\vec{d}_e=[d_c\ d_c]^T$ and $|\kappa|d_c\leq 1$.  For circular cylinder patches we use rectangular boundary: $\vec{u}$ must satisfy (\ref{eq-rect},\ref{eq-quad}) with $|\kappa|d_y\leq 1$.

\section{Patch Fitting} \label{Sec:patch_fitting}
\begin{figure*}[htb]
  \begin{center}
    \includegraphics[width=\textwidth]{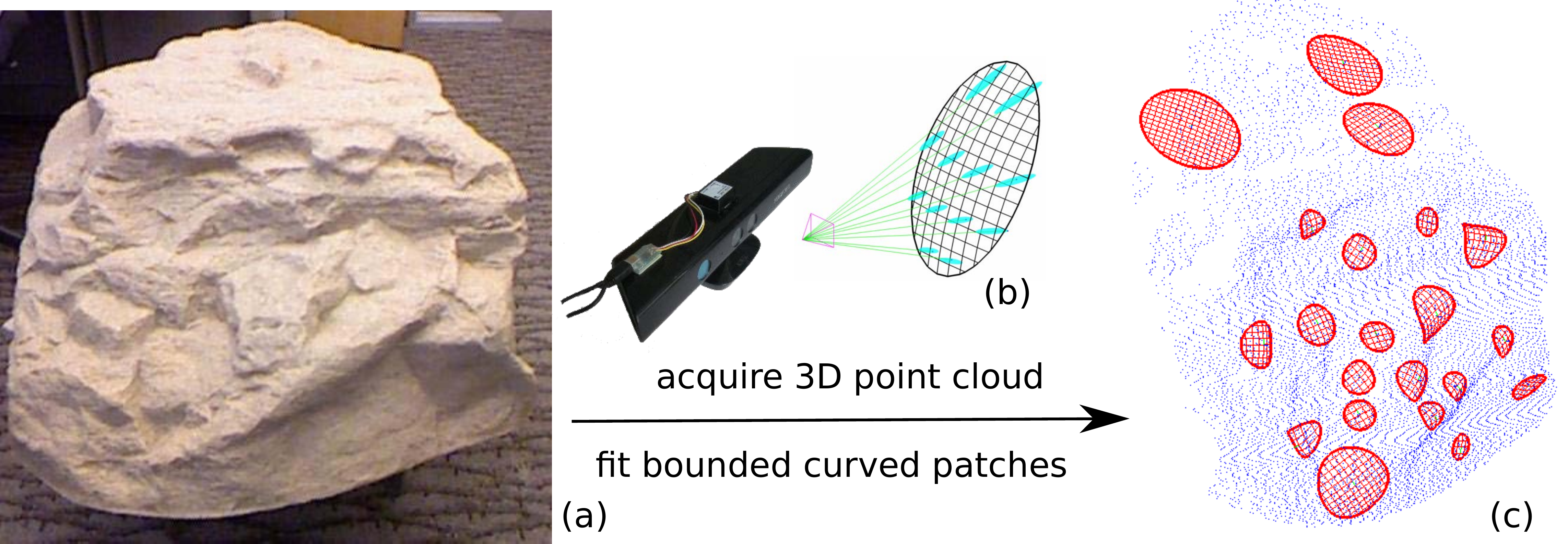}
  \end{center}
  \caption[Rock data acquisition, error ellipsoid estimation for each 3D point, and patch fitting for manually segmented point clouds]{(a) Experimental dataset: (fake) rock is $\sim$70$\times$30$\times$90cm W$\times$H$\times$D; $\sim$125k samples collected in a single scan with a Kinect at a distance of $\sim$1m (decimated for display); (b) 95\% probability error ellipsoids for stereo range sensing using the pointing/disparity error model of Murray and Little~\cite{ML05} (pointing error exaggerated for illustration); (c) 21 patches manually segmented and automatically fit.}
  \label{Fig:fitting}
\end{figure*}

Having defined the patch models, it is natural to consider recovering contact surface areas in the environment by fitting bounded curved patches to noisy point samples with quantified uncertainty both in the inputs (the points) and the outputs (the patch parameters), which is not a trivial problem\footnote{We have found very few prior reports on the particular fitting problem including boundaries and quantified uncertainty.} (Figure~\ref{Fig:fitting}).  Though linear least squares (LLS) can fit a quadric surface to points~\cite{DNC07}, and its extension to linear $\chi^2$ maximum likelihood fits data corrupted by white noise, the problem appears to become non-linear when the points are heteroskedastic (i.e. have nonuniform variance).  Also, we want to fit  bounded paraboloids, spheres, and circular cylinders, not just unconstrained quadrics.

In \cite{VK11} we give a non-linear fitting algorithm which handles these issues.  It is based on a variation of Levenberg-Marquardt iteration that fits a bounded curved patch to a set of 3D sample points.  The algorithm minimizes a sum-of-squares residual by optimizing the patch implicit and explicit parameters.  The residual for an individual sample point is computed by scaling the value of the implicit form by the inverse of a first-order estimate of its standard deviation, which is derived in turn from a covariance matrix modeling the sensor uncertainty for the point.  

Next we describe the patch fitting algorithm.  Elliptic, hyperbolic, circular, and cylindrical paraboloid as well as planar patches are fitted automatically depending on the detected curvatures of the underlying surface.  The non-paraboloids (cylindrical and spherical patches) are fitted only if requested.  Also, similarly, for planar paraboloids, the type selection for the boundary curve is only partially automatic --- rectangles and convex quads are only used if requested.

\subsection{Patch Fitting Algorithm} \label{Sec:fit}
\begin{figure*}[htb]
  \begin{center}
    \includegraphics{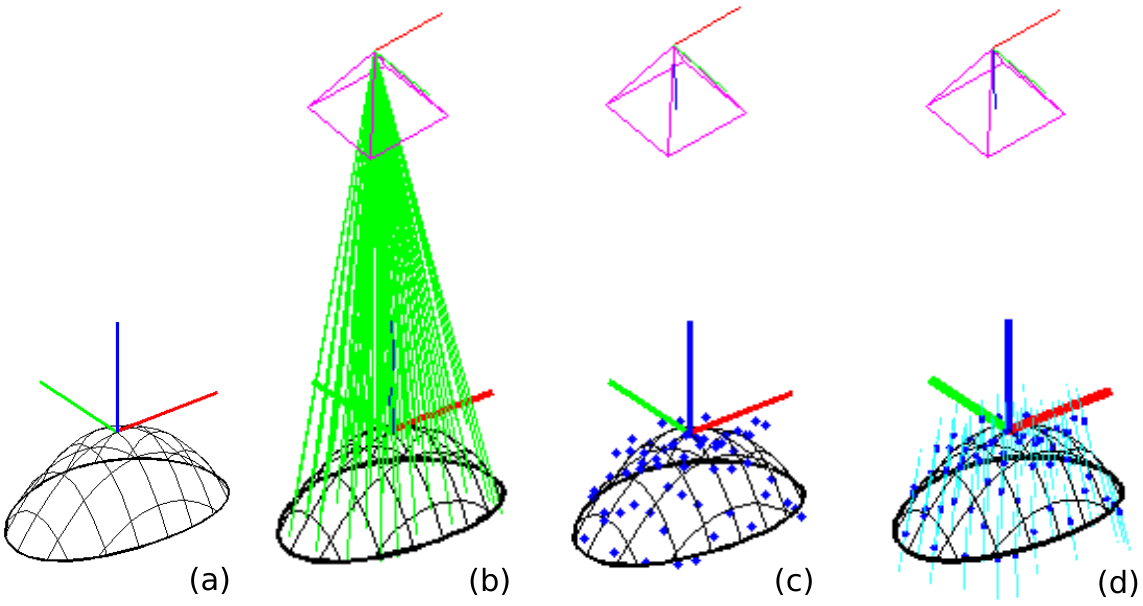}
  \end{center}
  \caption[Input for the paraboloid patch fitting process]{Input for the paraboloid patch fitting process: (a) the original paraboloid patch; (b) the viewing frustum and the measurement rays; (c) 61 data sample points (white Gaussian noise added); (d) error ellipsoids.}
  \label{Fig:fitting_steps_input}
\end{figure*}

The {\bf inputs} are (Figure~\ref{Fig:fitting_steps_input}):
\begin{itemize}
  \item $N$ sample points $\vec{q}_i\in\R^3$ with covariance matrices $\Sigma_i\in\R^{3\times3}$ (positive semi-definite)
  
  \item the general surface type to fit\footnote{Taking $s,b$ as inputs allows constrained fitting of specific types; they could be automatically found by checking all possibilities for the best fit.} $s\in\;$\{\emph{parab}, plane, sphere, ccyl\}

  \item the boundary type $b\in\;$\{ellipse, circle, aarect, cquad\} if $s=\text{plane}$\footnote{$b$ is implied if $s\neq\text{plane}$.}

  \item a boundary containment probability $\Gamma\in(0,1]$
\end{itemize}

The {\bf outputs} are:
\begin{itemize}
  \item the fitted patch type $(s,b)$
  
  \item parameters $\vec{p}\in\R^p$, where $p$ is the DoF of the patch type (Table~\ref{Tb:patches})
  
  \item covariance matrix $\Sigma\in\R^{p\times p}$
\end{itemize}

\begin{figure*}[h]
  \begin{center}
    \includegraphics{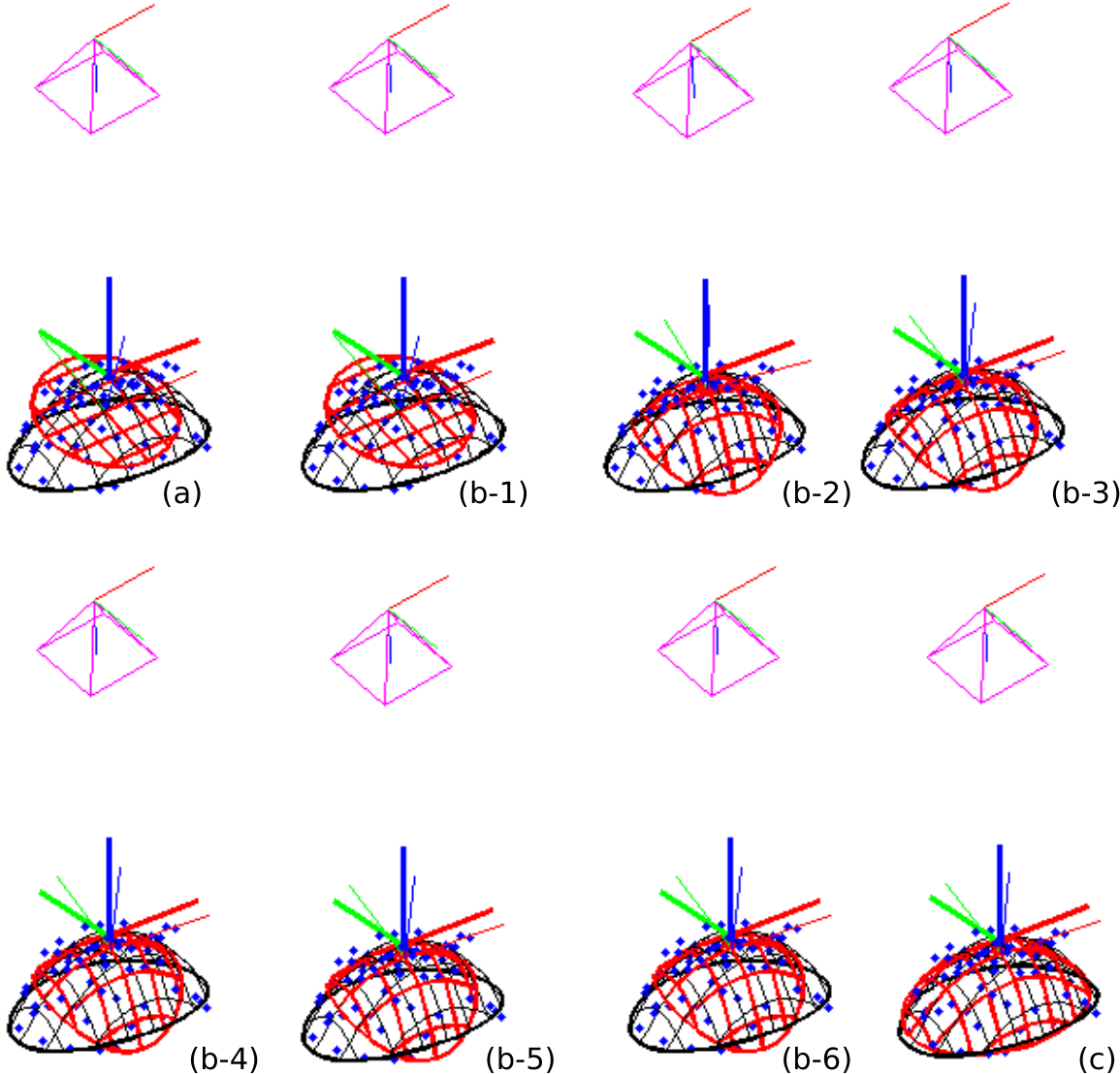}
  \end{center}
  \caption[Paraboloid patch fitting process]{Paraboloid patch fitting process: (a) plane fitting w/o input uncertainty by lls (\textbf{Step 1-i}); (b-1 to b-6) paraboloid fitting with input uncertainty by WLM; note that an unbounded surface is fit but for illustration we show the surface bounded; (c) fitting elliptic boundary to the 61 projected data points.}
  \label{Fig:fitting_steps}
\end{figure*}

The algorithm proceeds in 2 stages (9 total steps), which include heuristics for avoiding local minima when solving the non-linear system.  The first three steps fit an unbounded surface; the rest are largely concerned with fitting the boundaries, which can include final resolution of the patch center and orientation (in steps 6 and 9) where the bounding shape breaks symmetries of the underlying surface.  An illustration of the whole fitting process for a simulated paraboloid can be shown in Figure~\ref{Fig:fitting_steps}.

\newcounter{fit-stage}
\newcounter{fit-step}

\begin{list}{\bf Stage \Roman{fit-stage}:}{\leftmargin=1em\usecounter{fit-stage}}
  \item Fit an Unbounded Surface \label{Stage:fit-surface}
  
  \begin{list}{\it Step \arabic{fit-step}:}{\usecounter{fit-step}\setcounter{fit-step}{0}}
    \item \textit{Plane fitting} \label{fit-step-plane}
      \begin{enumerate}[(i)]
        \item Fit a plane with LLS, ignoring $\Sigma_i$. \label{fit-step-lls-plane}
        
        \item Unless $s=\text{\emph{parab}}$, re-fit the plane with \emph{weighted Levenberg-Marquardt} (WLM), detailed below, including $\Sigma_i$, using (\ref{Eq:swi}) with $\kappa=0$. \label{fit-step-wlm-plane}
        
        \item Set $\vec{t}\gets\bar{\vec{q}}-\uvec{z}_l^T(\bar{\vec{q}}-\vec{t})\uvec{z}_l$ (perp. proj. of $\bar{\vec{q}}\defeq\avg(\vec{q}_i)$ on plane).
      \end{enumerate}
      Note that:
      \begin{itemize}
        \item At the end of Step~\ref{fit-step-plane} the refined plane is defined by (i) the point $\vec{t}$ on it, which is the centroid of the data and (ii) by the plane normal which is the third column of the rotation matrix.  Note that the third element of the rotation vector $\vec{r}$ is zero, since it is currently a 2D orientation vector, because an \textbf{unbounded} plane is rotationally symmetric about its normal.  Note also that an unbounded plane has only 3-DoF, not 5-DoF.  The extra two DoF are constrained during fitting by keeping $\vec{t}$ at the projection of the centroid of the data points onto the plane.  There is a choice of boundary shapes for plane fitting, and all except circular will break the rotational symmetry of the unbounded plane.  This is handled later in Step~\ref{fit-step-plane-boundary}, where it may be necessary to extend the rotation vector $\vec{r}$ from 2D to 3D.
        
        \item (if $s=\text{\emph{parab}}$):  Since a general \textbf{assymetric} paraboloid will be fitted by WLM in Step~\ref{fit-step-wlm}, probably a relatively expensive WLM in Step~\ref{fit-step-plane} will not improve the refinement.  The plane fit by LLS in Step~\ref{fit-step-plane}-\ref{fit-step-lls-plane} will serve to initialize WLM to the correct region of parameter space (which is important, because WLM will only find a local optimum), but all components of $\vec{r}$ and $\vec{t}$ (the only parameters we have estimated so far) will be replaced by the WLM in Step~\ref{fit-step-wlm}.
        
        \item (if $s=\text{\emph{plane}}$):  Since the end goal is to fit a plane and there will be no WLM in Step~\ref{fit-step-wlm} for planes, it is required to refine it in Step~\ref{fit-step-plane}-\ref{fit-step-wlm-plane}, and importantly, to get its covariance matrix.  Note that in Step~\ref{fit-step-plane}-\ref{fit-step-wlm-plane} a \textbf{redundant} parameterization is used, since the point on the plane (defined by $\vec{t}$) has two extra DoF to slide around on the plane.  This redundancy is compensated by constantly adjusting $\vec{t}$ at the projection of the data centroid onto the plane (a point that is well centered in the data is preferable, since all the planar patch boundary shapes ($\text{\emph{ellipse, circle, aa rect, convex quad}}$) are origin-centered).

      \item (if $s=\text{\emph{sphere}}$):  Since in this case the end goal is a spherical patch, and in Step~\ref{fit-step-wlm} only an unbounded (i.e. complete) sphere will be fitted, there is a need to extract the orientation of the patch, which is not coming from the WLM.  It is determined, instead, by the plane normal calculated here in Step~\ref{fit-step-plane}.  Thus the plane normal estimation is essential, including the uncertainty of the input data points.  The LLS in Step~\ref{fit-step-plane}-\ref{fit-step-lls-plane} did not consider that uncertainty, so it is needed to refine the plane fit with WLM.
      
      \item (if $s=\text{\emph{ccyl}}$): For circular cylindrical patch, the logic is similar to the spherical patch.  The WLM in Step~\ref{fit-step-wlm} fits an unbounded cylinder and the orientation of the final bounded cylindrical patch will come from a combination of the plane normal that is calculated here and the direction of the cylinder symmetry axis recovered in the WLM in Step~\ref{fit-step-wlm}.
    \end{itemize}

    \item \textit{Surface Fitting} (if $s\neq\text{plane}$) \label{fit-step-wlm}
      \begin{enumerate}[(i)]
        \item With $\vec{k}=[0\ 0]^T$, $\vec{r}\defeq[\vec{r}_{xy}^T\ 0]^T$, and $\vec{t}$ from~1 as initial estimates, according to $s$ fit an unbounded paraboloid, sphere, or circ cyl with WLM on (\ref{Eq:pwi},\ref{Eq:swi},\ref{Eq:cwi}).
        
        \item If $s=\text{sphere}$ keep $\vec{r}_{xy}$ from~1.  If $s=\text{circ cyl}$ set $\vec{r}=\vec{r}([\uvec{x}_l\ \uvec{y}_l\ \uvec{z}_l])$ (log map) where $\uvec{z}_l$ is the normal of the plane from 1, $\uvec{x}_l$ is along the fitted cylinder axis, and $\uvec{y}_l\defeq\uvec{z}_l\times\uvec{x}_l$.
      \end{enumerate}

    \item \textit{Curvature Discrimination} (if $s\!=\!\text{\emph{parab}}$) \label{fit-step-refinement}
    
      Refine the patch based on the fitted curvatures $\vec{k}=[\kappa_x\ \kappa_y]^T$:\\
      \textbf{If} $\max(|\kappa_x|,|\kappa_y|)<\epsilon_k$ (a small threshold), set $s\!=\!\text{plane}$, $b\!=\!\text{ellipse}$, and $\vec{r}_{xy}$ using (\ref{Eq:r2}).\\
      \textbf{Else if} $\min(|\kappa_x|,|\kappa_y|)<\epsilon_k$ swap axes s.t. $|\kappa_y|>\epsilon_k$, then set $s=\text{cyl parab}$ and $\kappa=\kappa_y$.\\
      \textbf{Else if} $|\kappa_x-\kappa_y|<\epsilon_k$ set $s=\text{circ parab}$, $\kappa=\avg(\kappa_x, \kappa_y)$, and $\vec{r}_{xy}$ using (\ref{Eq:r2}).\\
      \textbf{Else if} $\sgn(\kappa_x)=\sgn(\kappa_y)$ set $s=\text{ell parab}$.\\
      \textbf{Else} set $s=\text{hyp parab}$.   
  \end{list}
  
  \item Fit the Boundary\\
  Having fitted an unbounded patch to the 3D data, the boundary needs to be determined.  Principal Component Analysis (PCA) is applied to the 2D projected points onto the local $xy$-plane for finding the enclosing boundary.  A closed-form solution~\cite{RVC02} for the corresponding eigendecomposition problem using 1D and 2D moments to fit approximate boundaries is used.  Note that the difference between planar and non-planar patches is that for the latter the principal direction axes are the $xy$ axes of the local reference frame (implied by the principal curvatures of the underlying surface), whereas for the planar patches the axes need to be determined by fitting the boundary shape.
  \begin{list}{\it Step \arabic{fit-step}:}{\usecounter{fit-step}\setcounter{fit-step}{3}}
    \item \textit{Determine the Boundary Type} (if $s\neq\text{plane}$)
    
    Set $b$ based on $s$.  If the patch is not a plane then the type of the boundary is uniquely defined from the type of the patch (Table~\ref{Tb:patches}).  Otherwise it is determined from the user as an input.
    
    Set $\lambda \defeq \sqrt{2}\erf^{-1}(\Gamma)$ for boundary containment scaling~\cite{Ribeiro04}.

    \item \textit{Initialize Bounding Parameters}
    
    Project the data $\vec{q}_i\in\R^3$ to $\vec{u}_i=[x_i\ y_i]^T\in\R^2$ using (\ref{Eq:u}).
    
    Set first and second data moments: $\bar{x}\defeq\avg(x_i)$, $\bar{y}\defeq\avg(y_i)$, $v_x\defeq\avg(x_i^2)$, $v_y\defeq\avg(y_i^2)$.

    \item \textit{Cylindrical Paraboloid and Circular Cylinder Boundary Fitting} \label{fit-step-cyl_parab-circ_cyl_bound}
    
    If $s\in\{\text{cyl parab},\text{circ cyl}\}$ set $\vec{d}_r=\lambda[\sqrt{v_x-\bar{x}^2}\ \sqrt{v_y}]^T$ and $\vec{t}\!\gets\!\!X_f(\bar{x}\uvec{x},\vec{r},\vec{t})$, where $\sqrt{v_x-\bar{x}^2}$ is the standar deviation along the $x$ axis of the local frame and $\sqrt{v_y}$ is the standard deviation along the $y$ axis (the data are already zero-mean in $y$ due to the nonzero principal curvature in that direction).

    \item \textit{Circular Paraboloid and Sphere Boundary Fitting}
    
    If $s\in\{\text{circ parab},\text{sphere}\}$ set $d_c=\lambda\max(\sqrt{v_x}, \sqrt{v_y})$.

    \item \textit{Elliptic and Hyperbolic Boundary Fitting}
    
    If $s\in\{\text{ell parab},\text{hyp parab}\}$ set $\vec{d}_e=\lambda[\sqrt{v_x}\ \sqrt{v_y}]^T$.

    \item \textit{Plane Boundary Fitting} \label{fit-step-plane-boundary}
    \begin{enumerate}[(i)]
      \item If $s=\text{plane}$, $\vec{r}_{xy}$ and $\vec{t}$ will be available from either 1 or 3.  The extents and rotation of the boundary curve are determined now by two-dimensional PCA.
      
      Set $\vec{t}\!\gets\!\!X_f(\bar{x}\uvec{x}+\bar{y}\uvec{y},\vec{r}_{xy},\vec{t})$ and (c.f.~\cite{RVC02})
      \beg \label{Eq:radii}
        l_{+,-}\defeq\sqrt{-\ln(1-\Gamma)(\alpha+\phi\pm\sqrt{\beta^2+(\alpha-\phi)^2})}\\
        \text{using }\alpha\defeq v_x-\bar{x}^2,\ \beta\defeq2(\avg(x_iy_i)-\bar{x}\bar{y}),\ \phi\defeq v_y-\bar{y}^2.\nonumber
      \eeg
      \item If $b\!\!=\!\!\text{circle}$ set $d_c\!\!=\!\!\max(l_+,l_-)$ and $\vec{r}_{xy}$ from $\vec{r}$ using (\ref{Eq:r2}).
    
      \item If $b\!\!\in\!\!\{\text{ellipse},\text{aarect}\}$ set $\vec{d}_{e,r}\!\!=\!\![l_+\ l_-]^T$.
    
      \item If $b\!\!=\!\!\text{conv quad}$ set\footnote{We currently fit convex quad the same as aa rect, but note that the former is still useful for designed (vs fit) patches, e.g. part of a foot sole.} 
        \beq
          \vec{d}_q=[d\ d\ d\ d\ \gamma]^T\!\!\!, d\defeq\sqrt{l_-^2\!\!+l_+^2}, \gamma\defeq\atantwo(l_-,l_+).
        \eeq
      \item \ifthenelse{\boolean{\postpubfixes}}
            {
              If $b\neq\text{circle}$, determine the in-plane rotation of the boundary shape by setting
              \beg
                \vec{r}=\vec{r}([\uvec{x}_l'\ \uvec{y}_l'\ \uvec{z}_l])\text{ (log map)} \\
                \text{using } \uvec{x}_l'\!\defeq\!\uvec{x}_l\cos\theta\!+\!\uvec{y}_l\sin\theta\!, \nonumber \ \
                \uvec{y}_l'\!\defeq\!\uvec{z}_l\times\uvec{x}_l',\nonumber \\
                \theta\!\defeq\!(1\!/2)\atantwo(\beta,\alpha\!\!-\!\!\phi), \nonumber \ \
                [\uvec{x}_l\ \uvec{y}_l\ \uvec{z}_l] \defeq R([\vec{r}_{xy}^T 0]^T).
              \eeg
            }{
              If $b\neq\text{circle}$, using plane normal $\uvec{z}_l\defeq R([\vec{r}_{xy}^T 0]^T)\uvec{z}$ set 
              \beg
                \vec{r}=\vec{r}([\uvec{x}_l\ \uvec{y}_l\ \uvec{z}_l])\text{ (log map)}\\
                \theta\defeq(1\!/2)\atantwo(\beta,\alpha\!\!-\!\!\phi)\nonumber \\
                \uvec{x}_l\!\defeq\![\cos\theta\ \sin\theta\ 0]^T\!\!\nonumber \\
                \uvec{y}_l\!\defeq\!\uvec{z}_l\times\uvec{x}_l.\nonumber
              \eeg
            }
    \end{enumerate}
  \end{list}
\end{list}

\begin{figure*}[!h]
  \begin{center}
    \includegraphics{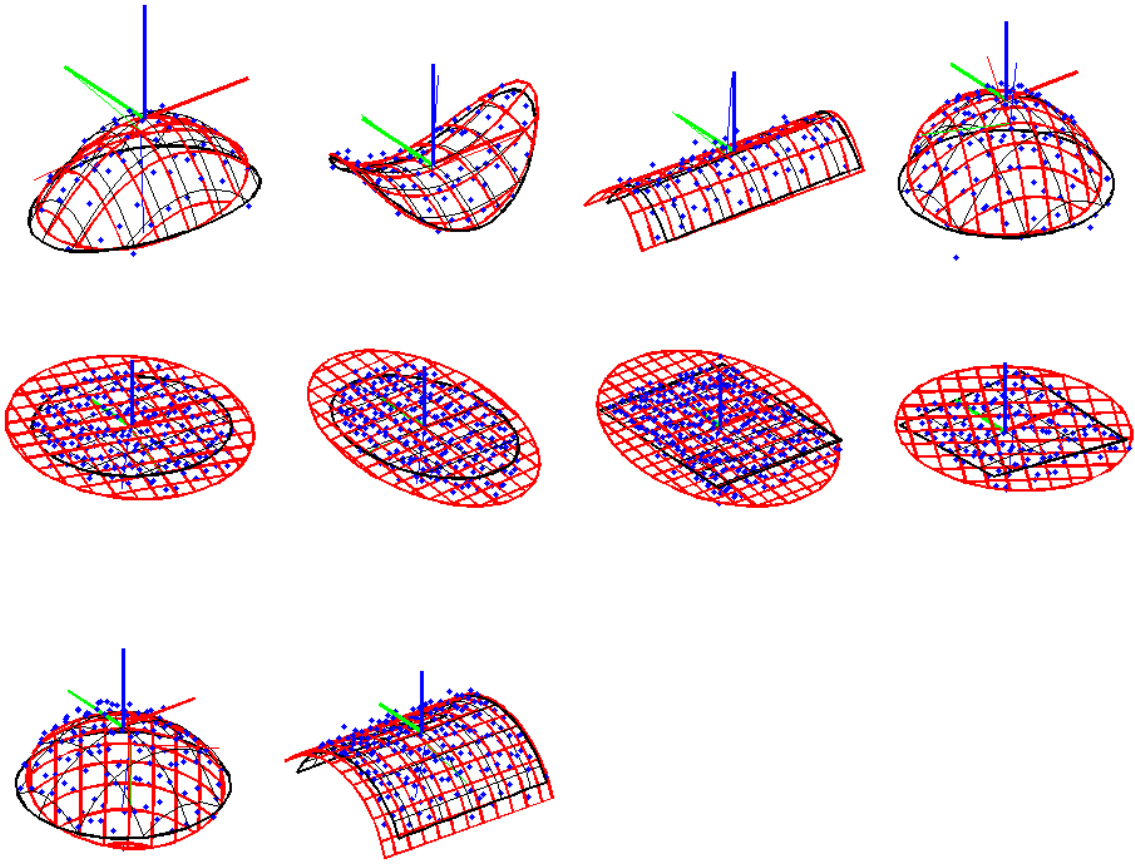}
  \end{center}
  \caption[Automatic fits for all patch types]{Automatic fits (red) for both paraboloid and non-paraboloid (lower right) patch types with requested elliptic boundary in simulated noisy range samples, using Kinect projection and error models.}
  \label{Fig:fitting_steps_all}
\end{figure*}

The fitting process for all types of curved bounded patch models is illustrated in Figure~\ref{Fig:fitting_steps_all}. %\subsection{Patch Uncertainty} \label{Sec:patch_error}
The covariance matrix of the patch parameters $\Sigma$ is calculated by first order error propagation through the above computations in each step (see Appendix~\ref{Sec:errprop}).

\subsection{The side-wall effect problem and flipping patch normals towards viewpoint} \label{patch_side-wall-viewpoint}
\begin{figure}[!b]
	\begin{center}
	\includegraphics{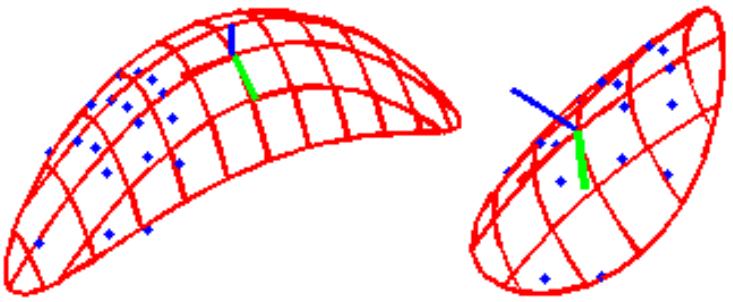}
	\end{center}
\caption[The ``side-wall'' effect reparameterization]{The reparameterization in Eq.~(\ref{Eq:sidewallconstr}) keeps the fitted paraboloid centered on the data (right).  This prevents the ``side-wall'' effect (left) and helps ensure good coverage, but can compromise the Euclidean residual.}
\label{Fig:sidewall}
\end{figure}

When the neighborhood points don't have a central symmetry then they may be unevenly distributed in the patch if left unconstrained.  We call this the \emph{side-wall effect} (Fig.~\ref{Fig:sidewall}).  To handle this, in \cite{KV13} we introduced a constrained fitting where the center of the patch $\vec{t}\in\R^3$ must lie on the line through the centroid $\vec{t}_p$ of the neighborhood parallel to the normal $\vec{\hat{n}}_p$ to an initial fit plane.  This is implemented as a reparameterization during the WLM incorporated in Step~\ref{fit-step-wlm}
\beq \label{Eq:sidewallconstr}
	\vec{t} = \vec{t}_p + a \vec{\hat{n}}_p
\eeq
where $a\in\R$ is the new patch parameter replacing $\vec{t}$.  Note that this constrained fitting affects the error propagation (see Appendix~\ref{Sec:errprop}).

The ``outward'' facing direction of the patch surface normal, which is the same as the local frame basis vector $\vec{\hat{z}}_{\ell}$, is ambiguous globally in the point cloud.  But considering that the the data were acquired from a single point of view $\vec{v}_p$ then the following equation should be satisfied:
\beg
  \vec{\hat{z}}_{\ell} \cdot (\vec{v}_p - \vec{t}) > 0
\eeg

If not we flip the patch by rotating it's local frame basis $\pi$ around $\vec{\hat{x}}_{\ell}$ and flipping its curvatures.
\subsection{Weighted Levenberg-Marquardt}
Levenberg-Marquardt (LM) is a standard iterative non-linear optimization~\cite{PTVF92}.  It can find a parameter assignment $\vec{p}_\text{opt}\in\R^p$ that locally minimizes the sum-of-squares residual $r$ of a differentiable objective function $f:\R^d\times\R^p\rightarrow\R$ applied to a dataset $\vec{q}_i\in\R^d$, $1\leq i\leq N$, starting from an initial estimate $\vec{p}_0$.  That is, it finds
\beq \label{Eq:lm}
  \vec{p}_\text{opt}=\argmin{\vec{p}\text{ near }\vec{p}_0}\ r,\ r\defeq\sum_{i=1}^N e_i^2,\ e_i\defeq f(\vec{q}_i,\vec{p}).
\eeq

Implementations typically take as inputs functions $f$ and $\partial f/\partial\vec{p}$, the data $\vec{q}_i$, and $\vec{p}_0$, and return both $\vec{p}_\text{opt}$ and a covariance matrix $\Sigma\in\R^{p\times p}$ representing its uncertainty.  A well known extension is to replace $e_i$ with $E_i\defeq e_i/\sigma_i$ where $\sigma_i>0$ are constant standard deviations modeling uncertainty in $e_i$.  The residual is then called $\chi^2$, and $\vec{p}_\vec{opt}$ locally maximizes the likelihood of the ``observations'' $e_i$.

For our use, $f$ is always the implicit form of a surface in world frame, i.e. $p_{wi}$, $s_{wi}$, or $c_{wi}$ (\ref{Eq:pwi},\ref{Eq:swi},\ref{Eq:cwi}).  The $\sigma_i$ are not constant, but can be estimated with with first order error propagation as\footnote{This assumes the $\Sigma_i$'s are positive-definite, which is a common requirement for covariance matrices.  A heuristic to allow \emph{semi}-definite $\Sigma_i$ is to clamp small $v_f(i,\vec{p})$ to a minimum positive limit.}

\beg
  \sigma_i=\sqrt{\var(f(\vec{q}_i,\vec{p}))}\defeq\sqrt{v_f(i,\vec{p})} \label{eq-sigma-i}\\
  v_f(i,\vec{p})\defeq\left(\frac{\partial f}{\partial\vec{q}}(\vec{q}_i,\vec{p})\right)\Sigma_i\left(\frac{\partial f}{\partial\vec{q}}(\vec{q}_i,\vec{p})\right)^T.\nonumber
\eeg

We define \emph{weighted} LM (WLM) to combine $\sigma_i$ and $f$ into a meta-objective function $F:[1\ldots n]\times\R^p\rightarrow\R$:
\beq \label{Eq:meta-obj}
  F(i,\vec{p})\defeq f(\vec{q}_i,\vec{p})/\sigma_i=f(\vec{q}_i,\vec{p})/\sqrt{v_f(i,\vec{p})}.
\eeq

Both $F$ and its gradient $\partial F/\partial \vec{p}$ are implied given $\vec{q}_i$, $\Sigma_i$, $f$, $\partial f/\partial\vec{p}$, $\partial f/\partial\vec{q}$, and $\partial^2f/\partial\vec{p}\partial\vec{q}$ (which is $d\times p$):
\beq \label{Eq:meta-obj-grad}
  \frac{\partial F}{\partial \vec{p}}(i,\!\vec{p})=
  \frac{\frac{\partial f}{\partial\vec{p}}(\vec{q}_i,\!\vec{p})}{\sigma_i}-
  e_i\frac{\frac{\partial f}{\partial \vec{q}}(\vec{q}_i,\!\vec{p})\Sigma_i\frac{\partial^2f}{\partial\vec{p}\partial\vec{q}}(\vec{q}_i,\!\vec{p})}{\sigma_i v_f(i,\!\vec{p})}.
\eeq

Given $\vec{q}_i$, $\Sigma_i$, $f$, $\partial f/\partial\vec{p}$, $\partial f/\partial\vec{q}$, and $\partial^2f/\partial\vec{p}\partial\vec{q}$, WLM synthesizes $F$ and $\partial F/\partial\vec{p}$ by (\ref{Eq:meta-obj},\ref{Eq:meta-obj-grad}) and then applies LM.  This is simplified further by the common implicit form of the world-frame surfaces (\ref{Eq:pwi},\ref{Eq:swi},\ref{Eq:cwi}), which are all variants of
\beg
  f_l(\vec{q}_l,\vec{k}_3)\defeq\vec{q}_l^TK\vec{q}_l-2\vec{q}_l^T\uvec{z} \label{Eq:fl}\\
  f_w(\vec{q}_w,\vec{p}_s)\!=\!f_l(X_r(\vec{q}_w,\vec{r},\vec{t}),\vec{k}_3) \label{Eq:fw}\\
  \vec{p}_s\!\defeq\![\vec{k}_3^T\ \vec{r}^T\ \vec{t}^T]^T,\ %
  \vec{k}_3\defeq[\kappa_x\ \kappa_y\ \kappa_z]^T,\ K\defeq\diag(\vec{k}_3)
  %,\ \vec{q}_l=X_r(\vec{q}_w,\vec{r},\vec{t})
  \nonumber
\eeg
where some components of $\vec{k}_3$, and for (\ref{Eq:swi}) the last component of $\vec{r}$, are held at zero.  The required derivatives of (\ref{Eq:fw}) are given by the chain rule from (\ref{Eq:dxr}) and derivatives of (\ref{Eq:fl})\ifthenelse{\boolean{\postpubfixes}}{
(using $R\defeq R(\vec{r})$):
\beg
  \frac{\partial f_w}{\partial\vec{q}_w}=\frac{\partial f_l}{\partial\vec{q}_l}\frac{\partial\vec{q}_l}{\partial\vec{q}_w},\ %
  \frac{\partial f_l}{\partial\vec{q}_l}=2(\vec{q}_l^TK\!\!-\!\!\uvec{z}^T),\ %
  \frac{\partial \vec{q}_l}{\partial\vec{q}_w}=R^T
\eeg

\beg
  \frac{\partial f_w}{\partial\vec{p}_s}=\left[\frac{\partial f_w}{\partial\vec{k}}\ \frac{\partial f_w}{\partial\vec{r}}\ \frac{\partial f_w}{\partial\vec{t}}\right],\ %
  \frac{\partial f_w}{\partial\vec{k}}=\vec{q}_l^T\diag(\vec{q}_l)
\eeg

\beg
  \begin{aligned}
    \frac{\partial f_w}{\partial\vec{r}}&=\frac{\partial f_l}{\partial\vec{q}_l}\frac{\partial\vec{q}_l}{\partial\vec{r}},\ \frac{\partial\vec{q}_l}{\partial\vec{r}}=\frac{\partial R^T}{\partial\vec{r}}(\vec{q}_w-\vec{t})\\
    \frac{\partial f_w}{\partial\vec{t}}&=\frac{\partial f_l}{\partial\vec{q}_l}\frac{\partial\vec{q}_l}{\partial\vec{t}},\ \frac{\partial\vec{q}_l}{\partial\vec{t}}=-R^T
  \end{aligned}\nonumber
\eeg
\beq
  \frac{\partial^2f_w}{\partial\vec{p}_s\partial\vec{q}_w} =
  \pard{}{\vec{p}_s}\left[\pard{f_w}{\vec{q}_w}\right]^T =
  \left[\pard{}{\vec{k}_3} \!\! \left[\pard{f_w}{\vec{q}_w}\right]^T \!\!\! \pard{}{\vec{r}} \!\! \left[\pard{f_w}{\vec{q}_w} \right]^T \!\!\! \pard{}{\vec{t}} \!\! \left[\pard{f_w}{\vec{q}_w} \right]^T
  \right]
\eeq
\beg
  \frac{\partial}{\partial\vec{k}_3}\!\!\left[\frac{\partial f_w}{\partial\vec{q}_w}\right]^T\!\!\!=2R\diag(\vec{q}_l),\ %
  \frac{\partial}{\partial\vec{t}}\!\!\left[\frac{\partial f_w}{\partial\vec{q}_w}\right]^T\!\!\!=-2RKR^T \nonumber\\
  \frac{\partial}{\partial\vec{r}}\!\!\left[\frac{\partial f_w}{\partial\vec{q}_w}\right]^T\!\!\!=2\frac{\partial R}{\partial\vec{r}}(K\vec{q}_l-\uvec{z})+2RK\frac{\partial R^T}{\partial\vec{r}}(\vec{q}_w-\vec{t}) \nonumber
\eeg
}{.}

\subsection{Experimental Results} \label{Sec:patch_fitting_er}
We tested the fitting algorithm both in real data from a Kinect viewing a rock and in simulation (Figure~\ref{Fig:fitting}).  For this initial experiment we implemented a simple interactive segmenter to manually select neighborhoods to fit patches in the 3D point cloud of the rock.  In Chapter~\ref{Ch:patch_mapping} we present algorithms for automatic neighborhood finding.  We used the two-parameter pointing/disparity stereo error model proposed by Murray and Little in~\cite{ML05} to estimate input sample covariances $\Sigma_i$ for all experiments (as described in Chapter~\ref{Ch:input}).

The results show that the algorithm can produce reasonable curved-surface patch models for local parts of non-flat environment surfaces.  Average times for our Matlab implementation are $\sim$20ms to fit $n\approx50$ sample points on a commodity workstation, while in a $C++$ implementation we reached $\sim$0.6ms to fit $n\approx50$ sample points.  SVD computations within LM are quadratic in $n$, though runtime also depends on the LM convergence rate.

\hfill

\textit{Note: In the rest of this thesis we will consider fitting only \textbf{paraboloid patches}, unless otherwise indicated.  Paraboloids are complete in that they form an approximation system for local regions on any surface with zero,one, or two nonzero local principal curvatures.}

\junk{
In~\cite{VK11,KV13} we introduced a set of bounded curved surface patch models and algorithms to fit and validate them to points sampling environment surfaces.  Here we use a subset of these, the paraboloids (Figure~\ref{Fig:patches}), which balance expressiveness with compactness of representation.\footnote{Modeling patches using paraboloids and other design choices are described in more detail in~\cite{VK11}.}
%\begin{figure}
%\begin{center}
%\includegraphics{patches/patches.pdf}
%\end{center}
%\caption{Paraboloid patch types and local coordinate frames.}
%\label{Fig:patches}
%\end{figure}
%We further simplify the patches used here by fixing their boundaries to circles (squares for cylindric paraboloids) with radius $r=10\mathrm{cm}$.  Elliptical or other boundary curves could also be used depending on the size and shape of the robot foot.
Various choices are possible with regard to the size and shape of the patch boundary curves, which can be adjusted in applications depending on the geometry of foot contact surfaces and the different contact configurations a particular footfall selection algorithm might consider.  Here we use a generic approach with rectangle bounds for cylindric paraboloids and ellipse (possibly degenerating to circular) bounds otherwise, and we fit the bounds to probabilistically contain 95\% of the points in a spherical foot-scaled neighborhood with $r=10\mathrm{cm}$~\cite{VK11}.

Each patch has two intrinsic parameters, the principal curvatures $\kappa_{x,y}$, and six extrinsic parameters $\vec{r}\in\R^3, \vec{t}\in\R^3$ giving the spatial pose of a local coordinate frame attached to the patch.  Eq.~\ref{eq:patch} gives an implicit form for the patch surface in local frame where $\vec{p_l}\in\R^3$ is a point on the patch, see~\cite{VK11} for details.
\beg \label{eq:patch}
	\vec{p_l}^T\diag(\kappa_x, \kappa_y, 0)\vec{p_l} - 2 \vec{p_l}^T [0~0~1]^T = 0
\eeg
Using 4-byte floats for the parameters requires 32 bytes per patch,\footnote{The boundary parameters are fixed here.} highlighting the compression of information from a dense $640\times480$ depth image (about 0.4MB with the Kinect's 11 bit depth resolution) to a corresponding sparse representation using e.g. 100 patches (about 3kB).}

\section{Patch Validation}\label{Sec:validation}
\begin{figure*}[htb]
  \begin{center}
    \includegraphics{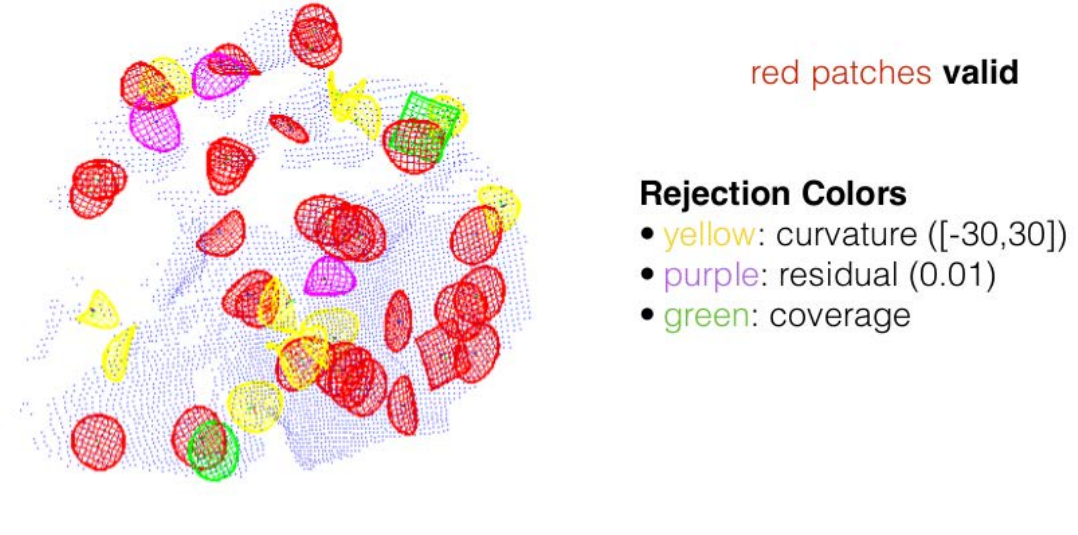}
  \end{center}
  \caption[Residual, coverage, and curvature patch validation]{Validation for 30 fitted paraboloids with respect to residual, coverage, and curvature.}
  \label{Fig:validation}
\end{figure*}

After fitting a patch to a set of point cloud data it is important to evaluate it, because it may fit the data but still not faithfully represent the surface.  In \cite{KV13} we introduced three measures based on the \emph{residual}, \emph{coverage}, and \emph{curvature}. Residual and curvature evaluate the surface shape, while coverage evaluates the boundary of the patch. (see Figure~\ref{Fig:validation}).

\subsection{Residual Evaluation} \label{Sec:patch_residual}
\begin{figure*}[htb]
  \begin{center}
    \includegraphics{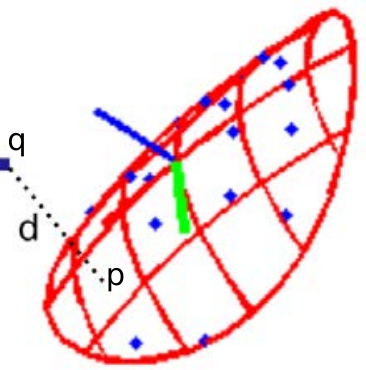}
  \end{center}
  \caption[Bad residual due to an outlier sample point]{Bad residual due to an outlier sample point $\vec{q}$ which is at distance $d = \|\vec{q}- \vec{p}\|$ from the patch.}
  \label{Fig:res_eval}
\end{figure*}

The patch residual measures the deviation between the sample points and the (unbounded) patch surface.  The residual can be bad either due to outliers (see Figure~\ref{Fig:res_eval}) or due to local minima in the WLM process.  We use the root-mean-square error (\emph{RMSE}) Euclidean residual $\varrho$ between the sample points $\vec{q}_i$ and their corresponding closest points $\vec{p}_i$ on the patch\footnote{Here we consider both $\vec{q}_i$ and $\vec{p}_i$ to be expressed in the patch local frame $L$.} (which must be calculated for each $q_i$)
\beq
	\varrho = RMSE(\{\vec{q}\},\{\vec{p}\}) = 
  \sqrt{\frac{\sum_{i=1}^N{\|\vec{q}_i- \vec{p}_i\|^2}}{N}}.\label{Eq:rmse}
\eeq

Whereas the patch fitting algorithm uses an algebraic residual for speed, $\varrho$ is a Euclidean residual and gives its result in the same physical units as the input data (e.g. meters)~\cite{PKPSTV93}, enabling it to be compared to a meaningful threshold.  However, calculating the $\vec{p}_i$ for each $\vec{q}_i$ can be computationally expensive.  We use a technique based on Lagrange multipliers~\cite{Eberly1999}.

When $\kappa_x\approx\kappa_y\approx0$ the paraboloid surface was fitted as a plane, so $\vec{p}_i = (I-\uvec{z}\uvec{z}^T)\vec{q}_i$, i.e. $\vec{p}_i$ is the projection of $\vec{q}_i$ onto the $xy$ plane of $L$.  Otherwise $\vec{p}_i$ is characterized as:
\beq
	\min_{\vec{p}_i\text{ satisfying~(\ref{Eq:fl})}}{\|\vec{q}_i - \vec{p}_i\|}.\label{Eq:min_p}
\eeq
Define a Lagrange function $\Lambda$ as
\beq
\Lambda(\vec{p}_i,\lambda) = (\vec{q}_i-\vec{p}_i)^T (\vec{q}_i-\vec{p}_i)
						+ \lambda (\vec{p}_i^T K \vec{p}_i - 2 \vec{p}_i^T\uvec{z}).
\eeq
with Lagrange gradient constraints
\beq \label{Eq:Lagrange}
	\nabla \Lambda(\vec{p}_i,\lambda) = \vecmb{0}^T
	\Leftrightarrow
		{\partial{\Lambda}}/{\vec{p}_i} = [0~0~0] \text{ and }
		{\partial{\Lambda}}/{\lambda} = 0.
\eeq
Expand the first gradient constraint from~(\ref{Eq:Lagrange})
\bea
  -2\vec{q}_i^T\!\!+\!2\vec{p}_i^T\!+\!\lambda(2\vec{p}_i^TK\!-\!2\uvec{z}^T) &= [0~0~0]\nonumber\\
  -\vec{q}_i\!+\!\vec{p}_i\!+\!\lambda(K\vec{p}_i\!-\!\uvec{z}) &= [0~0~0]^T\nonumber\\
  (I+\lambda K)\vec{p}_i &= \vec{q}_i+\lambda\uvec{z}\nonumber\\
  \vec{p}_i &= (I\!+\!\lambda K)^{-1}(\vec{q}_i\!+\!\lambda\uvec{z}) \label{Eq:p_i}
\eea
and substitute\footnote{The inverse of the diagonal matrix in~(\ref{Eq:p_i}) is evaluated symbolically and then denominators are cleared after substitution in~(\ref{Eq:fl}), avoiding any issue of non-invertability or division by zero.} for $\vec{p}_i$ in the second gradient constraint, which is the same as~(\ref{Eq:fl}).  

This leads to a fifth degree polynomial in $\lambda$, for which there is at least one real solution because imaginary solutions come in pairs. To solve the polynomial, we can either compute the eigenvalues of the companion matrix or we use Newton iteration.  For Newton's method an initial root guess for $\lambda$ (and thus for $\vec{p}_i$) is required.  We pick as $\vec{p}_i$ the point projected from $\vec{q}_i$ along the $z_{\ell}$-axis (local frame).  Newton's method appears to be $\sim\!50$ times faster than Eigendecomposition in tests with around $\sim\!1000$ sample points per patch.

Finally, backsubstitute\footnote{Division by zero can occur during this backsubstitution when $\vec{q}_i$ is on a symmetry plane or axis, but alternate forms can be used in those cases.} the real solution(s) in~(\ref{Eq:p_i}) and find the minimum as in~(\ref{Eq:min_p}). %TBD what about cases where \vec{q}_i is on a symmetry axis of the paraboloid such that there are infinite solutions?

\subsubsection*{Residual Approximations} \label{Sec:res_approx}
\begin{figure*}[h]
  \begin{center}
    \includegraphics{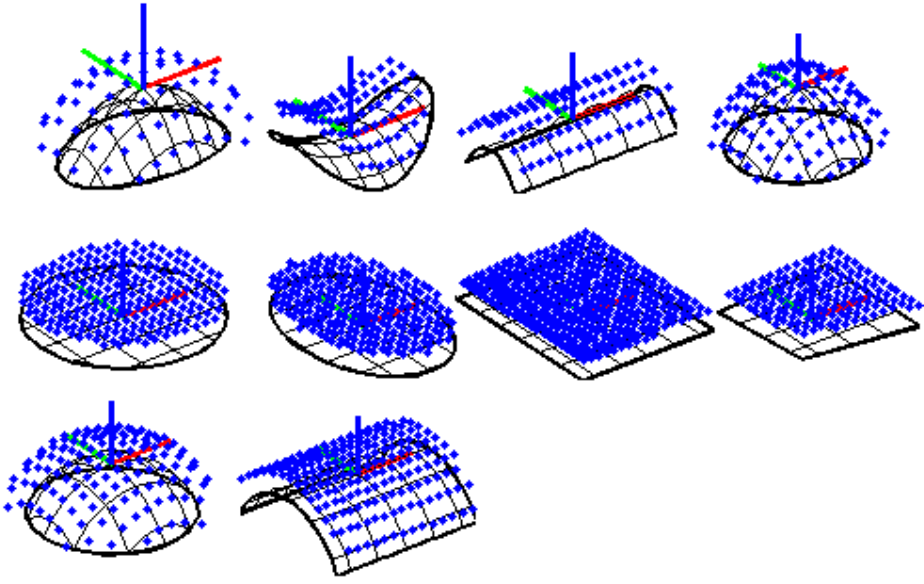}
  \end{center}
  \caption[Perturbed sample points for each patch type.]{Perturbed sample points in the direction of the surface normal for each patch type.}
  \label{Fig:res_eval_approx}
\end{figure*}

The problem of the Euclidean residual estimation is well studied, both for exact and approximate solutions; for instance Taubin's first and second order approximations \cite{Taubin91, Taubin93}, the 3L algorithm \cite{BLCC00}, the constrained minimization method \cite{ARCW02}, and MinMax \cite{HBM04} have been proposed.

We implemented three approximations.  The simplest approximation is to consider the vertical distance in the local $z_{\hat{\ell}}$-axis of the patch.  The other two are the first and second order Taubin approximations introduced in \cite{Taubin93} (see Appendix~\ref{Sec:Taubin}).  We tested these approximations in simulated data.  For each patch type we sampled a set of point data (a range of 60-200 points per patch) and we perturbed them in the direction of the surface normal at each point (Figure~\ref{Fig:res_eval_approx}).  All the approximations are $\sim\!100$ times faster than the exact solution of solving the fifth degree polynomial.  Both Taubin's approximations are very close to the exact solution, compared to the vertical distance one that does not give good results at all.

\subsubsection*{Residual Thresholds}
\begin{figure}[h]
  \begin{center}
	  \includegraphics{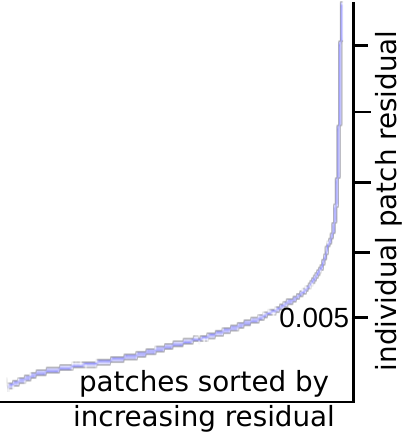}
  \end{center}
  \caption[Sorted residuals for 1000 random patches.]{Sorted residuals for 1000 random patches (see text), approximately $95\%$ of which are below $0.01$.}
  \label{Fig:res_eval_diag}
\end{figure}

To determine the residual threshold $T_r$ such that any patch with $\varrho > T_r$ will be dropped, we sorted all residuals (Figure~\ref{Fig:res_eval_diag}) for a sampling of 1000 random patches ($r=0.1$m, k-d tree neighborhoods), 100 on each of 10 rock datasets~\cite{KV13}.  The value $T_r=0.01$m was selected to include approximately $95\%$ of the patches.  In general $T_r$ can be set in an application dependent way.  Furthermore, the choice of RMSE residual is not essential.  For example, an alternate residual
\beq
  \varrho_{alt} = \max\|\vec{q}_i-\vec{p}_i\|
\eeq
could be used to check if any small surface bumps protrude more than a desired amount from the surface.

\subsection{Coverage Evaluation} \label{Sec:patch_coverage}
\begin{figure*}[!htb]
  \begin{center}
    \includegraphics{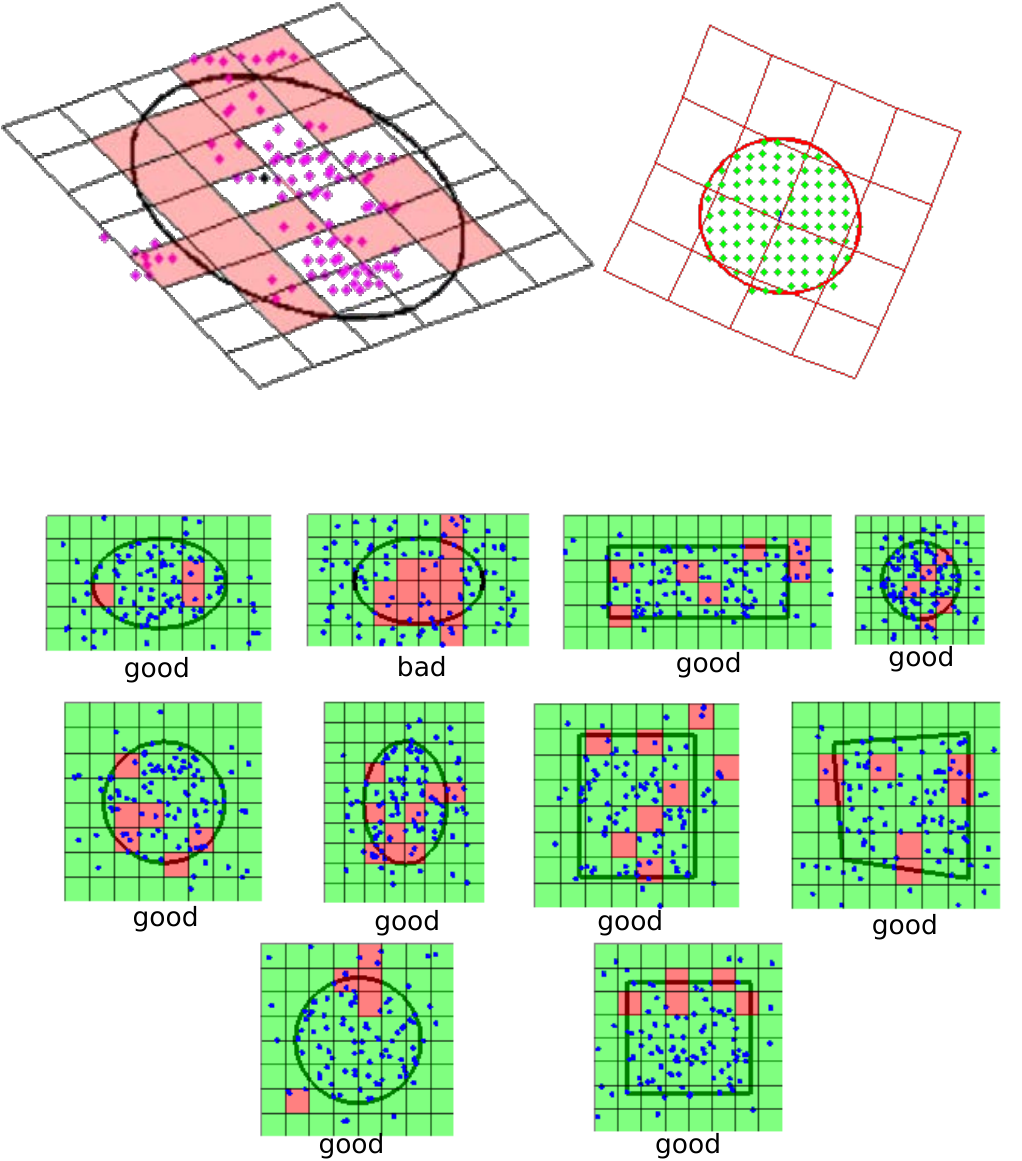}
  \end{center}
  \caption[Coverage evaluation for all types of patch models.]{\textbf{Top:} Coverage evaluation for a bad patch (left) where samples are not uniformly distributed into the boundary with bad cells in red and a good patch (right) with uniformly distributed sample points.  Thresholds: $\zeta_i = 0.9$, $\zeta_o = 0.2$, and $T_p = 0.3 N_p$.  \textbf{Bottom:} Coverage evaluation for all types of patch models.  Each patch has 100 points randomly distributed around them.  The boundaries have been split into 50 cells per patch.  A good cell (green) inside the boundary should have at least 5 points and a bad patch should have less than 80\% good cells. The patch coverage evaluation took 4ms/patch in Matlab.}
  \label{Fig:cov_eval}
\end{figure*}

A different evaluation is needed to take into account the patch boundary.  A patch may fit the data but still not faithfully represent the neighborhood, either because too many sample points are outside the patch boundary or there is too much area inside the boundary that is not supported by data points.  (Unlike~\cite{KKN2009}, we opt not to speculatively fill holes in the data.)

To detect these cases we generate an axis-aligned grid of fixed pitch $w_c$ on the $xy$ plane of the patch local frame $L$.  We generate only the required number of rows and columns in this grid to fit the projection of the patch boundary.

Define $I_c$ and $O_c$ to be the number of data points whose $xy$ projections are both inside a given cell $c$ and respectively inside or outside the projected patch boundary.  Define $A_i$ to be the area of the geometric intersection of the cell and the projected patch boundary, which will be detailed below.  The cell is considered \emph{bad} iff
\beq
\label{Eq:cov}
	I_c < \frac{A_i}{w_c^2} T_i \text{ or } O_c > (1-\frac{A_i}{w_c^2})T_o.
\eeq
for thresholds $T_i$ and $T_o$.  Here we fix these thresholds relative to the expected number of samples $N_e$ in a given cell if all samples were in-bounds and evenly distributed:
\beq
  T_i = \zeta_i N_e,\ \ T_o = \zeta_o N_e,\ \ N_e\defeq k/N_p,\ \ N_p\defeq\frac{A_p}{w_c^2}, 
\eeq
where $k$ is the number of sample points in the neighborhood and $A_p$ is the area of the patch approximated as the area inside the projected boundary.

The patch fails coverage evaluation iff there are more than $T_p$ bad cells.  After some experiments in fitting paraboloid patches with neighborhood radius $r=0.1$m, we set $w_c = 0.01$m, $\zeta_i = 0.8$, $\zeta_o = 0.2$, and $T_p = 0.3 N_p$.  Figure~\ref{Fig:cov_eval} illustrates patches that pass and fail coverage evaluation.

\begin{figure}
\begin{center}	%\includegraphics[width=0.25\textwidth]{intersection-cropped-ur.png}	%\includegraphics[width=0.17\textwidth]{intersection_rect-rect-cropped-ur.png}
\includegraphics[width=\textwidth]{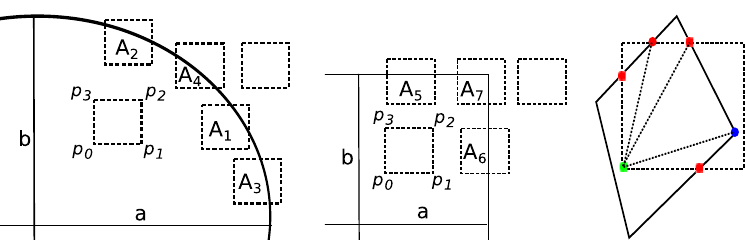}
\end{center}
\caption[Cell intersection cases for all types of boundaries.]{Left: the six possible placements of a cell for the top right quadrant of an elliptic boundary.  Middle: similar for the axis-aligned rectangular boundary.  Right: example of an intersection between a general convex quadrilateral boundary and a grid cell.}
\label{Fig:intersection}
\end{figure}

\subsubsection*{Intersection Area for Ellipse and Circle Boundaries} For an ellipse boundary with radii $a,b$, or for the degenerate case of a circle boundary with radius $r=a=b$, we compute the intersection area with a secant approximation since the exact computation involves a relatively expensive inverse trig function.  Wlog we describe only the top right quadrant (Fig.~\ref{Fig:intersection}, left); the other three are symmetric.  Let $\vec{p}_{0\ldots3}$ be the four corners of a grid cell in counter-clockwise order starting from the lower left.  The algorithm for computing the intersection area is:
\begin{enumerate}
	\item If $\vec{p}_2$ is inside the ellipse then $A_i = w_c^2$
	\item else if $\vec{p}_0$ is not inside the ellipse then $A_i = 0$
	\item else if $\vec{p}_1$ is inside the ellipse then\\
		    if $\vec{p}_3$ is inside the ellipse then $A_i = A_1$ else $A_i = A_2$
	\item else if $\vec{p}_3$ is inside the ellipse then $A_i = A_3$
	\item else $A_i = A_4$.
\end{enumerate}
%with
\bea
	& \begin{aligned}
    A_1 =\ &(x_b-x_0)w_c+(x_c-x_b)(Y(x_b)-y_0)+\\
          &((x_c-x_b)(y_0+w_c-Y(x_b)))/2
    \end{aligned}\\
	& \begin{aligned}
    A_2 =\ &(x_c-x_0)(Y(x_c)-y_0)+\\
          &(x_c-x_0)(Y(x_0)-Y(x_c))/2
    \end{aligned}\\
	& \begin{aligned}
    A_3 =\ &(y_c-y_0)(X(y_c)-x_0)+\\
          &(y_c-y_0)(X(y_0)-X(y_c))/2
    \end{aligned}\\
	& A_4 = (X(y_0)-x_0)(Y(x_0)-y_0)/2
\eea
\beg
	X(y) \defeq a \sqrt{1-y^2/b^2},\ Y(x) \defeq b \sqrt{1-x^2/a^2}\nonumber\\
  x_b \defeq X(y_0+w_c),\ x_c \defeq x_0 + w_c,\ y_c \defeq y_0 + w_c\nonumber\\
	[x_0,y_0]^T\defeq \vec{p}_0\nonumber
\eeg

\subsubsection*{Intersection Area for Axis-Aligned Rectangle Boundary} As above we consider only the top right quadrant (Fig~\ref{Fig:intersection}, middle).  Let the rectangle half-lengths be $a,b$, and define $[x_0,y_0]^T\defeq\vec{p}_0$.  The exact intersection area can be computed as follows:
\begin{enumerate}
	\item If $\vec{p}_2$ is inside the rect then $A_i = w_c^2$
	\item else if $\vec{p}_0$ is not inside the rect then $A_i = 0$
	\item else if $\vec{p}_1$ is inside the rect then $A_i=A_5=w_c(b-y_0)$
	\item else if $\vec{p}_3$ is inside the rect then $A_i=A_6=w_c(a-x_0)$ 
	\item else $A_i=A_7=(a-x_0)(b-y_0)$.
\end{enumerate}

\subsubsection*{Intersection Area for Convex Quadrilateral Boundary} To handle the case of a general convex quadrilateral (Fig.~\ref{Fig:intersection}, right), we use the fact that the intersection between a convex quad and a rectangle is always convex:
\begin{enumerate}
	\item Find the set of grid cell corner points that are inside the quad and vice-versa.
	\item Find the intersection points between the grid cell boundaries and the convex quad boundaries.
	\item Discard all points from steps 1 and 2 except those that lie in or on both figures.
	\item Sort all the points computed in the previous steps in counterclockwise order and connect the first point with each of the others in order, forming a triangle fan.  $A_i$ is the sum of the triangle areas.
\end{enumerate}

%\begin{figure}
%\begin{center}
%	%\includegraphics[width=0.15\textwidth]{intersection_convex-rect-cropped.png}
%	\includegraphics[width=0.2\textwidth]{SparsePatchExtraction/intersect_cq.pdf}
%\end{center}
%\caption{Intersection area computation between a rectangle and a convex quad. Point P1 (green) is rectangle's corner which is inside the convex quad, point P3 (blue) is convex quad's corner which is inside the rectangle. The rest of the points (red) are the intersection points. We pick P1 and we form the triangles $\{t1, t2, t3, t4\}$ in counterclockwise order. The intersection area is the sum of the triangle areas.}
%\label{Fig:inters-convex}
%\end{figure}

\subsection{Curvature Evaluation}
\begin{figure}[h]
  \begin{center}
    \includegraphics{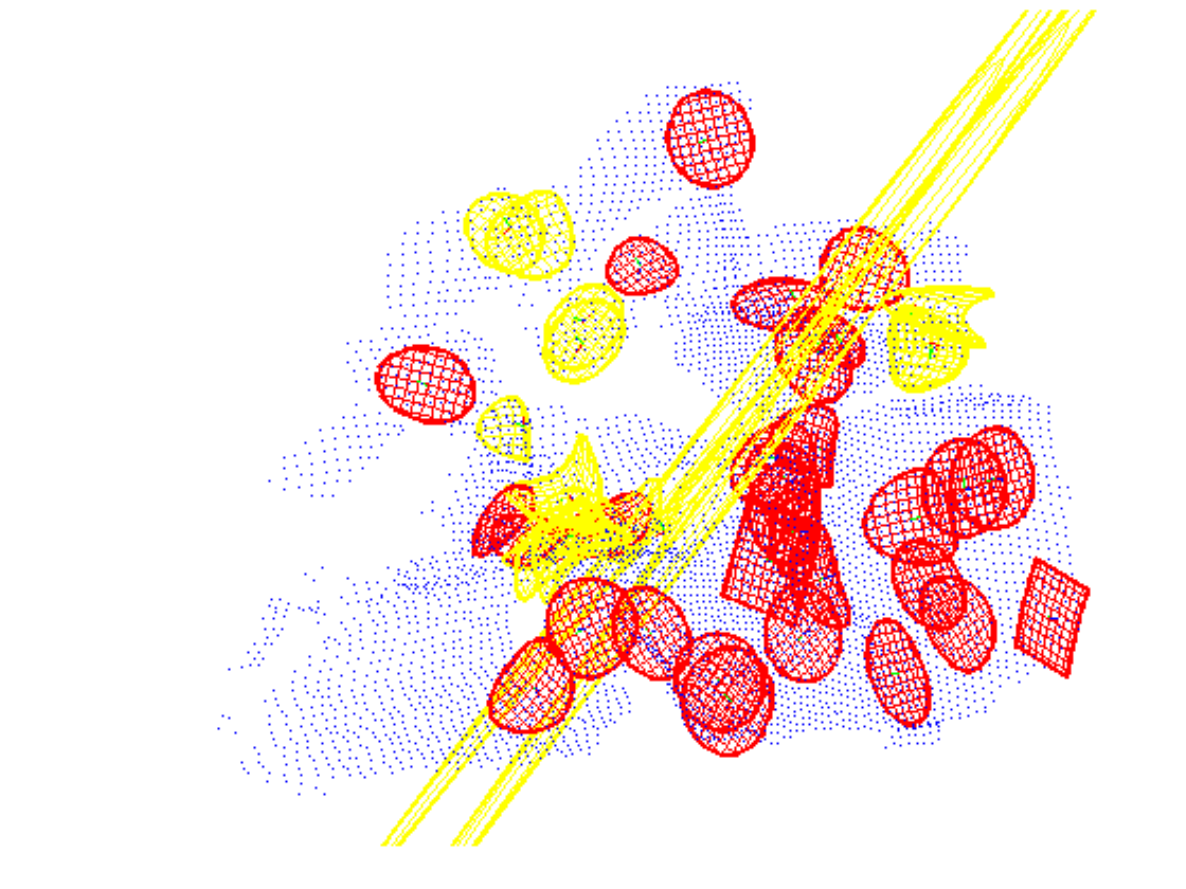}
  \end{center}
\caption[Curvature validation example]{In red are patches that pass all validations and in yellow patches that pass the residual and coverage validation but fail curvature validation with curvatures out of the [-30,30] range threshold.}
\label{Fig:overall_eval}
\end{figure}

Residual and coverage evaluation may still not be enough.  There may be cases where both residual and coverage checking passes, but the bounded patch does not represent the data correctly.  This may happen either when the point cloud data form a very curved surface or when the the LM non-linear fitting gets stuck in local minima as appears in Figure~\ref{Fig:overall_eval} in yellow.  A patch fails curvature evaluation iff its minimum curvature is smaller than a threshold $\kappa_{min,t}$ or its maximum curvature is bigger than a threshold $\kappa_{max,t}$.  We set this threshold experimentally to $\kappa_{min,t} = -1.5 max(\vec{d})$ and $\kappa_{max,t} = 1.5 max(\vec{d})$, where $\vec{d}$ is the patch boundary vector.

\smallskip
More fitting and validation experimental results are presented in Section~\ref{Sec:map_exp}.

\section{Related Work}
\subsection*{Modeling}
Modeling the environment and detecting potential contact surface areas around a robot using exteroceptive sensing is a common task, but still very challenging, especially for locomotion in uncertain environments.  The approach explored here contrasts with the traditional study of \emph{range image segmentation}, which also has a significant history~\cite{HBJFBGBEFF96}, where a partition of non-overlapping but potentially irregularly bounded regions is generated producing a dense labeling of the whole image.  In image segmentation, some work has been done with curved surfaces \cite{PBJB98}, but the main focus still appears to be on planes \cite{JB94, WS06, GFF08, PVB09}.

Many prior systems typically use \emph{dense} approaches in that they attempt to model all of the terrain in view.  Some are grid based~\cite{BHKKMSW89}, like those on Ambler \cite{KS96}, Dante II \cite{BW99}, and Mars Exploration Rovers (MER) \cite{MLB07} using laser scanners or stereo cameras to build an elevation map, find obstacles, and quantify traversability.  Usually these don't model detailed 3D contact features, though some attempt to recover surface models~\cite{PMPKRB09}.  A few other works do take a sparse approach but are restricted to planar patches, ranging from large flats in man-made environments \cite{WS06, GFF08, PVB09} down to small ``patchlets'' \cite{ML05}.

We proposed to only map a \emph{sparse} set of patches.  Also, our patch-based approach can homogeneously model contact surfaces both in the environment and on the robot itself, whereas most prior work considers modeling environment surfaces exclusively.  Irregularly bounded regions, which may be very large or non-convex, can present a challenge for higher-level contact planning algorithms which still need to search for specific contact areas within each region.  One aim of using regularly bounded regions of approximately the same size as relevant contact features on the robot is to trade potentially complex continuous searches \emph{within} patches for a discrete search \emph{across} patches.  Fewer and larger paraboloid patches can give a comparable fidelity of representation as the many small planar patches needed to cover a curved environment surface \cite{VBPS10}.  Of course other parts of the robot may also make contact in unintended ways. The patch model could help plan intentional contacts while other data structures are simultaneously used for general collision prediction \cite{SEMMEK08, PSBLY12, BVKEK13, HMB13}.

One challenge in modeling is dealing with missing data.  In~\cite{KKN09} texture synthesis was presented to deal with the problem of occluded terrain by filling in the missing portions.  Our approach avoids representing such missing areas where uncertainty is high.  As we described in Chapter~\ref{Ch:patch_tracking} we instead integrate multiple range scans taken from different perspectives (as the robot moves) to fill in missing areas with new observations using a volumetric fusion approach~\cite{NIHMKDKSHF11,RV12}.

%\cite{TX08}

\subsection*{Fitting}
One of our main results \cite{VK11} is an algorithm to fit curved, bounded patches to noisy point samples.  Fitting planes is well studied~\cite{WTHNY01}, including uncertainty~\cite{Kanatani05} and fitting heteroskedastic range data~\cite{PVB09}.  For curved surfaces quadrics are a natural option; Petitjean~\cite{Petitjean02} surveyed quadric fitting, but there were few results that (a) quantified uncertainty, (b) recovered geometric parameterizations, and (c) fit bounded patches.  In~\cite{DNC07}, Dai, Newman, and Cao describe recovery of paraboloid geometric parameters\footnote{They \emph{verify} that the fit result is a paraboloid and extract its parameters.  They do not consider \emph{constraining} the fit to paraboloids (vs other quadrics).} by linear least squares, without considering uncertainty.  In~\cite{WHJZL03} Wang, Houkes, Jia, Zhenga, and Li studied quadric extraction in the context of range image segmentation, including quantified uncertainty in the algebraic (not geometric) patch parameters, but not on the input points, while in \cite{WF90} superquadrics are fit using Levenberg-Marquardt considering variance in the range data.  Our fitting algorithm quantifies both input and output uncertainty and recovers geometric parameters of bounded patches. 

\section{Summary and Future Work}
We introduced a set of 10 particular bounded curved-surface patch types and algorithms to fit and validate patches to noisy point samples of a surface for sparsely representing contact surfaces in the environment near a robot, and also on the robot itself.  Paraboloid (including planar) patches can model portions of natural surfaces such as rocks; planar, cylindrical, and spherical patches can model common types of man-made surfaces as well as portions of the robot feet and hands.  The presented patch models all have minimal geometric parameterizations and quantified uncertainty in the form of covariance matrices.  Though surface modeling and surface fitting have been studied extensively, many prior works ignore uncertainty, are limited to planes, are dense vs sparse, and/or are concerned only with surfaces in the environment.  We address all of these issues to some extent and we demonstrated the effectiveness of our approach both in real data and in realistic simulation.

Fast patch fitting is a key aspect of the proposed method.  The presented algorithms are efficient enough for our application in bipedal locomotion (Section~\ref{Sec:patch_fitting_er}, Section~\ref{Sec:map_exp}, and Chapter~\ref{Ch:biped_exp}), but it is of great interest to check weather a fast normal vector and/or principal curvature extraction (e.g. using Integral Images~\cite{HRDGN12}) can replace the initial parameter estimation during WLM and reduce the number of iterations.  Moreover, a re-fitting process could be developed where an initially fitted patch (coming for instance from a previous frame) is refit to new data.  Validation and visualization of the covariance matrices of the patch parameters calculated by uncertainty propagation (Appendix~\ref{Sec:errprop}) is another possible next step for this work.  Finally, an extension to higher degree polynomials could be added for representing more irregular areas in the environment, while preserving geometrically meaningful minimal parametrization and quantified uncertainty.
\cleardoublepage

\part{Curved Patch Mapping \& Tracking}
%************************************************
\chapter{Patch Mapping} \label{Ch:patch_mapping}
%************************************************

\newcounter{stage}
\newcounter{step}

\begin{figure*}[h]
	\begin{center}
	\includegraphics[width=\textwidth]{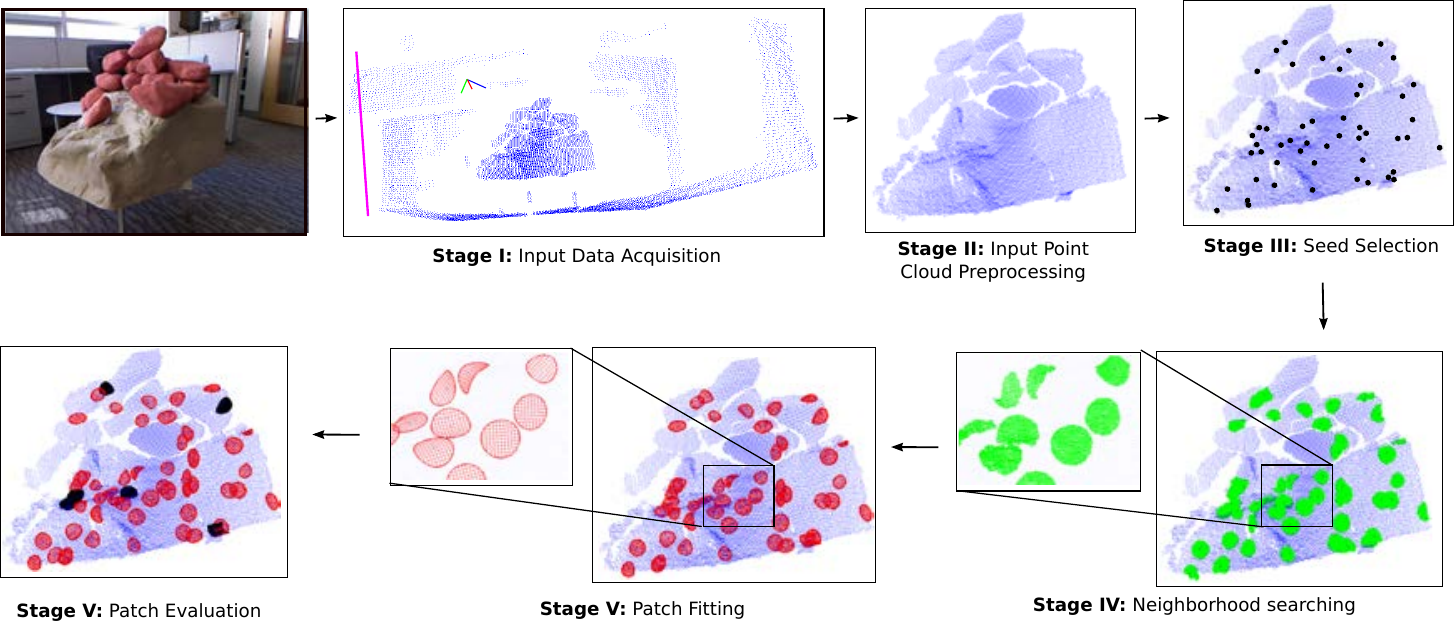}
	\end{center}
\caption[Patch Mapping system overview.]{Patch Mapping system overview, showing the main stages in the algorithm.  \textbf{Stage I:} the 3D point cloud data (blue points) and the gravity vector (pink) from the range and IMU sensors; \textbf{Stage II:} point cloud preprocessing including background removal; \textbf{Stage III:} 50 uniformly random seed points; \textbf{Stage IV:} $0.05m$ neighborhoods of each seed point; \textbf{Stage V:} patch fitting and validation for each neighborhood.}
\label{Fig:Segmentation}
\end{figure*}

Having introduced in Chapter~\ref{Ch:EnvRep} a new surface model for contact patches between a robot and local areas in the environment and algorithms to fit and validate patches to 3D point cloud data, algorithms to find potentially useful patches and spatially map them relative to the robot are now presented.  Patches are sparsely fit using the following five-stage approach\footnote{Though we describe each stage as a batch operation, in practice the implementation of Stages III to V is ``pipelines'' so that patches can be added to the map incrementally until a time or space limit is reached.} (Figure~\ref{Fig:Segmentation}):

\begin{list}{\bf Stage \Roman{stage}:}{\usecounter{stage}}
  \item Acquire Input Data from a Depth Camera and IMU (Section~\ref{Sec:map_input}).
  \item Preprocess the Input Point Cloud (Section~\ref{Sec:map_preprocess}).
	\item Select Seed Points on the Surface (Section~\ref{Sec:map_seeds}).
	\item Find Neighborhoods of Seed Points (Section~\ref{Sec:map_neighborhood}).
	\item Fit \& Validate Curved Bounded Patches to the Neighborhoods (Section~\ref{Sec:map_patch_model_fit}).
\end{list}
These functions dovetail with the patch tracking algorithms in Chapter~\ref{Ch:patch_tracking} to maintain a spatially and temporally coherent map (Section~\ref{Sec:spatial_map} of up to hundereds of nearby patches as the robot moves through the environment.

After acquiring RGB-D and IMU data from the sensors (\textbf{Stage I}), preprocessing (\textbf{Stage II}), like background removal, decimation, or saliency filtering, can be applied depending on the application.  Seed points (\textbf{Stage III}), and neighborhoods (\textbf{Stage IV}) are found, and finally patches are fit and validated to the neighborhoods (\textbf{Stage V}).  The neighborhood size $r$ is set to a fixed value derived from the size of the intended contact surface on the robot\footnote{In this thesis we mainly consider foot placement as an example contact task.  Thus we consider neighborhood sizes that are slightly bigger than the size of the robot's foot (e.g. 5--10cm for a mini-humanoid).}\multiplefootnoteseparator\footnote{The patch model could help plan intentional contacts while other data structures are simultaneously used for general collision prediction.}.  Using this algorithm a spatial patch map is defined in Section~\ref{Sec:spatial_map}.

Before describing the details of patch mapping we first introduce the notion of \emph{local volumetric workspace} (or simply the \emph{volume}), which will be extensively used from now on.

\subsection*{Local Volumetric Workspace} \label{Sec:Volume}
\begin{figure*}[h]
	\begin{center}
	\includegraphics[width=\textwidth]{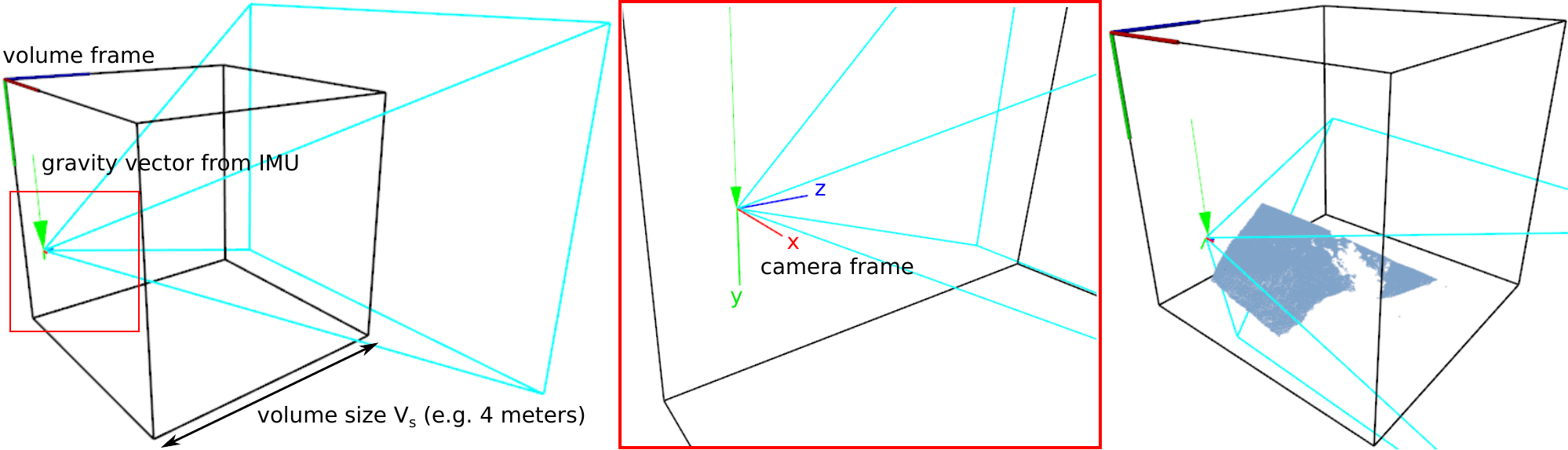}
	\end{center}
\caption[Local Volumetric Workspace]{A 4m cubic volume at initial pose (left), a zoom-in to the camera frame (middle), and a 4m cubic volume when camera acquires a point cloud (right).}
\label{Fig:map_volume}
\end{figure*}

When a robot moves in the environment, it constantly acquires new 3D point clouds and IMU data frames, typically at about $30Hz$ for the former and $100Hz$ or more for the latter.  Keeping all this information (even after fusion) to remove redundancies creates a huge amount of data over time, affecting both the performance of any downstream algorithm applied to them and memory requirements\footnote{An alternative is to keep only the information of the most recent frame, but cases like locomotion where the terrain under the robot's feet is required and the camera is not facing down (because its view would be obstructed by the legs and feet), requires data fusion.}.  Moreover, in many tasks, such as a biped robot locomoting on a rough terrain, the robot only needs to know an area around it for local 3D contact planning.  Thus, it is natural to consider only the (potentially fused) data in a moving volume around the robot.  Though there are several potentially useful definitions for such a volume, here we define it as a cube with a \emph{volume coordinate frame} at a top corner (Figure~\ref{Fig:map_volume}).  The $y$ axis of the volume frame points down (and may be aligned to the gravity vector derived from the IMU data) and the $x$ and $z$ axes point along the cube edges forming a right-handed frame.

We use the cubic volume model and this definition of the volume frame so that our \emph{local volumetric workspace} is the same as the TSDF\footnote{Truncated Signed Distance Function} volume in moving volume KinectFusion~\cite{RV12} that we will use in Chapter~\ref{Ch:patch_tracking}.  At any time $t$ the volume is fully described by: 1) its \textbf{size} $V_s$ (a constant), and 2) its \textbf{pose} relative to the camera with the following $4\times4$ rigid body transformation
  \beg \label{Eq:Vpose}
    C_t = \begin{bmatrix} R_t & \vec{t}_t\\
                          0   & 1
          \end{bmatrix}
  \eeg
where $R_t$ is the rotation matrix and $\vec{t}_t$ the translation vector that transforms from the camera to the volume frame at time $t$.

The volume pose relative to the environment may change as the robot moves around using one of the following policies:
\begin{enumerate} \label{vol_policies}
  \item \textbf{fv} (fixed volume): The volume remains fixed in the physical world.
  \item \textbf{fc} (fixed camera): Holds the camera pose fixed relative to the volume by applying 3D rigid transformations to the volume pose when the camera has moved beyond a distance $c_d$ or angle $c_a$ threshold.  Note that the thresholds can be specified as infinite, resulting in volume rotations or translations only, respectively.
  \item \textbf{fd} (fix down then forward): Rotates the volume first to keep the volume frame $y$-axis direction parallel to a specified down vector (which may be e.g. the gravity vector from the IMU), then holding the volume frame $y$-axis fixed, rotate the volume about it to align the $z$-axis as close as possible to a specified forward vector (e.g. the camera's $z$-axis vector). The volume is also automatically translated to keep the camera at the same point in volume frame.
  \item \textbf{ff} (fix forward then down): Does the same transformations as \textbf{fd} but in the opposite order.  In both cases the camera location remains fixed in the volume but the volume orientation is driven from specified down and forward vectors.
\end{enumerate}

In Chapter~\ref{Ch:patch_tracking} we review how KinectFusion can track the camera pose $C_t$ relative to the volume.
\section{Input Data Acquisition} \label{Sec:map_input}

The first stage of the algorithm is about acquiring the input data in each frame.  Chapter~\ref{Ch:input} describes in details this process that can be wrapped up in the following two steps.

\begin{figure*}[h,t]
	\begin{center}
	\includegraphics[width=\textwidth]{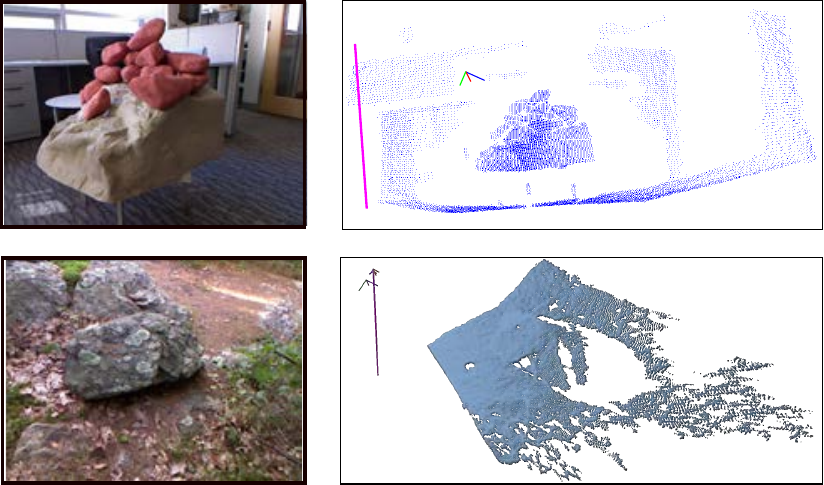}
	\end{center}
\caption[Dense point cloud input along with IMU-derived gravity vector.]{Dense $640\times480$ point cloud input (using Kinect) along with IMU-derived gravity vector (pink) for an indoor fake rock using MATLAB (upper) and an outside rocky trail using C++ code (lower).}
\label{Fig:input}
\end{figure*}

\begin{list}{\bf Stage \Roman{stage}:}{\usecounter{stage} \setcounter{stage}{0}}
  \item Acquire Input Data (Figure~\ref{Fig:input})
        \begin{list}{\it Step \arabic{step}:}{\usecounter{step} \setcounter{step}{0}}
          \item Receive image $Z$ from the depth camera and absolute orientation quaternion $\quat{q}$ from the IMU.  The depth camera may either be a physical sensor like Kinect or Carmine described in Chapter~\ref{Ch:input}, returning $640\times480$ images, or a virtual camera in the context of KinectFusion (see Chapter~\ref{Ch:patch_tracking}) which typically has a lower resolution, e.g. $200\times200$.  In the later case the virtual camera may also have a different pose in the volume than the physical camera.
          \item Convert $Z$ to an organized\footnote{Organized points have a 1:1 correspondence to an $M\times N$ depth image.} point cloud $C$ in camera frame and $\quat{q}$ to a unit gravity vector $\uvec{g}$ pointing down in camera frame.
        \end{list}
\end{list}
\section{Point Cloud Preprocessing} \label{Sec:map_preprocess}
Various types of preprocessing and filtering on the point cloud input may be applied depending on the task and the application requirements.  Some are related to the quality of the input data and some to performance.  In Section~\ref{Sec:range_sensing} we introduced some of these general filters, but apart from these we may have task-specific ones.  In this Section we introduce some preprocessing filters we developed for the rough terrain hiking task.  Note that we do not apply filtering that is not close to real-time performance (i.e. 30Hz).  Also, when filters remove points, we actually replace them with NaN\footnote{Not a Number.} values to maintain the organization of the point cloud with 1:1 correspondence to an $M \times N$ depth image, which is important for some later steps, like an optimized algorithm for finding neighborhoods (Section~\ref{Sec:map_neighborhood}).

\begin{figure*}[!h]
	\begin{center}
	\includegraphics{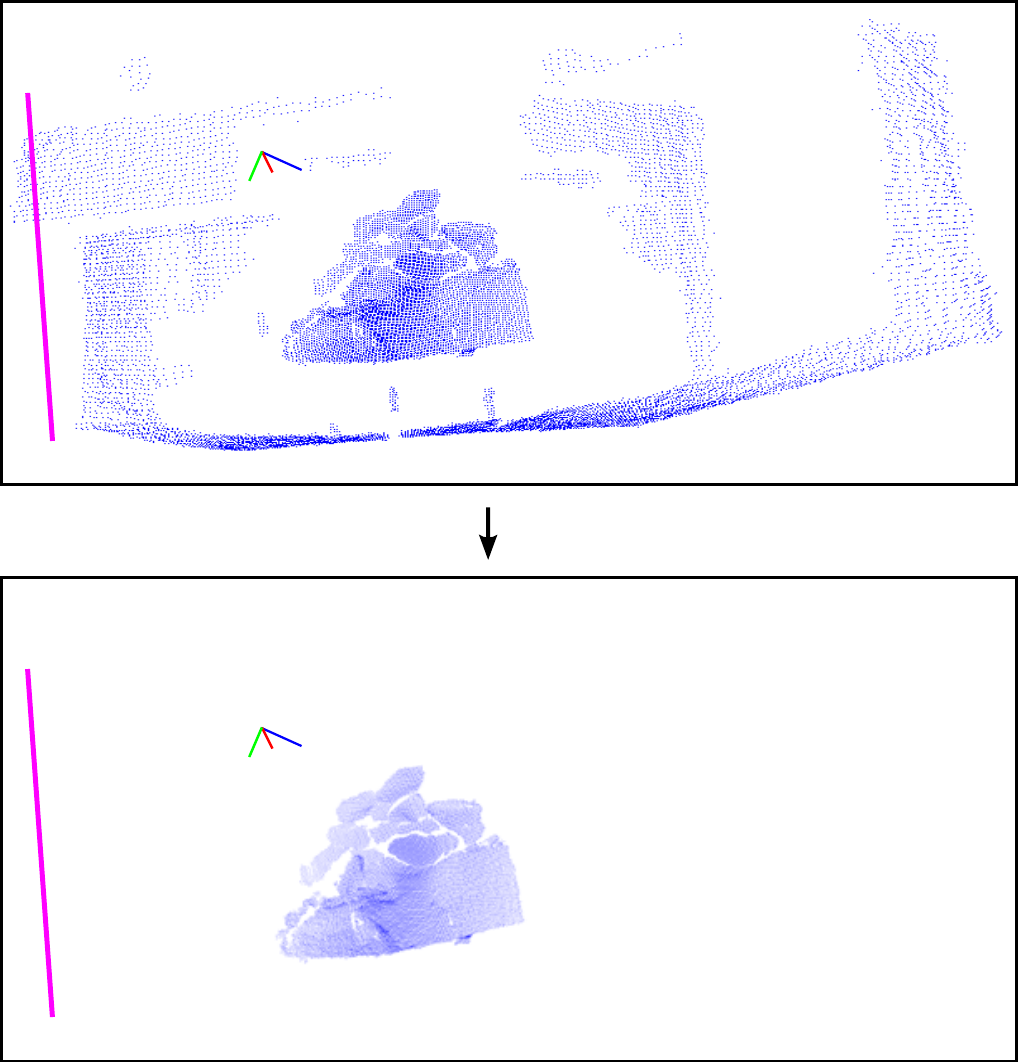}
	\end{center}
\caption[Background removal preprocessing]{Preprocessing of the input point cloud with background removal using a passthrough filter.}
\label{Fig:Background_Removal}
\end{figure*}

\begin{list}{\bf Stage \Roman{stage}:}{\usecounter{stage} \setcounter{stage}{1}}
  \item Preprocess the Input Point Cloud
        \begin{list}{\it Step \arabic{step}:}{\usecounter{step} \setcounter{step}{2}}
          \item Attempt to remove ``background'' points either using a passthrough filter thresholding the $z$-coordinate values in camera frame or by setting the volume size $V_s$ appropriately (Figure~\ref{Fig:Background_Removal}) and keeping only the points in it.
          \item Apply a discontinuity-preserving bilateral filter to $C$ to reduce noise effects~\cite{PF06}.
          \item Optionally downsample $C$ with a $2\times2$ median filter to create an auxiliary point cloud $D$.  We do this in the case that $C$ came from a $640\times480$ physical depth camera, but not when it came from a $200\times200$ virtual camera.  In the later case $D \defeq C$.
          \item Create a new point cloud $H$ by applying the hiking salienc filter (Section~\ref{Sec:map_saliency}), on point cloud $D$ .
        \end{list}
\end{list}

Point clouds $C$ and $H$ are kept until the next frame since they are used in later steps.  $C$ is going to be used for finding neighborhoods around seeds selected from $H$ (Sections \ref{Sec:map_seeds} and \ref{Sec:map_neighborhood}).

\subsection{Hiking Saliency Filter} \label{Sec:map_saliency}
In \cite{KV14} we introduced a real-time bio-inspired system for automatically finding and fitting salient patches for bipedal hiking in rough terrain.  A key aspect of the proposed approach is that we do not just fit as many patches as possible, but instead attempt to balance patch quality with sufficient sampling of the appropriate parts of upcoming terrain.

The term saliency has been used in computer graphics (e.g.~\cite{NGH04, LVJ05,LLKR07,WY09}) to describe parts of surfaces that seem perceptually important to humans.  Often these are locations of curvature extrema.  Such a definition may also be relevant here, as humans do sometimes step on e.g. the peak of a rock.  However, this seems relatively uncommon.  We thus introduce three new measures of saliency that relate to patches that humans commonly select for stepping and can be quickly applied in a point cloud to find good neighborhoods for fitting patches: Difference of Normals (DoN), Difference of Normal-Gravity (DoNG), and Distance to Fixation Point (DtFP).  These measures involve aspects of patch orientation and location.  The approach is bio-inspired both in that one of these relates to a known biomechanical property---humans tend to fixate about two steps ahead in rough terrain~\cite{MP07}---and also because we used observations of the patches humans were observed to select as a baseline for setting parameters of the measures.

\begin{figure*}[h,t]
	\begin{center}
	\includegraphics{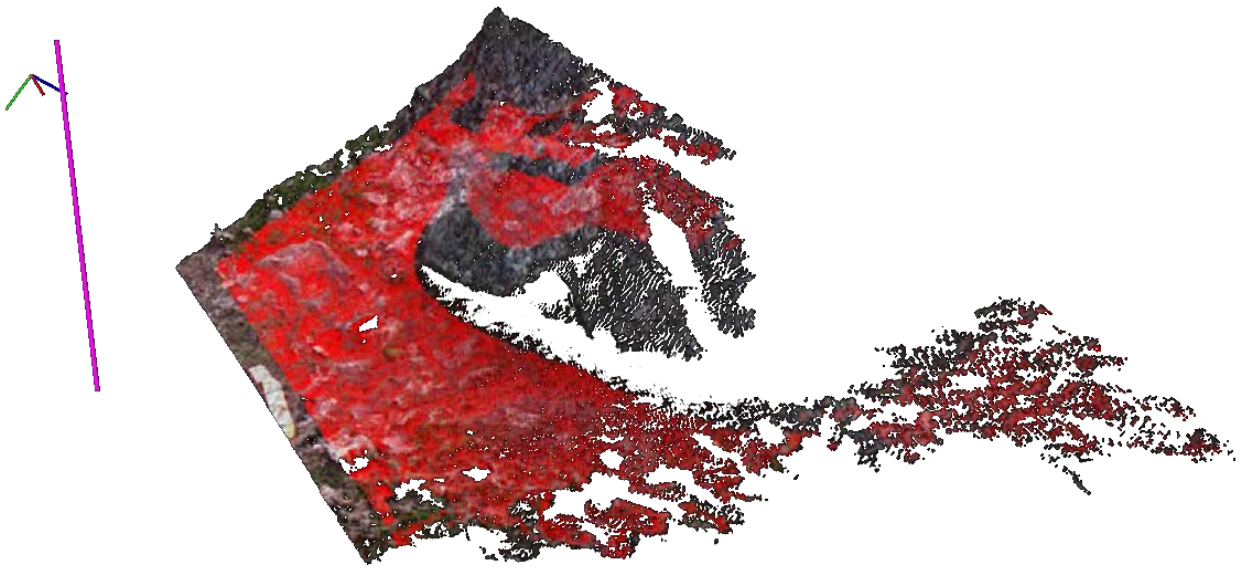}
	\end{center}
\caption[Hiking saliency filtering]{Hiking saliency filtering (salient points in red).  The thresholds for the DoN and the DoNG measures are set to $15^{\circ}$ and $35^{\circ}$ correspondingly, while the DtFP is set to infinite.}
\label{Fig:Sal_Points}
\end{figure*}

\subsubsection*{Difference of Normals (DoN)}
\begin{figure*}[h]
  \begin{center}
    \includegraphics[width=0.8\textwidth]{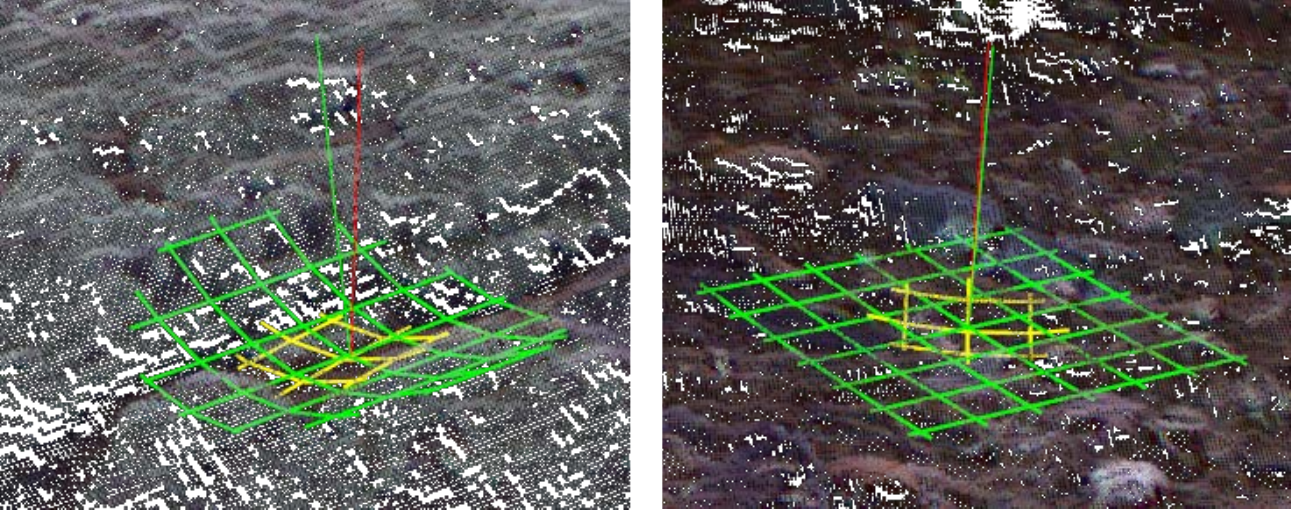}
  \end{center}
\caption[Illustration of the Difference of Normals measure]{Illustration of the Difference of Normals (DoN) measure for an irregular area (left) where the angle difference between the normals is bigger than a flat area (right).}
\label{Fig:DoN}
\end{figure*}

The difference of normals operator was introduced in~\cite{LG12} as the angle between the normals of fine scale vs coarse scale neighborhoods of a point (Figures~\ref{Fig:DoN} and~\ref{Fig:DoNG})\footnote{It was also used in \cite{ITHG12} as the norm of the normals difference.}.  This value relates to the irregularity of the surface around the point, and also to the local uniqueness of the point (following the same idea as the difference of Gaussians operator in 2D images).  We conjectured that points with low DoN may be salient for the purpose of footfall selection.  The coarse scale neighborhoods we use are of radius $r=10\mathrm{cm}$ and the fine scale are $r/2$ (for this and the next measure square neighborhoods are actually used to enable fast normal computation with integral images, see the algorithm below).

\subsubsection*{Difference of Normal-Gravity (DoNG)}
\begin{figure*}[h]
  \begin{center}
    \includegraphics[width=0.7\textwidth]{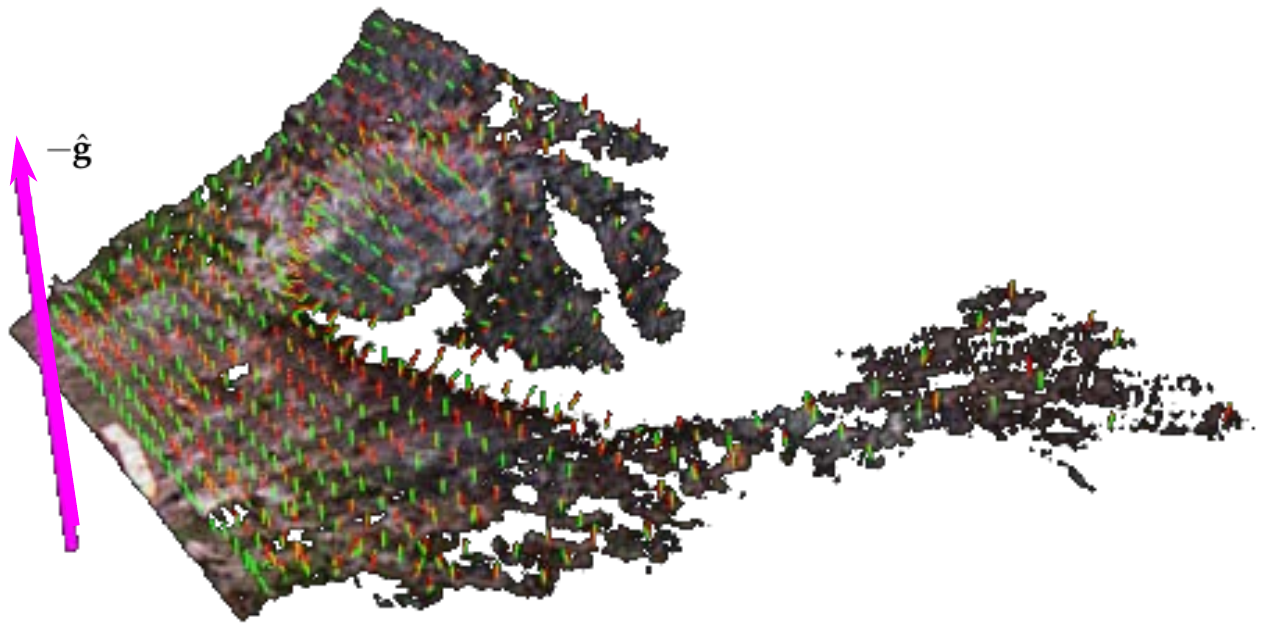}
  \end{center}
\caption[Illustration of the Difference of Normal-Gravity measure]{Illustration of the Difference of Normal-Gravity (DoNG) measure.}
\label{Fig:DoNG}
\end{figure*}

The angle between the $r$-neighborhood normal vector of each point and the reverse of the gravity vector $-\vec{\hat{g}}$ (from the IMU, $\vec{\hat{g}}$ points down) gives a measure of the slope of that area (Figure~\ref{Fig:DoNG}).  For fairly obvious reasons, points with low DoNG can be considered more salient for footfall selection.

\subsubsection*{Distance to Fixation Point (DtFP)}
\begin{figure*}[h]
  \begin{center}
    \includegraphics[width=0.9\textwidth]{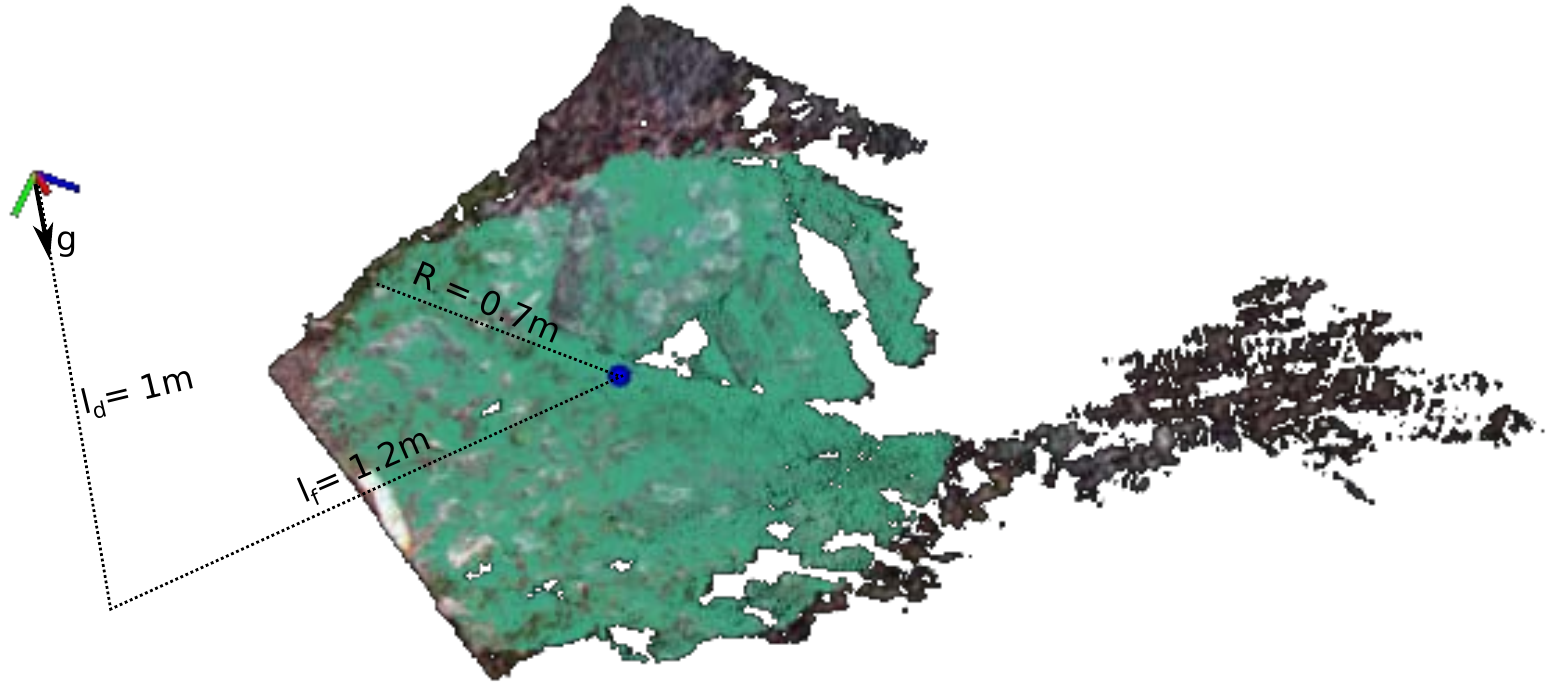}
  \end{center}
\caption[Illustration of the Distance to Fixation Point measure]{Illustration of the Distance to Fixation Point (DtFP) measure, where only point in distance of $4m$ (green points) from the fixation point (in blue) are kept.}
\label{Fig:DtFP}
\end{figure*}

Various biomechanical studies on vision for human locomotion (e.g.~\cite{MP07, MP08, Marigold08}) find that humans fixate approximately two steps ahead when locomoting in rough terrain.  We thus estimate a spatial fixation point near the ground approximately two steps ahead of the current position (Figure~\ref{Fig:DtFP}).  We define points with smaller Euclidean distance from the fixation point to have higher saliency.

\hfill

We now present the algorithm for calculating these three measures.  They can be calculated quickly for all points and so are useful to identify good seed points before fitting.

\begin{list}{\bf Stage \Roman{stage}:}{\usecounter{stage} \setcounter{stage}{1}}
  \item Preprocess the Input Point Cloud
    
    \textbf{DtFP saliency}\\  Parameters $l_d=1\mathrm{m},l_f=1.2\mathrm{m}$ are the distances down and forward from the camera to the estimated fixation point ($l_d$ is the approximate height at which we held the camera; $l_f$ is an approximation of two human step lengths~\cite{MP08}, minus the approximate distance from the body to the camera as we held it); parameter $R=0.7\mathrm{m}$ can be adjusted to capture the ground area to be sampled for upcoming steps.
    
    %\begin{list}{\it Step \arabic{step}:}{\usecounter{step}\setcounter{step}{6}}
    \begin{enumerate}
      \item[6.1] Estimate the fixation point $\vec{f}$ in camera frame \[\vec{f} \defeq l_d\uvec{g} + l_f([1~0~0]^T \times \uvec{g})\nonumber\] using the properties that $\uvec{g}$ points down and $[1~0~0]^T$ points right in camera frame.
      \item[6.2] Initialize $H$ as all points in $D$ within an $R$-ball region of interest of $\vec{f}$.
    %\end{list}
    \end{enumerate}       

    \textbf{DoN and DoNG saliency}\\  Parameter $r=10\mathrm{cm}$ is the patch neighborhood radius, which can be adjusted to match foot contact geometry; $f$ is the focal length of the depth camera in pixels; $\phi_d=15^\circ$ and $\phi_g=35^\circ$ are DoN and DoNG angle thresholds estimated from human-selected patches (Section~\ref{Sec:thresholds}).
    %\begin{list}{\it Step \arabic{step}:}{\usecounter{step}\setcounter{step}{8}}
    \begin{enumerate}
      \item[6.3] Compute surface normals $N,N_s$ corresponding to $D$ using integral images~\cite{HRDGN12}.  The normal $N(i)$ uses window size $2rf/Z(i)$ where $Z(i)$ is the $z$ coordinate (depth) of point $i$ in camera frame, and $N_s(i)$ uses window size $rf/Z(i)$.
      \item[6.4] Remove from $H$ all points $i$ for which \[N(i)^TN_s(i) < \cos(\phi_d).\nonumber\]
      \item[6.5] Remove from $H$ all points $i$ for which \[-N(i)^T\uvec{g} < \cos(\phi_g).\nonumber\]
    \end{enumerate}
    %\end{list}

The same integral image algorithm used for fast normal estimation can also produce ``surface variation'' values~\cite{PGK02} which are often related to local curvature, but this relation depends on the input and is not guaranteed.  We thus defer considering patch curvature for task-specific saliency until after patch fitting, which does give estimates of the true principal curvatures (see Sec.~\ref{Sec:map_patch_model_fit}).
\end{list}
\section{Seed Selection} \label{Sec:map_seeds}
\begin{figure}[h]
	\begin{center}
	\includegraphics[width=\textwidth]{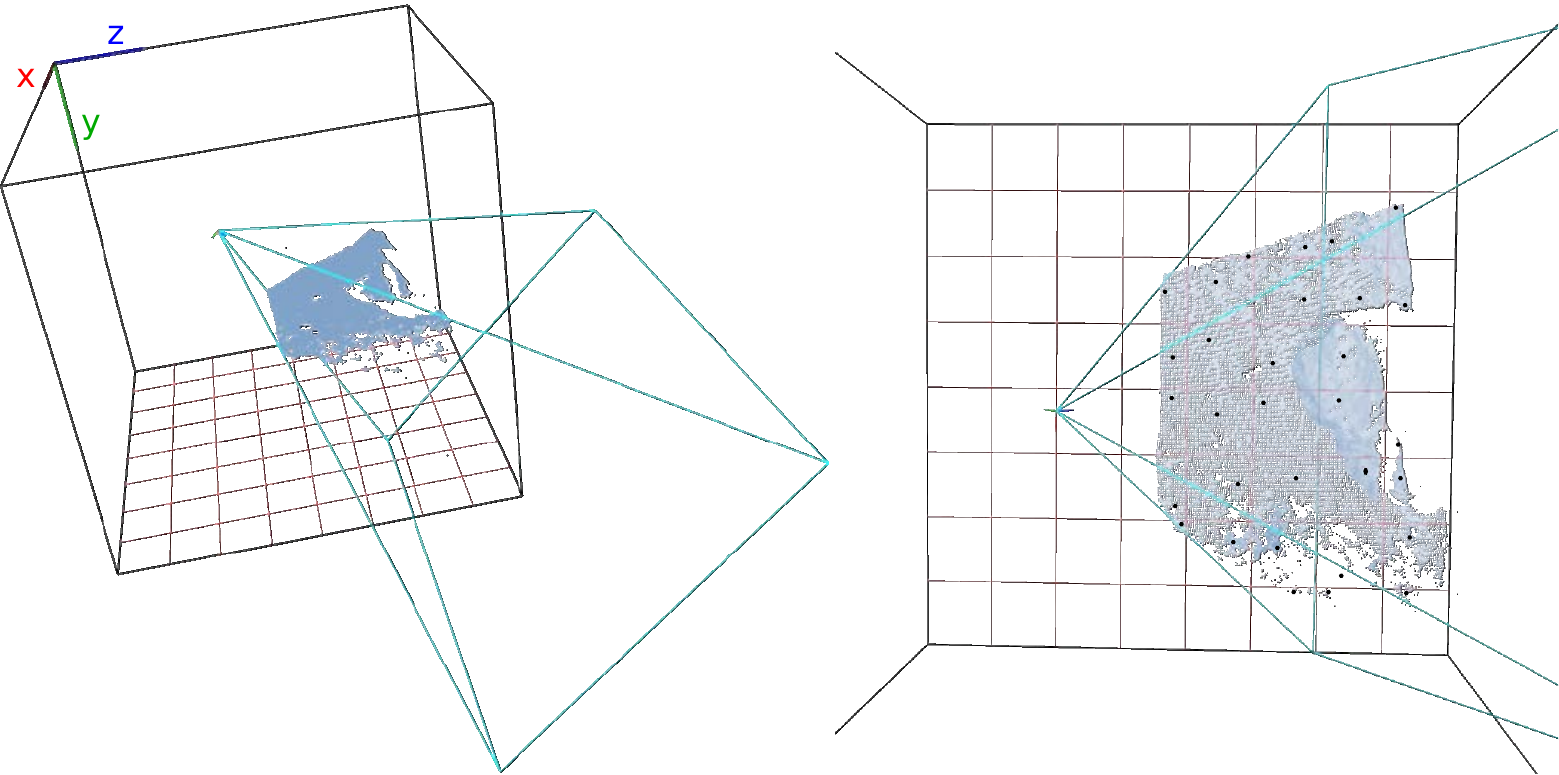}
	\end{center}
\caption[Grid cells in the volume]{Left: the xz-plane of the cubic volume is divided into an $8\times8$ grid.  Right: 31 seed points (one per cell) selected randomly in the environment; the cube is illustrated from the center of the upper cubic face.}
\label{Fig:grid_seeds}
\end{figure}

The selection of seed points around which patches will be fit is an important step in the algorithm.  We use uniformly random seed selection in $H$ relative to a coarse grid imposed on the $xz$ (horizontal) plane in volume frame.  We split the volume frame xz-plane into $V_{g} \times V_{g}$ grid cells (Figure~\ref{Fig:grid_seeds}).  We typically use $V_g = 8$.  The reason for splitting the space into grid cells is to sample the whole space more uniformly with seed points.  Using a random number generator only for selecting uniformly random points will not achieve the same effect since the density of the point cloud depends on the distance from the camera.  We next randomly pick up to $n_g$ points from each cell for a total of $n_s$ seed points.  We experimented with a non-maximum suppression algorithm~\cite{Pham10} instead of random subsampling, using a weighted average of the DoN and DoNG angles.  However the results were not clearly preferable.

Depending on real-time constraints we may use only a subset of the seeds.  We thus order the cells with respect to their distance from the projected camera position onto the volume frame $xz$ plane and we use the seeds in order of increasing distance until a time limit is reached.  As an option to limit the number of patches per cell, we can also ignore any new seed points for a cell that already has $n_g$ patches fitted to seeds within it.  Note that if the volume moves in the physical space following one of the moving volume policies introduced above the cloud remains in the same position and the seed points need to be remapped to new cells in the volume.  This remapping may move some prior seeds or patches out of the volume --- they will be removed from the map as described in Chapter~\ref{Ch:patch_tracking}.  It may also remap more than $n_g$ patches into a cell\footnote{A patch can be considered in a cell if its seed is in the cell.}; the extra patches can be culled if desired.  Figure~\ref{Fig:grid_seeds} (right) illustrates $n_s = 31$ seeds with the volume divided into an $8\times8$ grid ($V_g = 8$) and one seed point per grid ($n_g = 1$) was requested.  The seed selection proceeds as follows.

\begin{list}{\bf Stage \Roman{stage}:}{\usecounter{stage} \setcounter{stage}{2}}
  \item Select Seed Points on the Surface
  
  Let $n_g$ be the max number of seed points per grid cell.
  \begin{list}{\it Step \arabic{step}:}{\usecounter{step} \setcounter{step}{6}}
    \item Split the volume frame $xz$-plane into $V_{g} \times V_{g}$ grid cells.
    
    \item Project each point in cloud $H$ onto the $xz$-plane and find the cell it falls in by transforming the points from camera frame to volume frame and then setting their $z$ coordinate to $0$.
    
    \item Project the camera location on the $xz$-plane and order the cells in increasing distance of their center to the projected camera point.
    
    \item For each grid cell in order of increasing distance from the camera, randomly select new seed points from $H$ until at most $n_g$ seeds are associated to the cell.
  \end{list}
\end{list}
\section{Neighborhood Searching} \label{Sec:map_neighborhood}
\begin{figure}[h]
	\begin{center}
	\includegraphics[width=0.8\textwidth]{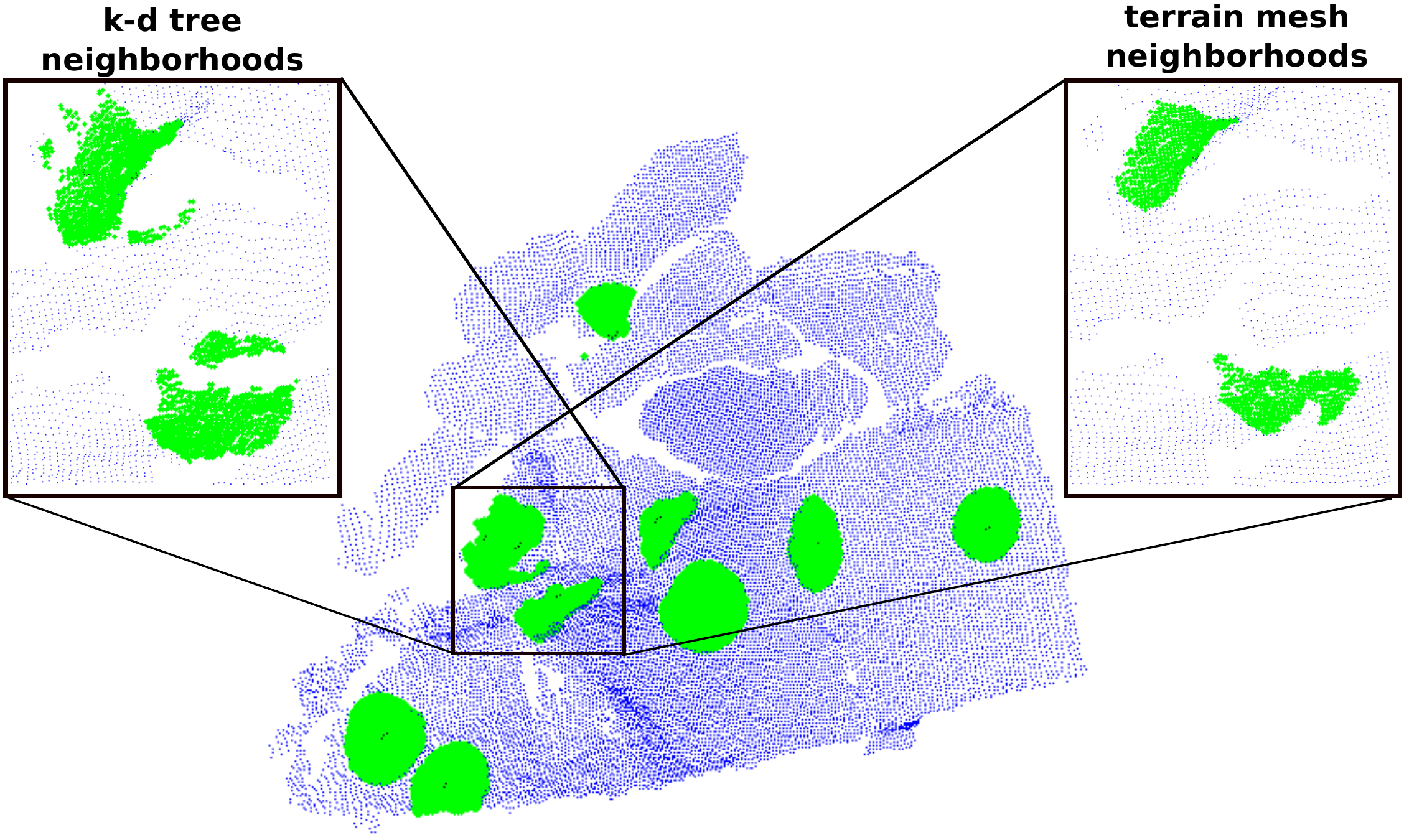}
	\end{center}
\caption[Neighborhood searching]{Ten neighborhoods with $r=0.05$m.  Unlike k-d tree neighborhoods, triangle mesh neighborhoods do not span surface discontinuities.}
\label{Fig:nearest_neighborhood}
\end{figure}

Having an ordered list of seed points, the next step is neighborhood searching in the original point cloud C for each of them.  Many methods have been introduced for finding local neighborhoods of 3D points, including approximations.  Two concepts of a neighborhood are: (a) $k$ nearest neighbors, i.e. the $k$ closest points to a seed; (b) all neighbors within distance $r$ from the seed, for some distance metric.  For fitting uniformly bounded patches we use the latter; the number of points $k$ in the recovered neighborhood thus varies depending on $r$ and the specifics of the distance metric and the search algorithm (Figure~\ref{Fig:nearest_neighborhood}).  We later uniformly subsample within each neighborhood if necessary to limit the total number of points used to fit each patch.  For general point clouds spatial decompositions like k-dimensional (k-d) trees~\cite{Bentley75} are commonly used, as well as triangle mesh structures for representing 3D sample points of surfaces.  For organized point clouds back-projection on the image plane has been used for a more efficient neighborhood extraction~\cite{RC11}.  We next present the two structures and the three methods that we have tested.

\subsection{Triangle Mesh} \label{Sec:MeshRepresentation}
\begin{figure*}
  \begin{center}	
    \includegraphics[width=0.8\textwidth]{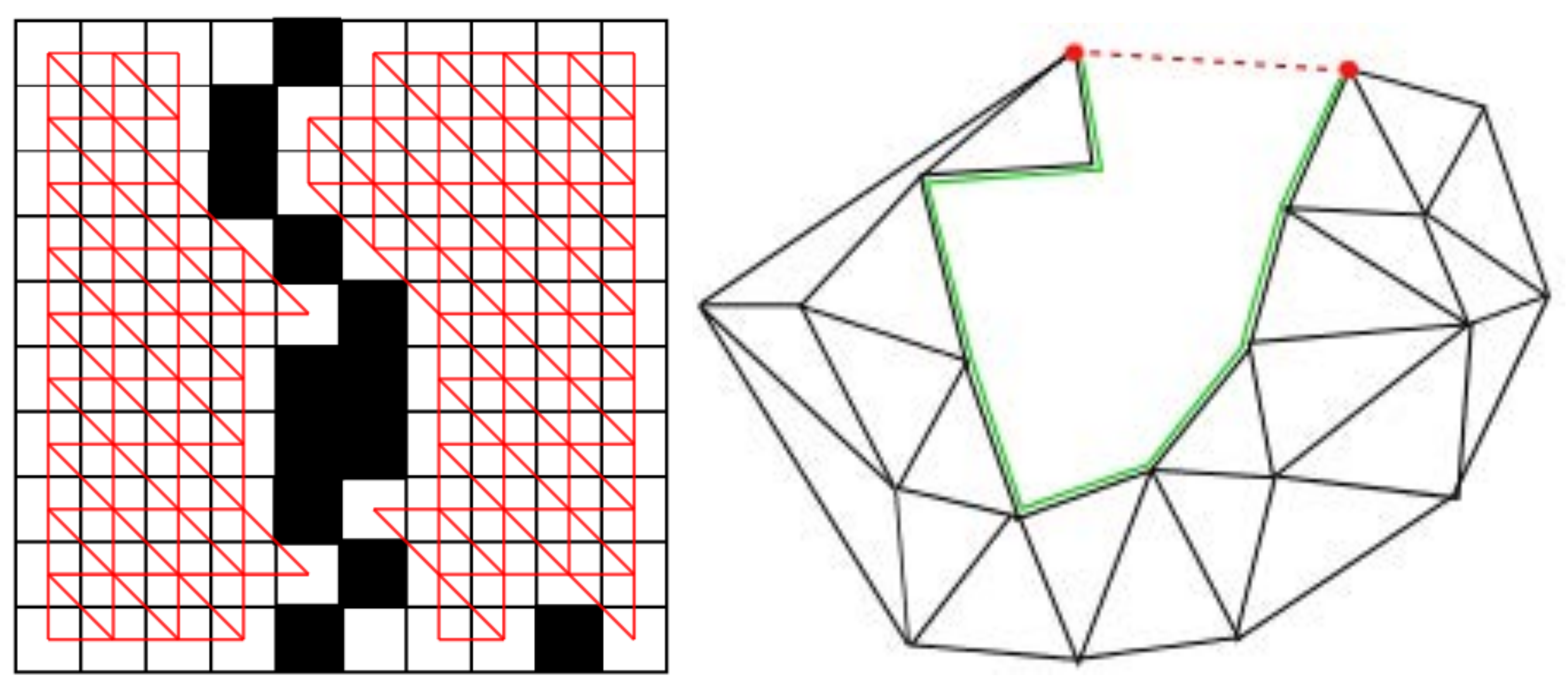}
  \end{center}
\caption[Triangle mesh construction]{Left: Terrain mesh in a 10-by-10 grid. The black pixels are those that either have invalid depth or belong to a Canny edge (see text).  Right: the chain distance (green) can distinguish points separated by a jump, whereas Euclidean 3D distance (red) may not.}
\label{Fig:TerrainTriGrid}
\end{figure*}

The triangle mesh structure can be constructed quickly since the input data is in the form of a grid.  The basic algorithm is to locally connect $(x,y)$ grid neighbors with triangle edges using only the presence or absence of valid depth data, but not the actual $z$ values.  We connect neighboring valid points in the same row and column and close triangles by adding diagonals (Figure~\ref{Fig:TerrainTriGrid}, left).%TBD: \cite{THA07, MDHWF08}

\begin{figure*}[h]
	\begin{center}
	\includegraphics[width=0.8\textwidth]{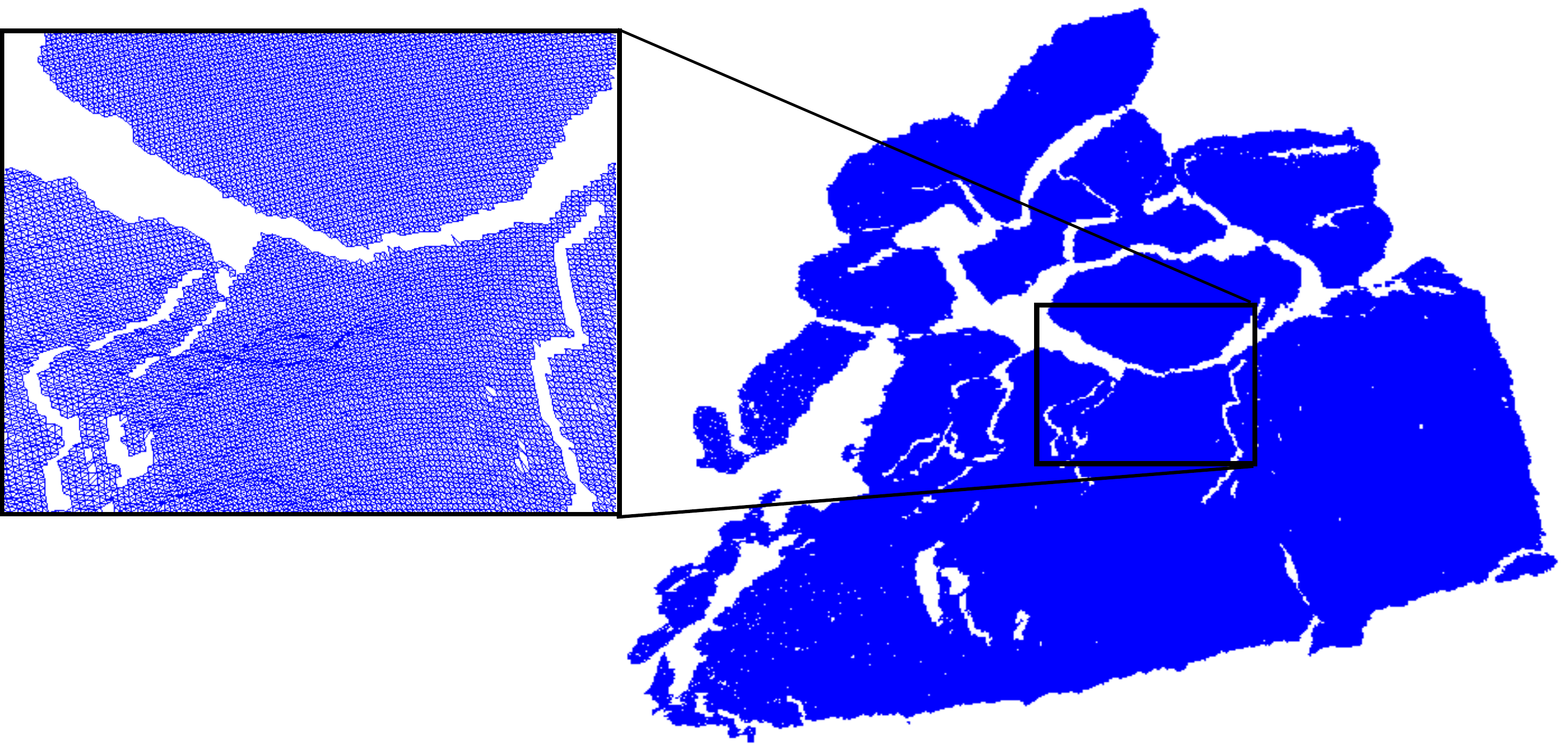}
	\end{center}
\caption[Depth jumps in triangle meshes]{Example of depth jumps between neighboring pixels.}
\label{Fig:Jumps}
\end{figure*}

A well known problem (Figure~\ref{Fig:Jumps}) is that depth discontinuities, i.e. \emph{jumps}, between $(x,y)$ neighbors could be bridged.  To address this we use Canny edge detection on the $z$ values~\cite{WSNBGH99}.  The resulting edge points are used to limit triangle construction, creating gaps in the mesh at jumps.  However, Canny edge detection does not guarantee continuous edges.  To help with this, we also remove both the triangles with sides longer than a threshold $T_{es} = 5$cm and those whose ratio of the longest side to shortest side (aspect ratio) is more than a threshold $T_{ar} = 5$.  %And when we are using the mesh to find spatial neighborhoods we still re-check the 3D distance between neighboring points. We could apply arc extrapolation, using for instance thresholding with hysteresis, to improve the Canny edges, but the runtime  would increase.

Mesh building, Canny edge detection, and removal of long triangles are all $O(N)$.  The cost for finding $k$ nearest neighbors (with breadth first search) is $O(k)$.

%In general meshes are good for static scenes, but require a more sophisticated update for dynamic scenes, because it is costly to reconstruct all the scene. We tested other triangulation techniques that are useful mainly for unorganized points, like 2.5D or 3D (constrainted) Delaunay triangulation, but the runtime was much bigger than terrain triangulation.

\subsection*{Neighborhood Searching Using the Triangle Mesh}\label{mesh}\label{mc}
First define \emph{chain distance} as the weighted edge path length between vertices in the mesh, with the weight between two vertices that share an edge equal to their Euclidean distance in 3D.  To find neighbors within distance $r$ from a seed point we apply a breadth-first search from the seed, pruning it when the chain distance exceeds $r$.  In that way we reduce the chances that the extracted neighbors cross discontinuities in the point cloud, even if they are spatially close (Figure~\ref{Fig:TerrainTriGrid}, right).

\subsection{K-D Tree} \label{Sec:kd-tree}
One of the most common data structures for spatial points is the k-d tree~\cite{Bentley75}.  Whereas the triangle mesh approach\footnote{At least the relative fast version given above.} depends on the grid organization of the data, k-d trees can be constructed from any point cloud.  However, k-d trees do not directly encode information about depth discontinuities.

%The main advantage of k-d trees is their efficiency for approximate nearest neighborhood searching. \emph{Static} k-d trees are efficient for insertions, deletions, and neighborhood queries to static point clouds, but unfortunately they also loose their efficiency in fully dynamic point clouds. Dynamic alternatives have been introduced and can be used in real time segmentation of point cloud data \cite{KWB09}.

The cost for building a k-d tree is $O(N \log^2{N})$ when using an $O(N \log{N})$ sorting algorithm for computing medians, or $O(N \log{N})$ with a linear median-finding algorithm~\cite{Bentley75}.  % Inserting and deleting a new point costs $O(\log{N})$. The cost for nearest neighbor searching is $O(\log{N})$ when the points are randomly distributed and  $O(3 N^{\frac{2}{3}})$ in the worst case. 
The cost for finding $k$ nearest neighbors is $O(k \log{N})$.

%The terrain triangulation seems better than the k-d tree structure for nearest neighborhood search, especially for finding neighbours in a ball of radius $r$ (see Section~\ref{nn}). Later in Section~\ref{experiments} we experimentally compare these two structures for our data point samples.

\subsection*{Neighborhood Searching Using the K-D Tree}\label{kdtree}
We search for neighbors within Euclidean distance $r$ of the seed using the classic method introduced in~\cite{Bentley75}.  The extracted neighborhood may span surface discontinuities (Figure~\ref{Fig:nearest_neighborhood}).

%We use an approximation to find all neighbors of distance at most $r$ from a query point. Under the assumption that the number of neighbors are proportional to the area of the circle whose radius is the searching distance from the query point (which is only strictly true if the points are uniformly distributed), we search for the $k_t$-nearest neighbors for a fixed number $k_t = 20$ and we compute the maximum distance $d_{max}$ over all the extracted $k_t$ neighbors from the query point. Then to find the points of distance at most $r$ we search for the $\floor{\frac{k_t r^2}{d_{max}^2}}$-nearest neighbors.

%In the algorithm we search for neighbourhoods within distance $r$, to keep the size of the patches approximately the same independent of the number of neighbours around a seed point.

\subsection{Image Plane Back-Projection}
This method has been used in PCL~\cite{RC11}, when the point cloud is organized, i.e. comes from a single projection point, which is the case in our system.  This method is faster than k-d trees and no extra data structure is required.  Given the 3D neighborhood-sphere around the seed point and the camera parameters, we can simply backproject it as a circle in the image plane centered at the seed's pixel.  The bounding square of pixels that the circle covers can be easily extracted.  For each one of these $O(r^2)$\footnote{The circle radius in pixels is proportional to the original sphere radius r in meters (the constant of proportionality depends on both the focal length and the distance of the sphere center to the camera center of projection).} pixels, the Euclidean distance of the corresponding 3D points (if any) to the seed point is checked to see if it is contained to the $r$-sphere.

The backprojection method has the same results as the k-d tree one, but its time and space complexity are improved in the common case by taking advantage of the fact that the point cloud is organized.  The sphere backprojection to a circle, as well as the bounding box of the circle in the image plane can be computed in constant time given the camera model, the seed point, and the neighborhood size $r$.  The Euclidean distance checking is linear in the number of checked pixels so the total cost to find an $r$-neighborhood is $O(r^2)$.  The neighborhood finding algorithm proceeds as follows.

\begin{list}{\bf Stage \Roman{stage}:}{\usecounter{stage} \setcounter{stage}{3}}
  \item Find r-Neighborhoods of Seed Points
  
  Parameter $n_f=50$ is the maximum neighborhood size for patch fitting, which can also be adjusted depending on patch size.
  
  \begin{list}{\it Step \arabic{step}:}{\usecounter{step} \setcounter{step}{10}}
    \item Use an organized search to find a neighborhood with at most $n_f$ points from $C$ randomly distributed within an $r$ ball of each seed $S(i)$.  In \cite{KV13} we studied the three different neighborhood methods described above.  Here we use the image plane backprojection method.
  \end{list}
\end{list}

\section{Patch Modeling and Fitting} \label{Sec:map_patch_model_fit}
\begin{figure}[h]
	\begin{center}
	  \includegraphics[width=\textwidth]{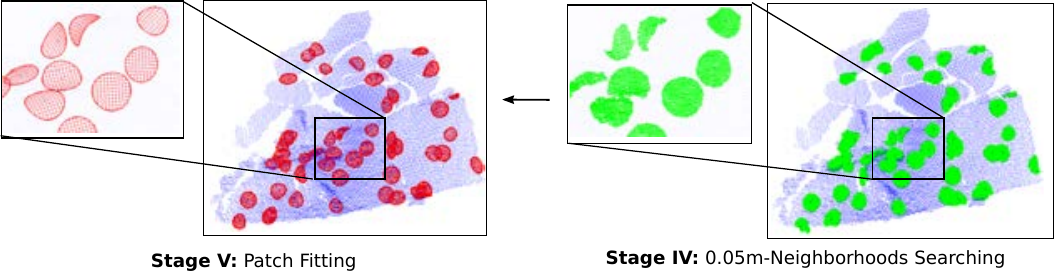}
	\end{center}
\caption[Patch fitting]{Patches fit to the $0.05m$-neighborhoods of 50 seeds.}
\label{Fig:map_fitting}
\end{figure}

\begin{figure}[h]
	\begin{center}
	  \includegraphics[width=\textwidth]{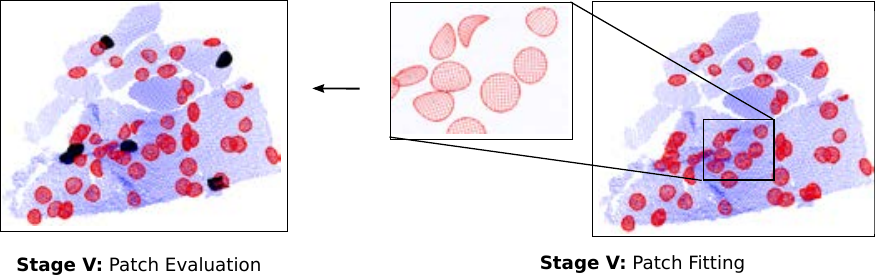}
	\end{center}
\caption[Patch validation]{50 fitted patches are validated with respect to residual, coverage, and curvature (discarded patches in black).}
\label{Fig:map_validate}
\end{figure}

Since we have a set of point cloud neighborhoods around each seed, we can proceed in patch fitting and validation as been described in detail in Chapter~\ref{Ch:EnvRep} (Figures~\ref{Fig:map_fitting},\ref{Fig:map_validate}), with the difference that during curvature validation we can also aply a fourth post-processing saliency measure for the hiking task\footnote{We are not aware of a method for finding true principal curvatures, short of fitting patches, that is as fast as we would like on the raw point cloud during pre-processing.}.

\subsection*{Minimum and Maximum Principal Curvature}
The smaller of the two principal curvatures $\kappa_\mathrm{min}\defeq\min(\kappa_x,\kappa_y)$ at a point is the inverse of the radius of the smallest osculating circle tangent to the surface there; similarly the largest osculating circle has radius $1/\kappa_\mathrm{max}$.  The signs of the principal curvatures also indicate whether the surface is concave (both positive), convex (both negative), or saddle (opposite signs) at that point.  These values can be used in a few different ways depending on the shape of the robot foot.  For example, for a flat footed robot (or to a rough approximation, for a human wearing a hiking boot), concave regions with more than slightly positive $\kappa_\mathrm{max}$ could be considered less salient, because the foot can't fully fit there.  A robot with spherical feet might prefer areas that are not too convex (as the foot would only make contact at a tangent point) but also not too concave to fit.

\begin{list}{\bf Stage \Roman{stage}:}{\usecounter{stage} \setcounter{stage}{4}}
  \item Fit \& Validate Curved Bounded Patches to the Neighborhoods
  
  Patch fitting, curvature saliency, and post-processing.  Parameter $\kappa_\mathrm{min}=-13.6\mathrm{m}^{-1}$, $\kappa_\mathrm{max}=19.7\mathrm{m}^{-1}$ are the min and max principal curvatures estimated from human selected patches (Section~\ref{Sec:thresholds}); $d_\mathrm{max}=0.01\mathrm{m}$ is the maximum RMS Euclidean patch residual.

  \begin{list}{\it Step \arabic{step}:}{\usecounter{step}\setcounter{step}{11}}
    \item Fit a patch $P(i)$ to each neighborhood as described in Chapter~\ref{Ch:EnvRep}.
    
    \item Discard patches with min principal curvature less than $\kappa_\mathrm{min}$ or max principal curvature greater than $\kappa_\mathrm{max}$ (curvature saliency).  This step could be adjusted depending on the application.
    
    \item Compute Euclidean patch residual (Section~\ref{Sec:patch_residual})~\cite{Taubin93,KV13} and discard patches where this is greater than $d_\mathrm{max}$.
    
    \item Apply the patch coverage algorithm (Section~\ref{Sec:patch_coverage})~\cite{KV13} to discard patches with areas not sufficiently supported by data.
  \end{list}

\end{list}
\section*{Termination Criteria} \label{Sec:map_termination}
Various termination criteria may be applied while adding patches to the map for each new data frame, for example: wall-clock time, total number of patches, or task-specific criteria.  Another approach is to specify a desired fraction $\nu$ of the total sampled surface area $S$ that should probabilistically be covered by patches.  Note that $\nu$ can be both less than 1, to sample sparsely, or more than 1, to oversample.  For instance with $r$-ball neighborhood search and ellipse-bounded paraboloid patch fitting we can estimate the expected number of patches for this criteria as
\beq \label{Eq:expnumpatches}
  \nu\frac{S}{\pi r^2}.
\eeq
Or, as we do in the experiments below, we can fit patches until the sum of their areas reaches or exceeds $\nu S$.  In practice it is nontrivial both to calculate the total sampled surface area $S$ and the area of any individual patch.  To approximate $S$ we compute the triangle mesh (Sec.~\ref{Sec:MeshRepresentation}) and sum the triangle areas.  We approximate the area of an individual patch as the area of the projection of its boundary on the $xy$ plane of the patch local frame $L$.
\junk{
\section{Algorithm} \label{Sec:map_algo}
The algorithm proceeds in four stages (Fig.~\ref{Fig:overview}): (I) preprocessing; (II) DtFP saliency; (III) DoN and DoNG saliency; (IV) patch fitting, curvature saliency, and postprocessing.

\begin{figure*}[tb]
\begin{center}
\includegraphics{overview/overview.pdf}
\end{center}
\caption{Algorithm overview: (I) dense point cloud input; (II) candidate seeds (green) near an estimated fixation point (blue); (III) dense normals are computed and used for saliency measures along with the IMU-derived gravity direction (purple); (IV) 100 patches are fit to a random subsampling of salient seeds (red) and are validated for quality of fit and acceptable curvature.}
\label{Fig:overview}
\end{figure*}

\begin{list}{\bf Stage \Roman{stage}:}{\leftmargin=1em\usecounter{stage}}
\item Input and preprocessing.
\begin{list}{\it Step \arabic{step}:}{\usecounter{step}\setcounter{step}{0}}
\item Receive $640\times480$ image $Z$ from the depth camera and absolute orientation quaternion $\quat{q}$ from the IMU.
\item Convert $Z$ to an organized\footnote{Organized points have a 1:1 correspondence to an $M\times N$ image.} point cloud $C$ in camera frame and $\quat{q}$ to a unit gravity vector $\uvec{g}$ pointing down in camera frame.
\item Apply a discontinuity-preserving bilateral filter to $C$ to reduce noise effects~\cite{PF06}.
\item Create a new $320\times240$ organized point cloud $D$ by downsampling $C$ with a $2\times2$ median filter kernel.
\end{list}

\item DtFP saliency.  Parameters $l_d=1\mathrm{m},l_f=1.2\mathrm{m}$ are the distances down and forward from the camera to the estimated fixation point ($l_d$ is the approximate height at which we held the camera; $l_f$ is an approximation of two human step lengths~\cite{MP08}, minus the approximate distance from the body to the camera as we held it); parameter $R=0.7\mathrm{m}$ can be adjusted to capture the ground area to be sampled for upcoming steps.
\begin{list}{\it Step \arabic{step}:}{\usecounter{step}\setcounter{step}{4}}
\item Estimate the fixation point $\vec{f}$ in camera frame \[\vec{f} \defeq l_d\uvec{g} + l_f([1~0~0]^T \times \uvec{g})\nonumber\] using the properties that $\uvec{g}$ points down and $[1~0~0]^T$ points right in camera frame.
\item Initialize seed points $S$ as all points in $D$ within an $R$-ball region of interest of $\vec{f}$.
\end{list}

\item DoN and DoNG saliency.  Parameter $r=10\mathrm{cm}$ is the patch neighborhood radius, which can be adjusted to match foot contact geometry; $f=(525/2)\mathrm{pixels}$ is the focal length of the depth camera at the downsampled $320\times240$ resolution; $\phi_d=15^\circ$ and $\phi_g=35^\circ$ are DoN and DoNG angle thresholds estimated from human-selected patches (Section~\ref{Sec:thresholds}).
\begin{list}{\it Step \arabic{step}:}{\usecounter{step}\setcounter{step}{6}}
\item Compute $320\times240$ surface normals $N,N_s$ corresponding to $D$ using integral images~\cite{HRDGN12}.  The normal $N(i)$ uses window size $2rf/Z(i)$ where $Z(i)$ is the $z$ coordinate (depth) of point $i$ in camera frame, and $N_s(i)$ uses window size $rf/Z(i)$.
\item Remove from $S$ all points $i$ for which \[N(i)^TN_s(i) < \cos(\phi_d).\nonumber\]
\item Remove from $S$ all points $i$ for which \[-N(i)^T\uvec{g} < \cos(\phi_g).\nonumber\]
\end{list}
The same integral image algorithm used for fast normal estimation can also produce ``surface variation'' values~\cite{PGK02} which are often related to local curvature, but this relation depends on the input and is not guaranteed.  We thus defer curvature saliency until after patch fitting, which does give estimates of the true principal curvatures.

\item Patch fitting, curvature saliency, and postprocessing.  Parameter $n_s=100$ is the maximum number of seed points and can be adjusted to balance coverage and patch overlap in relation to patch and region of interest size; $n_f=50$ is the maximum neighborhood size for patch fitting, which can also be adjusted depending on patch size; $\kappa_\mathrm{min}=-13.6\mathrm{m}^{-1},\kappa_\mathrm{max}=19.7\mathrm{m}^{-1}$ are the min and max principal curvatures estimated from human selected patches (Section~\ref{Sec:thresholds}); $d_\mathrm{max}=0.01\mathrm{m}$ is the maximum RMS Euclidean patch residual.
\begin{list}{\it Step \arabic{step}:}{\usecounter{step}\setcounter{step}{9}}
\item If $|S|>n_s$ then discard $|S|-n_s$ points randomly from $S$.  We experimented with a non-maximum suppression algorithm~\cite{Pham10} instead of random subsampling, using a weighted average of the DoN and DoNG angles.  However the results were not clearly preferable.
\item Use an organized search to find a neighborhood $H(i)$ with at most $n_f$ points from $C$ randomly distributed within an $r$ ball of each seed $S(i)$.  In \cite{KV13} we studied different neighborhood methods including breadth-first search on a triangle mesh and K-D tree spatial search.  Here we used the organized point cloud search method in PCL which is optimized using backprojections to the image plane.
\item Fit a patch $P(i)$ to each neighborhood $H(i)$.  Our fitting algorithm~\cite{VK11} uses a Levenberg-Marquardt nonlinear iteration to minimize an algebraic surface residual.  It can accommodate covariance matrices for input and output uncertainty, but we don't use that feature here.  The boundary curve is fit probabilistically (using moments) to include 95\% of $H(i)$ projected on the patch local frame $xy$ plane.
\item Discard patches with min principal curvature less than $\kappa_\mathrm{min}$ or max principal curvature greater than $\kappa_\mathrm{max}$ (curvature saliency).  This step could be adjusted depending on the application.
\item Compute Euclidean patch residual~\cite{Taubin93,KV13} and discard patches where this is greater than $d_\mathrm{max}$.
\item Our patch coverage algorithm~\cite{KV13} could also be applied to discard patches with areas not sufficiently supported by data, but we did not integrate it here yet.
\end{list}
\end{list}
}
\section*{Time Complexity and Runtime}
Stages I-III are $O(|Z|)$, i.e. linear in the input.  The implementation of Step 11 in Stage IV is $O(n_sr^2)$ (it could be improved to $O(n_sn_f)$ by switching to breadth-first search on a triangle mesh, but we found the constant factors favor the image backprojection method for neighborhood search in practice).  The runtime of stage V is dominated by $O(n_sn_f^2)$ for step 12.  Steps 14 and 15 are $O(n_sn_f)$.  The worst case time complexity for the whole algorithm is thus $O(|Z|+n_sn_f^2)$.

In practice on commodity hardware (one 2.50GHz core, 8GB RAM) the bilateral filter and downsampling (stage I) run in $\sim$20ms total.  Normal computation, DtFP, DoN, and DoNG saliency in Stage II take $\sim$35ms combined, dominated by $\sim$30ms for integral image computation using $640\times480$ input images from a hardware depth camera downsampled to $320\times240$ (the main reason for downsampling is that the required integral images take $\sim$150ms at $640\times480$~\cite{HRDGN12}).  Neighborhood finding in Stage IV takes $\sim$0.03ms per seed, and patch fitting and validation in Stage V are $\sim$0.8ms total per neighborhood with $n_f=50$.  The total time elapsed per frame when using $640\times480$ input images is $20+35+0.83n_p$ms, where $n_p$ is the number of patches actually added.  $n_p$ can range from $0$ in the case that the map is already full (or there are no new seed points) up to $n_g V_g^2$.  In practice we additionally limit the total time spent per frame to e.g. $100$ms, allowing up to around $50$ patches to be added per frame in this configuration.
\section{Homogeneous Patch Map} \label{Sec:spatial_map}
\begin{figure*}[h]
  \begin{center}
    \includegraphics{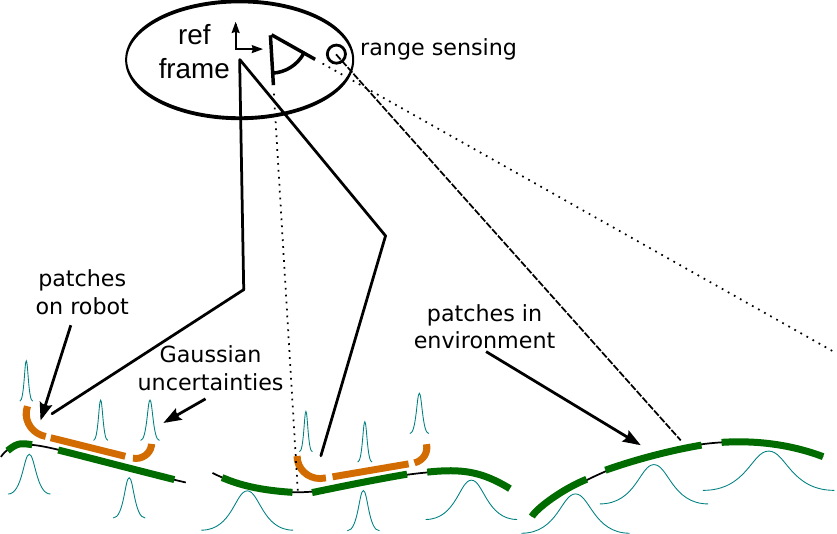}
  \end{center}
\caption[Homogeneous patch map]{Concept of the \emph{homogeneous patch map}: a sparse set of patches that locally approximate both environment surfaces (green) and key contact surfaces on a robot (brown).  All are spatially mapped with quantified uncertainty (blue Gaussians) relative to a body-centered reference frame.}
\label{Fig:map_patchmap}
\end{figure*}

Salient patches from the algorithm proposed in this chapter could form the basis for a \emph{homogeneous patch map}; a dynamically maintained local spatial map of curved surface patches suitable for contact both on and around the robot.  Figure~\ref{Fig:map_patchmap} illustrates the idea, including both environment surfaces and contact pads on the robot itself (potentially uncertain due to kinematic error).  Patches on the robot are not fully developed in this thesis since they would not be found and fitted by 3D exteroception, but would come from the robot model and proprioception.

The homogeneous aptch map could provide a sparse ``summary'' of relevant contact surfaces for higher-level reasoning.  As contacts are made the map could be further refined.  Exteroception can detect upcoming terrain patches from a distance, but with relatively high uncertainty.  Kinematic proprioception could sense the pose of contact patches on the robot itself---e.g. heel, toe, foot sole---potentially with relatively low uncertainty.  When a contact is made between a robot and environment patch, the latter could be re-measured \emph{exproprioceptively} through kinematics and touch, possibly with reduced uncertainty compared to prior exteroception.

All of the classic elements of SLAM~\cite{SC86} would apply to such a map: propagation of spatial uncertainty through kinematic chains, associating different observations of the same surface patch, and optimal data fusion.  Fusion by Kalman update is supported by the patch covariance matrices.  First-order propagation of uncertainty through a chain of transforms with $6\!\times\!6$ covariances $S_j$ is facilitated by the chain Jacobian $J_c$ given in Appendix~\ref{Sec:jacobians}:
\beq
\Sigma_c=J_c\mathcal{S}J_c^T,\ \mathcal{S}\defeq\diag(S_n,\ldots,S_1).
\eeq
$\Sigma_c\!\!\in\!\!\R^{6\times6}$ is the covariance of the pose of a patch at the end of the chain relative to the base.  For a 5-DoF patch,
\beq
\Sigma_{c_5}=J_5J_c\mathcal{S}J_c^TJ_5^T,\ %
J_5\defeq\begin{bmatrix}
\frac{\partial\vec{r}_{xy}}{\partial\vec{r}}\!\!&0\\
0\!\!&I_{3\times3}
\end{bmatrix}.
\eeq

\section{Experimental Results} \label{Sec:map_exp}
In this chapter we present two experiments to test the overall patch mapping approach.  In the first one we compare the triangle mesh and k-d tree data structures for neighborhood finding using our Matlab code.  In the second one we measured the patches that humans actually selected on several sections of rocky trail to establish the baseline saliency thresholds $\phi_{d,g}$ and $\kappa_{\mathrm{min},\mathrm{max}}$, using our C++ implementation.  Additional experiments using the patch map are presented in Chapter~\ref{Ch:patch_tracking}.

\subsection{Triangle Mesh vs K-D Tree Neighborhood Searching} \label{Sec:map_exp_struct}

\begin{figure*}[thpb]
	\begin{center}
	  \includegraphics[width=\textwidth,keepaspectratio]{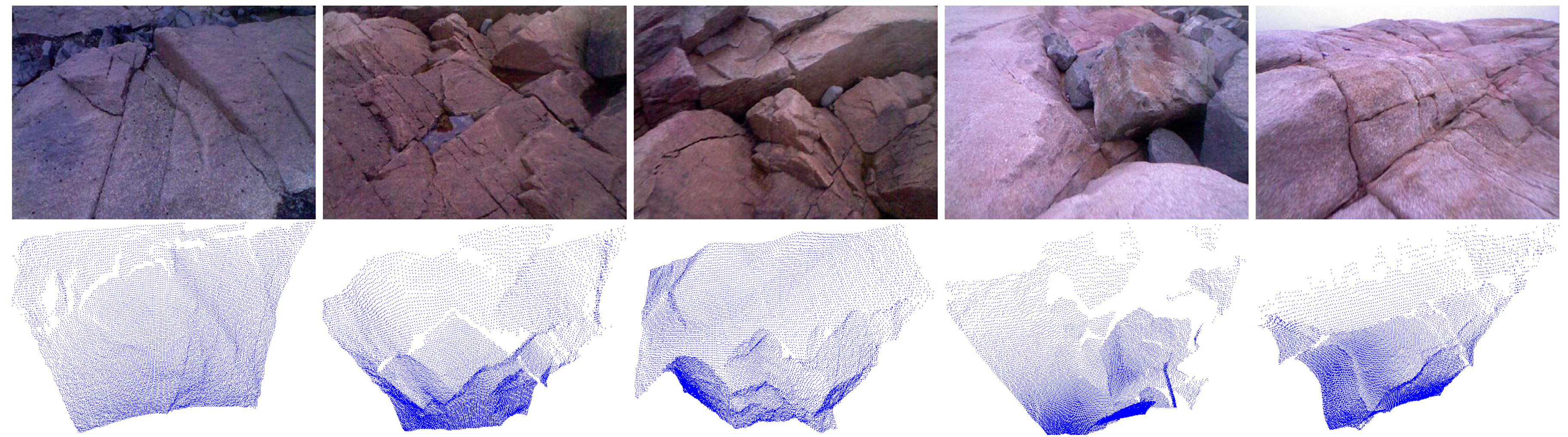}
	\end{center}
\caption[Five rock datasets]{Datasets rock 1--5.  All rock datasets were acquired with a hand-held Kinect outdoors on an overcast day.}
\label{Fig:RealRocks}
\end{figure*}

\begin{table}
\begin{center}

{\setlength{\tabcolsep}{0.9pt}
	\begin{tabular}{|c|c||c|c|c|c|c|c|c|}\hline
	\multirow{2}{*}[-1em]{\bf Data} & \multirow{2}{*}[-1em]{\bf Structure} & \multicolumn{2}{c|}{{\bf Patches}} & \multicolumn{3}{|c|}{{\bf \specialcell{Dropped\\patches}}} & \multirow{2}{*}{\bf \specialcell{Average\\residual\\$(mm)$}} & \multirow{2}{*}{\bf \specialcell{Total\\area\\$(m^2)$}}\\\cline{3-7}
	& & {\bf total} & {\bf valid} & {\bf \specialcell{due to\\residual}} & {\bf \specialcell{due to\\coverage}} & {\bf \specialcell{total}} & &\\\hline
	\multirow{2}{*}{rock 1}		& k-d tree 	& 167 & 142 & 15 & 10 & 25 & 4.6 & 4.57\\%\cline{2-9}
								& tri mesh 	& 164 & 144 & 6 & 14 & 20 & 4.3 & 4.57\\\hline
	\multirow{2}{*}{rock 2}		& k-d tree 	& 160 & 107 & 18 & 36 & 53 & 5.0 & 3.13\\%\cline{2-9}
								& tri mesh 	& 185 & 124 & 0 & 61 & 61 & 4.0 & 3.13\\\hline
	\multirow{2}{*}{rock 3}		& k-d tree 	& 231 & 183 & 24 & 24 & 48 & 5.2 & 5.53\\%\cline{2-9}
								& tri mesh 	& 227 & 199 & 1 & 27 & 28 & 4.7 & 5.53\\\hline
	\multirow{2}{*}{rock 4}		& k-d tree 	& 220 & 164 & 27 & 31 & 56 & 5.0 & 5.01\\%\cline{2-9}
								& tri mesh 	& 215 & 181 & 8 & 29 & 34 & 4.8 & 5.01\\\hline
	\multirow{2}{*}{rock 5}		& k-d tree 	& 195 & 157 & 17 & 24 & 38 & 5.4 & 4.86\\%\cline{2-9}
								& tri mesh 	& 188 & 163 & 5 & 20 & 25 & 5.1 & 4.86\\\hline
	\multirow{2}{*}{rock 6}		& k-d tree 	& 235 & 185 & 31 & 20 & 50 & 5.7 & 5.69\\%\cline{2-9}
								& tri mesh 	& 273 & 226 & 9 & 39 & 47 & 5.3 & 5.69\\\hline
	\multirow{2}{*}{rock 7}		& k-d tree 	& 267 & 213 & 30 & 25 & 54 & 4.4 & 6.54\\%\cline{2-9}
								& tri mesh 	& 266 & 219 & 17 & 30 & 47 & 4.2 & 6.54\\\hline
	\multirow{2}{*}{rock 8}		& k-d tree 	& 260 & 223 & 16 & 22 & 37 & 4.2 & 7.04\\%\cline{2-9}
								& tri mesh 	& 256 & 231 & 2 & 23 & 25 & 3.9 & 7.04\\\hline
	\multirow{2}{*}{rock 9}		& k-d tree 	& 187 & 159 & 13 & 16 & 28 & 4.8 & 4.95\\%\cline{2-9}
								& tri mesh 	& 189 & 162 & 9 & 19 & 27 & 4.6 & 4.95\\\hline
	\multirow{2}{*}{rock 10}		& k-d tree 	& 301 & 223 & 30 & 50 & 78 & 5.0 & 6.98\\%\cline{2-9}
								& tri mesh 	& 300 & 236 & 9 & 55 & 64 & 4.7 & 6.98\\\hline\hline
	\multirow{2}{*}{\bf \specialcell{rock\\average}}	& \textbf{k-d tree} 	& \textbf{222} & \textbf{176} & \textbf{22} & \textbf{26} & \textbf{47} & \textbf{4.9} & \textbf{5.43}\\%\cline{2-9}
								& \textbf{tri mesh} 	& \textbf{226} & \textbf{189} & \textbf{7} & \textbf{32} & \textbf{38} & \textbf{4.6} & \textbf{5.43}\\\hline\hline
	\multirow{2}{*}{fake rock} 		& k-d tree 	& 65 & 18 & 18 & 44 & 47 & 3.8 & 0.50\\%\cline{2-9}
								& tri mesh 	& 75 & 21 & 1 & 54 & 54 & 3.8 & 0.50\\\hline
	\end{tabular}
}
\end{center}
\caption{}
\label{tb:exp}
\end{table}

11 datasets were collected with the Kinect.  The first 10 are scenes of natural rocky terrain (Figure~\ref{Fig:RealRocks}) acquired with a hand-held Kinect outdoors on an overcast day (the Kinect does not work in direct sunlight).   The last is taken in the lab with synthetic rocks (Figure~\ref{Fig:Segmentation}).

The parameters were: neighborhood radius $r = 0.1$m, residual threshold $T_r = 0.01$m, coverage cell size $w_c = 0.01$m, coverage threshold factors $\zeta_i = 0.8, \zeta_o = 0.2$, and $T_p = 0.3 A_p/w_c^2$ (all motivated above).  We let the algorithm run for each dataset until the sum of the patch areas equaled or exceeded $90\%$ of the sampled surface area, both approximated as described in Section~\ref{Sec:map_termination}.
\begin{figure*}
	\begin{center}
	\includegraphics[width=0.8\textwidth]{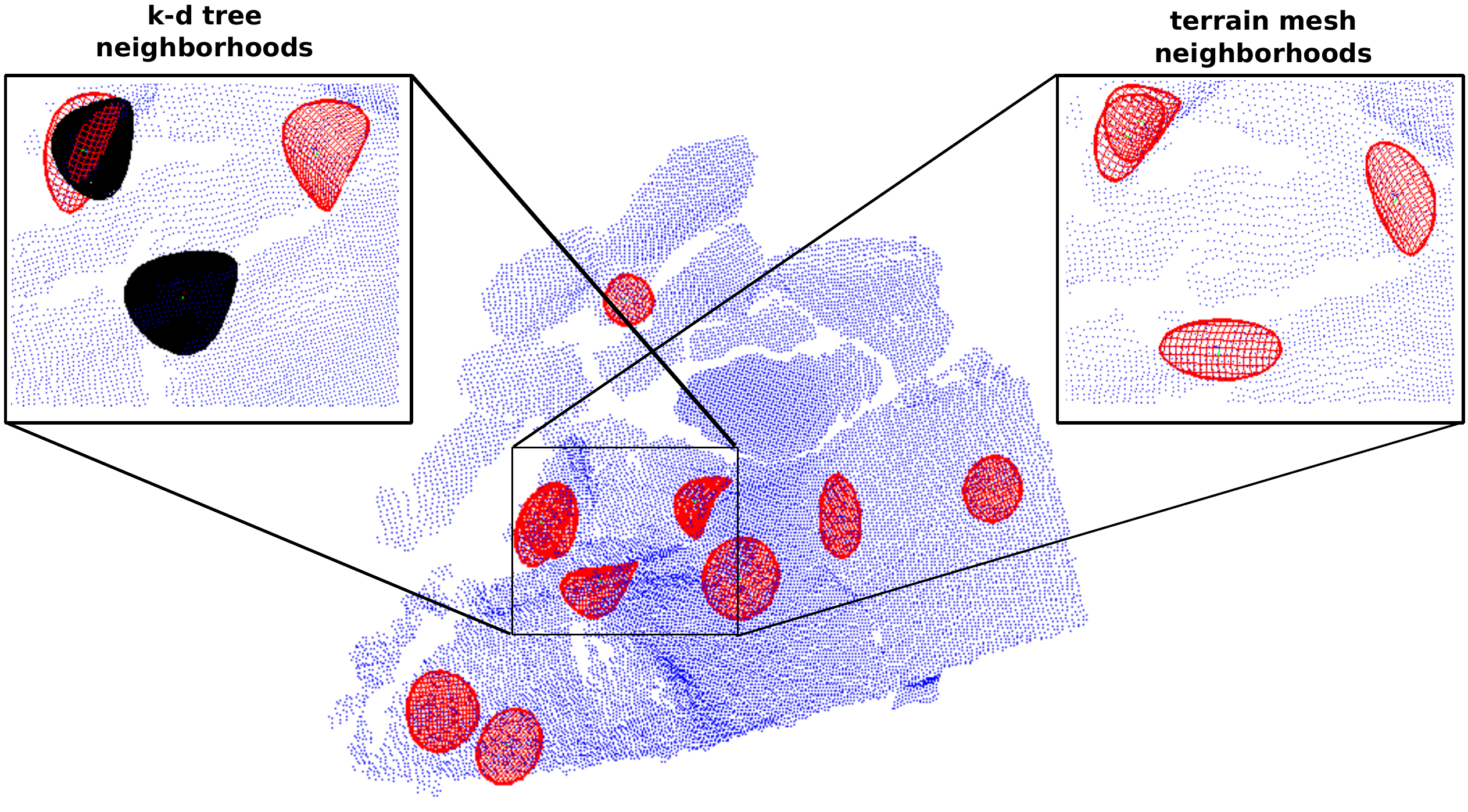}
	\end{center}
\caption[Qualitative patch fitting and validation results]{A subset of patches fit to the fake rock dataset corresponding to the neighborhoods in Figure~\ref{Fig:nearest_neighborhood}.  The black patches failed coverage evaluation.}
\label{Fig:fittingfakerock}
\end{figure*}

\begin{figure*}[h]
	\begin{center}
	\includegraphics[width=0.8\textwidth]{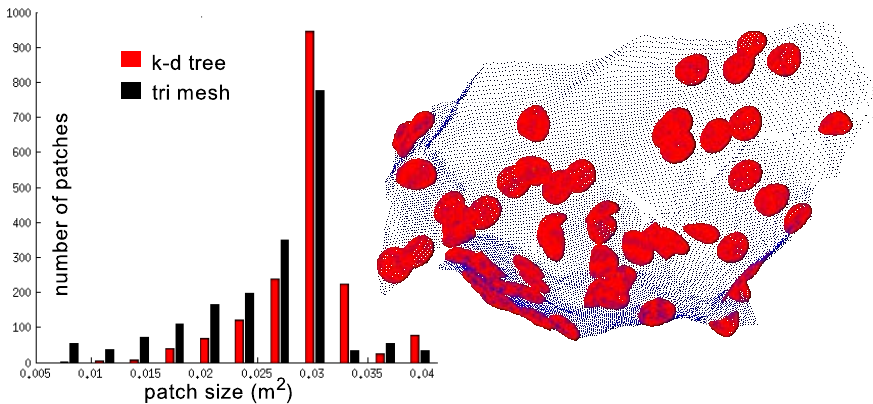}
	\end{center}
\caption[Patch size histogram comparison]{Left: Histogram of patch sizes for k-d tree (red) and triangle mesh (black) neighborhoods.  Right: 70 patches on the dataset rock 3.}
\label{Fig:FitRealRockAreaHist}
\end{figure*}

Qualitatively, as depicted in Figure~\ref{Fig:fittingfakerock} and~\ref{Fig:FitRealRockAreaHist}, the algorithm appears to give a reasonable representation of non-smooth environment surfaces.  Quantitatively, we measured the following statistics (Table~\ref{tb:exp}): the total number of patches before evaluation, the number of valid patches passing both residual and coverage evaluation, the number of dropped patches for each test, the average Euclidean residual (of the valid patches), and the total surface area for each dataset.

%Along all the datasets in Table-\ref{tb:exp}, we can first see that the geometrical residual per rock set is low with respect to the total surface area and the number of valid patches independent of the structure, which is reasonable since we filter out patches with big residual in the patch evaluation and it also proves a good surface representation.

There are generally more patches dropped due to residual for k-d tree neighborhoods, possibly because the triangle mesh neighborhoods avoid discontinuities which may not be fit well by a paraboloid.  We see the opposite effect for patches dropped due to coverage: more patches are generally dropped due to insufficient coverage when using triangle mesh neighborhoods.  The k-d tree neighborhoods may distribute samples more evenly, particularly near discontinuities.

Another interesting result is that more patches are required to reach $90\%$ of the surface area when using triangle mesh neighborhoods.  In Figure~\ref{Fig:FitRealRockAreaHist} we see that the distribution of patch areas created using mesh neighborhoods is skewed more to the low side than those created using k-d tree neighborhoods.  This can again be explained by the fact that k-d tree neighborhoods will span discontinuities but remain roughly circular, whereas triangle mesh neighborhoods may be less circular when the seed point is near an edge.

%The average run time among all datasets for our unoptimized Matlab implementation is $39.5sec$ for the k-d tree and $49.9sec$ for the mesh. The structure construction time ($\sim 0.2sec$), the fitting time per patch ($\sim0.15sec$), and the coverage and residual evaluation time per patch ($\sim 0.095sec$ and $\sim 0.007sec$) are similar between the two structures, whereas the neighbourhood search time is 10 times bigger for the mesh ($0.035sec$ vs $0.003sec$). Note that the times do not correspond to the theoretical analysis of Section~\ref{datastruct}, since we used the internal pre-compiled Matlab's k-d tree implementation which is more efficient than our mesh one.

\subsection{Human Subject Data for Hiking Saliency Thresholds}\label{Sec:Data}
For setting the saliency thresholds $\phi_{d,g}$ and $\kappa_{\mathrm{min},\mathrm{max}}$ used in Section~\ref{Sec:map_saliency}, patches that human subjects use when locomoting on rocky trails were analyzed.  Research on human locomotion shows that visual information is crucial when walking on uneven terrain~\cite{HM96, Patla97, PV97, HKKE06, RR06, MP07, Marigold08, MP08}), but so far only a few works (e.g.~\cite{LLP05}) have specifically applied this to perception for bipedal robots.

\subsubsection*{Method}
The trail sections were located in the the Middlesex Fells in Melrose, MA and were 9, 4, and 10.5 meters long.  All included rocks and other types of solid surfaces normally encountered outdoors.  We put strips of colored tape on the ground to mark nominal routes and to help establish visual correspondence among multiple video and RGB-D\footnote{The color data was used only for visual correspondence.} recordings.  The tape strips are intended to give subjects a rough idea of which route to pick but not the exact spots to place their feet.

\begin{figure}[h]
  \begin{center}
    \includegraphics{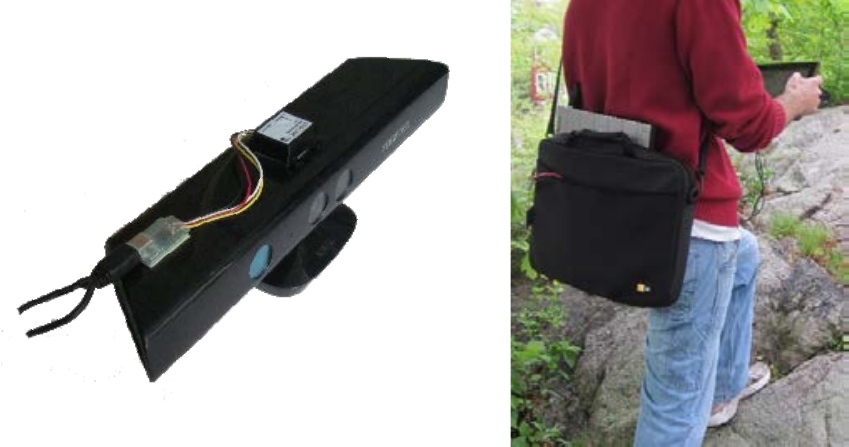}
  \end{center}
\caption[The sensing apparatus (Microsoft Kinect + CH Robotics UM6 IMU)]{Our sensing apparatus is a Microsoft Kinect RGB-D camera with a CH Robotics UM6 9-DoF IMU affixed to it.  It can be battery powered and works outdoors in shade or twilight.  Recording software runs on a laptop while a tablet strapped to the back of the sensor gives a heads-up display.}
\label{Fig:imucam}
\end{figure}

We collected 30Hz $640\times480$ RGB-D recordings of all trails with spatiotemporally coregistered\footnote{Though calibration methods have been developed (Section~\ref{Sec:imu_cal}), here spatial coregistration of IMU and depth data was based on the construction of the sensor apparatus.  As mentioned in Chapter~\ref{Ch:input} spatial registration of the depth and RGB data used built-in calibration in the Kinect sensor.  Temporal registration of all three datastreams was approximate.} 100Hz IMU data using a handheld Kinect camera with a CH Robotics UM6 9-DoF IMU (3-axis accelerometers, 3-axis gyroscopes, and 3-axis magnetometers) attached, including a Kalman filter to estimate absolute geo-referenced orientation (Figure~\ref{Fig:imucam}).  The structured light method used by the Kinect does not work well in full sunlight so we took this data at twilight.  Sunlight operation could be possible with other types of depth camera or stereo vision.  The camera was held facing $\sim45^\circ$ forward and down and $\sim1$m above the ground by a human operator who walked at normal pace along each trail section.  The data were saved in lossless PCLZF format~\cite{RC11}.

\begin{figure*}[h]
  \begin{center}
    \includegraphics[width=\textwidth]{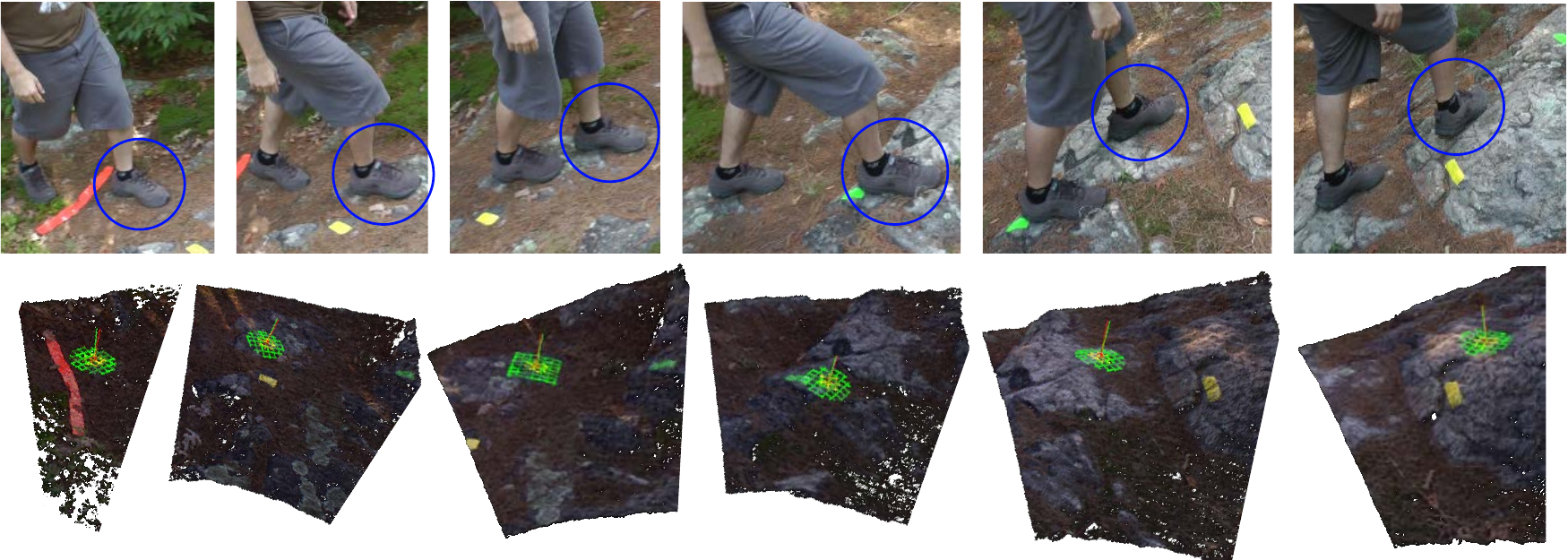}
  \end{center}
\caption[Human-selected patches]{Human-selected patches, manually identified in video and RGB-D recordings.}
\label{Fig:manseg}
\end{figure*}

We also took video recordings of the feet of five healthy human volunteers walking on these trails.  For each trail participants were asked to walk at normal pace twice in each direction, following the nominal marked route (60 recordings).  We visually matched all footsteps (total 867) in these recordings to corresponding (pixel, frame) pairs in the RGB-D+IMU data, and we fit patches (algorithm steps 11 and 12) at these locations (Figure~\ref{Fig:manseg}).

\subsubsection*{Results and Threshold Estimation}\label{Sec:thresholds}
\begin{figure*}[h]
\begin{center}
\includegraphics[width=\textwidth]{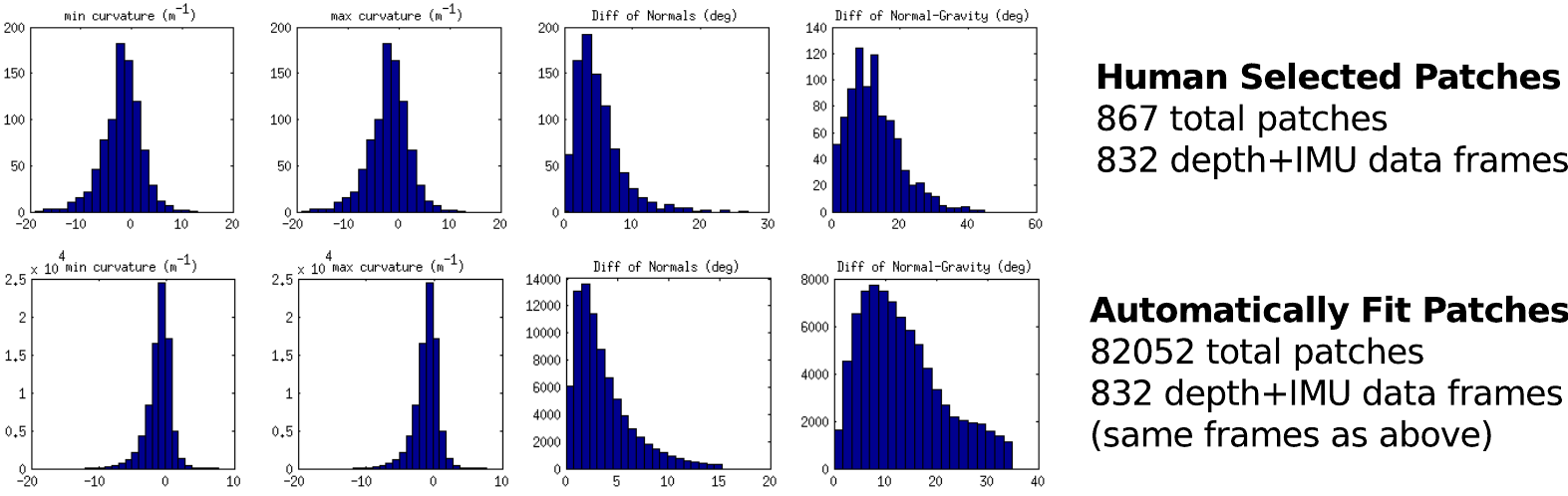}
\end{center}
\caption[Comparison of histograms of salient measures for human-selected and automatically identified patches]{Comparison of histograms of principal curvatures and our DoN and DoNG measures for human-selected and automatically identified patches.}
\label{Fig:data}
\end{figure*}

We took statistics\footnote{Min, max, median (med), average (avg), and standard deviation (std).} of properties of the human selected patches including the max and min curvatures, the difference angle between the two-level normals (DoN) and the difference angle (DoNG) between the full patch normal and the upward pointing vector $-\uvec{g}$ from the IMU (Fig.~\ref{Fig:data} top and rows labeled ``man'' in Table~\ref{Tb:stats}).  Thresholds $\phi_{d,g}$ and $\kappa_{\mathrm{min},\mathrm{max}}$ for the saliency algorithm were set to the corresponding averages from the human selected patches plus (minus for $\kappa_\mathrm{min}$) $3\sigma$, where is $\sigma$ is the standard deviation.

\begin{figure*}[h]
  \begin{center}
    \includegraphics[width=0.8\textwidth]{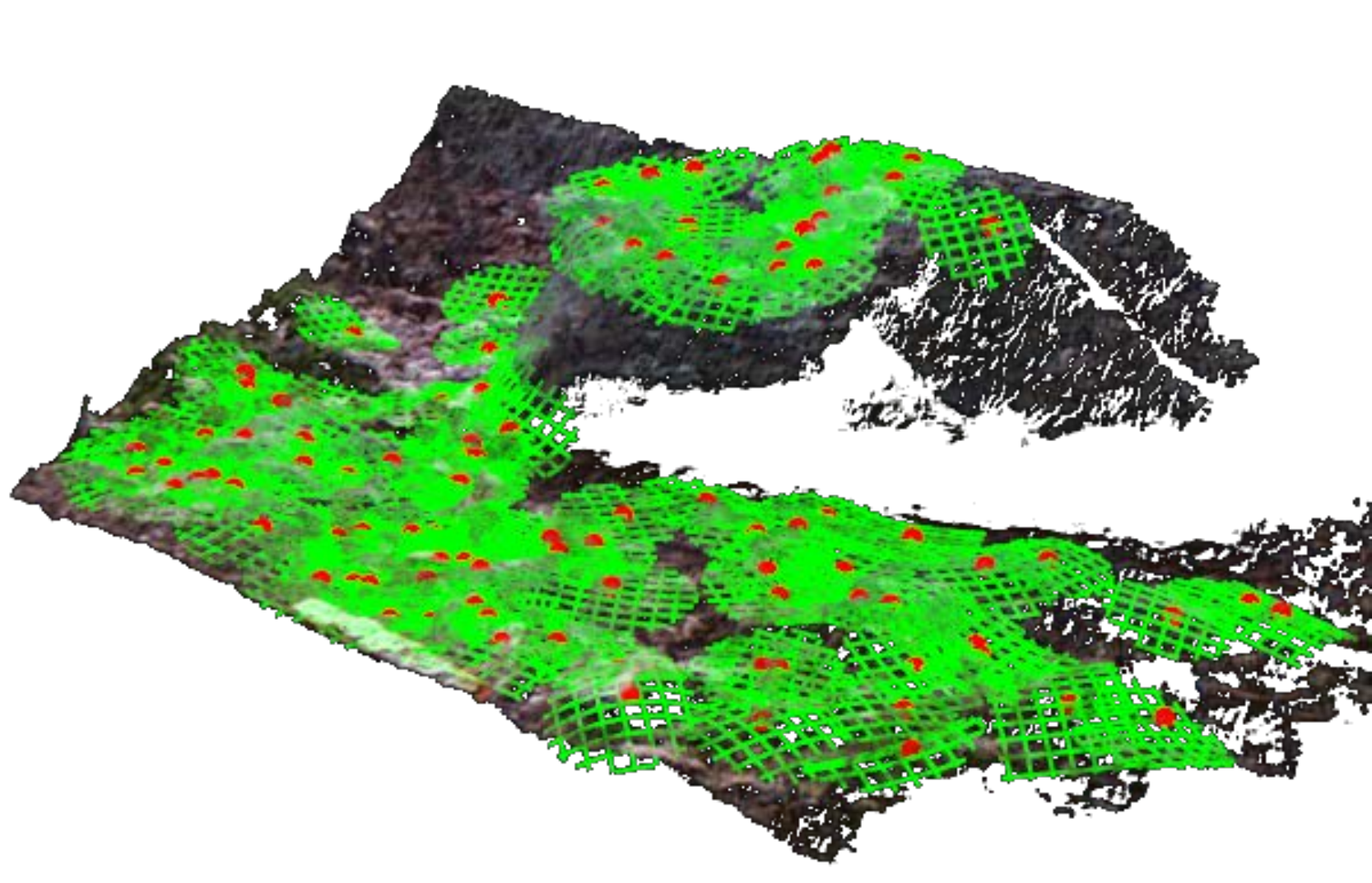}
  \end{center}
\caption[Patch fitting and validation]{100 patches are fit to a random subsampling of salient seeds (red) and are validated for quality of fit and acceptable curvature.}
\label{Fig:exp_fit}
\end{figure*}

We ran the full algorithm on the same data frames as the human-selected patches and collected statistics on the same patch properties (Figure~\ref{Fig:data} bottom and rows labeled ``auto'' in Table~\ref{Tb:stats}).  The results are similar to the human-selected patches.  In Figure~\ref{Fig:exp_fit} a set of 100 fitted patches using the human-derived saliency thresholds are illustrated.  Notice that: 1) there are no patches fitted further than $0.7m$ from the Fixation Point, 2) there are no patches in areas with big slope, and 3) there are no patches with big curvature.  In a way this is by construction,\footnote{All values for ``auto'' are defined to fall within the corresponding ``man'' average plus (minus for $\kappa_\mathrm{min}$) $3\sigma$.} but it does help establish that the algorithm can work as intended.  In total 82052 patches were fit across 832 data frames, meaning (since $n_s=100$) that about 1.4\% of patches were dropped due to the curvature, residual, and coverage checks (algorithm Steps 18, 19, and 20).  This relatively low number indicates that the saliency checks performed prior to fitting (DoN, DoNG, and DtFP) have an additional benefit in that they help reduce time wasted fitting bad patches.  In the experiments in Section~\ref{Sec:map_exp_struct} where patches were fit purely at random either 3\% (for triangle mesh-based neighborhoods) or 10\% (for K-D tree neighborhoods) of patches were dropped due to residual alone~\cite{KV13}.

\begin{table}
\begin{center}
	\begin{tabular}{|r||c|c|c|c|c|c|c|c|c|}
  	\hline
  	 	  {\bf measure} & {\bf min} & {\bf max} & {\bf med} & {\bf avg} & {\bf std} & \\ \hline
%  	$k_x$ & -7.81 & 0.00 & 0.06 & 10.09 & 1.03 & auto\\
%  	$k_x$ & -8.70 & 0.12 & 0.29 & 13.04 & 2.58 & man\\
%  	$k_y$ & -11.97 & 0.52 & 0.11 & 16.97 &2.85 & auto\\
%  	$k_y$ & -19.07 & 3.40 & 2.40 & 28.94 & 7.27 & man\\
	  DoN ($^\circ$) & 0.00 & 26.94 & 4.17 & 4.95 & 3.45 & man\\
                   & 0.00 & 15.31 & 2.83 & 3.65 & 2.85 & auto\\ \hline
	  DoNG ($^\circ$) & 0.24 & 44.85 & 11.37 & 12.34 & 7.54 & man\\
                    & 0.00 & 34.96 &  11.89 & 13.40 & 8.08 & auto\\ \hline
  	$k_{min}$ (m$^{-1}$) & -19.07 & 13.04 & -1.70 & -1.91 & 3.89 & man\\
                         & -11.97 & 7.64 & -0.87 & -1.14 & 1.75 & auto\\ \hline
  	$k_{max}$ (m$^{-1}$) & -7.87 & 28.94 & 3.78 & 4.62 & 5.02 & man\\
                         & -7.81 & 16.97 & 1.01 & 1.32 & 1.76 & auto\\ \hline
	\end{tabular}
\end{center}
\caption{}
\label{Tb:stats}
\end{table}

We observed humans walking on rocky trails and we took statistics of these four properties of the selected patches.  From these we calculated four thresholds (one per property).  A patch would be salient only if the values of its properties are in the corresponding interval of the average human-produced value plus-minus three times the standard deviation.  We then ran the full automated algorithm on the same data frames and collected statistics on the same properties (Figure~\ref{Fig:data}).  The results are statistically similar to the human-selected patches.

%\begin{figure*}[t]
%\begin{center}
%\includegraphics[width=\textwidth]{data/data.pdf}
%\end{center}
%\caption{Comparison of histograms of principal curvatures and our DoN and DoNG measures for human-selected and automatically identified patches.}
%\label{Fig:data}
%\end{figure*}
\section{Related Work}
To our knowledge the idea of sparsely mapping curved patches around a robot has not been significantly explored in prior studies.  A number of works have developed SLAM algorithms where the map features are flat surfaces in the environment~\cite{MDR04, WS06, BL08b, PBVPS10, TRC12, OE13}, though these are often either relatively large (e.g. the entire extent of a visible wall in a hallway) or relatively small (e.g. the local area around a visual point feature) compared to our approach which finds patches approximately the size of potential contact surfaces on the robot.  Patch mapping contrasts with prior work in range image segmentation~\cite{FH83, PM86, PBJB98, BG06} in that the latter seeks a maximal disjoint partition of the point cloud data into surfaces.  Our approach does not require patches on all surfaces and also allows overlapping patches so that the map can be biased to sample potentially useful parts of the environment more heavily.
\section{Summary and Future Work}
  In this chapter we presented the patch mapping system, where a spatial map of bounded curved patches are found, fit, and validated in the environment.  The map is intended to provide a reasonable sampling of potential contact patches near a robot, including task-specific criteria as we demonstrated for bipedal hiking on rocks.  Our real-time system takes dense point cloud inputs from a depth camera augmented with an IMU and outputs a sparse stream of salient patches that could be used by a task-specific contact planning algorithm (such as footfall selection for hiking).  We observed the patches that humans actually select in terrain and showed that the patches found by the system are statistically comparable.  We also investigated some other aspects of the system design, in particular, the seed selection algorithm and the choice of data structures for neighborhood searching.  K-D trees do not directly encode discontinuity information, resulting in more patches dropped due to large residual, but also produce more consistently sized neighborhoods than triangle mesh.  However, neither effect was large, and we ultimately preferred an accelerated method of neighborhood finding by image plane backprojection because of its superior runtime in practice vs k-d trees and triangle meshes.
  
  An open direction is to investigate how the pre-processing (for instance sub-sampling filtering) affects the fitting results, and whether a multi-resolution\footnote{In the style of a pyramid.} fitting method would be advantageous.  Also one of the most important aspects in the fitting and saliency process is the principal curvature calculation.  It is possible that a relatively fast approach based on integral images could be applied for estimating the curvature at every point prior to patch fitting.  This could improve the performance of the system by providing important information for patch fitting and saliency filtering.
%******************************************************************************
\chapter{Patch Tracking} \label{Ch:patch_tracking}
\begin{figure*}[!ht]
	\begin{center}
	  \includegraphics{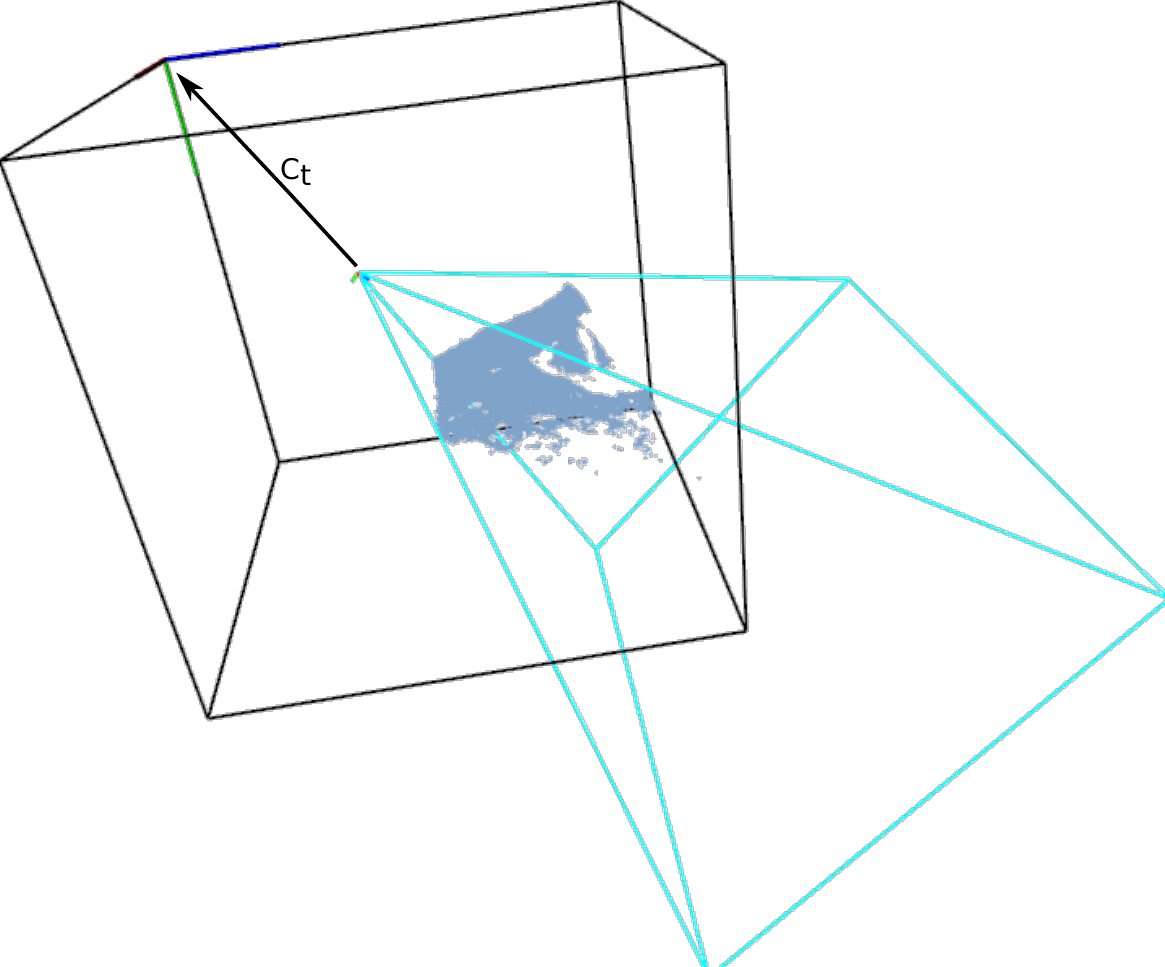}
	\end{center}
\caption[Camera pose with respect to the volume frame]{Camera pose $C_t$ with respect to the volume frame.}
\label{Fig:track_intro}
\end{figure*}

%******************************************************************************
In Chapter~\ref{Ch:patch_mapping} we introduced a method to create a map of patches in the environment around the robot.  Along with the map most locomotion applications will require \emph{patch tracking}, where patches are found and added to the map online, tracked as the robot moves and new frames are acquired, and then dropped when they are left behind.  This will complete the patch mapping and tracking system for creating and maintaining a dynamic patch map around a robot.  For solving the patch tracking problem, what is really needed to be tracked is the pose $C_t$ of the range sensor with respect to the volume frame at every frame $t$ (Figure~\ref{Fig:track_intro}).  Camera tracking is well-studied including in the context of Simultaneous Localization and Mapping (SLAM)~\cite{DB06}.  Various methods have been introduced depending on particular applications.  One challenge for a walking robot is the potential for shaking or jerky camera motion during walking.

In this Chapter we introduce a method for real-time camera tracking, using the Moving Volume KinectFusion system introduced by Roth and Vona in \cite{RV12} and implemented on a GPU\footnote{Our mini-biped robot is attached by tether to a control computer with a GPU.}, which extends the original KinectFusion system developed by Newcombe, Izadi, et al in~\cite{NIHMKDKSHF11,IKHMNKSHFDF11}.  We review this system in Section~\ref{Sec:track_amvkf} and we introduce some adjustments that were required for our walking robot system.  We then describe the patch mapping and tracking algorithm in Section~\ref{Sec:track_algo}, along with some experimental results on real rock data in Section~\ref{Sec:track_exp}.  We cover related work in Section~\ref{Sec:track_rw} and discuss the future directions in Section~\ref{Sec:track_sum}.
\section{Review of Moving Volume KinectFusion} \label{Sec:track_amvkf}
The original KinectFusion system for real-time 3D camera tracking and dense environment mapping was introduced in~\cite{NIHMKDKSHF11, IKHMNKSHFDF11}.  To briefly describe this system that was used for achieving very accurate and fast 3D mapping, we have to extend the notion of the volume which was introduced in Section~\ref{Sec:Volume} to a Truncated Signed Distance Formula (TSDF) volume~\cite{CV96}.  The TSDF volume is divided into small voxels, each one representing a portion of the physical world using two numbers: i) the distance from a physical surface (positive if it is in front, zero if it crosses, and negative if it is behind the closest surface), and ii) a confidence weight that represents the reliability of the data.  Ray casting~\cite{PSLHS98} or marching cubes~\cite{LC87} methods can then produce a point cloud that represents the surface represented in the TSDF Volume.  Two main processes alternate as new depth images are acquired (the KinectFusion system does not use the IMU, though that would be a possible extension):
\begin{enumerate}
  \item \textbf{Camera Tracking:} the camera is tracked using the Generalized Iterative Closest Point algorithm (GICP)~\cite{SHT09}, giving the camera-to-volume transformation $C_t$ at any frame $t$.
  \item \textbf{Data Fusion:} the distance and confidence values are updated in all TSDF voxels observed in the newly acquired depth image.
\end{enumerate}

\begin{figure*}[!ht]
	\begin{center}
	  \includegraphics[width=\textwidth]{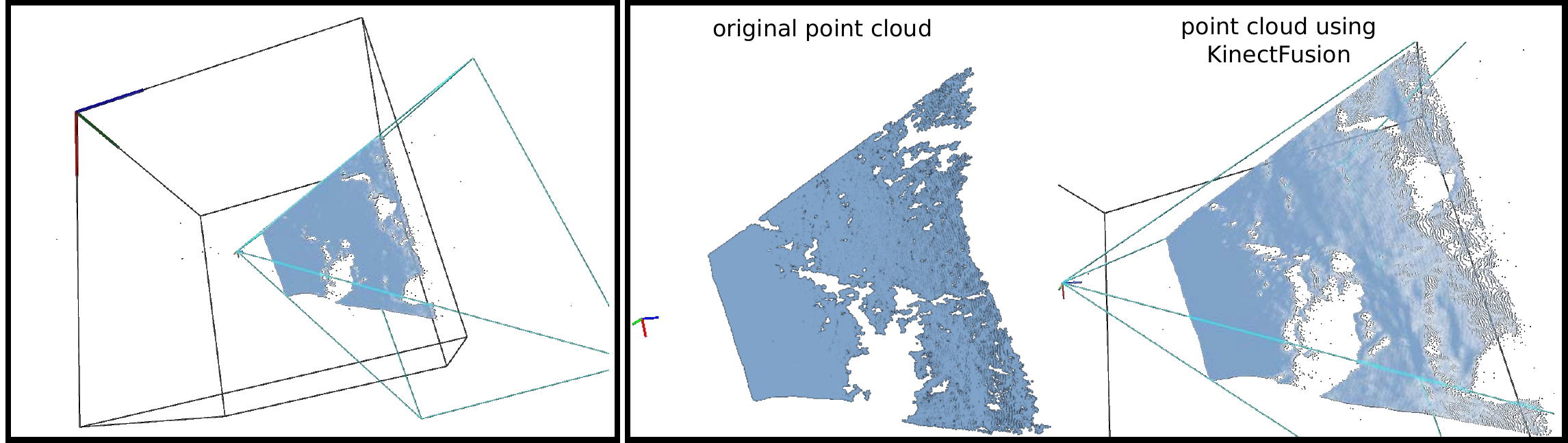}
	\end{center}
\caption[KinectFusion point clouds]{Left: the TSDF volume on a rocky trail.  Only the portion raycast from the camera viewport is visible.  Right: the original point cloud and the point cloud using KinectFusion data integration; note that the data with KinectFusion are smoother than the original and that small holes are filled.  The big empty spaces in the cloud are due to missing input data from the range sensor at the particular spots during all the previous frames.}
\label{Fig:track_kf}
\end{figure*}

\begin{figure*}[!h]
	\begin{center}
	  \includegraphics[width=\textwidth]{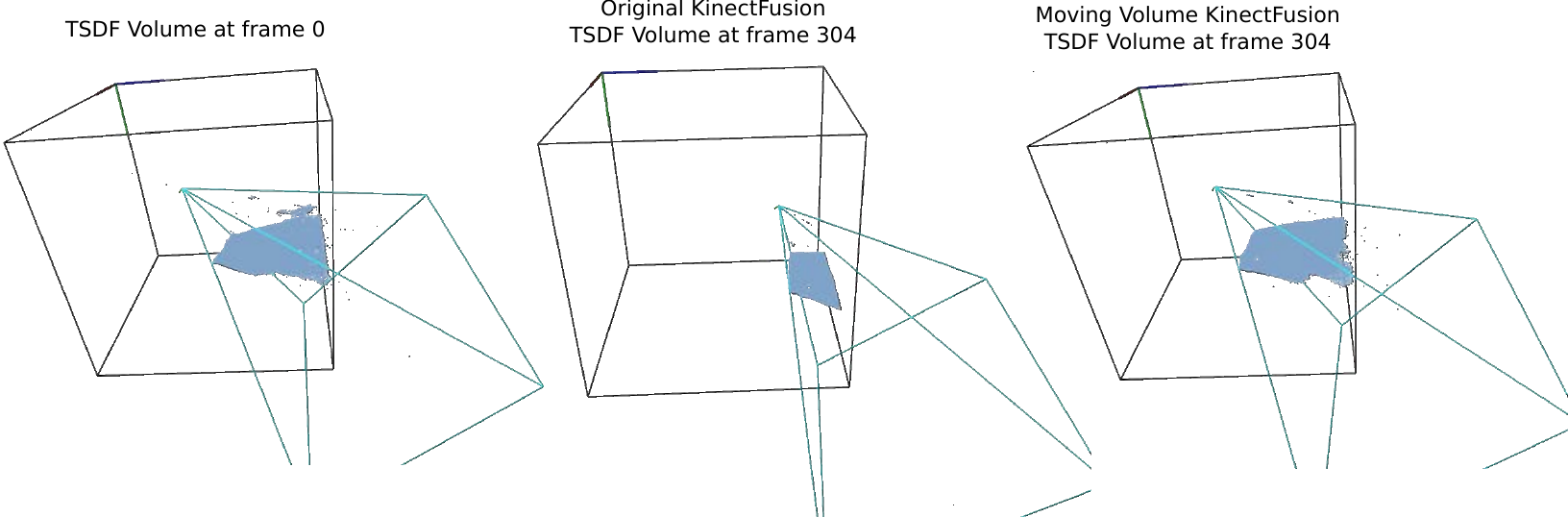}
	\end{center}
\caption[Moving volume KinectFusion vs the original KinectFusion system]{Moving volume KinectFusion, where the camera remains always close to the starting position vs the original KinectFusion system, where the camera moves in the Volume, failing to acquire more data when it reaches the edges.}
\label{Fig:track_mvkf}
\end{figure*}

This system was implemented on a GPU achieving real-time performance as well as impressive camera tracking results which can handle shaking during locomotion.  The data integration process fills holes in the cloud (Figure~\ref{Fig:track_kf}) and also provides outlier rejection.  The disadvantage of the original system is that the TSDF volume was fixed in the physical space.  For a robot moving in the environment, a volume that moves with it and keeps only the information around it for local locomotion purposes is required.  These features were introduced in Moving Volume KinectFusion~\cite{RV12}, where the TSDF volume is not fixed in the environment, but using the moving volume policies we introduced in Section~\ref{Sec:Volume}, it can move with the robot, by remapping (translating and rotating) the volume when needed.  The remapping leaves the camera and the cloud fixed relative to the physical world, but moves the TSDF volume by applying a rigid transform (Figure~\ref{Fig:track_mvkf}).  Note that as explained in~\cite{RV12} this is not a typical SLAM system~\cite{DB06}, but more a 3D Visual Odometry~\cite{SF11} one, since loop-closure is not handled.

This system is ideal for our purposes, except that 1) we require a task-specific way to set the inputs to the moving volume policies, and 2) a point cloud around and under the robot's feet will be required, and not only where the real camera is facing.  To handle these two requirements we modified the original system as described below.

\section{Adaptations to Moving Volume KinectFusion}
\begin{figure*}[!ht]
	\begin{center}
	  \includegraphics{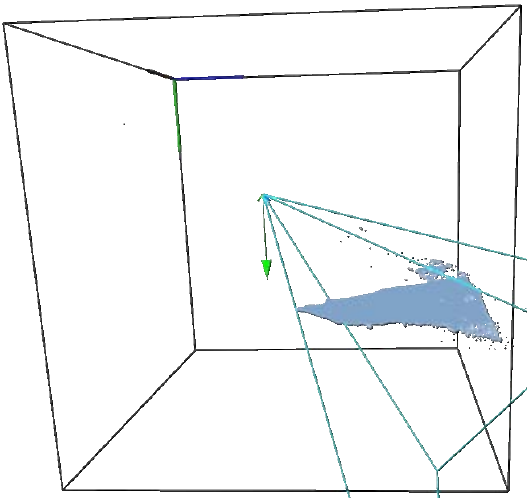}
	\end{center}
\caption[The gravity vector from the IMU sensor]{The gravity vector from the IMU sensor (green arrow) is considered as the down vector to be aligned with volume's $\vec{y}$-axis.}
\label{Fig:track_downgrav}
\end{figure*}

The TSDF Volume moving policies \textbf{fd} and \textbf{ff} as introduced in~\cite{RV12} and briefly described in Section~\ref{Sec:Volume} require a down vector for keeping the volume's $\vec{y}$-axis aligned to when the remapping takes place.  This down vector may be defined in various ways depending on the application.  In our purpose we would like the volume to be aligned with the gravity vector since we assume that the robot is locomoting in a standing-like pose.  For this purpose the first adaption in the original moving volume KinectFusion algorithm is to consider the gravity vector coming from the IMU as the down vector and not the volume's $\vec{y}$-axis which was used in~\cite{RV12} (Figure~\ref{Fig:track_downgrav}).

\begin{figure*}[!ht]
	\begin{center}
	  \includegraphics{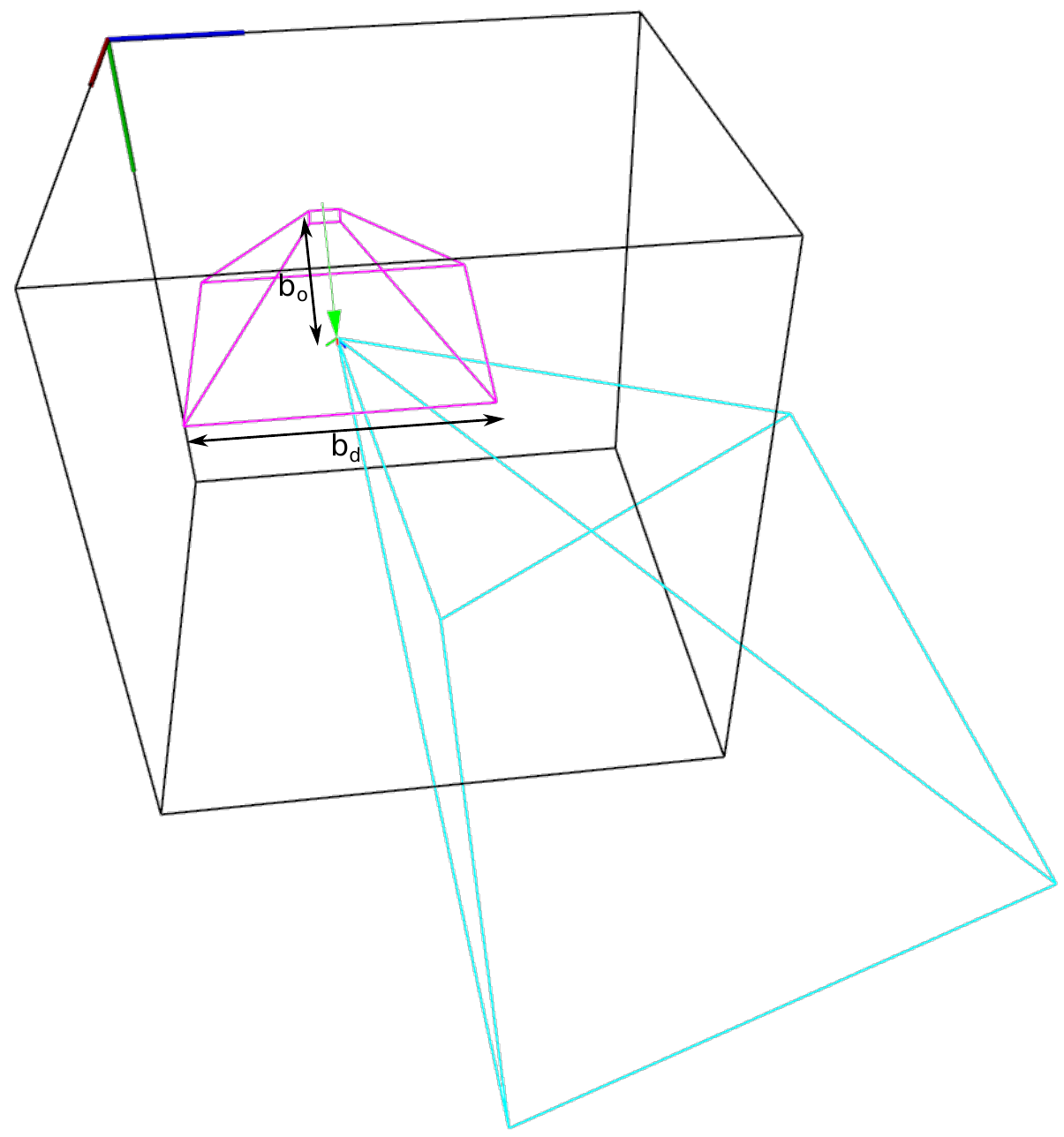}
	\end{center}
\caption[The frustum of the virtual birds-eye view camera]{The frustum of the virtual birds-eye view camera (magenta) is described by: i) its offset distance $b_o$ from the real camera, and ii) its resolution $b_r$.}
\label{Fig:track_bubble_prism}
\end{figure*}

The second adaption has to do with the raycast point cloud recovered from the TSDF volume, which by default is performed from the real camera viewpoint.  As we mentioned above this method does not produce points near the feet of the robot if the camera is not looking in that direction.  As long as the TSDF volume voxels have already captured some surface information from previous frames, we could raycast from a virtual birds-eye view (Figure~\ref{Fig:track_bubble_prism}).

To define a virtual birds-eye view camera, we first let its reference frame to be axis-aligned with the TSDF volume frame but with its $\vec{z}$-axis pointing down (along the volume frame $\vec{y}$-axis).  The center of projection of the virtual camera is at a fixed offset distance $b_o$ above the location of the real camera.  The width and height (in pixels) of the virtual camera are set as fixed resolution $b_r = 200$px.

\begin{figure*}[!ht]
	\begin{center}
	  \includegraphics[width=\textwidth]{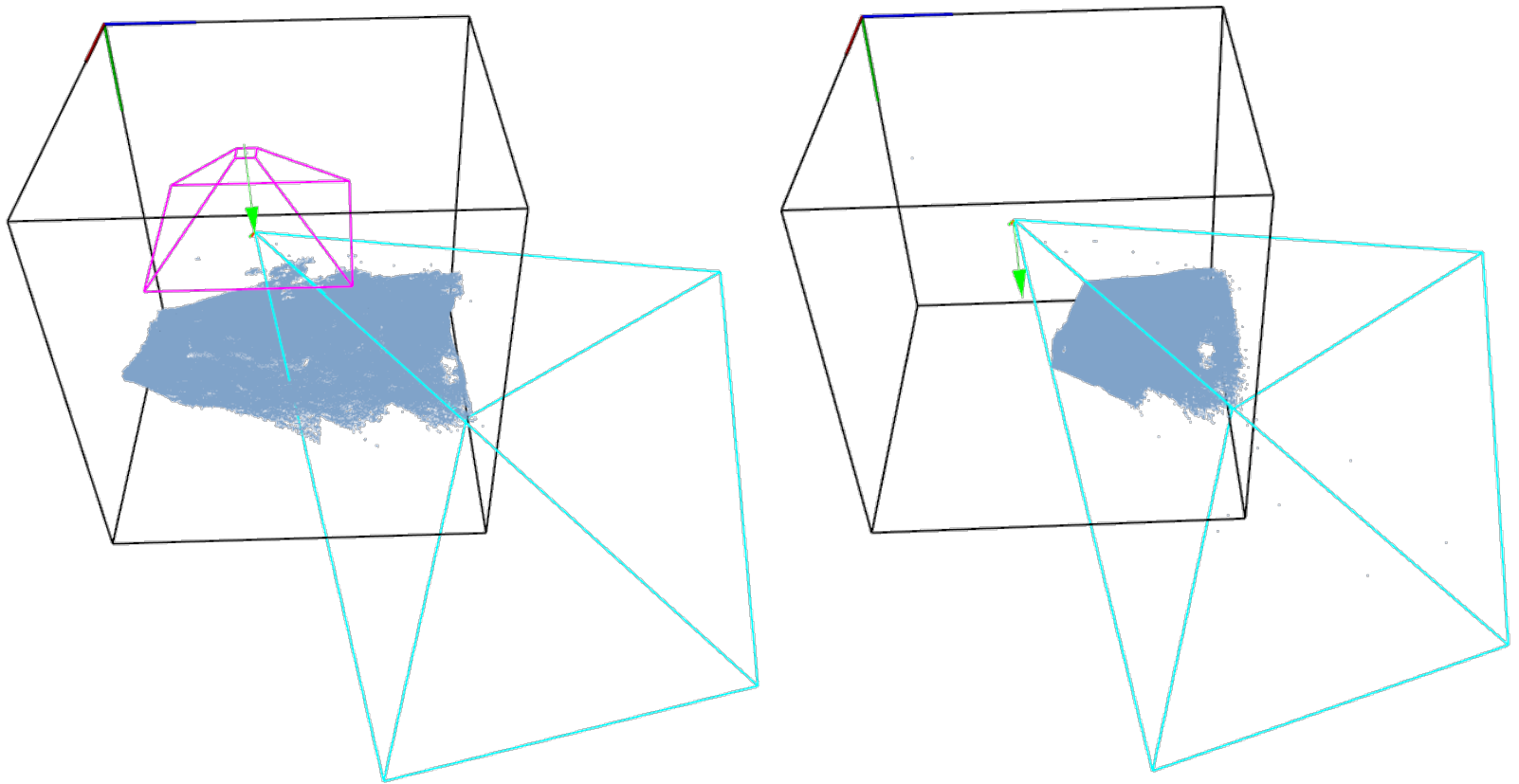}
	\end{center}
\caption[Bubble camera vs real camera ray casting]{Bubble camera (left) vs real camera point cloud ray casting (right).}
\label{Fig:track_bubble}
\end{figure*}

The result of using the virtual camera above the robot instead of the real one appears in Figure~\ref{Fig:track_bubble}, where the point cloud covers the surrounding area around and under the robot (since the physical camera is carried in the robot's head).  In that way patches can be fit under and around the feet even when the real camera is not facing in that direction.  Note that, as we mentioned in Section~\ref{Sec:map_saliency}, humans are performing perception in very similar ways, by fixating two steps ahead when locomoting in rough terrain, while considering step contact areas close to their feet, visually acquired before they reach them.
\section{Patch Mapping and Tracking Algorithm} \label{Sec:track_algo}
We now present the full patch mapping and tracking algorithm using the mapping system introduced in Chapter~\ref{Ch:patch_mapping} and the camera tracking and data fusion methods above.  The \textbf{inputs} are described in Table~\ref{Tb:stats}, while the \textbf{output} is a set of patches in the volume reference frame.

\begin{table}
\begin{center}
	\begin{tabular}{| p{7.5cm} || c |}
  	\hline
  	 	  {\bf Input} & {\bf Symbol} \\ \hline
  	 	  Initial camera pose in the volume frame. & $C_0$ \\ \hline
	      The TSDF Volume and cubic voxel size. & $V_s$, $X_{s}$ \\ \hline
	      The TSDF volume moving policy along with its (angle and position) thresholds, if any. & $\{fv,fc,fd,ff\}$, $c_d$, $c_a$ \\ \hline
	      The virtual camera frame offset from the real camera and resolution. & $b_o$, $b_r$ \\ \hline

	      Maximum (per frame) clock time for patch mapping. & $t_{s}$, $t_{m}$ \\ \hline
	      Maximum number of patches in the map and per cell. & $n_s$, $n_g$ \\ \hline
	      The distance for culling patches behind the heading camera vector ($\vec{z}$-axis). &  $d_{cp}$ \\ \hline	      
	      
	      The patch fitting options as described in Chapter~\ref{Ch:EnvRep}. & $r$, $s$, $b$, $\Gamma$, $V_g$, $n_f$, $d_{max}$, $\rho$, $w_c$, $\zeta_i$, $\zeta_o$, $T_p$\\ \hline
	      The saliency options as described in Section~\ref{Sec:map_saliency}. & $l_d$, $l_f$, $R$, $\phi_{d,g}$, $\kappa_{min,max}$ \\ \hline
	\end{tabular}
\end{center}
\caption{}
\label{Tb:inputs}
\end{table}

\newcounter{pmt-stage}
\newcounter{pmt-step}

\begin{list}{\bf Stage \Roman{pmt-stage}:}{\leftmargin=4em \usecounter{pmt-stage} \setcounter{pmt-step}{0}}
  \item System Initialization
    \begin{list}{\it Step \arabic{pmt-step}:}{\usecounter{pmt-step} \setcounter{pmt-step}{0}}
      \item Initialize the camera pose to the middle of the volume looking down at an angle corresponding to the viewpoint of the robot when standing in a default pose.
       
      \item Initialize the selection grid on the volume frame $\vec{xz}$ plane.
    \end{list}
\end{list}

\textbf{Runs for every new frame $t$} (up to the $30Hz$ input rate of the depth camera):
\begin{list}{\bf Stage \Roman{pmt-stage}:}{\leftmargin=4em \usecounter{pmt-stage} \setcounter{pmt-stage}{1}}
  \item Data Acquisition
    \begin{list}{\it Step \arabic{pmt-step}:}{\usecounter{pmt-step} \setcounter{pmt-step}{1}}
      \item Acquire a new frame of RGB-D and IMU data.
    \end{list}
        
  \item Patch Tracking
    \begin{list}{\it Step \arabic{pmt-step}:}{\usecounter{pmt-step} \setcounter{pmt-step}{2}}
      \item Get the new camera pose $C_t$ with respect to the Volume frame.
      
      \item Update the TSDF Volume voxels with the fused data.

      \item Remap (i.e. translate and/or rotate) the volume if it is required according to the moving policies, applying a rigid body transform.
            
      \item \textbf{If} the TSDF Volume was remapped:
      \begin{itemize}
        \item Update the position of each patch relative to the volume frame using the same rigid body transform.
        \item Remove the patches that have moved outside the volume.  Optionally remove patches that are further than $d_{cp}$ behind the camera (and thus behind the robot, with the assumption that the camera is forward-facing).
        \item Update the association of existing patches to grid cells in the volume frame $\vec{xz}$-plane.
        
      \end{itemize}
         
    \end{list}

  \item Patch Mapping
    \begin{list}{\it Step \arabic{pmt-step}:}{\usecounter{pmt-step} \setcounter{pmt-step}{7}}
      \item Create a point cloud by raycasting in the TSDF from the birds-eye view camera.
  
      \item Find, fit, and validate salient patches using the patch map method described in Chapter~\ref{Ch:patch_mapping}.  Note that if either the clock time limit $t_m$ exceeds or the maximum number of patches $n_s$ were fitted in the map, we proceed with the next frame. 
    \end{list}
\end{list}

If the moving volume KinectFusion looses track, the whole map gets reset and the system is initialized again and proceeds fro the beginning.
\section{Experimental Results} \label{Sec:track_exp}
\begin{figure*}[!t]
	\begin{center}
	  \includegraphics[width=\textwidth]{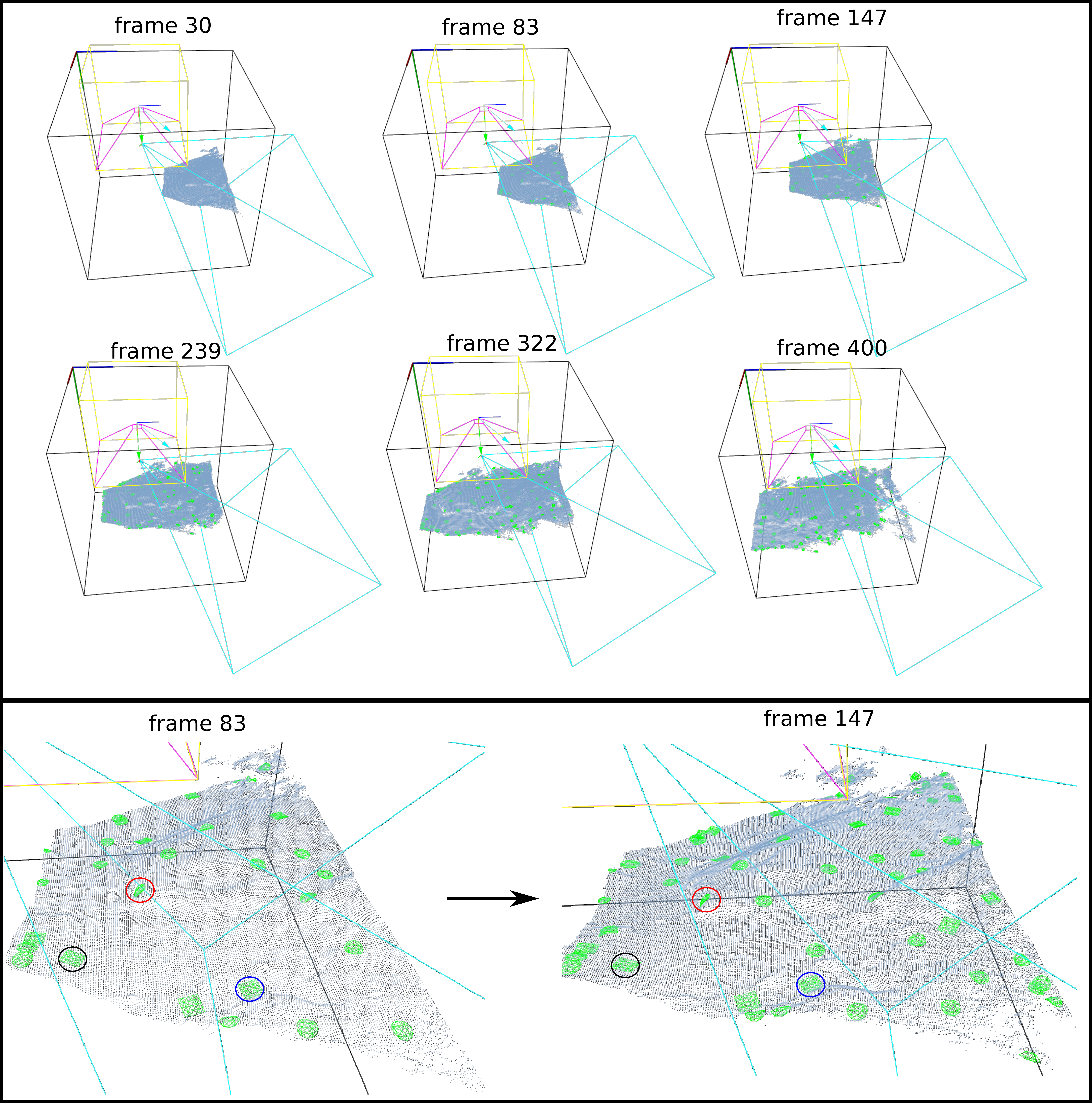}
	\end{center}
\caption[Patch Mapping and Tracking]{Patch mapping and tracking.  \textbf{Upper:} Salient patches are found, fit, and validated in a rocky trail environment, and then they are tracked while the sensor is moving.  Six frames give an overview of the overall approach.  \textbf{Lower:} A close up example of two frames.  Three patch tracking examples in black, red, and blue circles are indicated in the figure.}
\label{Fig:track_exp}
\end{figure*}

We run the patch mapping and tracking on the recording of rocky trails (Section~\ref{Sec:map_exp}) and we show some qualitative results in Figure~\ref{Fig:track_exp}.  The parameter values were: $C_0$ as described in Equation~\ref{Eq:Vpose} with $R_0 = I_{3\times3}$ and $t_0 = [2\ 2\ -0.4]^T$, $V_s = 4m$, $X_s = \frac{4}{512}m$, $fd$ moving volume policy with $c_d = 0.3m$ and $c_a = 0.05rad$, $b_o = [0, -1, 0]$, $b_r = 200$, $t_m = 60ms$, $n_s=1000$, $n_g=1$, $V_g = 8$, $d_{cp} = 4m$, and all patch and saliency parameters as defined in the experiment in Section~\ref{Sec:Data}.

\section{Related Work} \label{Sec:track_rw}
Building a map of an initially unknown environment, using a moving robot's sensor measurements, while simultaneously localizing in the map is known as the Simultaneous Localization and Mapping (SLAM) problem, and as we mentioned above it is well studied~\cite{DB06, SF11}.  The differences between the various methods are related to the type of the robot and the sensors, the hardware availability (e.g. CPU vs GPU), the application time requirements, and the map type.  In this work we used the KinectFusion system~\cite{NIHMKDKSHF11, IKHMNKSHFDF11}, and in particular the Moving Volume KinectFusion version~\cite{RV12}, implemented on a GPU, that builds a dense 3D volumetric map of the environment, using a 3D range sensor, while tracking the pose of the sensor.  Similar systems to moving volume KinectFusion were introduced in~\cite{WKFJLM12, HF12}.  An octree representation~\cite{ZZZL12, SFCS13} made KinectFusion memory efficient, while a real-time volumetric 3D mapping system implemented on a CPU, assuming known camera poses, has been recently introduced in~\cite{SSC14}.  In contrast to dense-map approaches, feature-based systems for RGB-D cameras were developed~\cite{EHESCB12}, though these may be less accurate than the dense approaches.  For monocular RGB cameras PTAM~\cite{KM07, KM09} and DTAM~\cite{NLD11} are state-of-the-art systems for CPU-based sparse and GPU-based dense mapping and tracking. 
\section{Summary and Future Work} \label{Sec:track_sum}
In this Chapter we presented a novel way to track patches in a map using the moving volume KinectFusion system.  This completes the patch mapping and tracking algorithm that is used by our bipedal robot in Chapter~\ref{Ch:biped_exp} for locomotion purposes.  Fitting patches in a single frame and tracking them may not be enough, since the data are continuously refined while the sensor is moving.  A patch refinement would be an interested future work for this system, where either the patch is re-fit to a new raycast at each frame or patch parameters are refined using TSDF voxel residuals.  It is also interesting to explore CPU-only implementations since GPUs are not always available in robotic systems.
\cleardoublepage

%******************************************************************************
\chapter{Application to Biped Locomotion} \label{Ch:biped_exp}
%******************************************************************************
Bipedal locomotion is one of the most challenging tasks in robotics.  Compared to quadrupeds and hexapods which usually have small point-like feet, bipeds usually have larger feet to support torques for balance\footnote{One exception are bipeds that are constrained by a boom, which often have small feet~\cite{ZKN12}.}  One challenge for bipedal locomotion in rough terrain is how to find potentially good footfall locations that can accommodate the feet.  In this thesis we proposed a novel patch mapping and tracking system that provides potential good areas for contact between the robot and a rough environment.  in this chapter we experimentally test our perception hypothesis with experiments on a real biped that steps on rocks.  Our lab has developed a mini 12-DoF biped robot (Section~\ref{Sec:biped_rpbp}), with a depth camera and an IMU attached, for applying our perception algorithm as part of a real-time foot selection system.  The focus of the experiments is perception and we thus use a very simple control system, where the robot uses predefined leg motion primitives (as in \cite{MLB13}) driven from the type of patch that is selected for contact.  We run two experiments.  In the first (Section~\ref{Sec:biped:exp1}) we let the robot walk open-loop on a flat area and create a spatial patch map.  In the second (Section~\ref{Sec:biped:exp2}) we train the robot to place its foot on four different types of patches on rocks.  We then place it in front of the same rocks again and let it create a patch map and find whether a match between the trained patches and those in the map exist.  If so we let it run the corresponding trained motion sequence and place its foot on the rock.

\section{RPBP Robot and Software Interface} \label{Sec:biped_rpbp}
For the locomotion experiments we use the Rapid Prototyped Biped (RPBP), which is a 12-DoF mini biped robot developed in our lab.  We briefly describe the design specifications of this platform as well as the software modules for connecting the patch mapping and tracking algorithms with the control system.

\subsection*{Rapid Prototyped Biped Robot (RPBP) Platform}
\begin{figure*}[!h]
  \begin{center}
    \includegraphics[width=\textwidth]{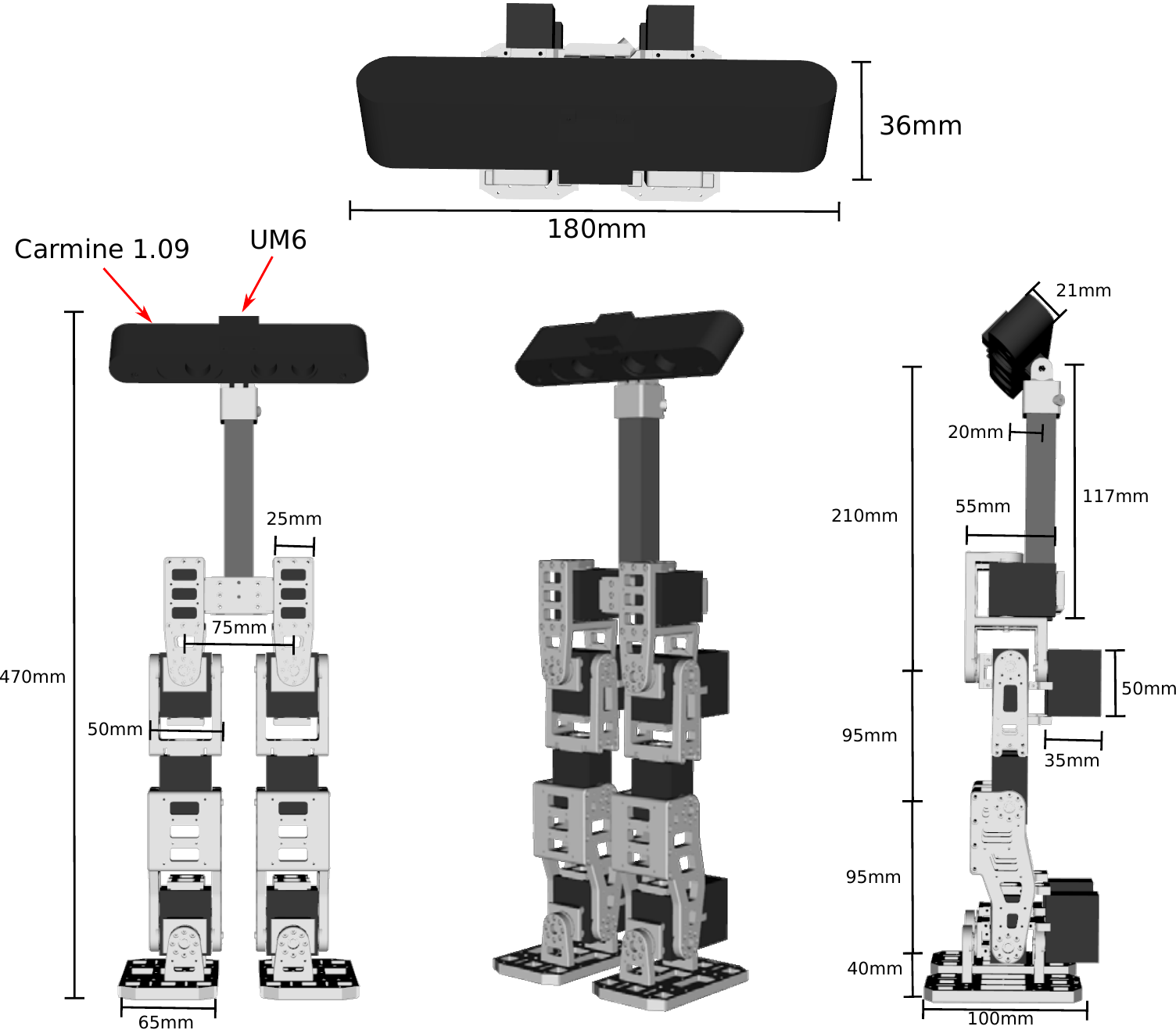}
  \end{center}
\caption{Kinematics specifications of the RPBP robot.}
\label{Fig:rpbp-model}
\end{figure*}

\begin{figure*}[!h]
  \begin{center}
    \includegraphics[width=\textwidth]{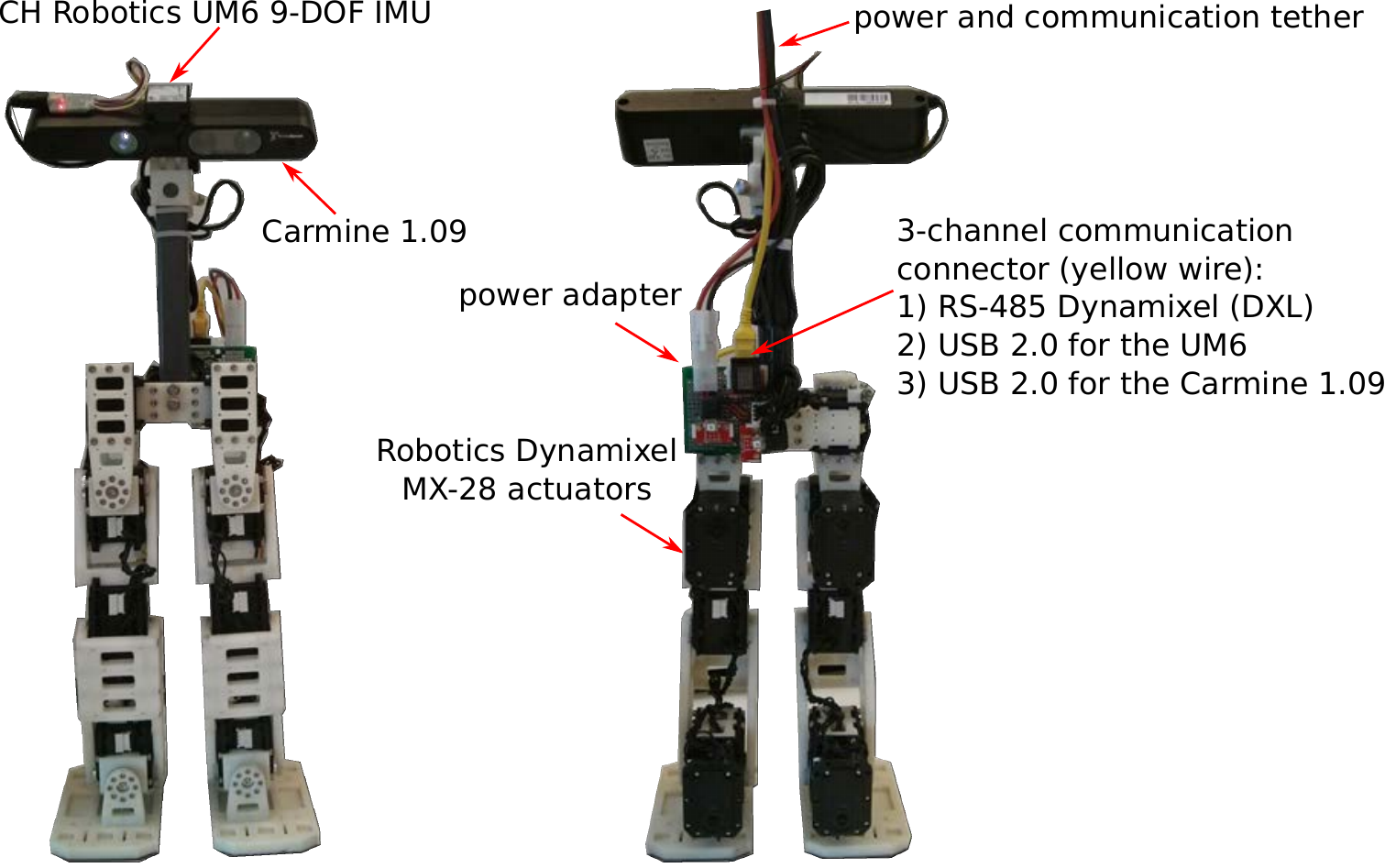}
  \end{center}
\caption{Physical hardware of the RPBP robot.}
\label{Fig:rpbp-hw}
\end{figure*}

RPBP (Figures~\ref{Fig:rpbp-model},~\ref{Fig:rpbp-hw}) is a 3D printed mini-biped robot.  It is $47cm$ tall and it weights around $1.7kg$.  It has two 6-DoF legs kinematically similar to the DARwIn-OP humanoid~\cite{HTA11}.  We use Robotics Dynamixel MX-28 actuators with high resolution magnetic rotation sensors and PID control.  We also use the short-range Carmine 1.09 depth camera with a mounted CH Robotics UM6 IMU sensor as described in Chapter~\ref{Ch:input}.  The robot does not have an on-board CPU.  We use off-board power and a 3-channel communication tether between the robot and an external computer that includes: 1) RS-485 Dynamixel (DXL) communication for controlling the motors, 2) USB 2.0 communication for the UM6 IMU, and 3) USB 2.0 for the Carmine 1.09 camera.

\subsection*{Software Interface}
\begin{figure*}[!]
  \begin{center}
    \includegraphics[width=\textwidth]{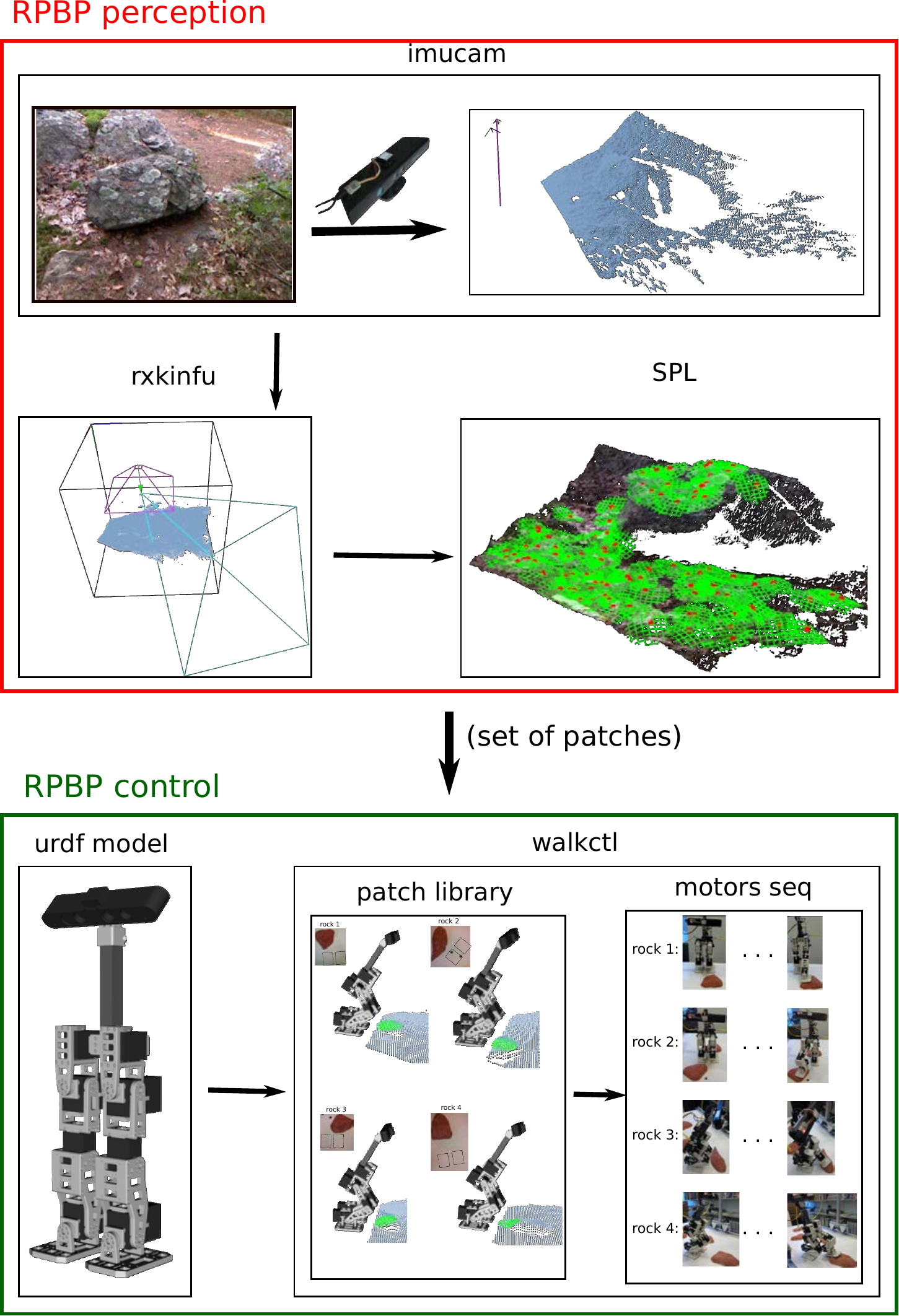}
  \end{center}
\caption{Software interface for the RPBP robot.}
\label{Fig:sys_interface}
\end{figure*}

The software interface for patch mapping and tracking and robot control system has been developed in C++, using the PCL~\cite{RC11} library.  It is divided into two big subsystems: perception and control (Figure~\ref{Fig:sys_interface}).  The perception system includes three libraries: i) imucam, ii) rxkinfu, and iii) SPL (Surface Patch Library)~\cite{SPL}.  The imucam library, developed in our lab, builds on PCL and implements an RGB-D+IMU frame grabber for the Carmine 1.09 camera and the UM6 IMU sensor, providing 30fps depth and 100fps IMU data (Chapter~\ref{Ch:input}).  The rxkinfu library implements the modified moving volume Kinect Fusion system (Chapter~\ref{Ch:patch_tracking}) providing a real-time dense 3D mapping and tracking system.  It was developed in our lab based on the kinfu code from PCL.  As input it gets the frames coming from imucam.  Finally SPL implements the patch mapping system (Chapter~\ref{Ch:patch_mapping}) where salient patches are fit to the environment and tracked using rxkinfu.

The perception system provides a set of patches to the control system, which is divided into two parts: i) a URDF (Unified Robot Description Format)~\cite{URDF} model of the robot and the dxrobot library, also developed in our lab, for Robotis Dynamixel-based communication, and ii) the RPBP walk control library.  The latter includes a patch library, i.e. patches in fixed positions relative to the robot and a library of corresponding predefined motions for each patch.  The walk control system is responsible for finding matches between the patches from the perception system and those in the library and executing the corresponding motion sequence.

\section{Rock Patch Tracking} \label{Sec:biped:exp1}
\begin{figure*}[h]
  \begin{center}
    \includegraphics[width=\textwidth]{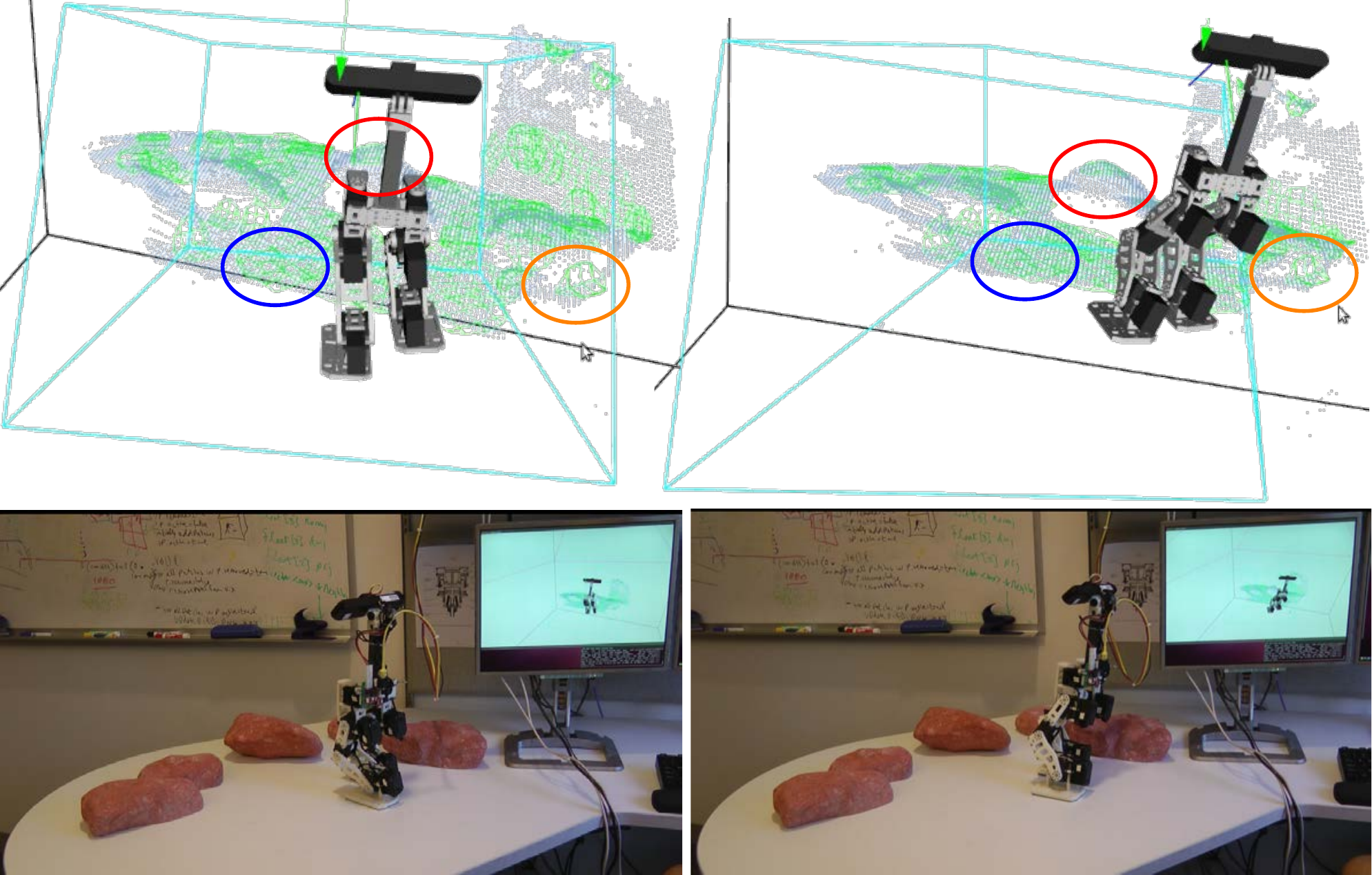}
  \end{center}
\caption[RPBP robot rock patch tracking]{The RPBP robot is walking on a flat table, while a patch map is created.  The figures represent different frames examples of patches tracked in circles.}
\label{Fig:circular_rocks}
\end{figure*}

For the first experiment we let the robot walk on a flat table using a predefined motion sequence.  The table includes four rocks that do not come in contact with the robot during locomotion.  When the robot is moving a map of patches is created.  We split the whole environment into an $8\times8$ grid and we let the map contain one seed point per cell.  The purpose of this experiment is to understand whether the shaking and the vibrations affect the patch mapping and tracking process.  For this we visually check particular patches (Figure~\ref{Fig:circular_rocks}) while the robot is moving, making sure that they are tracked correctly during the run.  A more precise evaluation would be to quantitatively measure the camera drifts in a way similar to how the original moving volume Kinect Fusion system was validated~\cite{NIHMKDKSHF11, RV12}.

\section{Foot Placement on Rock Patches} \label{Sec:biped:exp2}
\begin{figure*}[t]
  \begin{center}
    \includegraphics{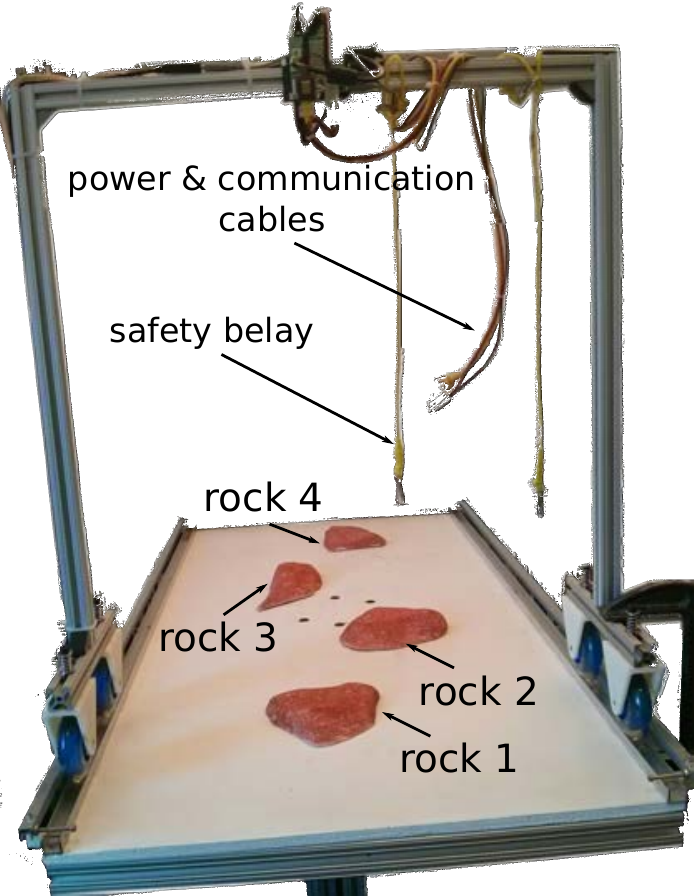}
  \end{center}
\caption[Four rocks table apparatus]{Table apparatus with four rocks in fixed positions, a safety belay for the robot, and the power and communication cables.}
\label{Fig:4rocks_apparatus}
\end{figure*}

In the second experiment we test the ability of the robot to use the real-time patch mapping system integrated in a foot placement application.  Our apparatus (Figure~\ref{Fig:4rocks_apparatus}) includes a table with 4 solid rocks in fixed positions.  The robot is always attached to a safety belay, but this does not affect its motion significantly, i.e. it does not hold it upright during the run.  We developed a simple control system where the robot executes a set of predefined motions, we manually trained by creating a library of patches and a motion sequence for each.

\subsection*{Foot Placement Training}
\begin{figure*}[!h]
  \begin{center}
    \includegraphics{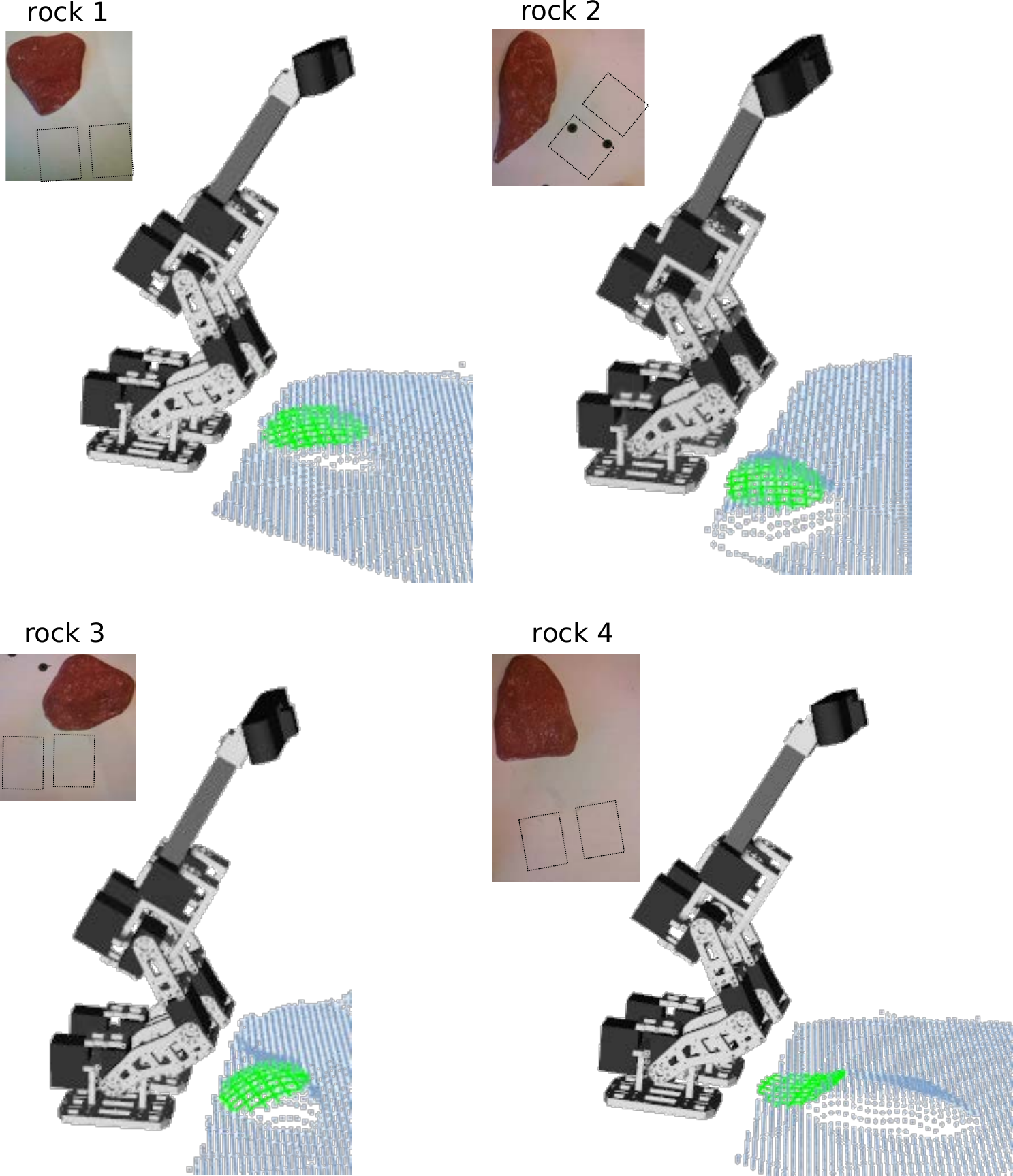}
  \end{center}
\caption[Trained patches for foot placement]{The robot at the lookdown pose, in front of four rocks, with the trained patches fit at the contact areas where foot placement will take place.}
\label{Fig:4rocks_training}
\end{figure*}

\begin{figure*}[!h]
  \begin{center}
    \includegraphics{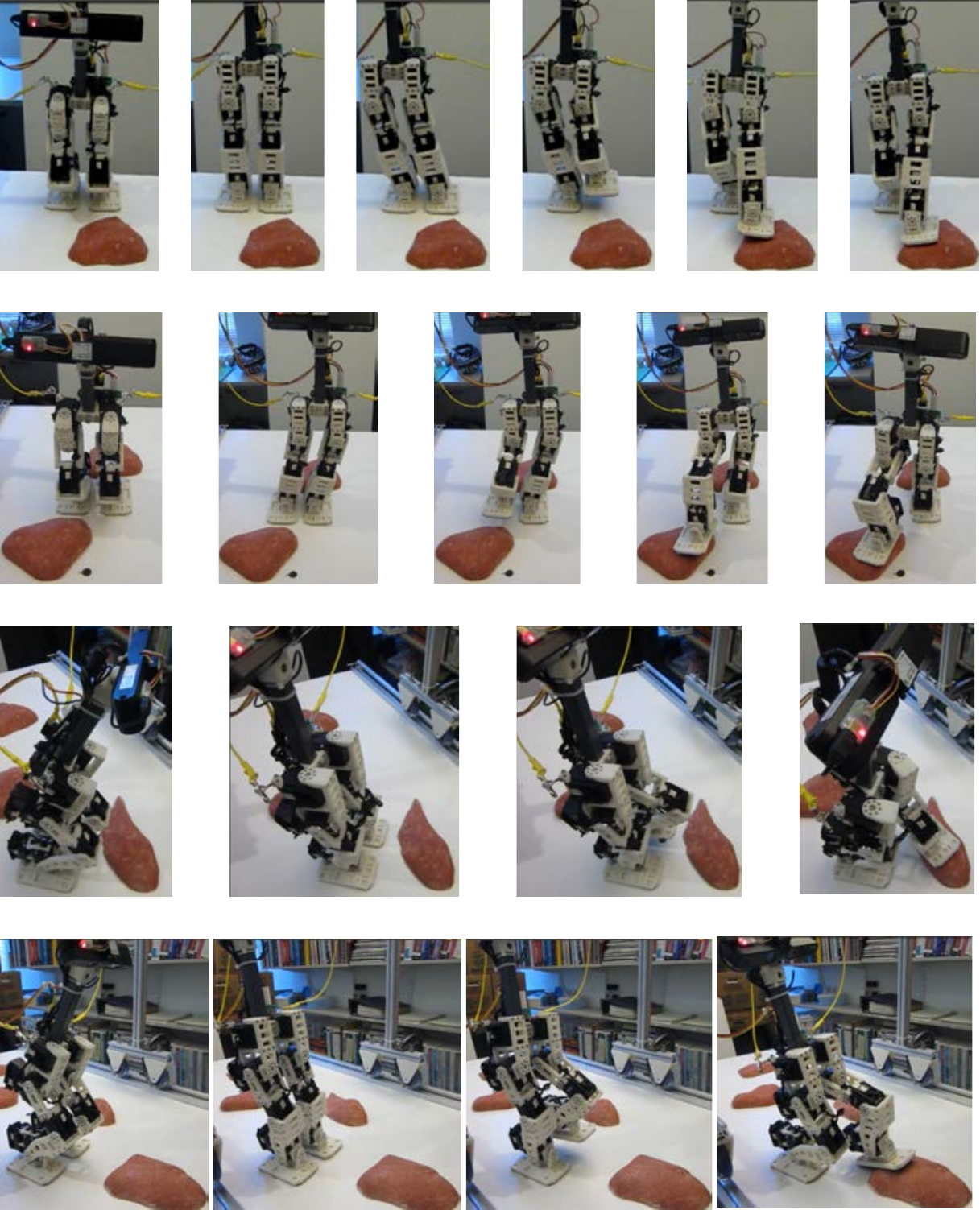}
  \end{center}
\caption[The motion sequences for foot placement on four rocks]{The motion sequences for foot placement on the four rocks using the trained patches in Figure~\ref{Fig:4rocks_training}.}
\label{Fig:steps}
\end{figure*}

We let the \emph{lookdown} robot pose, as  appears in Figure~\ref{Fig:4rocks_training}, be the starting point for training on each rock.  We place the robot in front of each of the four rocks in a defined position and we let the rxkinfu system provide us with a point cloud of the environment.  For each rock we manually select a neighborhood where we would like the robot to place its foot and we fit a patch to it.  The four patches we trained the robot to recognize, appear in Figure~\ref{Fig:4rocks_training}.  For each one of them we train the robot to place its foot with a corresponding motion sequence as shown in Figure~\ref{Fig:steps}.  For the training we used the the BRBrain library~\cite{BRBrain}, which is more convenient for that purpose.  In a similar way we could train the robot to place its feet on various other positions, but this goes beyond the intention of the experiment which focuses on perception, not control.

\subsection*{Foot Placement Test}
The foot placement experiment proceeds as follows.  We place the robot in the lookdown pose in front of each rock, roughly in the same position as it was trained in.  We then let the perception system to create a patch map as it was described in Chapter~\ref{Ch:patch_mapping}, but using a different method for seed selection.  Here we consider all the points within $5cm$ from the center of each trained patch\footnote{The trained patches are stored with poses defined relative to the robot}.  In this way we map patches close to what the robot is trained to step on.  We then perform a patch matching.  We compare every patch in the map with every trained patch.  The similarity comparison between two patches proceeds as follows:  
\begin{enumerate}
  \item Check whether the patches are of the same type (elliptic/hyperbolic/cylindric paraboloid or flat).
  \item Check whether the absolute difference between their boundary parameters are smaller than a threshold $d_s = 0.015m$.
  \item Check whether the absolute difference between the curvatures are smaller than a threshold $k_s=5 m^{-1}$.
  \item Check whether the angle between their normal vectors ($\vec{z}_{\ell}$ axis) is smaller than a threshold $a_s = 20^{\circ}$.
  \item Check whether the distance between their position $\vec{t}$ (translation vector) is smaller than a threshold $r_s = 0.01m$.
\end{enumerate}

For checking the angle between the normal vector we should consider any possible symmetry.  For all patch types except planar and circular paraboloids, we need in addition to compare the angle between the $\vec{y}_{\ell}$ axes using the same threshold $a_s$, as well as the $\vec{y}_{\ell}$ axes of the patches rotated $180^{\circ}$ around their $\vec{z}_{\ell}$ axes. If any of the patch in the map matches with a trained one we execute the corresponding motion sequence.

\begin{figure*}[!h]
  \begin{center}
    \includegraphics{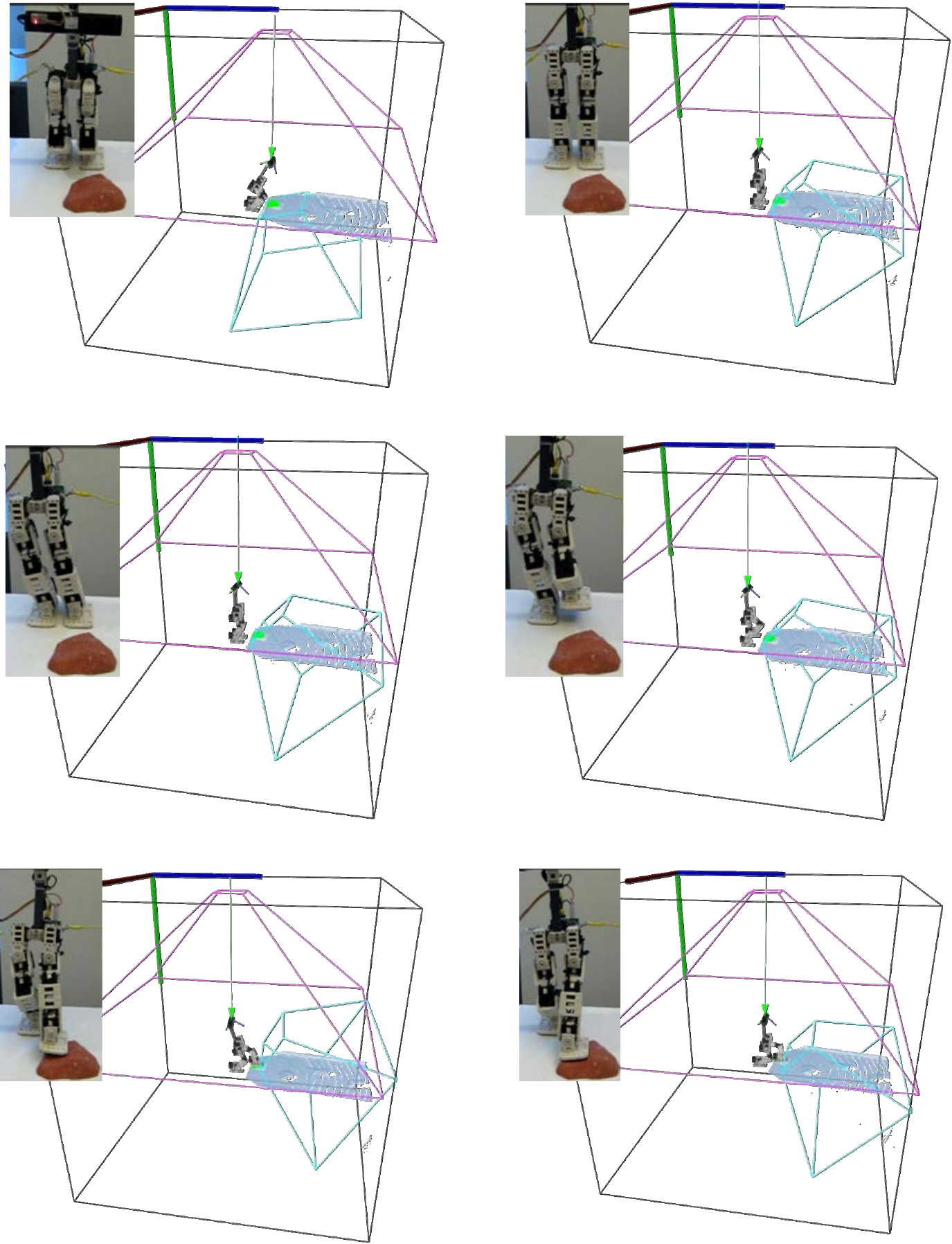}
  \end{center}
\caption[RPBP foot placement on rock 1]{RPBP detecting a patch on rock 1 for foot placement and proceeding with the predefined motion (Figure~\ref{Fig:steps}, first row).  The TSDF volume outline appears with the robot, the point cloud, the physical and virtual overhead camera frusta, and the patch at each step.}
\label{Fig:rock1_step}
\end{figure*}

We ran the experiment twenty times for each rock and the robot never failed to match the correct trained patch and successfully run the motion sequence for placing its foot every time.  Success was defined as maintaining balance and ending with the foot on the rock.  We also tried to place the robot in front of a few other rocks that it was not trained on and it stayed still, not having detected any patch match.  An example is visualized in Figure~\ref{Fig:rock1_step} for the first rock, where the robot detects a match with the corresponding trained patch and executes the motion sequence.  We also can see that the robot tracks very accurately a matched patch when it is moving.  In the last step of the visualization the foot is placed in contact with the patch.

\section{Related Work} \label{Sec:biped_rw}
The number of on-line perception systems for bipedal foot placement that are tested on real robots is very limited in the literature.  We are not aware of work on a biped that includes perception for significantly curved surfaces like natural rocks.  Though it is true that most current bipeds and humanoids have flat feet with limited ability to physically contact curved surfaces, this use case will become more important as more capable feet are developed~\cite{Gomez14}.  Some work has been done for the case of flat surfaces.  In a series of papers Okada et al.~\cite{OKII01, OII03, OOHI05} developed a perception system for detecting flat surfaces and having various humanoids stepping on them.  In their first paper a 16-DoF mini-humanoid that uses stereo vision data, detects planar segments that appear in various heights in front of the robot and performs a climbing step on them.  In their second paper they apply the same plane detection system on the HOAP-1 humanoid robot for detecting floor regions, creating a local map of polygon patches that belong to the floor and step on them.  In their third paper they apply their method on an HRP-2 robot for step climbing on horizontal flat surfaces.   Gutmann et al.~\cite{GFF04, GFF05, GFF08} have their QRIO robot detect and climb up and down on horizontal planar patches (mainly stairs) segmented using a point cloud.  Chestnutt et al.~\cite{CTSNKK09} used 3D laser point cloud data for detecting horizontal obstacles of different heights in the environment and climb on and off, using their own prototype humanoid robot, while more recently in~\cite{NCK12} HRP-2 was able to detect uneven flat surfaces with some slope and walk efficiently on them using dynamic ZMP-based walking motion.  In a series of papers~\cite{MHB12, HMB13, MLB13} and a thesis~\cite{Hornung14} the Humanoid Robots Lab in University of Freiburg has developed a system based on range sensing using a NAO robot for detecting flat obstacles and either step on or over them, using predefined motion primitives.
\section{Summary and Future Work}
In this chapter we presented experiments with our patch mapping and tracking system integrated on a real mini-biped for foot placement.  Using a simple control system where the walking motion is a human trained combination of sequences, we first train the robot to place its foot on four different patches and then we let the robot create a patch map of its environment, find matches between the patches and the trained ones, and if any exist run the predefined sequence motion.  A more advanced dynamic walking system requires balance control, possibly using the Zero Moment Point method~\cite{KKKFHYH03} and appropriate path planning.  For our experiments we assume that there are no collisions between the leg and any surface while executing the motion.  The problem of generating collision-free motions is well studied (e.g. \cite{HMB13}) and we are not considering it for these experiments, though it would be required in practical applications.

Various other directions are possible for experimental validation of the patch mapping system on the robot, for instance a comparison between our perception algorithm with a proprioceptive blind robot that tries to complete the same task~\cite{Vona11}.  It is also necessary to consider different types of foot for curved surfaces contact.  In our experiments we used flat feet, but a different design for a better contact, possibly using miltiple toes and/or compliance, may be preferable~\cite{Gomez14}.
\cleardoublepage

%******************************************************************************
\chapter{Conclusions and Future Work}
%******************************************************************************
In real world applications articulated robots need to come in contact with unstructured environments either for locomotion or manipulation.  In this thesis we introduced a novel perception system for modeling the contact areas between a robot and a rough surface and demonstrated its use in bipedal locomotion.  Our system creates in real-time a map of bio-inspired salient bounded curved patches that fit to the environment in locations of potential footfall contact and tracks them when the robot is moving using an RGB-D depth camera and an IMU sensor.

We envision our method to be part of a bigger system where not only foot placement, but also other types of contact (for instance dexterous manipulation) is driven using similar patches.  In this thesis we developed experiments which prove that the robot can find contact patches for footfall placement, but as explained in Chapter~\ref{Ch:biped_exp} high level path planning along with a more advanced control system is required for dynamic walking using these patches.  Furthermore, it is also interesting to understand how the patch uncertainty can play a role not only for data fusion while the robot is moving, but also for foot placement decisions.  For instance a highly uncertain patch may not be considered for foot placement, or the motor compliance may be adapted with respect to the level of patch uncertainty.  Last but not least, a mapping and tracking method (e.g.~\cite{SSC14, KSC13}) that does not use a GPU device (which is not always available in all autonomous robots) instead of the Kinect Fusion system may need to be used.
%\end{doublespacing}
\end{onehalfspacing}

\appendix
\cleardoublepage
\part{Appendix}
%************************************************
% Appendix
%************************************************
\chapter{Environment Representation}

\section{Jacobians} \label{Sec:jacobians}
We calculate the Jacobian of the exponential map (\ref{Eq:rexp}) as a $[3\times3]\times3$ row tensor\footnote{(\ref{Eq:drexp}) remains finite as $\theta \rightarrow 0$.  Small angle approximations for $\alpha$ and $\beta$ were given in Section~\ref{Sec:pose}; their derivatives can be approximated as $\pard{\alpha}{\vec{r}} \approx (\theta^2/30-1/3)\vec{r}^T$ and $\pard{\beta}{\vec{r}} \approx (\theta^2/180-1/12)\vec{r}^T$.}:
\beg \label{Eq:drexp}
  \pard{R}{\vec{r}} =
    [\vec{r}]_\times \pard{\alpha}{\vec{r}} + \pard{[\vec{r}]_\times}{\vec{r}} \alpha +\
    [\vec{r}]_\times^2 \pard{\beta}{\vec{r}} + \pard{[\vec{r}]_\times^2}{\vec{r}} \beta \\
  \pard{\alpha}{\vec{r}} = \frac{\theta\cos\theta-\sin\theta}{\theta^3} \vec{r}^T, \
  \pard{\beta}{\vec{r}} = \frac{\theta\sin\theta+2\cos\theta-2}{\theta^4}\vec{r}^T \nonumber\\
  \begin{aligned}\nonumber
    \pard{[\vec{r}]_\times}{\vec{r}} &= \begin{bmatrix}
      [  \vecmb{0}\  \uvec{z}\ -\!\!\uvec{y}]\ %
      [-\uvec{z}\   \vecmb{0}\  \uvec{x}]\ %
      [ \uvec{y}\ -\!\!\uvec{x}\   \vecmb{0}]
      %\begin{bmatrix}0&0&0\\0&0&-1\\0&1&0\end{bmatrix}\ %
      %\begin{bmatrix}0&0&1\\0&0&0\\-1&0&0\end{bmatrix}\ %
      %\begin{bmatrix}0&-1&0\\1&0&0\\0&0&0\end{bmatrix}
    \end{bmatrix}\\
  \pard{[\vec{r}]_\times^2}{\vec{r}}&= [\vec{r}]_\times \pard{[\vec{r}]_\times}{\vec{r}} + \pard{[\vec{r}]_\times}{\vec{r}} [\vec{r}]_\times.
  \end{aligned}
\eeg

The Jacobian of (\ref{Eq:r2}) is, with $\theta_{xy}$, $\alpha_{xy}$, and $\uvec{z}_l$ from (\ref{Eq:r2}),\footnote{For small $\theta_{xy}$, $\gamma_{xy}\approx2/(3-\theta_{xy}^2/2)$.}
\beg \label{Eq:dr2}
  \pard{\vec{r}_{xy}}{\vec{r}}=
    %\begin{bmatrix}1&0&0\\0&1&0\end{bmatrix}
    \begin{bmatrix}\uvec{x}^T\\\uvec{y}^T\end{bmatrix}
    \begin{cases}
      I&\text{if }\theta_{xy}\approx\pi\\
      \pard{}{\vec{r}}\frac{\uvec{z}\times\uvec{z}_l}{\alpha_{xy}}&\text{otherwise}
    \end{cases}\\
  \pard{}{\vec{r}}\frac{\uvec{z}\!\!\times\!\!\uvec{z}_l}{\alpha_{xy}} = [\uvec{z}]_\times\!\!\left(\!\!\frac{I}{\alpha_{xy}} - \!\!RZ(\gamma_{xy}R^T(I\!\!-\!\!Z)\!+\!\!I)\!\!\right)\!\!\frac{\partial R}{\partial\vec{r}}\uvec{z} \nonumber\\
  R \defeq R(\vec{r}),\ Z \defeq \uvec{z} \uvec{z}^T,\ \gamma_{xy} \defeq \frac{\theta_{xy} - \sin\theta_{xy} \cos\theta_{xy}}{\sin^3 \theta_{xy}}.\nonumber
\eeg

Viewing (\ref{Eq:xc}) as a vector function,
\beg
  [\vec{r}_n^T\ \vec{t}_n^T \ldots \vec{r}_1^T\ \vec{t}_1^T]^T\in\R^{6n}\rightarrow[\vec{r}_c^T\ \vec{t}_c^T]^T\in\R^6,\nonumber
\eeg
its Jacobian $J_c$ is $6\!\times\!6n$ where $6\!\times\!6$ block $j$ from right to left is, with $\phi_j$ from (\ref{eq-chain}), $R_j,X_j$ from (\ref{Eq:xc}), $\frac{\partial\vec{r}}{\partial R}$ from Appendix~\ref{Sec:rlog},
\beg \label{Eq:dxc}
  \pard{[\vec{r}_c^T\ \vec{t}_c^T]^T}{[\vec{r}_j^T\ \vec{t}_j^T]^T} =
  \begin{bmatrix}
    \pard{\vec{r}_c}{\vec{r}_j} & 0\\
    \pard{\vec{t}_c}{\vec{r}_j} & \pard{\vec{t}_c}{\vec{t}_j}\\
  \end{bmatrix}
\eeg

\beg
  \pard{\vec{r}_c}{\vec{r}_j} = \pard{\vec{r}_c}{R_c} R_l \pard{R_j}{\vec{r}_j} R_r
\eeg

\beg
  \begin{aligned}
    &\text{if }\phi_j=+1 & &\text{if }\phi_j=-1\\
    \pard{\vec{t}_c}{\vec{r}_j} &= R_l \pard{R_j}{\vec{r}_j} \vec{t}_r & \pard{\vec{t}_c}{\vec{r}_j} &= R_l \pard{R_j}{\vec{r}_j}(\vec{t}_r-\vec{t}_j)\\
    \pard{\vec{t}_c}{\vec{t}_j} &= R_l &
    \pard{\vec{t}_c}{\vec{t}_j} &= -R_l R_j
  \end{aligned} \nonumber
\eeg

\beg
  R_l \defeq R_n \cdots R_{j+1}, R_r \defeq R_{j-1} \cdots R_1, R_c \defeq R_n \cdots R_1 \nonumber\\
  \vec{t}_r \defeq (X_{j-1}\! \circ \cdots \circ \!X_1)(\vecmb{0}). \nonumber
\eeg \cleardoublepage
\section{The Logarithmic Map} \label{Sec:rlog}
Due to numerical issues with other equations we found in the literature, we developed the following numerically stable algorithm to calculate the log map
\beg
  \vec{r}(R),\ %
  R\defeq
  \left[\begin{smallmatrix}
          R_{xx}&R_{xy}&R_{xz}\\
          R_{yx}&R_{yy}&R_{yz}\\
          R_{zx}&R_{zy}&R_{zz}
        \end{smallmatrix}\right] \nonumber
\eeg
and its $3\times[3\times3]$ column tensor Jacobian $\partial\vec{r}/\partial R$.

\begin{algorithmic}
\If {$i\geq maxval$}
    \State $i\gets 0$
\Else
    \If {$i+k\leq maxval$}
        \State $i\gets i+k$
    \EndIf
\EndIf
\end{algorithmic}

\begin{algorithmic}
  \State $\vec{v}=[R_{zy}\!\!-\!\!R_{yz}\ R_{xz}\!\!-\!\!R_{zx}\ R_{yx}\!\!-\!\!R_{xy}]^T$
  \State $c=(\tr(R)-1)/2, s=\|\vec{v}\|/2, \theta=\atantwo(s,c)$
  \State $\text{choose }ijk\in\{xyz,yzx,zxy\}\text{ s.t. }R_{ii}\!=\!\max(R_{xx},R_{yy},R_{zz})$
  \State $\delta=1+R_{ii}-R_{jj}-R_{kk}$
  \If {$\delta>\epsilon_\delta$} $\com{$R$ not identity, $\theta$ not small}$
    \State $\gamma\!=\!\theta(3\!-\!\tr(R))^{\frac{-1}{2}}$
    \State $d=\sqrt{\delta}$
    \State $r_i\!=\!d\gamma$
    \State $r_j\!=\!\gamma(R_{ji}\!+\!R_{ij})/d$
    \State $r_k\!\!=\!\gamma(R_{ki}\!+\!R_{ik})/d$
    \State $\vec{r}=[r_x\ r_y\ r_z]^T\com{solution up to sign}$
    
    \If {$\theta<(\pi-\epsilon_\theta)$} $\com{resolve sign by testing action of $R$}$
      \State $\vec{p}=\vec{r}\!\times\![0\ 0\ 1]^T$
      \If {$\vec{p}^T\vec{p}<1/4$}
        \State $\vec{p}=\vec{r}\!\times\![0\ 1\ 0]^T$
      \EndIf
      \If {($R\vec{p})^T(\vec{r}\times\vec{p})<0$}
        \State $\vec{r}\gets-\vec{r}$
        \State $d\gets-d$
      \EndIf
    \EndIf
    \State $\com{solution for $\vec{r}(R)$ complete, now find $\partial\vec{r}/\partial R$}$
    \State $\uvec{r}=[\hat{r}_x\ \hat{r}_y\ \hat{r}_z]^T=\vec{r}/\theta$
    \State $\pard{\theta}{R} = (c[\uvec{r}]_\times-sI)/2$
    \State $w_i=1, w_j=-1, w_k=-1$
    \State $U=\diag([w_x\ w_y\ w_z]^T)$
    \State $\gamma \pard{d}{R} = \frac{\gamma}{2d} U$
    \State $d \pard{\gamma}{R} = \hat{r}_i\left(\pard{\theta}{R} \!+\! \frac{I\theta}{6\!-\!2\tr{R}}\right)$
    \State $\pard{r_i}{R} = \gamma \pard{d}{R} + d \pard{\gamma}{R}$
    \State $V=0_{3\times3}, V_{ji}\!\gets\!\!1, V_{ij}\!\gets\!\!1$
    \State $W=0_{3\times3}, W_{ki}\!\gets\!\!1, W_{ik}\!\gets\!\!1$
    \State $\pard{r_j}{R} = \frac{\gamma}{d} V \!+\! (R_{ji} \!+\! R_{ij})(d \pard{\gamma}{R} - \gamma \pard{d}{R}) / \delta$
    \State $\pard{r_k}{R} = \frac{\gamma}{d} W \!+\! (R_{ki} \!+\! R_{ik})(d \pard{\gamma}{R} - \gamma \pard{d}{R}) / \delta$
    \State $\pard{\vec{r}}{R} = [\pard{r_x}{R} \pard{r_y}{R} \pard{r_z}{R}]^T$
  \Else $\com{small $\theta$}$
    \If {$\theta>\epsilon_\theta$}
      \State $\alpha=s/\theta$
    \Else
      \State $\alpha=1-\theta^2/6$
    \EndIf
    \State $\vec{r}=\vec{v}/(2\alpha)$ $\com{solution for $\vec{r}(R)$, now find $\pard{\vec{r}}{R}$}$
    \If {$\theta > \epsilon_\theta$}
      \State $\lambda\!=\!(s\!-\!c\theta)/(2s^2)$
      \State $\pard{\theta}{R} = (c[\vec{r}/\theta]_\times\!\!-\!sI)/2$
    \Else
      \State $\lambda=\theta/12$
      \State $\pard{\theta}{R} = (c (1_{3\times3}-I)-sI)/2$
    \EndIf
    \State $\com{using $\pard{[\vec{r}]_\times}{\vec{r}}$ from (\ref{Eq:drexp}) and Kronecker product $\otimes$}$
    \State $\pard{\vec{r}}{R} = (1/(2\alpha)) (\pard{[\vec{r}]_\times}{\vec{r}})^T + \lambda \vec{v} \otimes (\pard{\theta}{R})$
  \EndIf
\end{algorithmic} \cleardoublepage
\section{Patch Fit Uncertainty Propagation} \label{Sec:errprop}

% MACROS
\newcommand{\norm}[1]{\left\lVert#1\right\rVert}

\newcommand{\cvec}{\vec{t}}
\newcommand{\rvec}{\vec{r}}
\newcommand{\kvec}{\vec{k}}
\newcommand{\mvec}{\vec{m}}
\newcommand{\qvec}{\vec{q}}
\newcommand{\dvec}{\vec{d}}
\newcommand{\rhovec}{\mathbold{\rho}}
\newcommand{\lvec}{\vec{l}}
\newcommand{\cnew}{\vec{t}'}
\newcommand{\rnew}{\vec{r}'}
\newcommand{\qnew}{\vec{q}'}
\newcommand{\qavg}{\bar{\vec{q}}}
\newcommand{\xavg}{\bar{\vec{x}}}
\newcommand{\yavg}{\bar{\vec{y}}}
\newcommand{\xell}{\hat{x}_{\ell}}
\newcommand{\yell}{\hat{y}_{\ell}}
\newcommand{\zell}{\hat{z}_{\ell}}
\newcommand{\Rell}{R_{\ell}}
\newcommand{\xhat}{\hat{x}}
\newcommand{\yhat}{\hat{y}}
\newcommand{\zhat}{\hat{z}}
\newcommand{\rxy}{\vec{r}_{xy}}
\newcommand{\vx}{v_\vec{x}}
\newcommand{\vy}{v_\vec{y}}
\newcommand{\vxy}{v_\vec{xy}}
\newcommand{\rxynew}{\vec{r}_{xy}'}
\newcommand{\Sigmaxell}{\Sigma_{\xell}}
\newcommand{\Sigmazell}{\Sigma_{\zell}}
\newcommand{\Sigmaqavg}{\Sigma_{\qavg}}
\newcommand{\Sigmarc}{\Sigma_{\rvec, \cvec}}
\newcommand{\Sigmakrc}{\Sigma_{k, \rvec, \cvec}}
\newcommand{\Sigmakrnewc}{\Sigma_{k, \rnew, \cvec}}
\newcommand{\Sigmarxyc}{\Sigma_{\rxy, \cvec}}
\newcommand{\Sigmarxycnew}{\Sigma_{\rxy, \cnew}}
\newcommand{\Sigmakrxyc}{\Sigma_{k, \rxy, \cvec}}
\newcommand{\Sigmakrxynewc}{\Sigma_{k, \rxynew, \cvec}}
\newcommand{\Sigmakc}{\Sigma_{k,\cvec}}
\newcommand{\Sigmamkrc}{\Sigma_{\vec{m}, \vec{k},\rvec, \cvec}}
\newcommand{\Sigmamkrxyc}{\Sigma_{\vec{m}, \vec{k},\rxy, \cvec}}
\newcommand{\Sigmaqkrc}{\Sigma_{\qvec_1 \ldots \qvec_N, \vec{m}, \vec{k},\rvec, \cvec}}
\newcommand{\Sigmadrkrc}{\Sigma_{\vec{\dvec_r}, \vec{k},\rvec, \cvec}}
\newcommand{\Sigmadckrxyc}{\Sigma_{\vec{\dvec_c}, \vec{k},\rxy, \cvec}}
\newcommand{\Sigmadekrc}{\Sigma_{\vec{\dvec_e}, \vec{k},\rvec, \cvec}}
\newcommand{\Sigmamrxyc}{\Sigma_{\mvec, \kvec,\rxy, \cvec}}
\newcommand{\Sigmaellrhorxyc}{\Sigma_{\lvec, \rhovec,\rxy, \cvec}}
\newcommand{\Sigmadcrxyc}{\Sigma_{d_c, \rxy, \cnew}}
\newcommand{\Sigmadrc}{\Sigma_{\dvec,\rvec, \cnew}}

\newcounter{error-stage} \newcounter{error-step}

During the fitting process the covariance matrix of the patch parameters $\Sigma$ is calculated by first order error propagation~\cite{Meyer92} using the Gaussian uncertainty model as follows.  In each step the input covariance matrix $\Sigma$ will either come from the WLM fitting or from the previous step.

\begin{list}{\bf Stage \Roman{error-stage}:}{\usecounter{error-stage}}
  \item Fit an Unbounded Surface
  
  \begin{list}{\it Step \arabic{error-step}:}{\usecounter{error-step} \setcounter{error-step}{0}}
    % Step 1 (s = plane)    
    \item \textit{Plane fitting}\\
      \textbf{Input}: $\rxy$, $\cvec$, $\Sigmarxyc \in \R^{5 \times 5}$ (from the WLM fitting)\\
      \textbf{Output}: $\Sigmarxycnew \in \R^{5 \times 5}$\\
      
      Let
      \beg
        \vec{r} = [\rxy\ 0], \
        \qavg = avg(\vec{q_i}) = \frac{1}{N} \sum_{i=1}^{N} \vec{q_i} \nonumber \\
        \cnew = \qavg - \zell^T (\qavg - \cvec) \zell = \qavg - (\zell^T \qavg) \zell +  (\zell^T \cvec) \zell \nonumber \\
        \zell = R(\vec{r}) \hat{z} \nonumber
      \eeg
      
      The propagated covariance is:
      \beg
        \Sigma_{\rxy, \cnew} = J \Sigma J^T \in \R^{5 \times 5}
      \eeg
        
      where
      
      \beg
        J = \begin{bmatrix}
              \vecmb{0}_{2 \times 3} & \vecmb{0}_{2 \times 3} & I_{2 \times 2} & \vecmb{0}_{2 \times 3}\\
              \frac{\partial{\cnew^T}}{\partial{\zell}} & \frac{\partial{\cnew^T}}{\partial{\qavg}} & \frac{\partial{\cnew^T}}{\partial{\rxy}} & \frac{\partial{\cnew^T}}{\partial{\cvec}}
            \end{bmatrix}
        \in \R^{5 \times 11}
      \eeg

      \beg
        \pard{\cnew^T}{\zell} = \zell (\cvec - \qavg)^T  + \zell^T (\cvec - \qavg) I_{3 \times 3} \nonumber
      \eeg

      \beg
        \pard{\cnew^T}{\qavg} = I_{3 \times 3} - \zell \zell^T \nonumber
      \eeg

      \beg
        \pard{\cnew^T}{\rxy} = \pard{\cnew^T}{\zell} \pard{\zell}{\rxy} = \pard{\cnew^T}{\zell} (\pard{R}{\rxy} \zhat) \nonumber
      \eeg

      \beg
        \frac{\partial{\cvec^T}}{\partial{\cvec}} = \hat{z}_{\ell} \hat{z}_{\ell}^T \nonumber
      \eeg

      and

      \beg
        \Sigma = \begin{bmatrix}
                   \Sigmazell             & \vecmb{0}_{3 \times 3} & \vecmb{0}_{3 \times 3} \\
                   \vecmb{0}_{3 \times 3} & \Sigmaqavg             & \vecmb{0}_{3 \times 3} \\
                   \vecmb{0}_{5 \times 5} & \vecmb{0}_{5 \times 5} & \Sigmarxyc
                 \end{bmatrix}
        \in \R^{11 \times 11}
      \eeg

      \beg
        \Sigmazell = J \Sigma_{\rvec} J^T, \text{ with } J = \pard{R}{\rvec} \zhat \in \R^{3\times3} \nonumber\\
        \Sigmaqavg = \frac{1}{N^2} \sum_{i=1}^N \Sigma_i, \text{ with $\Sigma_i$ point's $\vec{q}_i$ covariance matrix} \nonumber
      \eeg
      
    % Step 2 (s = plane)    
    \item \textit{Surface Fitting}\\
      \textit{If Equation~(\ref{Eq:sidewallconstr}) is enabled for the side-wall effect, then the input $\Sigma$ is replaced by $J \Sigma J^T$, where $J = I_{11 \times 11}$ with $J(6:8,6) = \vec{\hat{n}}_p$.}\\
    
      \textbf{If (s = sphere)}\\\\
      \textbf{Input}: $\rxynew = \rxy$, k, $\cvec$, $\Sigmakrxyc \in \R^{6 \times 6}$ (from the WLM fitting)\\
      \textbf{Output}: $\Sigmakrxynewc \in \R^{6 \times 6}$\\
      
      \beg
        \Sigmakrxynewc = J \Sigmakrxyc J^T \in \R^{6 \times 6} \text{, since } \rxynew = \rxy
      \eeg
      
      where
      
      \beg
        J = \begin{bmatrix}
              \vecmb{0}_{1 \times 2} & \vecmb{0}_{3 \times 3} & \vecmb{0}_{1 \times 3} \\
              I_{2 \times 2}         & \vecmb{0}_{2 \times 1} & \vecmb{0}_{2 \times 3} \\
              \vecmb{0}_{3 \times 2} & \vecmb{0}_{5 \times 5} & I_{3 \times 3}
            \end{bmatrix} \in \R^{6 \times 6}
      \eeg
      
      \textbf{If (s = circ cyl)}\\\\
      \textbf{Input}: $\rnew = \rvec(\Rell) = \rvec([\xell\ \yell\ \zell])$ (log map), $\zell = R(\rvec) \zhat$, $\xell R(\rvec) \xhat$, $\yell = \zell \times \xell = [\xell]_x^T \zell = [\zell]_x \xell$,  $\Sigmakrc \in \R^{6 \times 6}$\\
      \textbf{Output}: $\Sigmakrnewc \in \R^{7 \times 7}$\\
      
      \beg
        \Sigma_{\rxy, \cnew} = J \Sigma J^T \in \R^{7 \times 7}
      \eeg
        
      where
      
      \beg
        J = \begin{bmatrix}
              \vecmb{0}_{1 \times 3} & \vecmb{0}_{1 \times 3} & 1 & \vecmb{0}_{1 \times 3}\\
              \pard{\rnew}{\zell} & \pard{\rnew}{\xell} & \vecmb{0}_{3 \times 1} & \vecmb{0}_{3 \times 3}\\
              \vecmb{0}_{3 \times 3} & \vecmb{0}_{3 \times 3} & \vecmb{0}_{3 \times 1} & I_{3 \times 3}
            \end{bmatrix}
        \in \R^{7 \times 10}
      \eeg
      
      \beg
        \pard{\rnew}{\zell} = \pard{\rnew}{\Rell} \pard{\Rell}{\zell},\ \pard{\Rell}{\zell} = \begin{bmatrix} \pard{\xell}{\zell} & \pard{\yell}{\zell} & \pard{\zell}{\zell} \end{bmatrix} \nonumber \\
        \pard{\xell}{\zell} =  \vecmb{0}_{1 \times 3},\ \pard{\yell}{\zell} = [\xell]_x^T,\ \pard{\zell}{\zell} = I_{3 \times 3} \nonumber
      \eeg
      
      \beg
        \pard{\rnew}{\xell} = \pard{\rnew}{\Rell} \pard{\Rell}{\xell},\ \pard{\Rell}{\xell} = \begin{bmatrix} \pard{\xell}{\xell} & \pard{\yell}{\xell} & \pard{\zell}{\xell} \end{bmatrix} \nonumber \\
        \pard{\xell}{\xell} =  I_{3 \times 3},\ \pard{\yell}{\xell} = [\zell]_x^T,\ \pard{\zell}{\xell} = \vecmb{0}_{3 \times 3} \nonumber
      \eeg
      
      and
      
      \beg
        \Sigma = \begin{bmatrix}
                   \Sigmazell             & \vecmb{0}_{3 \times 3} & \vecmb{0}_{3 \times 4} \\
                   \vecmb{0}_{3 \times 3} & \Sigmaxell             & \vecmb{0}_{3 \times 4} \\
                   \vecmb{0}_{4 \times 3} & \vecmb{0}_{4 \times 3}  & \Sigmakc
                 \end{bmatrix}
        \in \R^{10 \times 10}
      \eeg
      
      \beg
        \Sigmazell = J \Sigma_{\rvec} J^T, \text{with } J = \pard{R}{\rvec} \zhat \in \R^{3\times3} \nonumber\\
        \Sigmaxell = J \Sigma_{\rvec} J^T, \text{with } J = \pard{R}{\rvec} \xhat \in \R^{3\times3} \nonumber
      \eeg
      
    % Step 3 (s = parab)
    \item \textit{Curvature Discrimination} (if $s\!=\!\text{\emph{parab}}$)\\
      \textbf{If} $\max(|\kappa_x|,|\kappa_y|)<\epsilon_k$ ($s\!=\!\text{plane}$)\\
        \textbf{Input}: $\rxy = \rxy(\rvec)$, $\Sigmarxyc \in \R^{8 \times 8}$\\
        \textbf{Output}: $\Sigmarxyc \in \R^{5 \times 5}$\\
        
        \beg
          \Sigmarxyc = J \Sigmakrc J^T \in \R^{5 \times 5}
        \eeg
        
        where
        
        \beg
          J = \begin{bmatrix}
                \vecmb{0}_{2 \times 2} & \pard{\rxy}{\rvec}     & \vecmb{0}_{3 \times 3} \\
                \vecmb{0}_{3 \times 2} & \vecmb{0}_{3 \times 3} & \vecmb{0}_{3 \times 3}
              \end{bmatrix} \in \R^{5 \times 8}
        \eeg
      
      \textbf{Else if} $\min(|\kappa_x|,|\kappa_y|)<\epsilon_k$ ($s=\text{cyl parab}$)\\
        \underline{If} $|\kappa_y|>\epsilon_k$\\
        \textbf{Input}: $\kappa=\kappa_y$, $\Sigmarxyc \in \R^{8 \times 8}$\\
        \textbf{Output}: $\Sigmakrc \in \R^{7 \times 7}$\\
        
        \beg
          \Sigmakrc = J \Sigmakrc J^T \in \R^{7 \times 7}
        \eeg
        
        where
        
        \beg
          J = \begin{bmatrix}
                [0\ 1]                 & \vecmb{0}_{1 \times 3} & \vecmb{0}_{1 \times 3} \\
                \vecmb{0}_{3 \times 2} & I_{3 \times 3}         & \vecmb{0}_{3 \times 3} \\
                \vecmb{0}_{3 \times 2} & \vecmb{0}_{3 \times 3} & I_{3 \times 3}
              \end{bmatrix} \in \R^{7 \times 8}
        \eeg
        
        \underline{Else} swap axes\\
        \textbf{Input}: $\kappa=\kappa_x$, $\rnew = \rvec (R(\rvec) [\yhat\ -\xhat\ \zhat])$ (log map), $\Sigmakrc \in \R^{8 \times 8}$\\
        \textbf{Output}: $\Sigmakrnewc \in \R^{7 \times 7}$\\
        
        \beg
          \Sigmakrc = J \Sigmakrc J^T \in \R^{7 \times 7}
        \eeg
        
        where
        
        \beg
          J = \begin{bmatrix}
                [0\ 1]                 & \vecmb{0}_{1 \times 3} & \vecmb{0}_{1 \times 3} \\
                \vecmb{0}_{3 \times 2} & \pard{\rnew}{\rvec}    & \vecmb{0}_{3 \times 3} \\
                \vecmb{0}_{3 \times 2} & \vecmb{0}_{3 \times 3} & I_{3 \times 3}
              \end{bmatrix} \in \R^{7 \times 8}
        \eeg
        
        \beg
          \pard{\rnew}{\rvec} = \pard{\rnew}{U} \pard{U}{\rvec},\ \pard{U}{\rvec} = \pard{R}{\rvec} W \nonumber \\
          U = R(\rvec) W,\ W = [\yhat\ -\xhat\ \zhat] \nonumber
        \eeg
      
      \textbf{Else if} $|\kappa_x-\kappa_y|<\epsilon_k$ ($s=\text{circ parab}$)\\
        \textbf{Input}: $\kappa=\frac{\kappa_x+\kappa_y}{2}$, $\rxy = \rxy (\rvec)$, $\Sigmakrc \in \R^{8 \times 8}$\\
        \textbf{Output}: $\Sigmakrxyc \in \R^{6 \times 6}$\\
        
        \beg
          \Sigmakrc = J \Sigmakrc J^T \in \R^{6 \times 6}
        \eeg
        
        where
        
        \beg
          J = \begin{bmatrix}
                [1/2\ 1/2]             & \vecmb{0}_{1 \times 3} & \vecmb{0}_{1 \times 3} \\
                \vecmb{0}_{3 \times 2} & \pard{\rxy}{\rvec}    & \vecmb{0}_{3 \times 3} \\
                \vecmb{0}_{3 \times 2} & \vecmb{0}_{3 \times 3} & I_{3 \times 3}
              \end{bmatrix} \in \R^{6 \times 8}
        \eeg

      \textbf{Else} ($s=\text{ell parab or hyp parab}$)\\
        No change in $\kvec$, $\rvec$, and $\cvec$
    \end{list}
    
    \item Fit the Boundary\\
      % Step 5    
      \begin{list}{\it Step \arabic{error-step}:}{\usecounter{error-step} \setcounter{error-step}{4}}
        \item \textit{Initialize Bounding Parameters}\\
        \textbf{Input}: $\mvec = \begin{bmatrix}\xavg & \yavg &  \vx & \vy & \vxy \end{bmatrix}^T$, $\Sigmakrc \in \R^{(n_k + n_r + 3)^2}$\\
        \textbf{Output}: $\Sigmamkrc \in \R^{(5 + n_k + n_r + 3)^2}$\\
        
        Let
        \beg
          \mvec = \begin{bmatrix} \xavg \\ \yavg \\  \vx \\ \vy \\ \vxy \end{bmatrix} =
                  \begin{bmatrix} \xhat^T X_r(\qvec_i,\rvec, \cvec) \\ \yhat^T X_r(\qvec_i,\rvec, \cvec) \\  (\xhat^T X_r(\qvec_i,\rvec, \cvec))^2 \\ (\yhat^T X_r(\qvec_i,\rvec, \cvec))^2 \\ (\xhat^T X_r(\qvec_i,\rvec, \cvec))(\yhat^T X_r(\qvec_i,\rvec, \cvec)) \end{bmatrix}
        \eeg
        where $\qnew_i \defeq X_r(\qvec_i,\rvec, \cvec) = R(-\rvec)(\qvec_i-\cvec) = (R(\rvec))^T(\qvec_i-\cvec)$
        
        \beg
          \Sigmamkrc = J \Sigmaqkrc J^T \R^{(5 + n_k + n_r + 3)^2}
        \eeg
        
        where
        
        \beg
          J = \begin{bmatrix}
                \pard{\mvec}{\qvec_1} \ldots \pard{\mvec}{\qvec_N} & \vecmb{0}_{5 \times n_k} & \pard{\mvec}{\rvec} & \pard{\mvec}{\cvec} \\
                \vecmb{0}_{n_k \times 3} \ldots \vecmb{0}_{n_k \times 3} & I_{n_k \times n_k} & \vecmb{0}_{n_k \times n_r} & \vecmb{0}_{n_k \times 3} \\
                \vecmb{0}_{n_r \times 3} \ldots \vecmb{0}_{n_r \times 3} & \vecmb{0}_{n_r \times n_k} & I_{n_r \times n_r} & \vecmb{0}_{n_r \times 3} \\
                \vecmb{0}_{3 \times 3} \ldots \vecmb{0}_{3 \times 3} & \vecmb{0}_{3 \times n_k} & \vecmb{0}_{3 \times n_r} & I_{3 \times 3}
              \end{bmatrix}
              \in R^{(5+n_k+n_r+3) \times (5N+n_k+n_r+3)}
        \eeg
        
        \beg
          \pard{\mvec}{\qvec_i} = \frac{1}{N} \begin{bmatrix} \xhat^T \\ \yhat^T \\ 2(\xhat^T \qnew_i) \xhat^T \\ 2(\yhat^T \qnew_i) \yhat^T \\ (\xhat^T \qnew_i) \yhat^T + \xhat^T (\yhat^T \qnew_i) \end{bmatrix} \pard{\qnew_i}{\qvec_i} \in \R^{5 \times 3}, \nonumber \\
          \pard{\qnew_i}{\qvec_i} = R(-\rvec) = (R(\rvec))^T \nonumber
        \eeg
        
        \beg
          \pard{\mvec}{\rvec} = \sum_{i=1}^N{\pard{\mvec}{\qnew_i} \pard{\qnew_i}{\rvec}},\ \pard{\qnew_i}{\rvec} = \pard{R^T}{\rvec} (\qvec_i - \cvec) \nonumber
        \eeg
        
        \beg
          \pard{\mvec}{\cvec} = (\sum_{i=1}^N{\pard{\mvec}{\qvec_i}}) \pard{\qnew_i}{\cvec},\ \pard{\qnew_i}{\cvec} = -R(-\rvec) = -(R(\rvec))^T \nonumber
        \eeg
      \end{list}
      
      % Step 6
      \begin{list}{\it Step \arabic{error-step}:}{\usecounter{error-step} \setcounter{error-step}{5}}
        \item \textit{Cylindrical Paraboloid and Circular Cylinder Boundary Fitting}\\
        \textbf{Input}: $\cnew = R(\rvec)(\xavg \xhat) + \cvec$, $\dvec_r=\lambda[\sqrt{v_x-\xavg^2}\ \sqrt{v_y}]^T$, $\Sigmamkrc \in \R^{12 \times 12}$\\
        \textbf{Output}: $\Sigmadrkrc \in \R^{9 \times 9}$\\
        
        \beg
          \Sigmadrkrc = J \Sigmamkrc J^T \in \R^{12 \times 12}
        \eeg
        
        where
        
        \beg
          J = \begin{bmatrix}
                \pard{\dvec_r}{\mvec}  & \vecmb{0}_{2 \times 1} & \vecmb{0}_{2 \times 3} & \vecmb{0}_{2 \times 3} \\
                \vecmb{0}_{1 \times 9} & 1 & \vecmb{0}_{1 \times 3} & \vecmb{0}_{1 \times 3} \\
                \vecmb{0}_{3 \times 5} & \vecmb{0}_{3 \times 1} & I_{3 \times 3} & \vecmb{0}_{3 \times 3} \\
                \pard{\cnew}{\mvec}  & \vecmb{0}_{3 \times 1} & \pard{\cnew}{\rvec} & \pard{\cnew}{\cvec}
              \end{bmatrix}
              \in R^{9 \times 12}
        \eeg
        
        \beg
          \pard{\dvec_r}{\mvec} = \lambda \begin{bmatrix} -\xavg(\vx-\xavg^2)^{-\frac{1}{2}} & 0 & \frac{1}{2}(\vx-\xavg^2)^{-\frac{1}{2}}) & 0 & 0 \\ 0 & 0 & 0 & \frac{1}{2} \vy^{-\frac{1}{2}} & 0 \end{bmatrix} \nonumber
        \eeg
        
        \beg
          \pard{\cnew}{\mvec} = \begin{bmatrix} R(\rvec) & \xhat & \vecmb{0}_{3 \times 1} \end{bmatrix}, \ \nonumber
          \pard{\cnew}{\rvec} = \pard{R}{\rvec} (\xavg \xhat), \ \nonumber
          \pard{\cnew}{\cvec} = I_{3 \times 3} \nonumber
        \eeg
        
        % Step 7
        \item \textit{Circular Paraboloid and Sphere Boundary Fitting}\\
        \textbf{Input}: $d_c=\lambda\max(\sqrt{\vx}, \sqrt{\vy})$, $\Sigmamkrxyc \in \R^{11 \times 11}$\\
        \textbf{Output}: $\Sigmadckrxyc \in \R^{9 \times 9}$\\
        
        \beg
          \Sigmadckrxyc = J \Sigmamkrxyc J^T \in \R^{11 \times 11}
        \eeg
        
        where
        
        \beg
          J = \begin{bmatrix}
                \pard{\dvec_c}{\mvec}  & 0 & \vecmb{0}_{1 \times 2} & \vecmb{0}_{1 \times 3} \\
                \vecmb{0}_{1 \times 5} & 1 & \vecmb{0}_{1 \times 2} & \vecmb{0}_{1 \times 3} \\
                \vecmb{0}_{2 \times 5} & \vecmb{0}_{2 \times 1} & I_{2 \times 2} & \vecmb{0}_{2 \times 3} \\
                \vecmb{0}_{3 \times 5} & \vecmb{0}_{3 \times 1} & \vecmb{0}_{3 \times 2} & I_{3 \times 3}
              \end{bmatrix}
              \in R^{7 \times 11}
        \eeg
        
        If $\vx > \vy$, then $\dvec_c = \sqrt{\vx}$
        \beg
          \pard{\dvec_c}{\mvec} = \lambda \begin{bmatrix} 0 & 0 & \frac{1}{2}\vx^{-\frac{1}{2}} & 0 & 0 \end{bmatrix} \nonumber
        \eeg
        
        Else if $\vx < \vy$, then $\dvec_c = \sqrt{\vy}$
        \beg
          \pard{\dvec_c}{\mvec} = \lambda \begin{bmatrix} 0 & 0 & 0 & \frac{1}{2}\vy^{-\frac{1}{2}}  & 0 \end{bmatrix} \nonumber
        \eeg
        
        Else $\vx = \vy$, then $\dvec_c = \frac{\sqrt{\vx}+\sqrt{\vy}}{2}$
        \beg
          \pard{\dvec_c}{\mvec} = \lambda \begin{bmatrix} 0 & 0 & \frac{1}{2}\vx^{-\frac{1}{2}} & \frac{1}{2}\vy^{-\frac{1}{2}} & 0 \end{bmatrix} \nonumber
        \eeg
        
        % Step 8
        \item \textit{Elliptic and Hyperbolic Boundary Fitting}\\
        \textbf{Input}: $\dvec_e=\lambda[\sqrt{\vx}\ \sqrt{\vy}]^T$, $\Sigmamkrc \in \R^{11 \times 11}$\\
        \textbf{Output}: $\Sigmadekrc \in \R^{13 \times 13}$\\
        
        \beg
          \Sigmadekrc = J \Sigmamkrc J^T \in \R^{13 \times 13}
        \eeg
        
        where
        
        \beg
          J = \begin{bmatrix}
                \pard{\dvec_e}{\mvec}  & \vecmb{0}_{2 \times 2} & \vecmb{0}_{2 \times 3} & \vecmb{0}_{2 \times 3} \\
                \vecmb{0}_{2 \times 5} & I_{2 \times 2} & \vecmb{0}_{2 \times 3} & \vecmb{0}_{2 \times 3} \\
                \vecmb{0}_{3 \times 5} & \vecmb{0}_{3 \times 2} & I_{3 \times 3} & \vecmb{0}_{3 \times 3} \\
                \vecmb{0}_{3 \times 5} & \vecmb{0}_{3 \times 2} & \vecmb{0}_{3 \times 3} & I_{3 \times 3}
              \end{bmatrix}
              \in R^{13 \times 10}
        \eeg
        
        \beg
          \pard{\dvec_e}{\mvec} = \lambda \begin{bmatrix} 0 & 0 & \frac{1}{2}\vx^{-\frac{1}{2}} & 0 & 0 \\ 0 & 0 & 0 &\frac{1}{2}\vy^{-\frac{1}{2}} & 0 \end{bmatrix}\nonumber
        \eeg
      \end{list}
      
      % Step 9
      \begin{list}{\it Step \arabic{error-step}:}{\usecounter{error-step} \setcounter{error-step}{8}}
        \item \textit{Plane Boundary Fitting} (if $s=\text{plane}$)\\
          \textbf{Input}: $\alpha$, $\beta$, $\phi$, $c{+,-}$, $\Sigmamrxyc \in \R^{10 \times 10}$\\
          \textbf{Output}: $\Sigmaellrhorxyc \in \R^{10 \times 10}$\\
            
          Let
          \beg
            \rhovec \defeq \begin{bmatrix} \alpha \\ \beta \\ \phi \end{bmatrix} = \begin{bmatrix} \vx-\xavg^2 \\  2 \vxy -\xavg \yavg \\ \vy-\yavg^2 \end{bmatrix} \nonumber \\
            \lvec \defeq \begin{bmatrix} l_+ \\ l_- \end{bmatrix} = \sqrt{-\ln(1-\Gamma)} \begin{bmatrix} \sqrt{\alpha + \phi + \sqrt{D})} \\ \sqrt{\alpha + \phi - \sqrt{D})} \end{bmatrix} \nonumber \\
            w_{\pm} \defeq \sqrt{\alpha + \phi \pm \sqrt{D})} \text{ and } D \defeq \beta^2+(\alpha-\phi)^2 \nonumber
          \eeg
            
          \beg
            \cnew = X_f(\xavg \xhat + \yavg \yhat, \rxy, \cvec) = R(\begin{bmatrix} \rxy \\ 0 \end{bmatrix}) (\xavg \xhat + \yavg \yhat) + \cvec \nonumber \\
            d_c = \max{(l_+, l_-)} \nonumber \\
            \dvec_e = \dvec_r = \lvec \nonumber \\
            \dvec_q = [\norm{\lvec} \ \norm{\lvec} \ \norm{\lvec} \ \norm{\lvec} \ \norm{\gamma}]^T \ \ \gamma \defeq \atantwo(l_-,l_+) \nonumber
          \eeg
            
           \beg
             \zell = R([\rxy 0]^T) \zhat \nonumber \\
             \xell = [\cos{\theta} \ \sin{\theta} \ 0],\ \theta = \frac{1}{2}\atantwo(\beta, \alpha-\phi) \nonumber \\
             \yell = \zell \times \xell = [\xell]_x^T \zell = [\zell]_x^T \xell \nonumber \\
             \rvec = \rvec([\xell \ \yell  \ \zell]) \text{ (log map)} \nonumber
           \eeg
        
        The covariance matrix is:
        \beg
          \Sigmaellrhorxyc = J \Sigmamrxyc J^T \in \R^{10 \times 10}
        \eeg
        
        where
        
        \beg
          J = \begin{bmatrix}
                \pard{\lvec}{\mvec}    & \vecmb{0}_{2 \times 2}  & \vecmb{0}_{2 \times 3}  \\
                \pard{\rhovec}{\mvec}  & \vecmb{0}_{3 \times 2}  & \vecmb{0}_{3 \times 3} \\
                \vecmb{0}_{2 \times 5} & I_{3 \times 2}          & \vecmb{0}_{2 \times 3} \\
                \pard{\cnew}{\mvec}    & \pard{\cnew}{\rxy}      & \pard{\cnew}{\cvec}
              \end{bmatrix}
              \in R^{13 \times 10}
        \eeg
        
        \beg
          \pard{\lvec}{\mvec} = \pard{\lvec}{\rhovec} \pard{\rhovec}{\mvec} \nonumber \\
          \pard{\lvec}{\rhovec} = \sqrt{- \ln{(1-\Gamma)}} \begin{bmatrix} \frac{1}{2} w_+^{-\frac{1}{2}} \pard{w_+}{\rhovec} \\ \frac{1}{2} w_-^{-\frac{1}{2}} \pard{w_-}{\rhovec} \end{bmatrix} \nonumber \\
          \pard{w_{+,-}}{\rhovec} = [(1 \pm D^{-\frac{1}{2}} (\alpha-\phi)) \ (\pm D^{-\frac{1}{2}} \beta) \ (1 \pm D^{-\frac{1}{2}} (\alpha - \beta))] \nonumber
        \eeg
        
        \beg
          \pard{\rhovec}{\mvec} =
            \begin{bmatrix}
              -2\xavg & 0       & 1 & 0 & 0  \\
              -\yavg  & -\xavg  & 0 & 0 & 2 \\
              0       & -2\yavg & 0 & 1 & 0
            \end{bmatrix} \nonumber
        \eeg
        
        \beg
          \pard{\cnew}{\mvec} = R([\rxy \ 0]) [\xavg \ \yavg \ 0 \ 0 \ 0 ], \ \nonumber
          \pard{\cnew}{\rxy} = \pard{R}{\rxy} (\xavg \xhat + \yavg \yhat), \ \nonumber
          \pard{\cnew}{\cvec} = I_{3 \times 3} \nonumber
        \eeg
        
        \begin{enumerate}
          \item If $b\!\!=\!\!\text{circle}$\\
            \textbf{Input}: $d_c\!\!=\!\!\max(l_+,l_-)$, $\Sigmaellrhorxyc \in \R^{10 \times 10}$\\
            \textbf{Output}: $\Sigmadcrxyc \in \R^{6 \times 6}$\\
          
            % Jacobians
            \beg
              \Sigmadcrxyc = J \Sigmaellrhorxyc J^T \R^{6 \times 6}
            \eeg
        
            where
        
            \beg
              J = \begin{bmatrix}
                \pard{d_c}{\lvec}      & \vecmb{0}_{1 \times 3}  & \vecmb{0}_{1 \times 2} & \vecmb{0}_{1 \times 3}\\
                \vecmb{0}_{2 \times 2} & \vecmb{0}_{2 \times 3}  & I_{2 \times 2}         & \vecmb{0}_{2 \times 3}\\
                \vecmb{0}_{3 \times 2} & \vecmb{0}_{3 \times 3}  & \vecmb{0}_{3 \times 2} & I_{3 \times 3}\\
              \end{bmatrix}
              \in R^{6 \times 10}
           \eeg
    
           if $l_+ > l_-$, then $d_c = l_+$ and $\pard{d_c}{\lvec} = [1 \ 0]$\\
           if $l_+ < l_-$, then $d_c = l_-$ and $\pard{d_c}{\lvec} = [0 \ 1]$\\
           if $l_+ = l_-$, then $d_c = (l_+ + l_-)/2$ and $\pard{d_c}{\lvec} = [1/2 \ 1/2]$
    
            \item If $b\!\!\in\!\!\{\text{ellipse},\text{aarect},text{conv quad}\}$\\
            \textbf{Input}: $\dvec$, $ \in \Sigmaellrhorxyc \R^{10 \times 10}$\\
            \textbf{Output}: $\Sigmadrc \in \R^{(n_d+6) \times (n_d+6)}$\\
            Let $R_l = [\xell \ \yell \ \zell]$.
            
            \beg
              \Sigmadrc = J \Sigmaellrhorxyc J^T \in \R^{(n_d+6) \times (n_d+6)}
            \eeg
        
            where
            
            \beg
              J = \begin{bmatrix}
                \pard{\dvec}{\lvec}    & \vecmb{0}_{n_d \times 3} & \vecmb{0}_{n_d \times 2} & \vecmb{0}_{n_d \times 3}\\
                \vecmb{0}_{3 \times 2} & \pard{\rvec}{\rhovec}    & \pard{\rvec}{\rxy}       & \vecmb{0}_{3 \times 3}\\
                \vecmb{0}_{3 \times 2} & \vecmb{0}_{3 \times 3}   & \vecmb{0}_{3 \times 2}   & I_{3 \times 3}
              \end{bmatrix}
              \in R^{6 \times 10}
           \eeg
           
           \underline{If  $b\!\!\in\!\!\{\text{ellipse},\text{aarect}\}$}
           \beg
             \pard{\dvec}{\lvec} = I_{2 \times 2} \nonumber
           \eeg
           
           \underline{If $b\!\!\in\!\!\{\text{conv quad}\}$}
           \beg
             \pard{\dvec_q}{\lvec} = \begin{bmatrix} \pard{\norm{\lvec}}{\lvec} & \pard{\norm{\lvec}}{\lvec} & \pard{\norm{\lvec}}{\lvec} & \pard{\norm{\lvec}}{\lvec} & \pard{\gamma}{\lvec} \end{bmatrix} \ \gamma\defeq \atantwo{(l_-, l_+)} \nonumber \\
             \pard{\norm{\lvec}}{\lvec} = \frac{\lvec^T}{\norm{\lvec}} \nonumber \\
             \pard{\gamma}{\lvec} = l_+ \pard{l_+}{\lvec} - l_- \pard{l_-}{\lvec} = l_+ [1 \ 0] - l_- [0 \ 1] = [l_+ \ l_-] \nonumber
           \eeg
           
           \beg
             \pard{\rvec}{\rhovec} = \pard{\rvec}{R_l} \pard{R_l}{\rhovec} \nonumber \\ 
             \pard{\rvec}{R_l} (from Appendix~\ref{Sec:jacobians}) \nonumber \\
             \pard{R_l}{\rhovec} = \begin{bmatrix}\pard{\xell}{\rhovec} & \pard{\yell}{\rhovec} &\pard{\zell}{\rhovec} \end{bmatrix} \nonumber
           \eeg
           
           \beg
             \pard{\xell}{\rhovec} = \begin{bmatrix} -\sin{\theta} \\ \cos{\theta} \\ 0 \end{bmatrix} \pard{\theta}{\rhovec} \nonumber \\
             \pard{\theta}{\rhovec} = \begin{bmatrix} \pard{\theta}{\alpha} \\ \pard{\theta}{\beta} \\ \pard{\theta}{\phi} \end{bmatrix} \nonumber \\
             \pard{\theta}{\alpha} = \frac{-\beta}{2}, \ \pard{\theta}{\beta} = \frac{\alpha-\phi}{2}, \ \pard{\theta}{\phi} = \frac{\beta}{2} \nonumber \\
             \pard{\yell}{\rhovec} = [\zell]_x \pard{\xell}{\rhovec} \nonumber \\
             \pard{\zell}{\rhovec} = \vecmb{0}_{[3 \times 1] \times 3} \nonumber
           \eeg
           
           \beg
             \pard{\rvec}{\rxy} = \pard{\rvec}{R_l} \pard{R_l}{\rxy} \nonumber \\
             \pard{R_l}{\rxy} = \begin{bmatrix} \pard{\xell}{\rxy} & \pard{\yell}{\rxy} & \pard{\zell}{\rxy} \end{bmatrix} \nonumber
           \eeg
           
           \beg
             \pard{\xell}{\rxy} = \pard{\xell}{\rhovec} \pard{\rhovec}{\mvec} \pard{\mvec}{\rxy} \nonumber
           \eeg
           
           \beg
             \pard{\yell}{\rxy} = (\pard{[\zell]_x}{\zell} \pard{\zell}{\rxy}) \xell + [\zell]_x \pard{\xell}{\rxy} \nonumber \\
             \pard{[\zell]_x}{\zell} = \begin{bmatrix} \begin{bmatrix}  0 & 0 & 0 \\ 0 & 0 & -1 \\ 0 & 1 & 0 \end{bmatrix} & \begin{bmatrix} 0 & 0 & 1 \\ 0 & 0 & 0 \\ -1 & 0 & 0 \end{bmatrix} & \begin{bmatrix} 0 & -1 & 0 \\ 1 & 0 & 0 \\ 0 & 0 & 0 \end{bmatrix} \end{bmatrix} \nonumber \\
             \pard{\zell}{\rxy} = \pard{R}{\rxy} \zhat \nonumber
           \eeg
           
          \end{enumerate}
      \end{list}
\end{list} \cleardoublepage
\section{Taubin's Residual Approximations} \label{Sec:Taubin}
The distance $\delta$ of a point $p=(p_x,p_y,p_z) \in \R^3$ from a paraboloid whose implicit (local) form is $f(x,y,z) = k_x x^2 + k_y y^2 -2z$, using the first order Taubin's~\cite{Taubin93} approximation is:
\beg
    \delta = \frac{F_{0}(p)}{\sqrt{F_{2}(p)}} \\
    F_{0}(p) = \| k_x x^2 + k_y y^2 -2z \| \nonumber \\
    F_{2}(p) = 4 (k_x^2 + k_y^2 + 1) \nonumber
\eeg

Using the second order Taubin approximation the distance $\delta$ is computed in three steps:
\begin{enumerate}
  \item Taylor series coefficient computation\\
  \beg
    F_{0,0,0}(p) = \| k_x x^2 + k_y y^2 -2z \| \nonumber \\
    F_{1,0,0}(p) = 2 k_x p_x, \nonumber \
    F_{0,1,0}(p) = 2 k_y p_y, \nonumber \
    F_{0,0,1}(p) = -2 \nonumber \\
    F_{1,1,0}(p) = \frac{\partial^2{f(p)}}{\partial{x} \partial{y}} = 0 \nonumber \\
    F_{1,0,1}(p) = \frac{\partial^2{f(p)}}{\partial{x} \partial{z}} = 0 \nonumber \\
    F_{0,1,1}(p) = \frac{\partial^2{f(p)}}{\partial{y} \partial{z}} = 0 \nonumber \\
    F_{2,0,0}(p) = k_x, \nonumber \
    F_{0,2,0}(p) = k_y, \nonumber \
    F_{0,0,2}(p) = 0 \nonumber
  \eeg
  
  \item Fh(p) computation for h=1,2\\
  \beg
    F_{0}(p) = \| k_x x^2 + k_y y^2 -2z \| \nonumber
  \eeg
  \beg  
    F_{1}(p) = -\begin{bmatrix} \bpm 1 \\ 1 \epm^{-1} F_{1,0,0}(p)^2 + \bpm 1 \\ 1 \epm^{-1} F_{0,1,0}(p)^2 + \bpm 1 \\ 1 \epm^{-1} F_{0,0,1}(p)^2\end{bmatrix}^{\frac{1}{2}} = \nonumber \\
    -\begin{bmatrix} 4k_x^2p_x^2 + 4k_y^2p_y^2 + 42\end{bmatrix}^{\frac{1}{2}} = -2 \sqrt{k_x^2p_x^2 + k_y^2p_y^2 + 1}\nonumber
  \eeg
  \beg     
    F_{2}(p) = -\begin{bmatrix} \bpm 2 \\ 1 \epm^{-1} F_{1,1,0}(p)^2 + \bpm 2 \\ 1 \epm^{-1} F_{1,0,1}(p)^2 + \bpm 2 \\ 1 \epm^{-1} F_{0,1,1}(p)^2 + \\ \bpm 2 \\ 2 \epm^{-1} F_{2,0,0}(p)^2 + \bpm 2 \\ 2 \epm^{-1} F_{0,2,0}(p)^2 + \bpm 2 \\ 2 \epm^{-1} F_{0,0,2}(p)^2\end{bmatrix}^{\frac{1}{2}} = \nonumber \\
    -\begin{bmatrix} k_x^2 + k_y^2\end{bmatrix}^{\frac{1}{2}} = - \sqrt{k_x^2 + k_y^2}\nonumber
  \eeg
  
  \item Solve for the min positive root of:
  \beg
    F_2(p) \delta^2 + F_1(p)\delta + F_0(p) = 0
  \eeg
  where $\delta$ is the approximate min Euclidean distance.
\end{enumerate}

\cleardoublepage%********************************************************************
% Bibliography
%*******************************************************
% work-around to have small caps also here in the headline
\manualmark
\markboth{\spacedlowsmallcaps{\bibname}}{\spacedlowsmallcaps{\bibname}} % work-around to have small caps also
%\phantomsection 
\refstepcounter{dummy}
\addtocontents{toc}{\protect\vspace{\beforebibskip}} % to have the bib a bit from the rest in the toc
\addcontentsline{toc}{chapter}{\tocEntry{\bibname}}
\bibliographystyle{plainnat}
\label{app:bibliography} 
\bibliography{Bibliography}

\end{document}